# Accelerating Entrepreneurial Decision-Making Through Hybrid Intelligence

# Accelerating Entrepreneurial Decision-Making Through Hybrid Intelligence

DESIGN PARADIGMS AND PRINCIPLES FOR DECISIONAL GUIDANCE IN ENTREPRENEURSHIP

**Dominik Dellermann**

## Dedication

*For my father Hans and my trainer Werner -*
*The persons who most influenced me throughout my life.*

# Table of Contents

# List of Abbreviations

| | | |
|---|---|---|
| AI | - | Artificial Intelligence |
| AGI | - | Artificial General Intelligence |
| ANN | - | Artificial Neural Network |
| ANOVA | - | Analysis of Variance |
| ANT | - | Actor Network Theory |
| API | - | Application Programming Interface |
| APX | - | Amsterdam Power Exchange |
| AVE | - | Average Variance Extracted |
| BU | - | Business Unit |
| CART | - | Classification and Regression Tree |
| CBMV | - | Crowd-based Business Model Validation |
| CR | - | Composite Reliability |
| CT | - | Computed Tomography |
| CVC | - | Corporate Venture Capital |
| DR | - | Design Requirement |
| DP | - | Design Principle |
| DSR | - | Design Science Research |
| DSS | - | Decision Support System |
| EEX | - | European Energy Exchange |
| FsQCA | - | Fuzzy-Set Qualitative Comparative Analysis |
| GUI | - | Graphical User Interface |
| HI-DSS | - | Hybrid Intelligence Decision Support System |
| HIT | - | Human Intelligence Task |
| IoT | - | Internet of Things |
| IS | - | Information System |
| IT | - | Information Technology |
| MCC | - | Matthews Correlation Coefficient |

| ML | - | Machine Learning |
| OCT | - | Opportunity Creation Theory |
| OGEMA 2.0 | - | Open Gateway Energy Management 2.0 |
| OS | - | Operating System |
| R&D | - | Research & Development |
| RE | - | Renewable Energies |
| RQ | - | Research Question |
| SVM | - | Support Vector Machine |
| SSD | - | Solid-State Drive |
| SDK | - | Software Development Kit |
| TCP/IP | - | Transmission Control Protocol/Internet Protocol |
| TCT | - | Transaction Cost Theory |
| UI | - | User Interface |
| VaR | - | Value at Risk |
| VC | - | Venture Capital |
| VPP | - | Virtual Power Plant |

# Chapter I

## Prologue

# 1. Introduction

## 1.1. Problem Definition

### Relevance and Motivation

Technological advances such as mobile computing, 3D printing, or cloud computing enable the creation of novel opportunities for entrepreneurs to create and capture value. However, previous studies revealed that around 75 percent of all start-ups fail at an early stage. This is also true for innovation projects and other forms of innovation related endeavour in incumbent firms (Blank 2013).

One main reason for this tremendous failure rate is that entrepreneurs are typically confronted with high levels of uncertainty about the viability of their proposed business idea. One prominent perspective is that opportunities for such novel business ideas cannot be just discovered by entrepreneurs in the market. Rather, they are endogenously created by actions of an entrepreneur who seeks to actively exploit it in a multistage and iterative process of interaction between herself and the environment (Alvarez et al. 2013). This is especially relevant in the age of digital innovation where entrepreneurial efforts become even more dynamic and dependent on the external ecosystem such as platform owners (Dellermann et al. 2016), partners and customers (Kolloch and Dellermann 2017), or other distributed stakeholders (Nambisan 2017).

Following this argumentation, entrepreneurial decision-making can be defined as complex decision-making problem under both risk and uncertainty (Knight, 1921). While risk includes quantifiable probabilities, uncertainty describes situations where neither outcomes nor their probability distribution can be assessed a priori (Diebold et al. 2010). Consequently, the entrepreneurial decision-making context is highly complex and contains lots of "*black swan events*" that seems to be unpredictable (Russell and Norvig 2016; Simon 1991; Funke 1991).

For this thesis, I identified several gaps in previous research, which I aim to address with my dissertation.

**Research Gap 1 –** *Limited Investigation of the Sources of Risk and Uncertainty in the Entrepreneurial Decision-Making Context.*

The first gap in previous research is related to the lack of understanding of the sources of risk and uncertainty in entrepreneurial decision-making. Little is known about the role of the ecosystem of users, suppliers, partners, and other stakeholders in making decisions. Most research in this field is rather descriptive or conceptual at all (e.g. Alvarez and Barney 2007, Alvarez et al. 2014). Consequently, the lack of empirical investigation of the sources of both risk and uncertainty in the entrepreneurial decision-making contexts as well as the role of the ecosystem as source of those, is the first research gap that was identified.

**Research Gap 2 –** *Limited Investigation of Scalable Mechanisms for Decisional Guidance in Entrepreneurial Decision-Making.*

The second research gap that I identified is related to the mechanisms applied for providing decisional guidance which supports and offers advice to a person regarding what to do (Silver 1991). To support entrepreneurs in making their decisions, feedback from social interaction with domain experts proved to be a valuable strategy in managerial practice. Consequently, the dominant form of decision support that emerges is human mentoring (Hochberg 2016) . However, human generated decisional guidance holds also various limitations that can be subsumed under two dimensions: cognitive limitations (e.g. limited information processing capabilities, expertise, flexibility, or biases) that prevent individual experts from providing optimal guidance, and resource constraints (e.g. time constraints, financial resources, social capital, and demand side knowledge) (Zhang and Cueto 2017; Shepherd 2015; Shepherd et al. 2015; Dellermann et al.

2018a). Both limitations prevent from providing optimal, scalable, and iterative decisional guidance for entrepreneurial decision-making and limit the integration of stakeholders in this process. Consequently, I identified the lack of investigation of scalable mechanisms that allows iterative integration of stakeholders in guiding entrepreneurial decision-making as the second gap in previous work.

**Research Gap 3** – *Limited Investigation of IT-supported Decisional Guidance and DSS for Complex Decision-Making under Uncertainty and Risk.*

The third gap in the current body of knowledge is related to the design of IT-supported decisional guidance for classes of complex decision-making problems under both risk and uncertainty. Decisional guidance has been proven as a suitable approach in research on decision support systems in various contexts of IS research (Silver 1991; Morana et al. 2017; Parikh et al. 2001; Limayem and DeSanctis 2000).

Although the adaption of these findings to the context of entrepreneurial decision-making is promising, previous research provides little knowledge on both design principles (abstracted design knowledge) and design paradigms (general rational for the decisional guidance provided) for complex decision-making problems under both uncertainty and risk. While DSS that are based on statistical models are consistent (experts are subject to random fluctuations), are potentially less biased by a non-random sample, and optimally weigh information factors, previous work on DSS provides little knowledge on systems that can deal with a such complex class of problems like entrepreneurial decision making. First, despite of advances in deep learning techniques (LeCun et al. 2015), such systems are constrained by a lack of adaptability and are not capable to capture the complex dynamic interactions between elements that are required for providing decisional guidance for situations that require dealing with extreme uncertainty (Slovic and Fischhoff 1988; Zacharakis and Meyer 2000).

Second, such methods are having troubles with processing *"soft information"* (e.g. creativity) or tacit learning experience, which is required to provide decisional guidance for complex problems. Finally, statistical methods struggle with so called *"black swan"/" broken leg"* events (Dawes et al. 1989) in in which humans are surprisingly good at predicting with a combination of intuitive and analytical reasoning. Consequently, I identified the lack of investigation design knowledge on decisional guidance and DSS for complex decision-making problems under uncertainty and risk such as in the entrepreneurial context as the third major gap in previous work.

## Purpose and Scope

Guidance in general proved to be valuable to accelerate entrepreneurial decision-making despite its limitations. Consequently, the idea of this dissertation is to design mechanisms for providing efficient and effective decisional guidance to entrepreneurs that can constraints of human mentoring, integrate stakeholders, and alleviate limitations of recent statistical methods of intelligent decision support systems.

For this thesis, I use the term **design paradigm** as the general rational for the decisional guidance provided, which is collective intelligence/ crowdsourcing (Chapter III) and hybrid intelligence (Chapter IV). Finally, the term guidance **design principles** (DP) then define the abstract DSR knowledge contribution and learning of the design of Section 5.3, 5.4 and 6.5.

For this purpose, I suggest and discuss two directions to overcome those limitations. First, I propose the design paradigm of collective intelligence (e.g. Malone and Bernstein 2015; Wooley et al. 2010) and IT enabled crowdsourcing (e.g. Leimeister et al. 2009) to overcome cognitive and resource constraints of individual human mentoring and allow the integration of stakeholders, which constitute a main source of uncertainty for entrepreneurs. Second, I suggest the design paradigm of hybrid intelligence that can enhance the limited capability of decision support systems based on machine learning (e.g. Jordan and Mitchell 2015; Goodfellow et al. 2016; LeCun et al. 2015) and leverages the complementary capabilities of humans and machines in making both intuitive and analytical decisions under uncertainty.

As the context of entrepreneurial decision-making is a highly idiosyncratic class of problem, I focus the first part of my thesis on the decision-making context itself and examine how both uncertainty (e.g. Section 4.1) and risk (e.g. Section 4.3) are created as well as the general logic and design of systems that provide decisional guidance (e.g. Section 5.3 and 6.5).

## 1.2. Research Questions

This thesis aims at answering three distinctive RQ related to providing decisional guidance for entrepreneurial decision-making. The general purpose of this dissertation is, therefore, to first examine the decision-making context and then provide design paradigms and design principles for the problem domain.

RQ 1 aims at exploring the sources of risk and uncertainty in the entrepreneurial decision-making context by investigating the role of the ecosystem (i.e. involved stakeholders) in creating such. The general goal of this RQ is to provide a better understanding of the decision-making context in general as well as an in-depth examination of the ecosystem as source of risk and uncertainty. This examination of the problem is required to develop suitable solutions that aid entrepreneurial decision-makers.

**RQ 1:** *What are the sources of risk and uncertainty in the entrepreneurial decision-making context?*

**Method:** Case study research and FsQCA.

**Results:** Exploration of ecosystem dynamics as source of uncertainty in entrepreneurial actions; examination of the negative effects of uncertainty and dependence on innovation success; investigation of the mechanism of uncertainty and analysis the mechanisms of both uncertainty and stakeholders in the ecosystem in generating risks for entrepreneurs.

Based on the findings from RQ 1, I identified the integration of the ecosystem as generic valuable strategy to manage risk and uncertainty.

**RQ 2:** *How to design for the integration of the ecosystem as guidance in entrepreneurial decision-making?*

Following this logic, RQ 2 investigates the design for the integration of the ecosystem as guidance in entrepreneurial decision-making and consists of two parts: First, I conceptually develop a design paradigm for the integration of the ecosystem as guidance in entrepreneurial decision-making.

**RQ 2a:** *What are design paradigms for the integration of the ecosystem as guidance in entrepreneurial decision-making?*

**Method:** Interdisciplinary literature review and conceptual development.

**Results:** Crowdsourcing to access collective intelligence as design paradigm for decisional guidance; identification of requirements to adapt crowdsourcing for providing guidance in entrepreneurial decision-making.

Second, it is necessary to develop design principles for the integration of the ecosystem as guidance in entrepreneurial decision-making to build DSSs.

**RQ 2b:** *What are design principles for the integration of the ecosystem as guidance in entrepreneurial decision-making?*

**Method:** Design science research projects and conceptual development.

**Results:** Developing conceptual design principles for a CBMV system for in entrepreneurial decision-making; development of mechanisms for providing feedback and expert matching to apply crowdsourcing for decisional guidance in entrepreneurial decision-making.

Based on the design paradigm and design principles identified in RQ2, the aim of RQ3 is to create knowledge on the design of DSS for providing guidance under uncertainty and risk in entrepreneurial decision-making.

**RQ 3:** *How to design DSS for providing guidance under uncertainty and risk in entrepreneurial decision-making?*

RQ 3 again consist of two related parts. The first part RQ 3a extends the findings beyond the scope of ecosystem integration through crowdsourcing and has the purpose of developing more generalizable and superior design paradigms for providing guidance under uncertainty and risk in entrepreneurial decision-making.

**RQ 3a:** What are design paradigms for providing guidance under uncertainty and risk in entrepreneurial decision-making?

**Method:** Interdisciplinary literature review and taxonomy development.

**Results:** Hybrid intelligence as superior design paradigm for decisional guidance to deal with uncertainty and risk; identification of design knowledge for providing guidance in entrepreneurial decision-making.

The second part RQ3b then uses this design paradigm of hybrid intelligence to propose design principles for providing guidance under uncertainty and risk in entrepreneurial decision-making.

**RQ 3b:** What are design principles for providing guidance under uncertainty and risk in entrepreneurial decision-making?

**Method:** Design science research projects.

**Results:** Developing a data ontology and examination of successful decision patterns for entrepreneurial decision-making; development of design principles for a HI-DSS for decisional guidance in entrepreneurial decision-making.

## 1.3. Structure of the Dissertation

The holistic logic of my dissertation is structured along the RQs and its intended contribution: the examination of the problem context (i.e. entrepreneurial decision-making) and the proposed solution (decision support systems and decisional guidance).

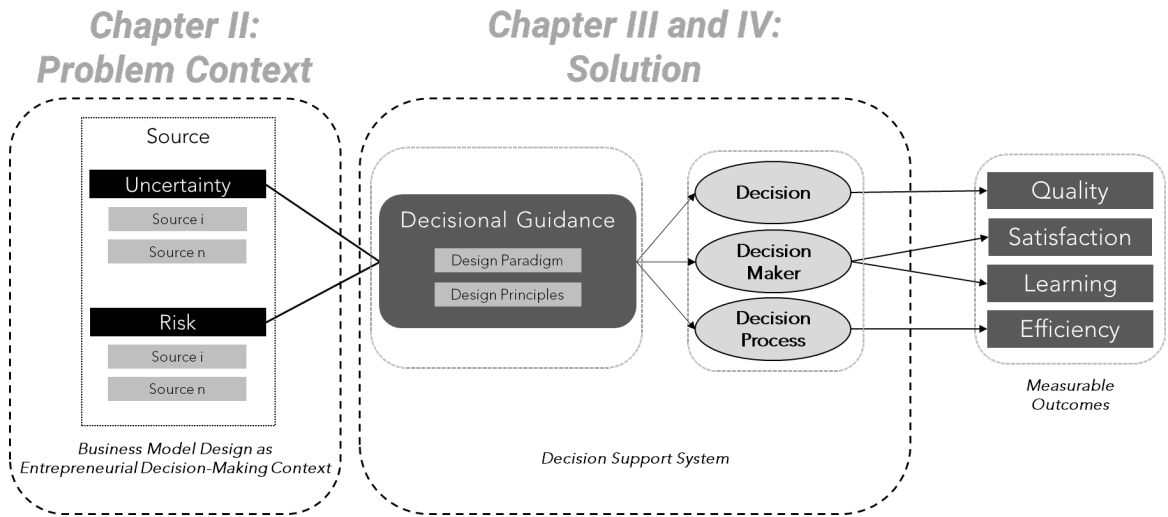

**Holistic Logic of this Thesis**

Chapter II of this dissertation focuses on the (entrepreneurial) decision-making context. Chapter III first explores collective and crowdsourcing as design paradigm for decisional guidance and then develop design principles for decisional guidance that follow this paradigm. Chapter IV then further develops hybrid intelligence as superior design paradigm for decisional guidance and concluding with design principles for DSS for the entrepreneurial decision-making context.

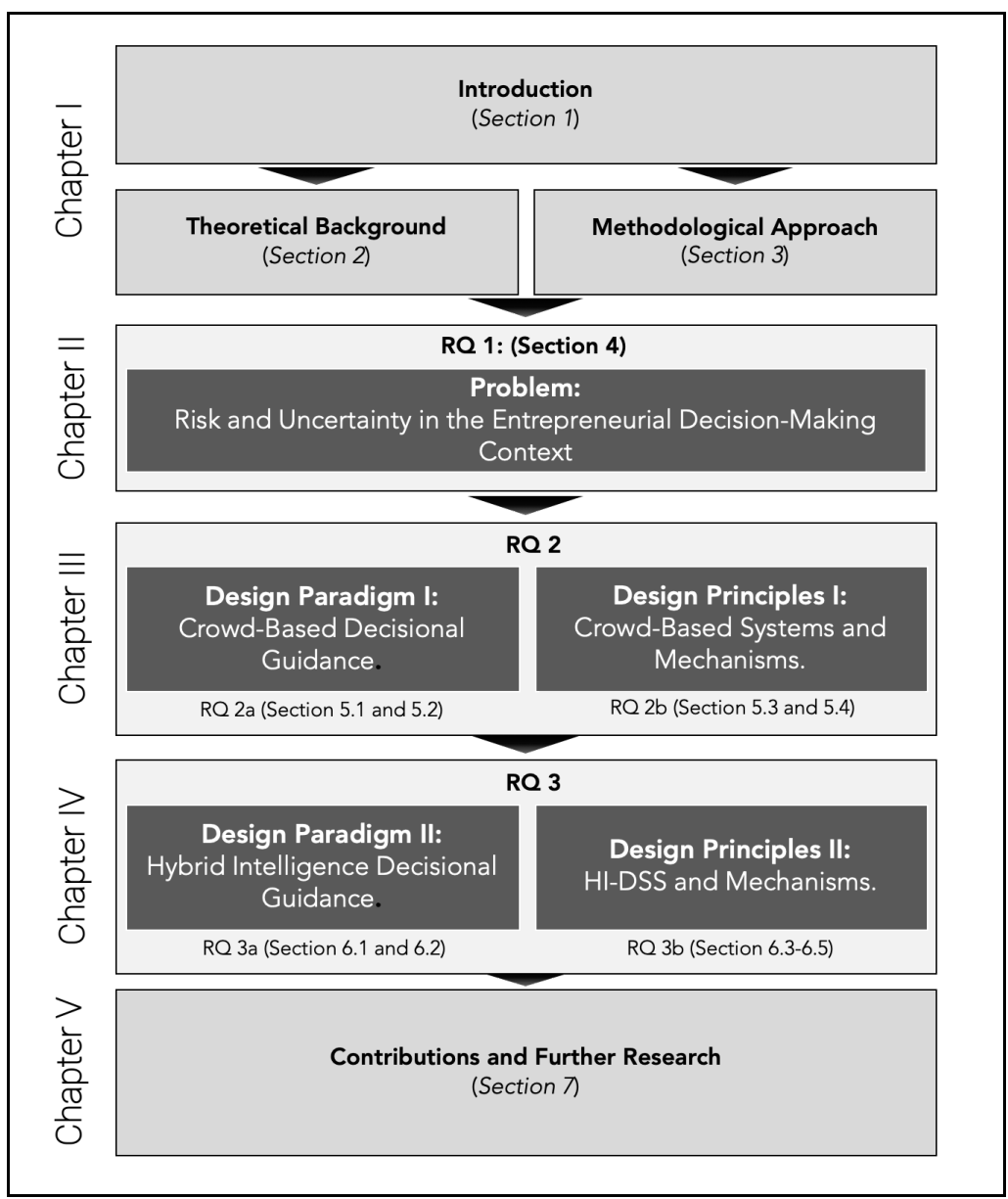

*Structure of the Thesis*

My thesis starts with an Epilogue in **Chapter I** by reviewing the theoretical and conceptual background of this work in *Section 2*. I start in Section 2.1 by reviewing the existing body of knowledge on decisional guidance and DSS, concluding with a detailed explanation of how the following chapters use those concepts. In Section 2.2, I outline the context of entrepreneurial decision-making, its challenges, and strategies how entrepreneurs deal with uncertainty and risk. Finally, Section 2.3 explain business model design as core of entrepreneurial decision-making and its role as research context when investigating entrepreneurial actions. *Section 3* then provides an overview of the applied methodological procedures and its rational. I highlight all various research approaches that were applied in the individual studies.

In **Chapter II** of this dissertation, I investigate the decision- making context, by exploring the ecosystem of an entrepreneur as source of uncertainty and risk as well its effect on entrepreneurial success. I conclude with the integration of the ecosystem as valuable strategy for decision-making under risk and uncertainty.

**Chapter III** then proposes crowdsourcing as a mechanism to access the collective intelligence of the ecosystem. I suggest this as first design paradigm for decisional guidance in entrepreneurship and conceptually derive requirements of crowdsourcing for this context. The second part of this Chapter (Section 5.3 and 5.4) develops DP for decisional guidance in entrepreneurial decision-making.

In **Chapter IV** of this thesis, I build on those findings and suggest hybrid intelligence as superior design paradigm for decisional guidance in this context. This is followed by the development of DP for a hybrid intelligence method to provide guidance under uncertainty and risk and a HI-DSS for supporting entrepreneurial decision-making.

The dissertation concludes in **Chapter V** with the summary of my contributions from both a theoretical and practical perspective, as well as outlining directions of future research avenues for interdisciplinary research related to the topic of this thesis.

# 2. Theoretical Background

## 2.1. Entrepreneurial Decision-Making

### 2.1.1. Risk and Uncertainty in Entrepreneurial Decision-Making

The context of entrepreneurial decision-making describes a specific class of managerial decision-making problem. It is inherently complex as it is uncertain in a Knightian definition (Knight 1921).

More recent research has framed such situations of extreme uncertainty as unknowable risks or unknown-unknowns. Those scholars divide between risk with quantifiable probabilities; uncertainty, which describes risks that are known but cannot be quantified; and the most complex form of unknowable risks or unknown-unknowns where neither outcomes nor their probability distribution can be assessed a priori (Diebold et al. 2010). The latter type of unknowable risk is the dominant form of uncertainty in early stage tech start-ups although all forms exist (Dellermann et al. 2017d). For the purpose of this thesis, I rely on this form of unknown-unknowns when referring to uncertainty.

This facet of entrepreneurial decision-making can be explained as entrepreneurs plan their actions on markets that do not even exist yet or developing novel value propositions which technological feasibility is still unknown. Following this argumentation, the data that would be needed to estimate the probability distributions of certain outcomes or to make assumptions about outcomes does not yet exist (Alvarez and Barney 2007).

This means that even if an entrepreneur would have unlimited cognitive capacity and resources to collect data, she would be unable to correctly quantify the risk (which is the quantified form of uncertainty) associated with certain actions such as the design of a business model (Burke and Miller 1999). Consequently, decision makers are confronted with situations of *"unknown-unknowns"* (Diebold et al. 2010), *"[...] that include both uncertainty and noise due to a large amount of unsystematic risk and conditions of evolving certainty around systematic risk [...]"* (Huang and Pearce 2015): 636).

Making decisions in such context is highly complex for several reasons. First, not all outcomes of a decision cannot be assessed a priori (Huang and Pearce 2015). Second, even if this was the case it would remain impossible to estimate a probability distribution for such outcomes (Knight 1921). Third, as entrepreneurial decisions and the related outcome highly depend on the ecosystem in which entrepreneurs operates, the decision context is extremely dynamic and dependent on complex interactions (Alvarez et al. 2015). Fourth, entrepreneurial decision-making problems are ill-structured, as not one *"correct"* solution exists (Simon 1991). Finally, the feedback on weather a decision was good or bad is time-delayed, requiring years to uncover (Alvarez et al. 2013).

Following this argumentation, I define entrepreneurial decision-making as complex decision-making task that requires to deal with both, uncertainty (unknown-unknowns) and risk.

## 2.1.2. Entrepreneurial Decision Strategies

Dealing with such complex decision-making tasks is particularly difficult, as decision makers are not perfectly rational, but bounded rational (Cyert and March 1963; Newell and Simon 1972; Simon 1955). Such bound rationality typically has two dimensions that result in human deviations from optimal action: cognitive bounds and cognitive biases. The first dimension, covers limitations such as basic computational constraints of the human brain such as working memory, information processing etc. The second dimension is related to idiosyncratic human errors that lead to systematic deviations from rationality in judgment and choice (Kahneman 2011). This bound rationality prevents decision makers from optimizing their actions and is the most basic rational for the need of decisional guidance in general (e.g. Silver 1991). Nevertheless, human decision makers use various strategies to solve such problems.

To understand how individual entrepreneurs, deal with such contexts and make decisions, one must zoom into the individual cognitive strategies of decision-making under uncertainty and risk (Tversky and Kahneman 1983; Dane and Pratt 2007). For this study, individual cognitive properties entrepreneurs (Mitchell et al. 2002) will not be integrated in this discussion as this is beyond the scope of this thesis. Rather I will focus on the generic cognitive processes that are applied for making decisions under extreme uncertainty.

The most dominant streams of cognitive psychology assumes that individual decision-making is influenced by two different systems of decision processing (Glöckner and Witteman 2010; Evans 2008). The first mode of reasoning is rather unconscious, rapid, and holistic, more popular under the term of *"system 1"* thinking. The second type is conscious, slow, and deliberative better known as *"system 2"* thinking (Kahneman and Frederick 2002; Stanovich 1999). The first mode of thinking is also frequently termed as intuition, which describes a *"non-rational"* and *"non-logical"* mode of thinking based on simple heuristics, and mental shortcuts (Epstein 1994; Kahneman and Tversky 1982). The second mode of thinking can be defined as analytical reasoning, which should follow strict rules of probabilistic statistics (Griffiths et al. 2010).

There is a long-standing discourse on which mode of thinking is superior. For instance, intuition is frequently associated with inaccurate or suboptimal choices (Kahneman and Egan 2011; Bazerman and Moore 2008). In contrast, other scholars argue that intuition is often superior as analytical reasoning is limited by working memory, which is especially relevant when decision complexity increases (Gigerenzer 2007).

For the context of entrepreneurial decision-making, previous research argues that the most valuable approach is a combination of analysing and quantifying all available data on the one hand and dealing with unknown-unknowns through intuitive decision-making at the same time (Huang 2017; Huang and Pearce 2015).

Decision makers in the context of entrepreneurship, such as angel investors rely on *"algorithm-based"* factors to integrate objective and quantifiable information such as financial statements, risk analysis, return on investment calculation, market information, and other forms of *"hard"* data (Zacharakis and Meyer 2000; MacMillan et al. 1987).

This strategy is typically complemented with a subjective and affective judgement of an entrepreneurial opportunity that is based on intuition and prior experience (Hisrich and Jankowicz 1990). The integration of soft and cognitive factors such as human intuition is a valuable strategy for making decisions under extreme uncertainty (Huang and Pearce 2015).

Consequently, on the individual level of entrepreneurial decision makers a combination of intuitive and analytical reasoning is most valuable for making decisions under extreme uncertainty (Huang and Pearce 2015; Huang 2016).

### 2.1.3. Guidance in Entrepreneurial Decision-Making

To address both modes of reasoning and making assumptions about certain actions, entrepreneurs must collect empirical evidence. Using decisional guidance in this vein can support decision makers in situations that consist of both uncertainty and risk (e.g. Silver 1990).

For making analytically supported decisions this means gathering information such as financial data, or market reports (Maxwell et al. 2011; MacMillan et al. 1987). Statistical models that use large amount of data as input are, thus, capable of predicting parts of the outcome and value of certain decisions. Such *"[..] actuarial (statistical) models refer to the use of any formal quantitative techniques or formulas, such as regression analysis, for . . . [supporting] clinical tasks [...]"* (Elstein and Bordage 1988). Therefore, they proved to be a valuable form of decisional guidance in the context of early stage ventures (Zacharakis and Meyer 1998). The use of actuarial models as an analytic for of decisional guidance is valuable as its guidance is consistent, not biased by a non-random sample of prior experience and its *"optimal"* information factors ( (Fischhoff et al. 1977; Fischhoff 1988; Slovic 1972). Therefore, I focus on ways to integrate such form of decisional guidance in entrepreneurial decision-making through the mechanisms of AI and ML in Chapter IV.

Additionally, for dealing with situations of uncertainty the interaction with an entrepreneur's external environment (ecosystem) proved to be the most valuable strategy for decisional guidance (Alvarez et al. 2013; Alvarez and Barney 2007).Therefore, I identify the form of guidance that emerges from social interaction with the ecosystem as a proven complementary strategy to improve decision-making through analytical decisional guidance.

This form of dealing with uncertainty are gathering feedback from peers, family members, or friends or validating one's idea by consultants and mentors (Tocher et al. 2015).  Thereby, entrepreneurs test their assumptions against their ecosystem to receive feedback on the viability of their actions. This allows entrepreneurs to cognitively objectify their idea in situations of unknown-unknowns (Alvarez and Barney 2010; Ojala 2016) and persuade a reasonable number of stakeholders of the viability of the opportunity to gain access to further valuable resources that support the entrepreneur in enacting the opportunity (Alvarez et al. 2013). Therefore, I focus on ways to integrate such form of decisional guidance in entrepreneurial decision-making through the mechanisms of crowdsourcing in Chapter III.

## 2.2. Business Model Design as Core of Entrepreneurial Actions

### 2.2.1. The Business Model Concept

For this thesis, I used the business model is as core object when studying entrepreneurial actions and decision-making. Therefore, I will start by defining this term and provide an understanding of the interpretations of the concept that are used for this thesis.

Although lots of different definitions regarding the concept of a business model exist, it provides a holistic framework for the economic model of a firm (Morris et al. 2005; Zott et al. 2011). In general, this model is focused on how value is created and capture (Gassmann et al. 2014). Thus, the business model describes the logic *"[…] by which the enterprise delivers value to customers, entices customers to pay for value, and converts those payments to profit […]"* (Teece 2010:172). The business model can, thus, be characterized as organizational design choices that define the *"[…] an architecture for product, service and information flows, including a description of the various business actors and their roles […]"* (Timmers 1998) and examines *"[…] the content, structure, and governance of transactions designed so as to create value through the exploitation of business opportunities[…]"* (Amit and Zott 2001): 511).

Therefore, the business model is *"[…] a statement of how a firm will make money and sustain its profit stream over time […]"* (Stewart and Zhao 2000). Thereby, it is arranging the operational logic such us internal processes of a firm and its strategy (Casadesus-Masanell and Ricart 2010) and requires decisions on service delivery methods, administrative processes, resource flows, knowledge management, and logistical streams (Afuah 2014).

First, the business model can therefore be used for classifying certain types of firms (Zott et al. 2011; Magretta 2002), which allows to classify new ventures and define similarity among them. This application of the concept is relevant for the expertise requirements and matching of this thesis (Section 5.3; Section 5.4).

Second, the configuration of design choices can be used as antecedent of heterogeneity in firm performance. Therefore, we use the business model to examine its design choices as an important factor contributing to firm performance (Zott et al. 2011). This application of the business model is relevant for this thesis in Section 6.1, where I examine the effect of design choices in defining entrepreneurial success and in Section 6.5, where I use ML techniques for providing guidance on design choices that lead to start-up success.

### 2.2.2. The Business Model as Core of Entrepreneurial Actions

The business model is core of entrepreneurial actions and related decision-making (Demil et al. 2015). Previous work in entrepreneurship heavily focused on how entrepreneurs create novel opportunities to create value (Shane and Venkataraman 2000). The business model is, thus, applied to provide an explanation and structuring framework for examining entrepreneurial actions by adding *"[…] a more holistic, fit-based view of strategic management […]"* (Priem et al. 2013). Therefore, it explicitly focuses on the role of users and the ecosystem in explaining entrepreneurial actions by discussing the value proposition (e.g. Chesbrough and Rosenbloom, 2002) or by including the firm's ecosystem in the process of creating and capturing value from an entrepreneurial opportunity (Amit and Zott 2001; Zott et al. 2011; Zott and Huy 2007; Plé et al. 2010).

Moreover, the business model concept provides a perspective on the relevance and role of implementation when entrepreneurs try to< benefit from an opportunity (Demil et al. 2015). Consequently, the business model can be used to as a kind of action plan for entrepreneurs. The design of a business model is, thus, one of the most pivotal tasks when entrepreneurs aim at capitalizing from an opportunity (Ojala 2016). Therefore, I define the highly uncertain process of iteratively making design choices, testing it against the market and other stakeholder, and reassess the proposed design as core of entrepreneurial action in early stage ventures. Following this logic, I use the design of a business model as phenomenon of interest for examining entrepreneurial decisions and suggesting guidance mechanisms to support such decisions.

### 2.2.3. The Business Model Interpretations for this Thesis

Previous research gives a wide array of different interpretations of the concept of business model (Massa et al. 2017). I will therefore provide a discussion on how the concept is used for this thesis.

First, the business model concept defines attributes of a real firm (Casadesus-Masanell and Ricart 2010; Casadesus-Masanell and Zhu 2010; Markides 2013). This interpretation leverages the business model concept as schema for classifying real-world manifestations of ventures and allows the identification of business model archetypes (Johnson 2010; McGrath 2010; Rappa 2001; Gassmann et al. 2014). For this thesis, this interpretation has a dual role. On the one hand, it is used for connecting concrete design choices to firm performance (e.g. Section 6.1). On the other hand, I apply this interpretation for providing decisional guidance on real-world manifestations of a start-up (e.g. Section 5.3; Section 6.5).

Second, the business model is interpreted as cognitive schema (Magretta 2002; Martins et al. 2015; Chesbrough and Rosenbloom 2002). Previous research argues that entrepreneurial decision makers have an image or a mental model of the firm, not the firm itself (Eggers and Kaplan 2009; Eggers and Kaplan 2013; March and Simon 1958). Consequently, Martins et al. (2015: 105) conceptualize business models as *"[...] cognitive structures that consists of concepts and relations* *among them that organize managerial understanding about the design of activities and exchanges that reflect the critical interdependencies and value-creation relations in their firms' exchange networks [..]".* For this thesis, I use the cognitive schema interpretation of business models to communicate an entrepreneurs mental model of a start-up to its ecosystem. The business model is for instance used to communicate the mechanisms of value creation and value capture to the crowd (e.g. Section 5.3; Section 6.5).

Finally, the business model has an important role as formal conceptual representation (Osterwalder 2004; Osterwalder and Pigneur 2010). This interpretation connects both the attributes of a firm and the cognitive schema interpretation and highlights the role of the concept in providing a simplified representation of reality (Massa et al. 2017). Thus, it defines an explicit formalization of the firm, written down in pictorial, mathematical, or symbolic form. In the context of this thesis, this interpretation is applied to use ML techniques for examining business model design choices and bringing a human mental model in data representation for the ML part of providing decisional guidance (e.g. Section 6.4; Section 6.5). The use of such formal problem representations that allow to structure knowledge comparable as used in the human mind (Ha and Schmidhuber 2018; Stuhlmüller 2015) is especially relevant for solving AI-complete problems and create a shared understanding between humans and machines (Evans et al. 2018).

## 2.3. Decision Support Systems and Guidance

### 2.3.1. Decision Support Systems

Decision support systems (DSS) have a long tradition in IS research and is one of the most pivotal systems that were explored in this field (Todd and Benbasat 1999; Gregor and Benbasat 1999; Alter 2013; Alter 1980; Benbasat and Schroeder 1977).

DSS are a special type of IS that are focused on supporting and improving managerial decision-making (Arnott 2004). Such systems use decision rules, decision models, and knowledge bases to support managerial decision makers in solving semi- and unstructured problems ((McCosh and Morton 1978). Therefore, DSS design an environment in which human decision makers and IT-based systems interactively collaborate. This is especially relevant for providing cognitive aids that assist managers in complex tasks that still require human judgement (Keen 1980). In this collaborative problem solving, human focus on the unstructured part of the problem, while the IT artefact provides an automatic structuring of the decision context (Arnott and Pervan 2005).

More recently, IS research has focused on the application of AI for the purpose of DSS, thus, starting a sub-domain of intelligent DSS (Remus and Kottemann 1986; Bidgoli 1998). These intelligent DSS are for instance rule-based expert systems and more previously ML supported systems that apply for instance ANN, genetic programming and fuzzy logic (Turban et al. 2005). Contrary to the general application of AI in automating tasks and replacing human judgment, DSS aims to supporting the human decision-maker rather than replace her (Arnott and Pervan 2005).

While research on intelligent DSS is a steadily evolving field, knowledge on DSS that are capable to solve highly unstructured and complex problems (i.e. wicked problems) is still nascent (Meyer et al. 2014).

Therefore, there is a clear gap in previous work in solving tasks such as providing guidance to entrepreneurial decision makers.

## 2.3.2. Decisional Guidance as Design of DSS

In general, decisional guidance is a concept that describes any aids or advice that tells a human decision-maker what to do (Morana et al. 2017). This is not limited to technological aids, but also other forms of advice such as mentoring etc. For instance, in the context of entrepreneurial decision-making, so far, guidance is provided as face-to-face mentoring in institutions such as business incubators (Dellermann et al. 2018b).

In the context of IS research, decisional guidance Information describes design features of a DSS that provides such advice to the user (Silver 1991, 1990; Arnold et al. 2006). (Silver 2006) defines decisional guidance as *"[...] the design features of an interactive computer-based system that have, or are intended to have, the effect of enlightening, swaying or directing its users as those users exercise the discretion the system grants them to choose among and use its functional capabilities [...]"*. Such advice (e.g. explanations or suggestions) then helps users to achieve a certain goal ( (Benbasat and Wang 2005; Wang and Benbasat 2007)Gregor and Benbasat 1999) and allow to *"[...] integrate the expertise of one or more experts in a given decision domain [...]"* (Arnold et al. 2006:2). Decisional guidance can be described as both a *"[...] decision aid as technological intervention [that] should assist in the implementation of normative decision-making strategies; or [a] decision aid as a behavioural approach with the aim of extending the capabilities and overcoming the limitations of decision-makers [...]"* (Todd and Benbasat 1999:11). Consequently, decisional guidance provides recommendations for solving problems or supports the user in making decisions (Silver 1991).

Decisional guidance can be characterized along ten distinctive dimensions (Morana et al. 2017). Each of the dimensions relevant for this thesis will be discussed in the related Section (e.g. Section 5.4 and Section 6.5).

| Dimension | Definition | Characteristic | References |
|---|---|---|---|
| **Target** | Target of guidance describes what distinct activity is enlightened. | (1) choosing which activity to perform; (2) making choices when engaging in a given activity. | Silver (1991); Silver (2006) |
| **Directivity** | Directivity of guidance describes what form of guidance is and how it aims to influence the users' activity. | (1) suggestive guidance, which makes judgmental recommendations; (2) informative guidance, which provides pertinent information that enlightens the users' decision; (3) quasi-suggestive guidance, which does not explicitly provide recommendations but from which one can directly infer recommendations. | Silver (1991); Silver (2006) |
| **Mode** | Mode of guidance describes how the guidance works. | (1) predefined mode, meaning the system designer prepares the provided guidance; (2) dynamic mode, meaning an adaptive mechanism *"learns"* as the system is used; (3) participative mode, in which users participate in determining the guidance they receive. | Silver (1991); Silver (2006) |
| **Invocation** | Invocation of guidance describes how the guidance is accessed. | (1) automatically by the system based on redefined usage events;<br>(2) user-invoked after the users' request;<br>(3) intelligently adapting the guidance based on usage context. | Gregor and Benbasat (1999); Silver (1991); Silver (2006) |
| **Timing** | Timing of guidance describes when the guidance will be provided to the user. | (1) concurrently, during the actual activity;<br>(2) prospectively, before the actual activity;<br>(3) retrospectively, after the actual activity. | Silver (2006) |
| **Format** | Format of the guidance describes how the provided guidance is formatted. | (1) text, when using primarily written words;<br>(2) images, when using pictures and depictions;<br>(3) animation, when using videos and moving pictures;<br>(4) audio, when using speech and verbal instructions. | Gregor and Benbasat (1999) |
| **Intention** | Intention of guidance describes the context for why guidance is provided. | (1) clarification, used to illuminate a perceived anomaly;<br>(2) knowledge, used to provide additional information;<br>(3) learning, used to support learning and training;<br>(4) recommending, used to suggest a certain decision or activity. | Arnold et al. (2004a); Arnold et al. (2004b); Gönül et al. (2006); Parikh et al. (2001) |

| **Content** | Content type of guidance describes the purpose of the guidance provision. | (1) trace, when providing the line of reasoning; <br> (2) justification, when outlining the reasoning with an additional line of argumentation; <br> (3) control, when providing evidence for a successful strategy; <br> (4) terminological, when providing expert knowledge on concepts of a certain domain. | Gregor and Benbasat (1999) |
|---|---|---|---|
| **Audience** | Audience of guidance describes which types of users are addressed by the guidance. | (1) novices, users with no or only limited knowledge and expertise of the domain of interest; <br> (2) experts, users with a (high) amount of knowledge and expertise of the domain of interest. | Gregor and Benbasat (1999) |
| **Trust-Building** | Trust building describes whether the guidance affects the user's confidence in it. | (1) passive, when the guidance is not deliberately affecting the trust of the user in it; <br> (2) proactive, when the guidance is purposefully affecting the trust of the user in it. | Wang and Benbasat (2005); Wang and Benbasat (2007) |

*Taxonomy of Decisional Guidance*

When provided to human decision makers, decisional guidance can influence certain aspects of the context, thereby leading to measurable outcomes (Parikh et al. 2001). First, it influences the **decision** itself, thus, increasing the *quality* of a decision (Meth et al. 2015) . Second, it can affect the **decision maker**, thereby, increasing user *satisfaction* and *learning* (Gönül et al. 2006). Finally, decisional guidance can also have an impact on the **decision-making process**. Consequently, it alters the *efficiency* of the process itself (e.g. Parikh et al. 2001).

### 2.3.3. Decisional Guidance in the Context of this Thesis

This thesis uses **decisional guidance** as central phenomenon of interest. For the purpose of this thesis, I use the aggregated definition of Morana et al. (2017: 33), who define decisional guidance as *"[...] the concept of supporting users with their decision-making, problem solving, and task execution during system use by providing suggestions and information [...]"* while *"[g]uidance design features refer to the actual implementation of the guidance concept [...]".*

As the context of entrepreneurial decision-making is a highly idiosyncratic class of problem, I focus my thesis on the decision-making context itself and examine how both uncertainty (e.g. Section 4.1) and risk (e.g. Section 4.3) are created as well as the general logic and design of systems that provide decisional guidance (e.g. Section 5.4 and 6.5).

I use the term **design paradigm** as the general rational for the decisional guidance provided, which is collective intelligence/crowdsourcing (Chapter III) and hybrid intelligence (Chapter IV). Finally, the term guidance **design principles** (DP) then define the abstract DSR knowledge contribution and learning of the design of Section 5.4 and 6.5.

The effects of decisional guidance on measurable outcomes, however, is beyond of the scope of this thesis. I made these decisions for two distinctive reasons, which I will discuss at the end of the thesis in more detail. First, a lot of previous work examined the effect of decisional guidance in various contexts (e.g. Arnold et al. 2004; Parikh et al. 2001; (Limayem and DeSanctis 2000); Gönül et al. 2006) and non-IT-based decisional guidance is a common form of supporting entrepreneurs (Dellermann et al. 2017c), which leads me to make the assumption that those effects might be similar in this context. Second, measuring the effects of decisional guidance in entrepreneurial decision-making is extremely complicated as the outcome of such decisions are typically several years time-delayed (Maxwell et al. 2011).

# 3. Methodological Paradigms

Within this chapter, I outline the applied methodological approaches that were applied for this thesis. I used various data collection and analysis techniques from the four generic methodological approaches literature review, qualitative methods, quantitative research, and design science research (DSR). Each method applied was dependent on the RQ that should be answered as well as its methodological requirements. Therefore, the selection of methods was rather pragmatic than philosophically influenced. The detailed application of research paradigms is explained in each individual section of the manuscript.

## 3.1. Literature Review

The first methodology that I used was the analysis of the existing body of knowledge on related topics. Such literature reviews are an essential approach to structure a specific knowledge domain (Rowe 2014). This approach was especially relevant for the topic of this thesis as it spans interdisciplinary research from the fields of IS, strategic management, entrepreneurship, and computer science.

In general, a literature review can be defined as a systematic, explicit, and reproducible method for identifying, assessing, and synthesizing the current body of knowledge in a certain domain (Vom Brocke et al. 2009).

Most commonly, a literature review is applied by exploring literature through related scientific database queries based on keywords regarding the topic under study. Using a forward search in existing search engines such as Google Scholar allows to identify additional papers that cite papers from the search query. Finally, applying a backward search from the identified papers enables the researcher by reviewing the scientific background on which each paper was built on (Webster and Watson 2002).

This approach has a threefold relevance for this dissertation (Webster and Watson 2002). First, it was used to summarize existing research and serve as basis for developing taxonomies (e.g. Section 5.3 and Section 6.3). Second, the literature review allowed me to conceptually develop the idea of combining crowdsourcing for entrepreneurship (e.g. Section 5.1) as well as conceptualizing hybrid intelligence (e.g. Section 6.1). Thus, I was able to derive theoretical propositions from applying this method. Finally, I used the literature review methodology to identify directions for further research and develop a research agenda on crowdsourcing for leveraging collective intelligence to solve complex entrepreneurial problems (e.g. Section 5.2).

## 3.2. Qualitative Methods

Qualitative are a popular approach of social science to explore complex social interactions between and the behavior of human actors (Recker 2012). Such research approaches are especially valuable when little a-priori knowledge about a certain phenomenon exist. Thus, qualitative research typically has an exploratory character (Creswell 2014). In this vein, the quality of the conducted research highly depends on the analytic and interpretive skills of the individual researchers as she plays a major role in the social context of qualitative studies (Bhattacherjee 2012).

For conducting qualitative research, the most common data collection techniques include interviews with key stakeholders of the phenomenon under study, observations, publicly available documentations, and audiovisual material such as websites, social media data (Creswell 2014). For rigorous transparency and traceability reasons, various data analysis techniques such as coding, identifying and exploring critical incidents or the analysis of content (Myers 1997).

For this thesis, I applied qualitative research methods to explore the evolutionary process of digital business models and entrepreneurial actions (e.g. Section 4.1) as well as identifying the problem that guides my design-oriented research in Sections 5.3 and Section 6.5.

## 3.3. Quantitative Methods

On the other hand, quantitative research methods are typically applied to test hypothesis that were derived from the current body of knowledge. Such explanatory studies are characterized by generalizability of the findings due to representativeness of the study population and the precision of the applied measurement instruments and analysis (McGrath 1981).

The generic linear research process of quantitative studies typically starts by developing theoretical models and deriving hypotheses. In the next step, measurement instruments are developed. Then data is collected and analysed. Finally, the results are evaluated and interpreted in the context of previous work. (Recker 2012).

Within this study, quantitative methods have a threefold role. First, I applied quantitative confirmatory studies to evaluate artefacts that were designed in this research (e.g. Section 5.4). Second, I used quantitative methods to derive explanatory knowledge on mechanisms (e.g. Section 4.2). Finally, I leveraged machine learning (ML) techniques to explore mechanisms on a large scale based on the development of a-priori theoretical models (i.e. a taxonomy) in Section 6.1.

## 3.4. Design Science Research

The third research paradigm of this thesis is design science, which *"[…] creates and evaluates IT artifacts intended to solve identified organizational problems […]"* (Hevner et al. 2004).

The central goal of DSR is to solve a generic class of problems by designing a specific artefacts (e.g. processes or IT) solution and then derive generalizable knowledge from that approach (Gregor and Hevner 2013).

Idiosyncratic for DSR is that such research projects are typically initiated by a relevance cycle of real-world problems. Therefore, requirements of stakeholders are crucial for such studies. On the other hand, a rigor cycle connects those real-world problems back to the scientific body of knowledge that can be used to support the design of an artefact and to discuss finding from the design in the light of previous research. The design process then follows an iterative back and forth between relevance and rigor (Hevner 2007).

Methodological papers on how to conduct DSR suggest various processes to conduct DSR. For this study, I followed the suggestions of Peffers et al (2007) and Vaishnavi and Kuechler (2015).

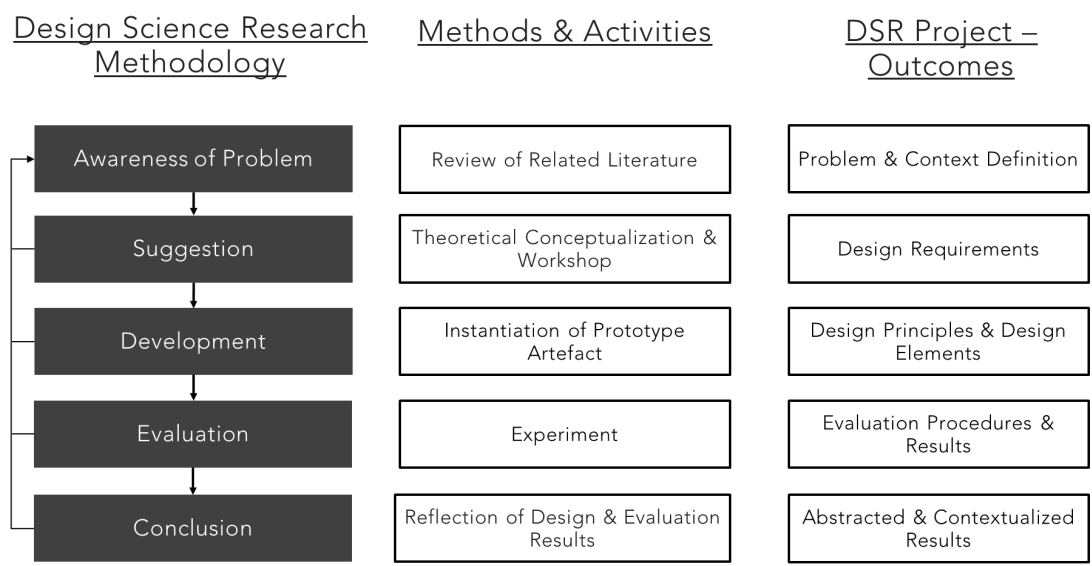

**DSR Process**

The DSR process, that is applied for this dissertation, consists of five iteratively conducted steps that typically are related to certain methods and activities and leading to certain DSR project outcomes.

First, the awareness of the problem step identifies and defines a real-world problem. In the second step, the objectives for the proposed solution are suggested. The third step then focuses on the design and development of the actual artefact. Depending on the framework applied, (Peffers et al. 2007) suggest an additional demonstration step that allows to gather feedback on the usefulness of the artefact, while and (Vaishnavi and Kuechler 2015) directly propose the evaluation of the instantiated solution. This steps covers a holistic evaluation along pre-specified criteria (Sonnenberg and Vom Brocke 2012; Venable et al. 2016).

Depending on the specific RQ and the number of design iterations conducted for developing the artefact in this dissertation, I used both the framework of Peffers et al. (2007) (e.g. Section 6.5) and Vaishnavi and Kuechler (2015) (e.g. Section 5.3).

The design of the artefact then can be considers as contributing knowledge by offering insights in the design problems and its solution (Gregor and Jones 2007). Those knowledge contributions can then be categorized along three levels (Gregor and Hevner 2013).

| | Contribution Types | Example Artifacts |
|---|---|---|
| More abstract, complete, and mature knowledge<br><br>↑ ↑ ↑ ↑<br>↓ ↓ ↓ ↓<br><br>More specific, limited, and less mature knowledge | **Level 3:** Well-developed design theory about embedded phenomena | Design theories (mid-range and grand theories) |
| | **Level 2:** Nascent design theory—knowledge as operational principles/architecture | Constructs, methods, models, design principles, technological rules |
| | **Level 1:** Situated implementation of artifact | Instantiations (software products or implemented processes) |

*DSR Contribution Types*

The knowledge contribution that can be derived from DSR can then be classified the maturity of the solution and its application (Gregor and Hevner 2013).

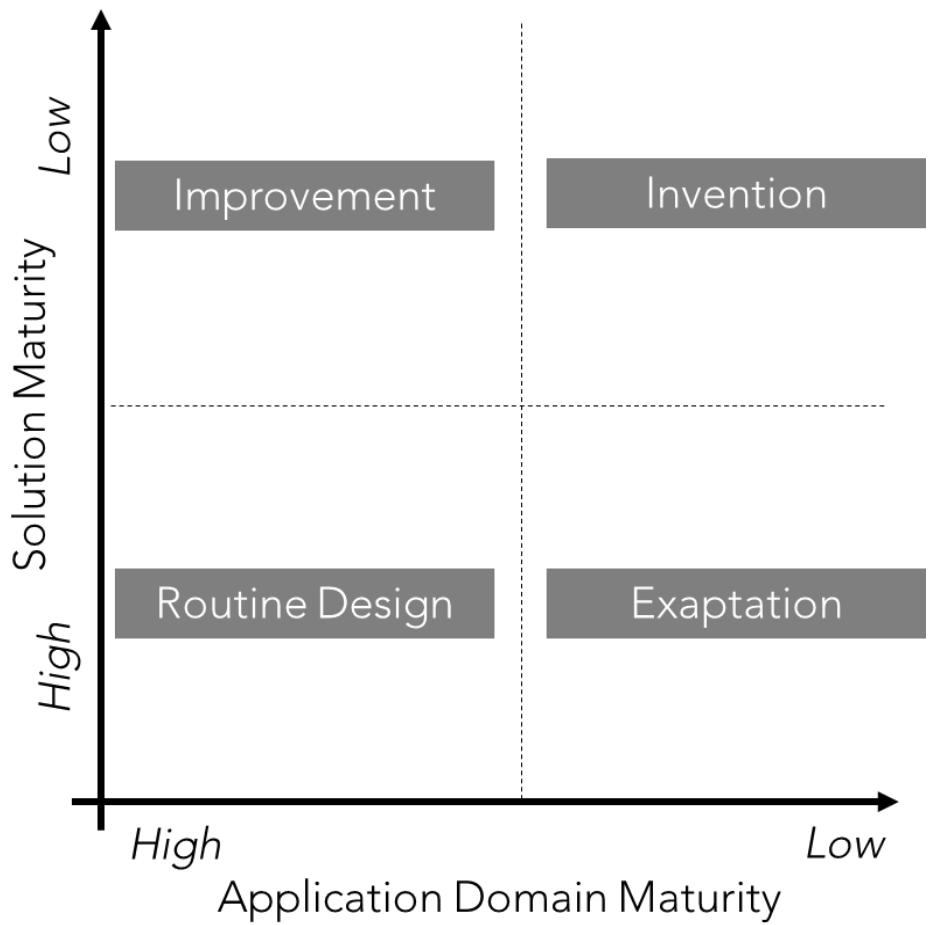

*DSR Knowledge Contribution Framework*

The knowledge contributions of my dissertation can be characterized as both exaptation and invention. The design of mechanisms (e.g. Section 5.4) and the CBMV (Section 5.3) itself in Chapter III of this dissertation constitute an exaptation with a high solution maturity (e.g. ML based filtering or crowd-based decision support) and a low application domain maturity (i.e. support for entrepreneurial decisions).

On the other hand, the design of the hybrid intelligence method (Section 6.4) and the HI-DSS (Section 6.5) can be characterized as invention as both solution maturity and application domain maturity are relatively low.

# Chapter II
## Risk and Uncertainty in the Decision-Making Context

# 4. Problem: Risk and Uncertainty in the Entrepreneurial Decision-Making Context The Role of the Ecosystem

## Purpose and Findings

The purpose of Chapter II is to examine the context of entrepreneurial decision-making. Within this part of the thesis, I identify the the ecosystem and various stakeholders as source of both risk and uncertainty in entrepreneurial decision-making. Therefore, I start by exploring the effect and interactions of different stakeholders from the ecosystem when entrepreneurs design digital business models. (Section 4.1). This study indicates that the ecosystem and its actors co-evolve and heavily influence each other in digital business model design, thus, leading to highly uncertain outcomes. Following this logic, the ecosystem itself can be identified as source of uncertainty that influences entrepreneurial actions.

Section 4.2 then analysis the mechanisms of both uncertainty and stakeholders in the ecosystem in generating risks for entrepreneurs. In this section, I analysed how the actions of entrepreneurs in developing applications and the behavior and governance of a platform owner influence each other, thus, creating varying levels of risk influence and likelihood in entrepreneurial actions.

Finally, the first Chapter of this thesis closes by identifying mechanisms for managing risks by dealing with uncertainty and the ecosystem in business model design (Section 4.2). This part of the thesis identifies risks resulting from both uncertainty and the actions of stakeholders and then suggests strategies for managing such risks and uncertainty by integration the ecosystem in entrepreneurial actions.

## Relevance for Dissertation

The findings of this chapter first identify two main reasons why entrepreneurial actions fail: uncertainty and risk that arise from the ecosystem. These findings examine the general problem that this dissertation covers. Therefore, the integration of stakeholders and the ecosystem is a valuable strategy to reduce both uncertainty and the risk of entrepreneurial actions and, thus, increasing the probability of success. Those issues can be seen as requirements to guide entrepreneurial actions (i.e. decision-making). Consequently, the need for mechanisms to integrate the ecosystem in entrepreneurial decision-making arises as RQ 2 and set the foundation for Chapter III.

## 4.1. Stakeholders as Source of Uncertainty in Business Model Design

The findings of this chapter were previously presented at the R&D Management Conference (Dellermann and Kolloch 2016) and published as (Kolloch and Dellermann 2017) and examine how the actions of entrepreneurs and the ecosystem influence each other, thereby, increasing the uncertainty of outcomes in entrepreneurial business model design. This study is based on a single case study in the German energy industry.

### 4.1.1. Introduction

The emergence of the Internet-of-Things (IoT), creates a technological network of connectivity with self-configuring capabilities that are enabled by standardized and interoperable formats and connecting heterogeneous digitized objects via the internet (Atmore et al. 2010). Digital technology, therefore, is combining digital and physical components into novel value propositions. Furthermore, ubiquitous computing enables the interconnection of multiple devices (Yoo et al. 2012).

Along with this digitization of technology, the organizing logic of innovation is changing. Schumpeters model of the lone entrepreneur ( (Schumpeter 1942) that brings a certain value proposition to the market must be rethought, as innovation are increasingly created in networks (i.e. ecosystems) of produces, users, complementors and several other institutions that create a social system consisting of multiple and heterogeneous actors (Adner 2006; Moore 1993). The high level of openness in innovation makes firms more dependent on each other as well as dynamics within the firm's environment (Adner and Kapoor 2010; Chesbrough 2007; Chesbrough 2006). Therefore, innovation ecosystems are an ensemble of interdependent and heterogeneous actors (e.g. suppliers, distributors, competitors, customers, government, and other institutions) (Teece 2007) that emerge around an innovation (i.e. a technological network) and are dynamic and steadily evolving (Iansiti and Levine 2004).

Understanding how such ecosystems evolve over time is becoming critically important for many firms. Hence, research on ecosystem evolution gains increasing attention (Henfridsson and Bygstad 2013). Drawing on the metaphor of a biological ecosystem, one suitable way to explain the path-dependent and frequently chaotic dynamics within such a system is Darwin's notion of evolution and co-evolution (Darwin 1859). While evolution describes the change of a system over time on a more holistic level, co-evolution explicitly focuses on the interaction between entities within a system that creates conflict or cooperation and therefore creates dynamics.

One aspect that has not been considered by research on ecosystem dynamics is an integrated view on how such interaction between both, technological and human entities in an ecosystem affect the relationships among them and influence the dynamics of an innovation ecosystem. However, integrating the technological as well as the social perspective is required to gain a deeper understanding of the dynamics of innovation ecosystems.

I therefore argue that Actor Network Theory (ANT) is a suitable theoretical lens (Callon 1987; Latour 2005) for analysing an innovation ecosystem as network of human (e.g. organizations) and non-human (e.g. technological) actors. The dynamics of an ecosystem are defined as a socio-technological process in which various organizations translate and inscribe their interests into a technology, creating an evolving network of human and non-human actors (Henfridsson and Bygstad 2013). Controversies are situations in which formerly fixed opportunities are challenged and contradict the status quo (Latour 2005; Venturini 2010). Such changes in the status quo of a socio-technological system frequently lead to ripple effects, which result in an overall system's evolution. In the sense of ANT this can be positive controversies such as the emergence of novel ideas or technologies or negative like in the sense of conflicts. This argumentation is in line with previous research that highlighted the role of dialectic objectives and conflicts in organization or groups as source of innovative outcomes (Harvey 2014). However, this research was neither focusing on the interorganizational level of ecosystems nor did it examine the crucial role of technology in such settings. Therefore, the concept of controversies in socio-technological actor networks are a suitable mechanism to explain ecosystem dynamics.

For this purpose, I organized the paper as follows. The upcoming sections review present work on the emergence and characteristics of digital ecosystems and my conceptual framework based on ANT. I then argue for virtual power plants (VPPs) as suitable objects for examining digital ecosystems. To investigate the impact of controversies on digital innovation ecosystems, I apply a case study approach examining a project of setting up a VPP within the German energy industry. A discussion of the results derived from the case analysis draws the contribution to the mechanisms of controversies on the evolution of the ecosystem. The contribution and the limitations of the paper are highlighted in the concluding section.

## 4.1.2. Digital Ecosystems and Entrepreneurial Actions

As digital technology is combining digital and physical components into new value propositions, firms can no longer rely on enhancing features and the quality of their products by solely focusing on their individual innovation efforts. Digital disruption in various traditional industries requires the blurring of industry boundaries and converging knowledge bases. Such convergence brings together previously separated user experiences (e.g. adding mobile internet), physical and digital components (e.g. smart products) and previously separated industries (e.g. software and hardware industry) (Yoo et al. 2010).

In general, the properties of digital technology implicate a layered architecture (Adomavicius et al. 2008), which is a specific functional design hierarchy that initiates the modular design of digital innovation (Baldwin and Clark 2006). This allows an effective division of labour among different actors during the design and production of complex systems (Sosa et al. 2004; Staudenmayer et al. 2005). Thus, pervasive digital technology can be seen as an enabler of new market dynamics as well as increased exchange of specialized competences (e.g. knowledge and skills) between heterogeneous actors in complex network structures (Yoo et al. 2010). The modularity of digital innovation is therefore changing the traditional value chain into value networks and fundamentally reshaping the traditional innovation logic (Garud and Karnøe 2003; Sosa et al. 2004). The combinable developmental process of novel digital technology explains how components interact with other components and reshape an ecosystem of human and non-human actors.

The concept of such ecosystem helps to analyse interdependencies more explicitly. Innovation ecosystems are defined as a "[…] loosely interconnected network of companies and other entities that coevolve capabilities around a shared set of technologies, knowledge, or skills, and work cooperatively and competitively to develop new products and services […]." (Nambisan and Baron 2013).

Organizations increasingly participate in ecosystems to capitalize on knowledge outside the boundaries of the single firm (Simard and West 2006). The companies' single innovation efforts therefore reciprocally influence each other making the relationships among the actors of the ecosystem central to its success (Iansiti and Levine 2004). Digital ecosystems are not homogenous constructs but include different actors with different kinds of relations and variable strength of ties among them (Teece 2007). Vice versa, an ecosystem is not a stable construct but a dynamic and steadily evolving entity, which is changed by the relationships between the individual actors and their interdependencies (Ghazawneh and Henfridsson 2013; Selander et al. 2013), changing the direction and strength of ties among them (Basole 2009).

### 4.1.3. Conceptual Framework

## Actor-Network Theory

I argue that the interaction within the innovation ecosystem of a VPP, is strongly affected by human (i.e. organizational) and nonhuman (i.e. technological) actors. Thus, ANT is an appropriate starting point for the intercourse to my research design as it explicitly highlights this interplay (Latour 1990, 2005). Despite being criticized, it is lately used to study innovation especially in the field of information systems (IS), which fits my perspective on the context of digital innovations (Dery et al. 2013; Hanseth and Monteiro 1997). In fact, several authors emphasized the importance of ANT in analysing the interaction between stakeholders, particularly to address the crucial role of technology  (Pouloudi et al. 2004; Vidgen and McMaster 1996)Luoma-aho and Paloviita 2010).

The origin of ANT, which lies within the field of socio-technological systems, implies that *"[…] the study of any desired technology itself can be developed into a sociological tool of analysis […]"* (Callon 1987:83). Thus, the view of technology as a socially constructed system caused by several interactions perfectly fits my understanding (Hughes 1987). Following this logic, the underlying concepts of ANT are inscription and translation (Callon 1987). Engineers inscribe their intentions or imaginations of how it fits best to the desired scope into a developed or designed technical artefact (e.g. software, application). Callon (1987) titles such engineers as *"engineer-sociologists"* since they become sociologists in the way of inscribing their technical vision in the real world (organizational) context. To illustrate this, I give a clear example: Why do drivers trust their navigation systems at least as much as tourist information centres when searching a street? This is due to engineers inscribing navigation systems with specific respect to how drivers reach their way best as by those who once decided to develop a city guide (map). This plausible illustration highlights the central aspect of ANT of treating human and nonhuman actors equally. Throughout an innovation process, especially a digital innovation effort, it becomes increasingly difficult to recognize the frontiers between technical and social influence variables. In case of commonly acknowledged technology as social artefacts (see example above), the technology itself becomes an actor in terms of ANT. This actor then inherits the same characteristics as human ones (Callon 1987; Latour 2005). In fact, this feature of not distinguishing between human and non-human actors is *"condition sine qua non"* for Latour (2005), to test every study's valid claim of applying ANT.

## Conceptualizing Digital Ecosystems

In this paper, I decided to apply ANT for several distinctive reasons. Firstly, ANT can be utilized as a framework for conceptualizing an innovation effort as an emerging network or ecosystem, which is exactly what I am aiming at. Despite having primarily, a social notation since persons are mainly responsible for the success or failure of innovation efforts, non-human actors like software, technology, grids etc. are also crucial in such projects. In this context, non-humans (following Latour 2005) are a series of heterogeneous inanimate actors called *"agents"* and must be extended to the understanding of *"actors".* Alternatively, the actor or agent is someone or something that produces an effect or change (Giddons 1984). Examples for such *"actants"* are for instance technology, software, platforms as well as information content.

Secondly, only traditional actor roles as subject for investigation are considered, which is problematic since they exclusively act at the frontend of innovation. Hence, ANT allows an in-depth understanding of the dynamics and the interaction among all actors influencing the outcome of an innovation and therefore the dynamics of an ecosystem. Ideally, ANT yields a basis for the examination on how actors form alliances, promote their ideas in front of other actors and use technology (artefacts) to work on their respective ties (Lee and Oh 2006). In this context, Callon (1986) establishes the second principle of ANT, the translation. He defines it as the *"[...] methods by which an actor enrols others [...]"* and is typically depicted via the four stages of problematization, intersegments, enrolment, and mobilization. As translation is not always successful, Callon (1986) further conducts that each entity could choose to either accept or refuse the translation, which is of course a significant aspect for the evolution of an innovations' ecosystem. Because it defines, which actors or entities are parts of the final ecosystem, which do have an active/supporting part, and which do not.

In sum, networks are no pre-existing entities solely consisting of pre-defined actors that collaborate to fulfil the project mission. Instead, they account for a volatile property emerging from relationships, which are the essence of the interaction between several actors. ANT is therefore most suitable to analyse relationship of heterogeneous actors in an ecosystem during value creation. Due to my research focusing on interorganizational innovation ecosystems, I define an organization as an entity of humans and thus a human actor. As the architecture of digital innovation is typically following a layered modular logic, the product can be decomposed into loosely coupled components interconnected through standardized interfaces (Schilling 2000). These characteristics make the boundaries of the innovation fluid while the meaning is pre-specified. Following actor-network theorists such as Callon (1986) and Latour (1990), I assume that the digital innovation itself is a network of technological actors (components). Around this, a social network of human actors (i.e. organizations) emerges and coevolves with the technological network in reciprocal manner. Hence, human and non-human actors translate and inscribe their interests into a technology, creating an evolving network of human and technical entities (Hanseth and Monteiro 1997; Aanestad and Jensen 2011). I therefore define such an innovation ecosystem as a social technological system (actor network) consisting of two inseparable parts: a social system (human actor network) and a technological system (non-human actor network).

In the sense of ANT, human actors inscribe their beliefs into a technological artefact. Vice versa, if a human actor uses a technological artefact, thus interacting with it, the affordances of the technological actor frame the initial beliefs of a human (Faraj et al. 2004). Therefore, the interactions between human and non-human actors can take several distinctive forms.

| Type of Interaction | Practical Example |
|---|---|
| **Human/human actor** | Social interaction between two human entities |
| **Non-human/non-human actor** | Server-web browser interaction via TCP/IP |
| **Human/non-human actor** | Use of a technological artifact (software use) |

*Type of Interactions in Digital Ecosystems*

## Controversies as a Source of Uncertainty

Ever since research started dealing with the management of innovation, different aspects were identified that have a crucial impact on the succession of the distinctive efforts. Hereby, a strong focus was naturally set on the framework in which an innovation is urged to act and thus influenced by several aspects that are not considered in the beginning. Such models view creative outcome like innovation as a process of random variation and selective retention (Campbell 1960; Simonton 1999). However, more recent research highlights the role of dialectics as a source and shaper of innovative outcome (Harvey 2014). In this view on the evolution of creative artefacts conflicts and disagreement between actors provides opportunities for diverse viewpoints to be integrated in creative synthesis. In such settings dialectics arise through the social interaction between single actors that have divergent goals but converge their opinions in a creative synthesis (Kolb and Putnam 1992).

Following this assumption of dialectics as a source and shaper of innovation, I relied on the concept of controversies from an ANT perspective to analyses opposing interests of several actors within a digital innovation ecosystem that contribute to the innovation and their effects on the very same (Latour 2005; Venturini 2010). Through an ANT lens', controversies are any aspects that contradict the status quo and thus influence the interaction and relationships between various actors within the innovation ecosystem (Latour 2005). They span a broad range from the perception of the need for reciprocal consideration to the development of a compromise. Venturini (2010) defines controversies as dynamic conflicts that emerge when formerly fixed ideas and things are challenged and discussed. Extending the dialectic approach between social actors by applying ANT and the concept of controversies consequently provides a major benefit. ANT allows to include both human and technological actors as source of divergent viewpoints that can be integrated through synthesis and thus foster innovation and thus applies a socio-technological rather than a solely social perspective on the dialect perspective as driver of ecosystem evolution. I therefore propose that controversies are a suitable mechanism to explain the co-evolutionary interaction between human and non-human actors in an innovation ecosystem and thus reveal the dynamic of the actor network.

### 4.1.4. Methodology

# Research Design

As mentioned above, this paper presents the findings of an exploratory research based on an in-depth case study of a German utility setting up a VPP. The main objective of the project was to implement an innovative solution for the utility, which generates profits, and strengthen the image of the utility in the region alike. Hereby, my research has an exploratory character aiming at a deeper understanding of the underlying controversies of human and non-human actors within digital innovation ecosystems. For this reason, I choose the case study approach that particularly allows to research into little explored topics with the purpose of theory building (Dul and Hak 2007; Eisenhardt 1989). Contrary to other research strategies, the case study methodology is not intended to make predictions about statistical relationships and frequencies (Eisenhardt and Graebner 2007; Yin 2017). Instead, the conclusions drawn from case study results are *"[…] generalizable to theoretical propositions and not to populations or universes […]"* (Yin 2017: 13). In other words, I conducted this case study to gain new and useful insights on digital innovations in the energy industry and pointing to gaps in the existing theory on both the coevolution of digital innovation and ecosystems and beginning to fill them (Siggelkow 2007). In line with this argumentation Lee et al. vote for applying ANT to new (digital) technologies to better understand the phenomena surrounding the technologies as well as to set up ANT as an empirical lens (Lee et al. 2015).

Thus, I choose the present case due to its exemplarity, which enabled me to apply my framework (Yin 2017). In fact, a VPP provides me with a precious opportunity since the innovation on various actors across traditional industry boundaries and therefore highlights the importance of an innovation ecosystem (e.g. suppliers, complementors, national institutions, application interfaces etc.). In addition, the case offers the opportunity not just to identify but gain an in-depth understanding of controversies between the involved actors (human and non-human) alike.

The project kick-off started in May 2012 and finished in February 2015 as a whole. One of the authors had first-hand access since she participated in the project management team of the utility. Despite the involvement, the researcher aggregated the collected information and undertook participant observation to use the case study approach (Missonier and Loufrani-Fedida, 2014). To end up with unbiased results, the researchers were introduced to the project members as *"neutral beholders"* and did not intervene or act in any way. I found several controversies between human and non-human actors that have a crucial impact on the innovation and the ecosystem.

## Data Collection

Regarding the collection of data, the triangulation of different methods is recommended in the literature to increase internal validity and to obtain a comprehensive description of the cases (Eisenhardt 1989; Yin 2017). For an in-depth examination of the process and gaining valuable insights from mixed sources (Yin 2017), I analyzed interviews, press articles and observations. In my study, data was primarily collected especially through semi-structured interviews with key actors (i.e. platform owner, suppliers, customers, marketers, grid operators) who were directly involved in the VPP project. I therefore ensure the acknowledgement of various perspectives. First, I conducted 20 semi-structured interviews to gain access to rich empirical data (Eisenhardt and Graebner 2007). Second, I collected, clustered, and listed 36 press articles and official documents along with internal and private documents (partnership agreements, supplier conditions, legal documents etc.). Third, the experience of observations in 121 days of participation in the project was beneficial.

## Data Analysis

I transcribed all interview and asked the interviewees to review my case write-ups to verify my analysis. In the next step, I followed the analysis principles of Glaser and Strauss (Glaser and Strauss 1967). First, I used open coding to identify single controversies, involved actors, their corresponding links and key events. After that, I applied I relied on principles of Latour (2005) and recommendations by Venturini (2010) to identify and analyze controversies within the innovation ecosystem. Furthermore, I used the underlying logic of the markers of (Missonier and Loufrani-Fedida 2014), which were developed for the special case of innovation in information systems. Slightly differing but nonetheless comparable my markers include the following five dimensions namely **(1)** the subject of the controversy, **(2)** the involved actors (along with their respective interests), **(3)** the synthesis that solves the controversy, **(4)** the effect on the actor network and **(5)** consequential controversies. Hereby, my major goals were not only to identify the controversies but also to examine the type of controversy, the mechanism, and the pathway of evolution of the innovation ecosystem.

## Case Description

The challenge to plan and to implement the German energy transition (GET) causes fundamental changes in the energy industry. However, another even more significant upheaval, which affects the sector, is the emergence of digitalization. Hereby, digitalization offers chances not only to revolutionize the market but also to worsen the transformation of the logic of energy business in general. Industry dynamics explain how actors in a distinctive industry interact through either collaboration or competition with each other. Hereby, the German energy industry is no exception. For German energy suppliers, times of singular electricity selling as solid sources of income are over. Traditionally, the success of companies in this industry was defined by the ownership of big power plants, which become increasingly obsolete. Realizing this trend and including it in one's own innovation strategy is crucial for firms competing in energy markets. Especially, municipal utility companies that rely heavily on conventional (fossil) power plants are facing significant disruption through the increasing capacity additions of renewable energies (RE). One major issue, which is often not considered, is the strong need of RE's for a strong infrastructure, which enables reliable power transmission in times of no wind and sun. Mainly because of the highly fluctuating feed-in times of RE, a stable energy supply and in special a reliable base load has become a greater challenge than in previous years. Additionally, the formerly rigid German energy industry is characterized by an increasing decentralization as thousands of small and local units instead of big centralized plants pre-dominantly generate energy. In line with this development, an increasing number of novel actors entered the market and even the role models for traditional players in the industry shifted. For example, private households with a PV-panel on the roof and storage units evolve from single consumers to prosumers (producers and consumers alike). As a follow, tens of thousands of such small power producers must be managed and their electricity flows must be gathered and orchestrated to feed it into the grid and to the market. VPPs aim at providing a solution; since they connect several decentralized power-generating units (foremost RE such as photovoltaic, wind farms and biogas plants). In general, VPPs must be regarded as an emerging technological trend in the energy industry. A VPP can be defined as a cluster of grid connected micro-power units that is monitored and controlled on an aggregate level by a VPP operator for commercial or technical goals. A VPP is used to participate in trade on energy markets (APX, EEX), which is enables by the technological distribution network management such as providing regulating and reserve power. As stated above a VPP (or VPPs in general) are technological innovation that combines various stakeholders (actors) in an innovation ecosystem. Therefore, the case study uses one technological instantiation of a VPP as setting, while VPPs in general refer to one type of steering distributed power systems. Due to the virtual connection (via tele control boxes and APIś) and creation of a generation mix they can balance and compensate the different decentral knots.

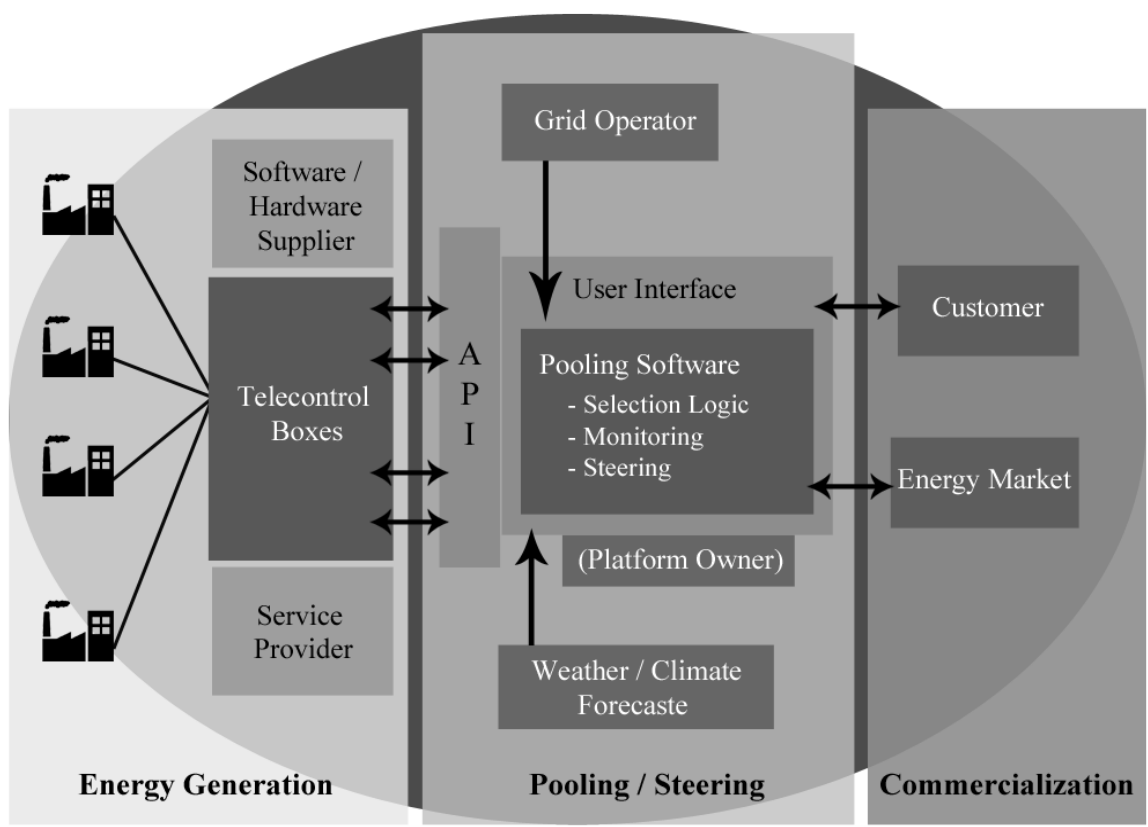

*VPP Architecture*

Since this novel and decentralized way of generating, steering and commercialize energy requires digital connection between all actors via IT, this innovation effort can be marked as a digital innovation. Further, the very central concept of this business model is relying on collaboration or connectedness while the ecosystem defines this dependence on each other. In conclusion, my case is appropriate since it combines the feature of digitization (technology-dependence) as well as a highly dependence on the innovation ecosystem and thus is a perfect setting for my analysis.

## 4.1.5. Findings

The apprehended case can be sub-divided into two controversy loops, which must be regarded as consecutive with respect to the total ecosystem dynamics.

The first loop arose from a mundane problem. The platform owner who was eager to coordinate the single actors of the ecosystem prompt faced several complaints of customers. Caused by crucial delays of the first tele-control boxes supplier the customers were not connected to the pool in the promised period, which led naturally to dissatisfaction and sunken revenues. The problem was caused by understaffed work force of the supplier and delays in the development of the network compatibility. Thus, the platform owner was forced to drain the responsibility of the installation process from hardware supplier to ensure a successful business case. Fortunately, an agreement with a new service provider could be reached who took over the whole hardware delivery and installation part.

Another problem arose when the new hardware boxes were not suitable for synchronization with the software of the platform. While the hardware boxes tend to connect to the grid via an own application interface (API) the software was designed to connect all devices for a centralized control via its own API. This was a system-threatening issue since the focal value proposition of the VPP was under attack. The platform owner was forced to substitute the hardware supplier, which had an impact on the actor network in form of an adjusted component (supplier). While all this actions and coordination took its time, heavy delays of the project were inevitable. The launch date to the grid, which equally marks the start of value generation, postponed several times. A remarkable amount of plant operators opt-out during this process and the remaining actors were forced to strengthen their relationships to avoid the failure of the project. During this process, the platform owner had many controversial discussions with plant operators and customers trying to smooth the moods.

Besides these technical controversies, a second loop could be recognized which had more of a conceptual nature. First, the commercial exploitation and the way to market it was a basis of a conflict. The platform owner who is himself embedded in an organizational network pledged for a collaborative approach for the marketing efforts. In contrast, the first marketer, who is simultaneously the central service unit of the network the platform owner is embedded, was eager to develop a stand-alone solution. After several unfruitful debates over the pros and cons of each approach both opinions were incompatible. Thus, the ecosystem changed again since the software vendor (who was connected to the marketer) as well as the marketer were replaced by the platform owner.

Second, the substitution of the software vendor led to a broad variety of APIs in the system and thus belittle the ease of use for the customers. Customers denied the use of the offered software and the market software itself as well as supplier were again replaced. Third, the connection to the grid was affected by this replacement too. While the grid operator was willing to exploit the heterogeneity of the individual standards (Germany can be divided in four grid zones), the government planned to set up a regulatory standard for the whole of Germany. Thus, all of the four transmission grid operators collaborated to set a German-wide standard protocol for platform operators. Since this was a different standard from the one of the case company, all installed boxes at the plant operators must be reworked while the costs were not refunded.

My findings reveal that throughout the whole process of establishing a VPP several controversies occur and influence the coevolution of the innovation ecosystem as well as the innovation. As the respondents highlight, controversies do not only shape the configuration of the innovation ecosystems evolution; they additionally define the whole composition of the structure as well as the target corridor responsible for the choice of each individual actor. These influenced the ecosystem in a way that resulting controversies consecutively needed an atmosphere of re-adjustments and re-configuration.

During the interviews, it became obvious that two distinctive controversy tracks were pre-dominant within my case. The first one evolved on a technical level since it deals with the appropriateness of components and the choice of suitable suppliers. In contrast, controversies shaped the second track on the personal and technical requirements of actors and their respective distribution within the ecosystem including responsibilities and task definitions.

With respect to the first track, the focal controversy arose between two organizational entities (human actors), the hardware supplier and the platform owner covering the in-ability of the hardware supplier to provide the assured control boxes and installation in the agreed quantity and time. Hereby, a re-adjustment of the constellation of the innovation ecosystem was necessary and the platform owner re-allocated the responsibility for the installation from the supplier towards a service provider. As the Head of the Project stated *"[...] for me, it was absolutely inacceptable to stave off the customers, since I vouched with my reputation to ensure the installation in-time. Due to the delays I was forced to evaluate other alternatives and while I mandated the service provider the relationship with the hardware supplier was strained [...]"*. In the follow, a new actor who was previously not considered entered the ecosystem and the hardware supplier was complemented. For this reason, a change in the technological components of the VPP compared to the first innovation setting was implemented and required an adjustment by the other established actors. This evolution of the ecosystem and the resulting reconfiguration of the VPP's components resulted in a consecutive controversy.

The application interface between two non-human actors, the hardware boxes and software platform were not compatible anymore. Hence, the connection of the decentralized power suppliers to the grid was no longer possible. Consequentially, the platform owner could not provide the proposed value proposition to the customer. The opposing interests between the single components of the VPP, the hardware boxes and the software layer, led to a controversy that affected the need for a reconfiguration of the components of the VPP to ensure the functionality of the innovation. Consequently, the reconfiguration of the constellation of the innovation ecosystem and hence the entrance of a new provider for hardware boxes was required. Since the platform owner was forced to re-design the network, this internal process was delaying the installation deadline promised to the customers. The customers were not contented by the ongoing delays and began to discuss the contractual agreements with the platform owner. This controversy exclusively between human actors (the platform owner and the customers) led to a re-adjustment of the planned costs for the platform owner, losses of already signed customers and losses of orders for the service provider. In contrast, the remaining actors intensified their collaboration, which ensured a better and more amicable understanding. A Sales Manager highlighted, that *"[…] it was quite difficult to keep the customers in line since they were understandably not happy with the delays. They were afraid of financial losses and actively searching for new alternatives and marketers. Thus, handling the inner controversies had a high priority in order to avoid market losses […]".* A Project Manager also added *"[…] I was forced to solve this controversy as soon as possible since it connotes for significant financial losses on my side. Unfortunately, there was no majority opinion on who was responsible for the installation (service provider alone or in collaboration) to fulfil the designated dates. Each party was eager to receive the responsibility for providing this service after the designated actor (hardware box supplier) dropped out since it involves earning additionally revenues […]".* This controversy resulted in an evolution of solely the innovation ecosystem as some customers (human actors) dropped out.

The second track of controversies was more on a generalized level influencing the configuration of the ecosystem. As often in business cases, when several heterogeneous partners are forced to co-ordinate their actions which each other, controversies arose. First to mention are the conflicting interests on a human actor's level. For example, between the platform owner and the platform operator, the conflict inherits the question of how deeply the marketing and accounting activities are merged.

*"[...] It comes as no surprise that every enterprise whether operating in a collaboration or in a single-handed business venture tries to maximize their profits. Unfortunately, the value chain is most often limited and predefined so that an allocation of the profits and revenues is inevitable. For this reason, most of the struggles or controversies as you would put it, arose in the forefront of an innovation project. Each party tries to scavenge the biggest piece of pie and naturally the distribution of tasks which is related to this allocation is heavily embattled [...]"* as the Senior Product Manager stated.

As no compromise, could have been reached, the platform owner replaced both the platform operator and the marketers leading to a change in the constellation of the ecosystem while the innovation remains untouched. In a follow, the new software supplier had difficulties to ensure the compatibility of her software solution to a broad variety of API's while the customers favoured the usability of the software interface. The consequence was that this human vs. non-human controversy led to a rejection of the customers to use the solution. The software supplier was substituted, and a more customer-friendly but less complex and applicable software solution was implemented. A new actor and results in the reconfiguration of the technological components of the VPP enlarged the evolution of the innovation ecosystem. The last issue for itself was a sum of controversial interests for the grid operators whose aim was to synchronize all transmission codes across the German energy market and standardize the connection for all feed-in of VPPs. As the responsible manager of the grid operator puts it *"[...] the main difficulty was the co-ordination between all actors. The four transmission grid operators aimed at standardizing the feed-in options for all power plant operators. On the other side, most VPP's are handled very heterogeneous in terms of software codes, connection points and commercial exploitation. I am obliged to offer every system the feed-in but since there are so many different solutions it can took time. A retrofitting to a standardized system, which would on the other hand accelerate the process, is most often too expensive for the system operators [...]".* The foremost agreements were in consequence not valid anymore and the grid operator was forced to re-code the connection ports, which worsens the delay discussion with the customers for the platform owner. Hence, the changes of the transmission standards of the VPP required an adaption of the technological components and lead to an evolution of the digital innovation, namely the VPP.

## 4.1.6. Discussion

The gained insights enabled me to regard controversies not only as concomitant in setting up the VPP but also rather as a constitutional factor determining the coevolution of the innovation ecosystem as well as the configuration of the digital innovation itself.

### A Typology of Controversies

The layered modular architecture of digital innovation including single components that interact and are provided by heterogeneous firms, lead to the fact that not only human actors (organizations) but also technological actors can create controversies due to antithetical interests. This effect is particularly important for the layered modular architecture of digital innovation (Yoo et al. 2010).

First, controversies can arise due to disagreement of organizations (human/human) about certain topics like contractual agreements, target dimensions, participation, and resource allocation (Type I). Such controversies are for instance interpersonal conflicts. Therefore, ANT explains the process in which actors form alliances and promote their point of views to convince other actors in the ecosystem.

Second, following this logic, even non-human/non-human conflicts are possible as digital technology is combining digital and physical components and enables the interconnection of loosely coupled components through standardized interfaces (Type II). The non-interoperability of technological components is a shared uncertainty between actors as the requirement specification may exceed the range of functions. Therefore, for instance technological protocols for communication might not be interoperable and create controversies between technological artefacts.

Third, especially in digital innovation, non-human actors (technology) can trigger controversies (Type III). As, for instance, the engineers encoded their respective visions about application into a software code, the software itself used by the platform owner becomes an individual actor within the innovation ecosystem with own interests and requirements. Therefore, human/non-human controversies can arise as for instance the usability of technology might then lead to opposing interest with the user. Therefore, uses might resist using a software artefact leading to controversies between human and non-human actors.

In addition, a match between my paper and the study for standardization and ANT can be stated (Lee and Oh 2006; Lee et al. 2015). In line with this argumentation and my findings, standard wars must be considered as a framework heavily affected by technological actors. Standardization which is a significant factor for digitization since it aims at contributing world-wide services and applications in the same *"language"* is often area for alliances favouring a specific technology which is to become global standard.

## The Mechanisms Co-Evolution

The coevolution within the innovation ecosystem is determined by dynamic interactions between actors that try to achieve a common goal. As the characteristics of the ecosystem itself (e.g. coopetition) and the process of setting up common goals is marked by the pondering of which way is the most appropriate to achieve them they are often shaped by disagreement, negotiation, and alliance formation. Hence, these mechanisms, known as controversies, are a suitable way to describe the dynamics of the ecosystem, as the digital innovation is not a static construct. My findings reveal that controversies not only shape the coevolution but also even originate it. Following this argumentation, controversies are not only a negative aspect but also the various interests of human and non-human actors actively shape and improve the composition of the ecosystem as well as the innovation as they foster the way of reflective consideration toward the most suitable outcome. The findings within my case reveal that controversies can have different effects on the social system (human actor network) as well as the technological system (non-human actor network).

In the context of interdependent actors within an ecosystem, the outreach of a controversy goes beyond dyadic relations. Hence, a controversy can also have indirect effects on actors within the ecosystem and create consequential conflicts. The reconfiguration of an ecosystem can create a helix of consequential conflicts that lead to further evolution of the ecosystem as well as the innovation. Due to the multiplexity of relations and interdependencies within ecosystems, the change of constellations of actors may affect the whole ecosystem as the reconfiguration of interests and reallocation of resources can lead to new controversies.

## The Logic of Ecosystem Dynamics

Based on my case study findings I have examined that the coevolution of human and non-human actor networks, which is induced by controversies within innovation ecosystems follows three pathways.

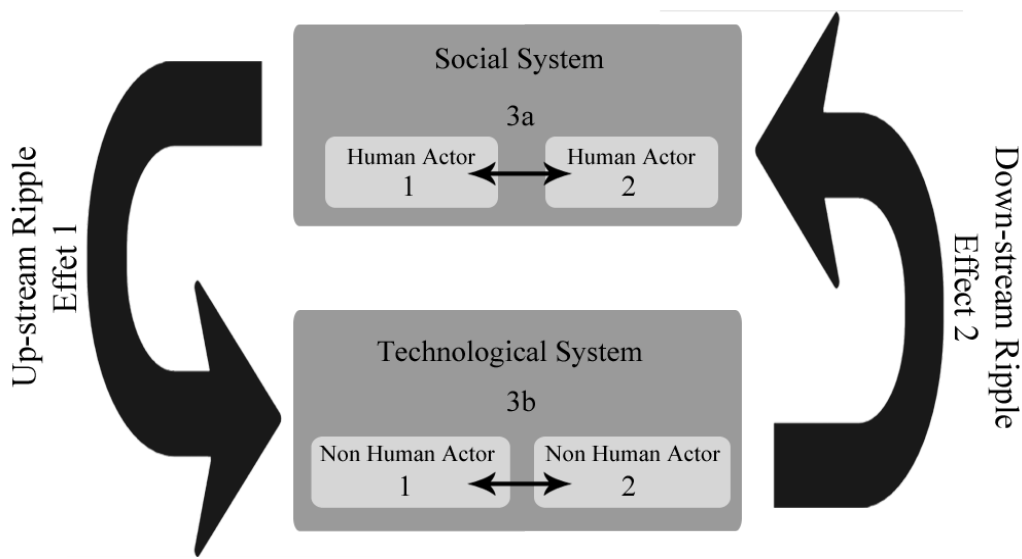

*Evolution of Ecosystems*

First, the logic of coevolution resulting from a controversy can have an upstream ripple effect. This means controversies that result in the reconfiguration of the technological actor network can affect the constellation of the human actor network and therefore lead to ecosystem dynamics (path 1).

Second, the controversy can shape the ecosystem by adding new actors, removing actual ones, or exchanging human and non-human actors. The reconfiguration of the network of human actors (i.e. organizations) can further directly affect constellation of technological components. Adding new actors that inscribe their respective interests into the technological components is often changing the architecture of the digital innovation. Thus, a downstream ripple effect can create the coevolution logic of the ecosystem. For instance, the reconfiguration of human actors resulting from a controversy might induce a consecutive controversy among non-human actors (path 2).

Finally, the third logic of co-evolution must be regarded twofold. On the one side, a controversy may result in a change of the constellation of the human actor network without change of the technological network. This is mostly the case when similar actors substitute actors, which do not need changes in the technological network (path 3a).

On the other hand, controversies can induce a redefinition of the technological network without changes in the respective human network. Hereby, the flexibility of the current organizations within the ecosystem is challenged. This is the case if they can fulfil adjustments needed from technological changes without any profound evolution of the organizational constellation (path 3b).

### 4.1.7. Conclusion

My primary aim in this paper is to analyse the dynamics of innovation ecosystems. I attempt to shed light on how controversies between human (organizational) and non-human (technological) actors coevolve and shape the evolution process of an innovation ecosystem. The results of my qualitative case study of a VPP project in the German energy industry provide several interesting insights for both theory and practice.

By applying ANT, this paper contributes to research on ecosystem dynamics (Um et al. 2013)Ravasz and Barabási 2003) and innovation evolution (Audretsch 1995; Nelson 2009). In particular, I extend previous research on dialectics and creative synthesis in groups as source of innovative outcomes (Harvey 2014) by applying an interorganizational level of analysis. Moreover, I extend this research by highlighting the role of technological actors in the dialectic process of ecosystem evolution. My results provide an integrated view of the interaction between both, technological and social entities and how these affect dynamics of an innovation ecosystem. I therefore show different typologies of controversies, their mechanisms as well as their pathway of influencing ecosystems. My work reveals that social and technological networks within an innovation ecosystem are reshaped by controversies between human and non-human actors, which underline the postulate of (Orlikowski and Iacono 2001) for theorizing the role of (information) technology.

Based on my key findings implications for managers as well as practitioners can be derived. Firstly, it is worth acknowledging that technology is not only a tool but has a far more significant role. While several key inscriptions via the designers lead to a more active role, technology itself must be valued more in-depth since it is often fundamental for the collaboration of most actors und thus becomes a non-human actor itself. I discovered that these non-human (technological) actors play a crucial role as they can also create controversies due to their in scripted interests. Secondly, managers must be aware that the substitution of one actor within the innovation ecosystem not necessarily leads to a frictionless procedure. Even if the substitute can cover the workload on an operational basis, the adjustments and ties towards other actors must be established to prevent further conflicts. Vice versa, changes of the technological components of the innovation can influence the ecosystem. Therefore, I suggest that managers should pay critical consideration on the role of technology.

## 4.2. The Ecosystem as Source of Risk for Entrepreneurs

The findings of this chapter were previously published as (Dellermann and Reck 2017) and examine the configurational mechanisms of how uncertainty, the ecosystem and actions of the entrepreneur in software startups influence each other in driving the success of entrepreneurial success and failure by conducting a FsQCA study with cloud platform app developer startups that built their business model on the participation in the platform ecosystem.

### 4.2.1. Introduction

Pervasive digital technology significantly changes the logic of innovation. One of the most important aspects of organizing such innovation processes is shifting the locus of innovation on technological platforms (Tiwana 2015a; Tiwana et al. 2010). A digital platform, i.e. an extensible code base, allows the development of complementary products or services (e.g. applications) that augment a platform's native functionality (Lyytinen et al. 2016). Companies offering such complementary applications are called software startups or third-party developers (Ghazawneh and Henfridsson 2013). To accelerate innovation on digital platforms, platform owners must create and sustain vibrant ecosystems of third-party developers, which consist mainly of software startups and entrepreneurs that build their business models on the participation in the platform ecosystem (Boudreau 2012). Modular platform architecture enables software startups to develop their own apps independently, yet platform interfaces ensure their interoperability. This tendency towards a disintegrated architecture is mirrored by an increasing degree of interorganizational modularity, distributing the partitioning of innovation among many heterogeneous firms (Baldwin and Clark 2006).

Digital technology therefore creates several idiosyncrasies in the organizational logic of innovation (Yoo et al. 2012). First, the loosely coupled relationships between actors like the platform owner and single third-party entrepreneurs represent a hybrid form of organizations which exhibits characteristics of both markets and formal alliances in the traditional sense of economic exchange theories (Williamson 1985). Second, following this logic, control and knowledge is distributed between various actors (Lyytinen et al. 2016). Finally, such relations are frequently characterized by coopetition (i.e. simultaneous cooperation and competition). For instance, although platform owners encourage the development of third-party innovations, they might compete with software startups in certain market niches (Ceccagnoli et al. 2012).

Although organizing digital innovation around a technological platform has created new business opportunities by providing complementary resources, it also introduced essential new risks. I refer to this phenomenon as risk of third-party innovation. In comparison to traditional risks of software engineering (Barki et al. 1993; Wallace et al. 2004), the locus of this form of risk is not within the own organizational boundaries but on platforms as well as within the focal software startup's relationship multiple and heterogeneous actors. Exogenous and relation-specific factors like for instance opportunistic behaviour of the platform owner, market related factors as well technological dependencies on the platform, thus constitute crucial threats which lay outside the direct control of a software startup.

To theoretically explain the emergence of software development risks and provide IS management with means for its management, previous research proposes that successful organizations establish a fit between the degree of uncertainty of their environment and their structural and control approaches (Bourgeois III 1985). This perspective extensively examined the role and interplay of control mechanism and environmental factors in influencing the risk of IT projects (Rochet and Tirole 2003).

In the context of third-party development on technological platforms, this perspective runs its limits for two main reasons. First, the contingency approach assuming the existence of a single state of fitness between control mechanisms and potential exogenous hazards is not able to capture the increasing dynamics and complexity of an ecosystem as the focus of IT innovation is shifting to platforms. I therefore utilize configuration theory (Ragin et al. 2006; Ragin and Inquiry 2008) as theoretical lens to overcome the traditional reductionism problem (Meyer et al. 1993) and examine the equifinality of different solutions for managing risk in ecosystems where a different set of elements can produce the same outcome.

Second, entrepreneurial software startups are typically not able to apply direct control mechanisms to govern third-party innovation in platform ecosystems for reducing their risk. Congruent with previous work, which highlights the role of modular architecture as a control function for alliances (Tiwana 2008) or to reduce opportunistic behaviour (Williamson 1985). I argue that the modularization of application-platform linkages is the useful mechanism for software startups to manage the relation with the platform owner.

Addressing these two shortcomings of previous research, the purpose of my work is therefore to shed light on software startups' third-party innovation risk by explaining its prevalence based on different configurations exogenous hazards from the platform ecosystem as well as the microarchitecture of single applications which may serve as a safeguard against those hazards.

To explore these issues, my research analyses data from a survey of 42 software startups on five leading cloud platforms using fuzzy set Qualitative Comparative Analysis (FsQCA) (Ragin and Inquiry 2008).

The FsQCA approach is a case-oriented method that enables analysing asymmetric and complex causal effects by extracting configurations that consistently lead to the outcome of interest (El Sawy et al. 2010).

My study offers three noteworthy contributions. First, it outlines the influence of environmental hazards on the risk related to a major form of organizing digital innovation, platform-based application development. Second, it empirically validates the inseparability of environmental dynamics and architectural choices in such digital innovation settings. Third, it offers insights on how digital architecture can be utilized as a coordination device of software startups to manage interorganizational relations and to reduce risk.

### 4.2.2. Conceptual Background

## Risk of Entrepreneurial Innovation

In IS research, risk represents a function of both uncertainty and loss or damage, which is experienced by a decision maker (March and Zur Shapira 1987). A further crucial concept in this context is hazards, which is defined as a source of danger (Kaplan and Garrick 1981). Consequently, if an actor is not able safeguard against such hazards, they create a potential loss, i.e. risks.

Previous approaches examining risk in inter-organizational arrangements like for instance R&D alliances (Das and Teng 1996) or IT outsourcing (Aubert et al. 2004) are theoretically grounded in theories of economic exchange (i.e. transaction cost theory). Following the logic stated in the introduction, however, I argue that the specific characteristics of digital technologies create also significant changes in the nature and analysis of risk. The loosely coupled relationships between the platform owner and a software startup represent a hybrid between characteristics of a market and an alliance. Therefore, significantly new uncertainties evolve for the participants of platform ecosystems. The distribution of control and knowledge among heterogeneous actors accelerates uncertainty regarding the technology itself or the behaviour of the alter (Ceccagnoli et al. 2012). For instance, the platform owner's control over boundary resources (i.e. software development kit (SDK) application programming interfaces (APIs)) makes software startups increasingly dependent (Ghazawneh and Henfridsson 2013).This limits third-party developers' space to control the exchange with the platform owner itself. Furthermore, as this new organizing logic of digital innovation frequently requires coopetition (i.e. simultaneous cooperation and competition) to drive innovation, software startups may suffer from platform owners to adopt and modify their applications to capture attractive market niches. While platform owners encourage the development of third-party innovations, the loss of intellectual property is therefore a common threat in this context (Ceccagnoli et al. 2012).

The risk of third-party innovation as an outcome variable is therefore defined as the potential failure of the software startups' innovation effort in a loosely coupled and co-opetitive relationship with the platform owner. This concept has two distinctive sub dimensions (Nooteboom et al. 1997): risk likelihood (i.e. the probability that the digital innovation effort will fail) and risk impact (i.e. the perceived loss in the form of missing or underachieving the goals of the innovation effort). While the first sub dimension is resulting from uncertainty, the latter is accelerated by the specificity of a digital platform and the resulting migration costs to another technology.

## A Configurational Perspective on Risk

In IS, researchers adopt a contingency approach risk management to examine the role and interplay of control mechanism and environmental factors in influencing the risk of IT projects (Raz et al. 2002; Ropponen and Lyytinen 2000). This approach has been strongly influenced by research in organizational contingency theory, which proposes that successful organizations ensure an originality of fit between the degree of uncertainty of their environment and their structures (Bruns and Stalker 1961). Rather than assuming the existence of best-fitting combinations of predictor variables, I assume equifinality of different configuration of variables Thereby, I take a holistic viewpoint which abstains from evaluating net effects of single variables but treats such configurations in whole as explanatory factors for the outcome of interest. Such an application of configurational theory in the context of digital innovation in platform ecosystems is suitable for two reasons.

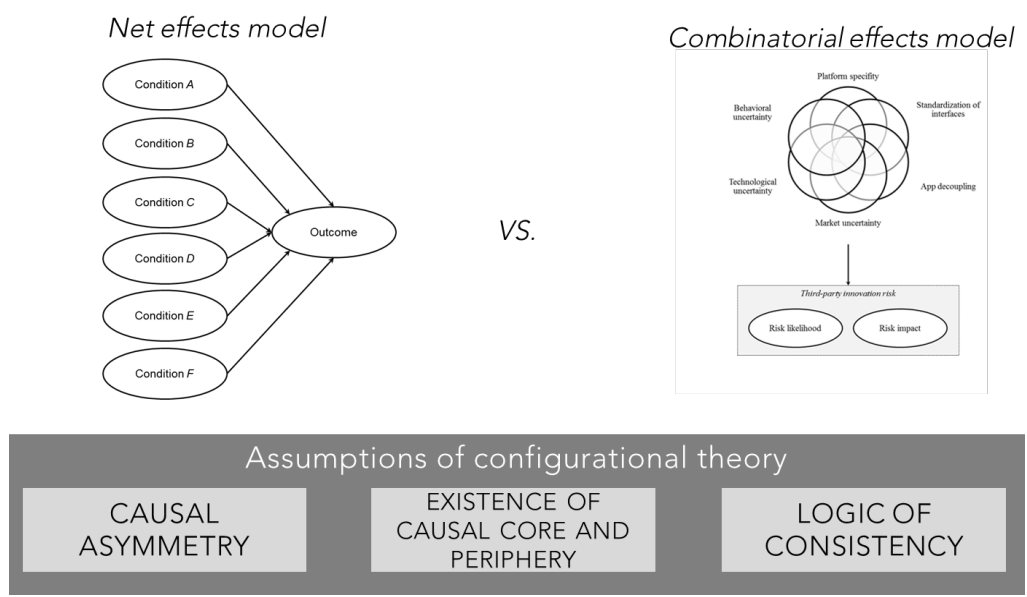

*Combinatory vs. Net Effect Models*

First, in configurational approaches whole sets of elements serve to simultaneously explain the outcomes of interest (El Sawy et al. 2010). Because of that, configurational theory is particularly appropriate to explain synergetic and complementary causalities (Ragin et al. 2006). This resonates well with current theoretical perspectives on the organizing logic of digital innovation in general and platform and ecosystem management in specific. Research in this field highlights the inseparability of ecosystem dynamics from app architectures and their mutual effect on innovation outcomes. Therefore, examining variable in isolation therefore is no reasonable approach towards explaining risk in third-party development.

On the other hand, recent organizational and information systems research suggests that the assumption of symmetric causal relationships might not display organizational realities (Fiss 2011). In contrast, configurational theories imply equifinality between different sets of initial conditions and assume asymmetric rather than symmetric relations between conditional variables and outcomes (El Sawy et al. 2010). Consequently, corresponding analysis procedures allow for the detection of sufficient or necessary causes of a dependent variable. For instance, while the existence of a hazard might consistently lead to high risk for software startups, this does not mean that its absence will lead to low levels of risk (e.g. there might be other hazards which substitute for it). Considering these advantages of configurational perspective, I argue that understanding organizational outcomes of the distributed organizing logic of digital innovation strongly depends on configuration of several design choices with its environment.

### 4.2.3. Research Framework

I divided the concept of third-party innovation risk into two distinctive dimensions: risk likelihood (i.e. the probability that the digital innovation effort will fail) and risk impact (i.e. the perceived loss). The framework comprises two facets of causal conditions for risk. It proposes that from the perspective of software startups, the configuration of four exogenous hazards (i.e. platform specificity; behavioural, market & technological uncertainty) and two endogenous choices to manage their innovation effort (i.e. app decoupling and standardization of interfaces) influence the risk of third-party innovation.

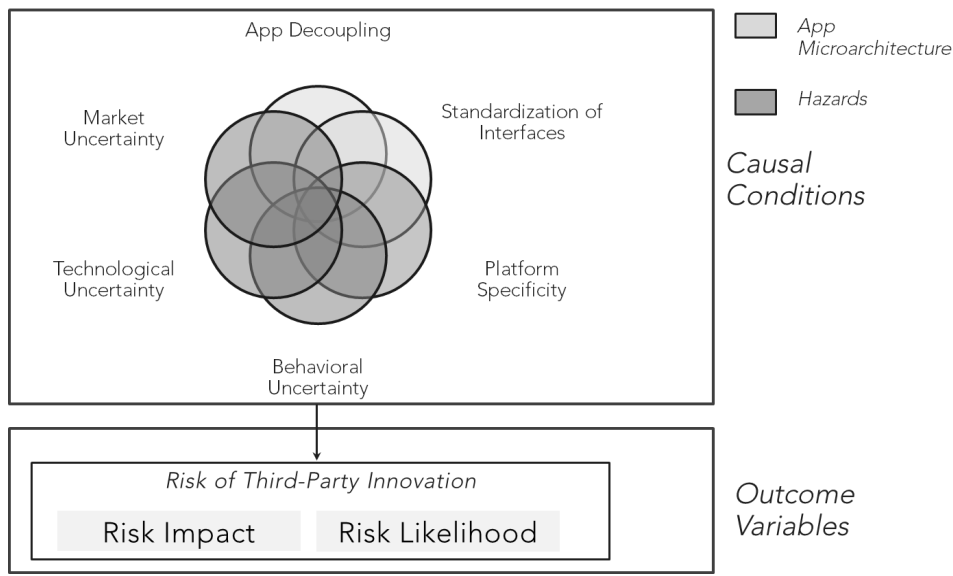

*Configurational Research Framework*

In the selection of my causal conditions, I follow notions of Tiwana et al. (2015a) on intra-platform dynamics and the required fit of architecture and environmental dynamics to process strategic outcomes. My set of causal conditions therefore includes design elements outside (hazards of the ecosystem) as well as within (app decoupling and standardized interfaces) the range of software startups' influence and is theoretically guided by the dimensions of transaction cost theory (Williamson 1985).

**Platform Specificity:** The specificity of a certain platform represents the first hazard for a software startup. Platform specificity refers to the transferability of a software startupś application to a different platform as well as the value of software startupś assets within alternative partner relations (Rindfleisch and Heide 1997). For instance, platforms require investments in relation-specific knowledge to participate in the platform ecosystem and capitalize from the access to complementary resources and capabilities (Aubert et al. 2004). Specific assets can be for instance, human assets, technological assets or knowledge about platform architecture, interface specifications and market characteristics. High levels of asset specificity and the related investment requirements create dependence between partners, lead to lock-in effects which make it difficult for the software startups and move to another platform (Kude and Dibbern 2009). A high specificity of assets required for building complementary products therefore results for instance in high multi-homing costs (Armstrong and Overton 1977; Armstrong and Wright 2007). Therefore, the amount of a potential loss is likely to be higher under conditions of high platform specificity.

The second exogenous hazard for software startups in platform ecosystems is uncertainty, which is commonly defined as the absence of complete information about the contextual environment. This in turn leads to an inability to predict it accurately. The concept of uncertainty is crucial in organization theory and frequently applied in studies on risk in IS (Milliken 1987). For my study I define uncertainty on the interorganizational environment than on the project level. On this level I apply an environmental perspective on uncertainty, which explains the unpredictability of the firm's environment surrounding a relationship between firms (Gatignon and Anderson 1988).

**Market Uncertainty:** Market conditions are crucial drivers for the risk of software startups, as for instance the sustainability of the specific niche is required to succeed. Volatile customer demand, the unpredictable emergence of new substitute products or changes in the competitive environment might increase the threat of failure during the development of complementary products (Rindfleisch and Heide 1997).

**Technological Uncertainty:** Furthermore, technological unpredictability covers the inability to accurately forecast the technological requirements within the relationship, which is especially important in complementary platform markets. Technological complexity and changes are the most significant sources of uncertainty (Nidumolu 1995). Technological uncertainty is also frequently related to a lack of experience with the technologies employed in the ecosystem (Nooteboom et al. 1997), which increases the threat of failure due to inadequate capabilities. Furthermore, the unpredictability of technological evolution might constitute a source of risk during third-party innovation (Lyytinen et al. 2016).

**Behavioural Uncertainty:** In contrast to environmental uncertainty, which is not directly related to the partner, behavioural uncertainty arises from the complexity and difficulty of evaluating each other's actions within a relationship. Taken to the platform context, the platform owner might follow its individual interests and cause hidden costs by inefficient and ineffective behaviour (Williamson 1985). Moreover, although platform owners encourage the development of complementary products to nurture the overall value of the ecosystem (Rochet and Tirole 2003), there is often a tension between them and software startups. This tension arises from the software startupś threat of opportunistic behaviour of the platform owner by for instance exploiting resources or competing in the partner's niche (Kude and Dibbern 2009).

Building on Tiwana (2015a), who outlines the required fit of application architecture and platform dynamics I extend this line reasoning to the risk of third-party innovation. Prior works highlight that the role of modular architecture as control mechanism to influence the outcome of interorganizational arrangements (Tiwana 2008) or to reduce opportunistic behaviour (Stump and Heide 1996). Therefore, third-party developers possess design alternatives based on which they can influence the governance of their relation to the platform. Concretely, the microarchitecture (in contrast to the macro-architecture of the overall platform) of their apps allows software startups to minimize risk by exploiting the benefits of modularization (Tiwana et al. 2013). On the micro level of application architecture, I focus on the modularization of the app-platform linkages rather than internal modular app architectures. App modularization therefore minimizes the application–platform dependencies on the degree to which an app is required to be conforming to the specified interface that is vice versa determined by the platform owner (Tiwana 2015b). Hence, applications within the same ecosystem can significantly vary in their level of modularization (Mikkola and Gassmann 2003) as its micro-architecture reflects an endogenous choice of the software startup.

**App Decoupling:** Decoupling allows for changes within a module which do not require parallel changes in the platform and vice versa. App decoupling reduces dependencies at the boundary between app and platform and minimizes the interactions between both (Tiwana 2015a). Hence, the technological volatility of a platform does not necessarily require changes in the single application. It enables the flexible and independent development of apps. Third-party developers are therefore able to adapt the applicatioń internal implementation without the need of knowledge about internal details of the platform (Tiwana et al. 2010).

**Standardized Interfaces:** Standardization refers to the use of standards and protocols predefined by the platform owner (e.g. platform specific APIs) that are applied to meet conformance between the platform and the software startupś applications. A platform owner introduces such standards to manage the relationships between the app and the platform. Standardization reduces the need for iteration between the software startup and the platform owner and ensures interoperability between the platform and the app. This underlines the role of standardized interfaces as a control mechanism (Tiwana et al. 2013).

Both mechanisms allow software startups to developed apps independently and ensure interoperability with the platform and represent an architectural control mechanism to manage their innovation activities in the ecosystem.

## 4.2.4. Research Methodology

**Data Collection and Sample**

My sample consist of 750 startup firms which built their business model on the participation in the ecosystem of five leading cloud platforms (i.e. Microsoft Azure, Oracle Cloud Platform, Amazon Web Services, SAP HANA, and Salesforce Force.com). There were two reasons for choosing these platforms. First, all platforms are well-established and have solid traction among third-party developers. Second, in all five platforms, a high level of power imbalance is prevalent, so that they perfectly meet my requirements for analysing asymmetric third-party relationships.

Key informant data was collected via a web crawling approach which randomly gathered startup contacts from the platforms´ app stores. This approach is consistent with previous surveys of third-party developers. The potential respondents were contacted via an e-mail containing information on the research project, a link to the online questionnaire as well as the request to complete the survey or to forward the questionnaire to other executives (C-level; IT executives) as further potential key informants (Kumar et al. 1993).

In total, I obtained complete data on N=42 cases. This equals a response rate of 5.6 %, a common value in such settings (Benlian et al. 2015). I assessed this possibility by comparing responses of early and late respondents (Armstrong and Overton 1977). T-tests did not reveal any significant differences ($p > 0.05$) rejecting the presence of non-response bias in my dataset.

Software startups from all five platforms replied (Microsoft Azure: 9; Oracle Cloud Platform: 4; Amazon Web Services: 2; SAP HANA: 9; and Salesforce Force.com: 14). Most of them were high-level executives (C-level: 71.4 %; BU executives: 19 %) and indicated high experience in managing platform-based software development (>10 years: 83.3 %).

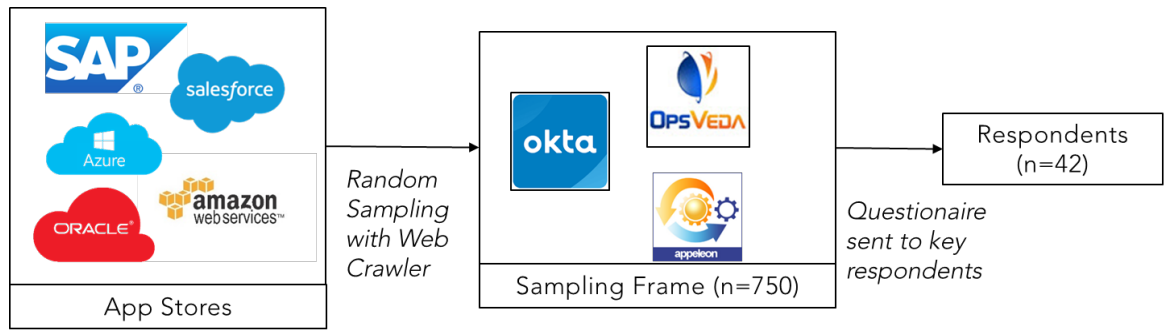

▪ Questionnaire based on literature and refined through extensive discussions and pre-tests with 6 senior managers from the software industry
▪ Use of 7 point Likert scales
▪ Reliability and validity of all constructs (Cronbach´s α>.86; CR>.9; AVE>.72)

*Sampling Approach*

## Measurement Validation

Based pilot study with managers in the software industry, I constructed my measurement instrument. To ensure validity, reliability as well as rigor of my research (Lewis et al. 2005), I adapted existing scales to the platform context and refined them based on the insights from the pilot study. Subsequently, these refined items were evaluated in a pre-test procedure. This helped me ascertaining that the formulation of all items was unambiguous and comprehensible.

The psychometric statistics (see Appendix) of the measured constructs indicate a strong evidence for adequate reliability with Cronbach's α greater than .85 for all variables. Furthermore, I can assert discriminant validity as confirmatory factor analysis yielded high factor loadings concerning so that the Fornell/Larcker criterion is fulfilled for all my study variables.

To reject the possibility of common method bias, I conducted Harman's single-factor test (Podsakoff et al. 2003). The unrotated factor solution resulted in 5 factors explaining 77 % of the variance (35 % was the largest variance explained by one factor). Hence, common method bias is unlikely to be a problem.

## Fuzzy-Set QCA

I chose FsQCA as means to analyse the obtained data. This set-theoretic approach is utmost suitable to configurational theories as it aims at extracting whole configurations rather than single factors that help to explain outcomes of interest (Fiss 2011). Thereby, FsQCA draws on set-based measures of consistency and coverage to evaluate the predictive power of the potentially possible conditional configurations. Consistency values display to which degree cases that share a certain combination of conditions also lead to a specific outcome (Liu et al. 2017). Hence, this indicator is analogous to correlation estimates in statistical methods. The other indicator of quality, coverage, represents the degree to which a configuration covers the instances on which a specific outcome is realized. Defined as such, the meaning of coverage values resembles that of R-square values in regression analysis. The FsQCA procedure consists of three steps through which consistent configurations are detected (Ragin and Inquiry 2008): calibration, construction of truth tables, truth table analysis.

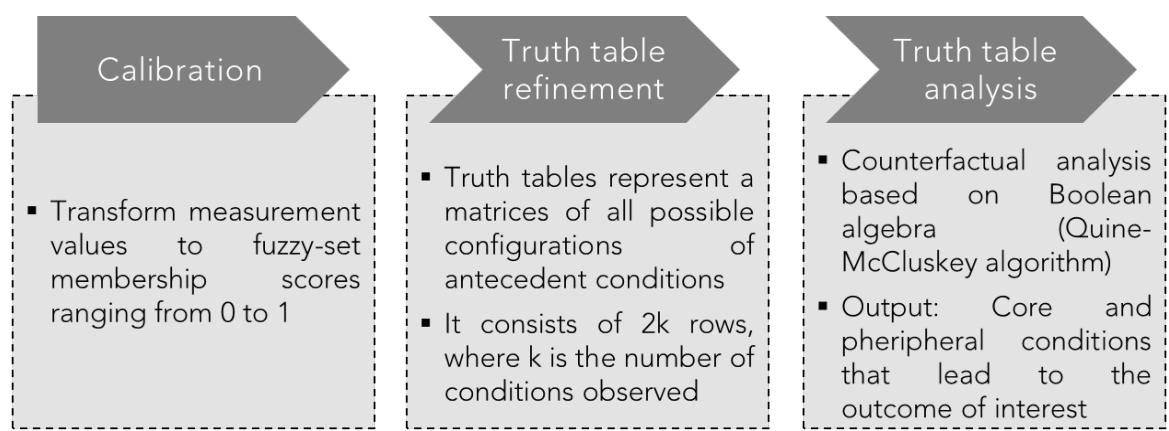

**FsQCA Process Steps**

Calibration of construct measures is necessary because FsQCA as a set-theoretic analysis approach draws on membership scores (here, e.g. membership in the group of firms with highly decoupled apps) rather than values on interval or ratio scales. In my study, I thus transformed the Likert scale measures into fuzzy set membership scores. These range between 0 and 1 with 0 indicating full non-membership, 1 indicating full membership and 0.5 marking the crossover point (Schneider and Wagemann 2012). I follow the calibration approach outlined by Fiss (2011) and chose the observed maximum and minimum values within the sample to specify full membership and full non-membership for all variables. The median of observed values served as cross-over point. Based on these three values, the calibration procedure in the FsQCA software program (version 2.5) (Schneider and Wagemann 2012) transforms all obtained measures to membership scores.

The second step of FsQCA is the construction and refinement of a matrix of all configurations of antecedent conditions (in my case a 64x6 matrix; in general, $2^k$xk, with k as the number of conditions observed (Ragin 2008). To fit the requirements of FsQCA, this truth table must subsequently be refined. This procedure evaluates each configuration based on two criteria: frequency and consistency. The frequency assesses which of the configurations appear in the dataset. In Large samples, it is often reasonable to exclude infrequent cases so that it is necessary to set a frequency threshold for the inclusion of configurations in the further analysis procedure. As my sample is medium-sized in terms of FsQCA literature, I chose the standard threshold of 1 which is suitable for samples of this size. The consistency criterion captures if a truth table row consistently yields an outcome. The consistency value thereby should outreach at least .8, so I chose a conservative threshold of .9 (Schneider and Wagemann 2012). Overall, in 28 cases, configurations exceeded the frequency threshold of which 13 also exceeded the consistency threshold for risk likelihood and 17 for risk impact.

In the third step, the truth tables are analysed via counterfactual analysis. This approach is based on Boolean algebra in general and applies the Quine-McCluskey algorithm. This algorithm strips away factors which are not consistently present concerning an outcome (Fiss 2011) to identify the conditions within a configuration which cause the outcome. Hence, the algorithm excludes conditions that are no essential part of a sufficient configuration for the respective outcome and produces two distinct solutions: the parsimonious solution and the intermediate solution. The parsimonious solution on the one hand draws on all simplifying assumptions derived from counterfactuals. It passes a more thorough reduction procedure, so that the data provides strong empirical evidence for the causality of these conditions. Therefore, the parsimonious solution encompasses the causal core of conditional variables. In contrast, the intermediate solution only includes simplifying assumptions based on easy counterfactuals (Ragin 2008). The conditional variables which appear in the intermediate solution but do not appear in the parsimonious solution thus represent the causal periphery of a configuration (Fiss 2011).

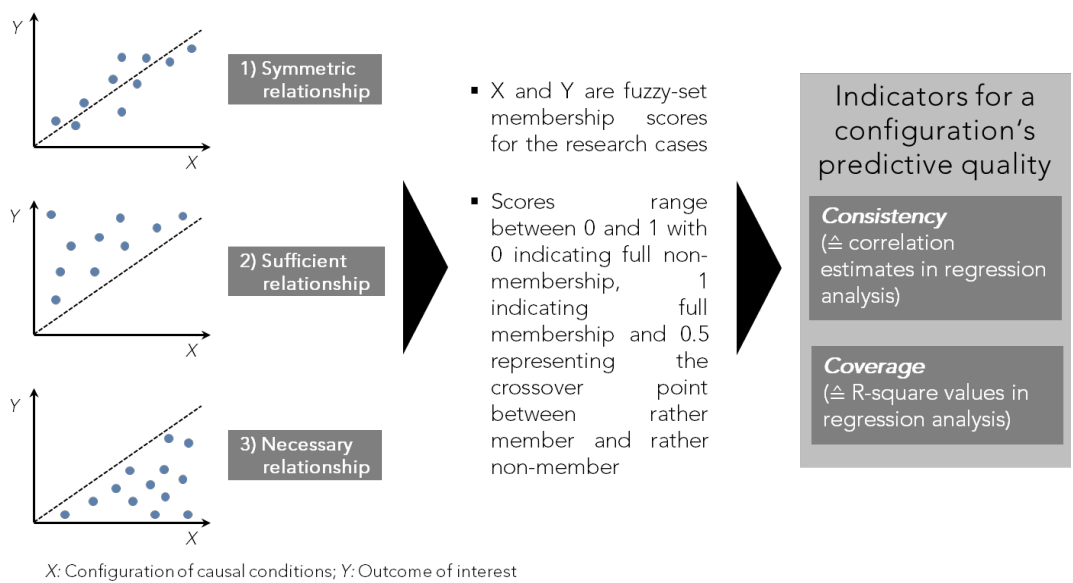

X: Configuration of causal conditions; Y: Outcome of interest

## 4.2.5. Results

The results of the FsQCA reveal several patterns that explain how different configurations of app architecture and environmental hazards result in high or low levels of both risk likelihood and risk impact. I extracted these patterns by comparing structures of different configurations. Black circles indicate the presence of a condition, crossed-out circles indicate the absence of a condition, large circles indicate core condition, and small circles indicate peripheral conditions. Blank spaces indicate a condition may be either present or absent.

| | Solutions for high risk likelihood | | | | | | Solutions for high risk impact | | | | | | |
|---|---|---|---|---|---|---|---|---|---|---|---|---|---|
| | 1a | 1b | 2a | 2b | 3a | 3b | 1 | 2a | 2b | 2c | 3 | 4a | 4b |
| *Hazards* | | | | | | | | | | | | | |
| Platform specifity | ⊗ | ● | ⊗ | | ⊗ | ● | ⊗ | ● | ⊗ | ● | ⊗ | | ● |
| Behavioral uncertainty | ⊗ | ⊗ | ⊗ | ⊗ | ⊗ | ● | | ⊗ | ● | ● | ⊗ | ● | |
| Technological uncertainty | ● | ● | ⊗ | ⊗ | ⊗ | ⊗ | ● | ⊗ | ⊗ | | ⊗ | ● | ● |
| Market uncertainty | ⊗ | ● | | ⊗ | ● | ● | ⊗ | ● | ● | ● | | ● | ● |
| | | | | | | | | | | | | | |
| *App microarchitecture* | | | | | | | | | | | | | |
| App decoupling | ⊗ | ● | ⊗ | ● | ⊗ | ⊗ | ⊗ | ⊗ | ⊗ | ● | ⊗ | ● | ● |
| Standardized interface | ⊗ | ● | ● | ● | ⊗ | ⊗ | ⊗ | ⊗ | ⊗ | ● | ● | ⊗ | ● |
| | | | | | | | | | | | | | |
| *Consistency* | .92 | .97 | .95 | .96 | .95 | .97 | .94 | .96 | .95 | .99 | .91 | .92 | .98 |
| *Raw coverage* | .30 | .30 | .34 | .49 | .25 | .29 | .35 | .24 | .30 | .48 | .32 | .37 | .48 |
| | | | | | | | | | | | | | |
| *Overall solution consistency* | | | .91 | | | | | | .89 | | | | |
| *Overall solution coverage* | | | .72 | | | | | | .81 | | | | |

Notes: Black circles indicate the presence of a condition, circles with "x" indicate its absence. Large circles indicate core conditions, small ones indicate peripheral conditions. Blank spaces indicate may be either present or absent.

## Configurations for High Risk Probabilities

I identified seven different configurations that result in a high likelihood of risk. Consistency for configurations ranges from 0.90 to 0.99. Raw coverage, which describes the importance of a certain configuration in explaining the intended outcome, range from 0.26 to 0.46. The overall solution consistency shows these seven solutions can consistently result in high likelihood of risk with 89 %. The overall solution coverage indicates that the extent to which these seven configurations cover high likelihood of risk cases is 76 %. I compared the seven configurations of my analysis to extract two strong patterns:

**Pattern (1):** In platform ecosystems with a high level of market uncertainty software startups are very likely to perceive a high likelihood of risk in third-party innovation (2a&b; 3a&b), which can be explained by the increased likelihood for market disruption or instability of the software startups niche.

**Pattern (2):** If the interfaces are not standardized and market uncertainty is high (3a&b), especially with lack of app decoupling as peripheral condition, the likelihood of risk for software startups is high as changing market conditions might increase the need for adaptions in the application. However, if apps are not modularized, software startups are not able to improve the application fast and independently. Therefore, lack of modularization reduces the flexibility to react to changes within the market environment.

## Configurations for High Risk Impact

Furthermore, I identified seven different configurations that result in a high impact of risk that exceed minimum consistency threshold. These seven solutions consistently result in high risk impact with 89 % and cover 81 % of cases with this outcome. Comparing the seven configurations reveals two further important patterns:

**Pattern (3):** The impact of software startupś risk in third-party innovation is high when the environment is volatile. Market uncertainty (2a, b, c; 4a & b) and technological uncertainty (1; 4a&b) are the main hazards to result in a high impact of risk.

**Pattern (4):** The interplay of high interface standardization and low app decoupling (3) represents the second pattern to create a high impact of risk. This can be explained as high standardization requires high investment of the software startup to adhere platform-specific interface standards while a lack of decoupling reduces flexibility and increases the threat of cascading ripple effects that might disrupt its interoperability with the platform.

**Configurations for Low Risk Probabilities**

I compared the sets of causal conditions of low risk with the configurations that lead to high risk to detect relevant differences. Consequently, I identified six configurations that result in a low likelihood of risk.

These solutions consistently result in a low likelihood of risk with 91 % and cover 72 % of cases with this outcome. Comparing the six sets of causal conditions I extracted three further patterns:

**Pattern (5):** If behavioural uncertainty is missing, software startups perceive a low likelihood of risk (1a&b; 2a&b), although technological uncertainty is high (1a&b). This shows that software startups that can monitor the behaviour of the platform owner face a lower likelihood of risk as they reduce the space for opportunism.

| | Solutions for low risk likelihood | | | | | | | Solutions for low risk impact | | | | | |
|---|---|---|---|---|---|---|---|---|---|---|---|---|---|
| | 1a | 1b | 1c | 2a | 2b | 3a | 3b | 1 | 2 | 3a | 3b | 4 | 5 |
| *Hazards* | | | | | | | | | | | | | |
| Platform specifity | ⊗ | ⊗ | ● | ⊗ | ● | ● | ⊗ | ⊗ | ● | ⊗ | ⊗ | ⊗ | |
| Behavioral uncertainty | ⊗ | | ● | ⊗ | ● | ⊗ | ● | ● | ⊗ | | | ⊗ | ⊗ |
| Technological uncertainty | | ● | ● | ⊗ | ⊗ | ⊗ | ⊗ | ⊗ | ⊗ | ● | ● | ⊗ | ⊗ |
| Market uncertainty | ⊗ | ⊗ | ● | ● | ● | ● | ● | ● | ● | ⊗ | ⊗ | | ⊗ |
| | | | | | | | | | | | | | |
| *App microarchitecture* | | | | | | | | | | | | | |
| App decoupling | ⊗ | ⊗ | | ⊗ | ● | ⊗ | ⊗ | ⊗ | ⊗ | ⊗ | ⊗ | ⊗ | ● |
| Standardized interface | ⊗ | ⊗ | ⊗ | ● | ● | ⊗ | ⊗ | ⊗ | ⊗ | ⊗ | ⊗ | ● | ● |
| | | | | | | | | | | | | | |
| *Consistency* | .90 | .92 | .99 | .92 | .96 | .98 | .98 | .98 | .97 | .97 | .98 | .89 | .93 |
| *Raw coverage* | .40 | .39 | .46 | .32 | .34 | .29 | .36 | .37 | .29 | .42 | .44 | .37 | .55 |
| *Unique coverage* | .03 | .01 | .15 | .03 | .08 | .01 | .01 | .02 | .01 | .03 | .03 | .03 | .26 |
| | | | | | | | | | | | | | |
| *Overall solution consistency* | | | | .89 | | | | | | | .90 | | |
| *Overall solution coverage* | | | | .76 | | | | | | | .83 | | |

Notes: Black circles indicate the presence of a condition, circles with "x" indicate its absence. Large circles indicate core conditions, small ones indicate peripheral conditions. Blank spaces indicate may be either present or absent.

***Configurations for Low Risk***

**Pattern (6):** Configurations of market uncertainty in presence with an absence of technological uncertainty account for low risk likelihood (3a&b) if the company does not draw on app decoupling. This fact can be explained as technological stability allows the software startup to reduce risk by offering ability to react to changes in the market quickly. Under these circumstances app decoupling does not offer additional benefits.

**Pattern (7):** Likelihood of third-party innovation risk is low when interfaces are highly standardized (2a&b), which reflects the role of interfaces to standardize rules that apps ought to obey and can expect the platform to obey. This underlines the role of app architecture as a control mechanism for risk.

**Configurations for Low Risk Impact**

By analysing cases for a low impact of risk, I uncovered six different configurations that result in a low impact of risk. These solutions consistently result in that outcome with 90 % and cover 83 % of cases with a low level of risk impact. By comparing these configurations for low risk impact, I found two final patterns:

**Pattern (8):** Surprisingly, the specificity of a platform is not a main driver of risk impact but its missing predicts low impact of potential losses (1; 3a&b; 4). From this finding I can derive that software startups do not perceive failure to have a high impact on them when they did not heavily invest in knowledge and other resources that are idiosyncratic for this certain platform or app migration to another platform can be easily achieved.

**Pattern (9):** If uncertainty in the ecosystem is low, software startups face a low level of risk impact. Especially, when behavioural and technological uncertainty are missing (2; 4; 5). This shows the interplay of a reduced space for opportunism and the stability of the platform in reducing risk.

## 4.2.6. The Drivers of Risk

From the nine pattern that I identified in the comparison of configurations that lead to high and low risk, I can reveal holistic insights of the drivers of third-party innovation risk and the role of app architecture as a control mechanism. Based on the commonalities among the patterns, I identified three holistic findings to explain the risk of third-party innovation and its management.

First, uncertainty of the platform owneŕs behaviour as well as the specificity of a platform, are no main drivers of software startuṕs risk. Instead configurations in which both are absent display a low impact and likelihood of risk during digital innovation. Hence, while environmental hazards are needed to turn specific assets and opportunistic partners into considerable drivers of risk, engaging with reliable partners or acting on platform with low asset specificity might at least partially mitigate the impact of environmental hazards.

Second, market and technological uncertainty are the main drivers of risk in digital innovation. Unstable market conditions and technological volatility are crucially influencing the impact and likelihood of risk during third-party innovation.

Third, application architecture represents not a direct control mechanism to govern the platform dependencies during digital innovation. Standardization of interfaces might represent a necessary condition to achieve a low level of risk under certain circumstances. Consequently, the use of standardized interfaces is required to minimize risk. However, if apps are highly modularized, this does not necessarily imply low levels of risk, but the effect depends on the environment.

### 4.2.7. Conclusion

By comparing different configurations that result in high and low risk, I identified nine patterns that describe the role of environmental hazards and app architecture in shaping risk. From these patterns I derive the role of technological and market uncertainty as core drivers of risk. Furthermore, my findings reveal that behavioural uncertainty and platform specificity are not drivers of risk per se. However, their absence is required to achieve low levels of risk. In addition, I detect the role of app architecture as a control mechanism for third-party innovation. As the absence of app modularity is always implying a high level of risk, it is a necessary condition for minimizing risk.

Therefore, the contribution of my study is threefold. First, it contributes to research of risk in IS by applying a configurational perspective on the new organizing logic of digital innovation and providing evidence for the equifinality of different paths in reducing risk. Second, my research contributes to past work on platform dynamics (Ceccagnoli et al. 2012) and intra-platform management  (Tiwana 2015a,b) by uncovering the interplay of environmental factors and technological architecture in achieving organizational outcomes. Third, I contribute to previous studies on modularity as interorganizational control mechanism (Tiwana 2008; Tiwana et al. 2013) by revealing app modularization as necessary condition to minimize risk.

From a practical point of view, my results show that app developers should use app decoupling and standardized interfaces to reduce risk in uncertain environments.

## 4.3. How to Manage Risk and Uncertainty in Business Model Design

The findings of this chapter were previously published as (Dellermann et al. 2017a) and develop a framework for managing both risk and uncertainty in digital business model design through stakeholder integration. This study is based on a multiple case study in the German energy industry.

### 4.3.1. Introduction

The concept of business model design gained significant attention over the last years, as companies like Apple and Uber disrupted whole industries and generated tremendous returns offering not just new products or services but designing new concepts of doing business. Two key features characterize an essential portion of these business model designs: They are enabled by digital technology and embedded in complex inter-organizational networks. In such ecosystems, firms do not solely rely on internal innovation and value creation endeavours. Instead, they are involved in innovation activities with partners and thus are highly dependent on resources and contributions of suppliers, vendors of complementary offerings, consumers, and other actors (Adner 2006). The emergence of the Internet of Things (IoT) creates a global network of connectivity that are enabled by standardized and interoperable formats and connecting heterogeneous digitized objects via the internet. Also, traditional industries, like for instance the German energy industry, are therefore encouraged to combine digital and physical components into novel value propositions. The accelerating interdependence between innovation partners, however, has not only created new business opportunities but also introduced essential new risks. Such risks are not sufficiently covered by traditional approaches of risk management.

This paper addresses this critical gap by offering important insights from digitally enabled business models in the German energy industry that can guide practitioners in managing the process of digital business model transformation. To explain how managers should treat risks related to digital business model design together with multiple partners, this paper analyses a specific digital business model design in the energy sector – the VPP. As a result, a new multi-step framework for the strategic management of risks in digital business model design is proposed.

### 4.3.2. Business Model Design in the IoT

As digital technology is combining atoms and bits to turn digital and physical components into novel products, ubiquitous computing enables the interconnection of multiple devices (Iansiti and Lakhani 2014). In particular, the Internet-of-Things (IoT) has a strong potential to transform products, services and whole industries (Manyika et al. 2015) since it constitutes a *"[..] dynamic global network infrastructure with self-configuring capabilities based on standard and interoperable communication protocols where physical and virtual "things" [...] use intelligent interfaces and are seamlessly integrated into the information network [...]"* (Vermesan and Friess 2014b).This allows the connection of heterogeneous digitized objects that are integrated into the Internet. Companies like Nest, SmartThings or Axeda, for instance, link billions of devices worldwide. Moreover, established firms like General Electrics and Cisco have started to develop and offer numerous IoT-based products and services, increasingly extending to all areas of everyday life. More and more, smart, connected products are questioning the traditional logic of how value is created and captured, offering firms new possibilities for business model designs (Porter and Heppelmann 2014).

In the context of digital business models, diverse devices and IT infrastructures allow multiple actors to interoperate and distribute value creation across various companies. These relations link suppliers, software startups, integrators and customers and enable new logics to create mutual value. Thus, to successfully design and deploy the business model, a firm must clarify which resources it has to acquire from its business partners and which main activities these partners perform and attract and maintain effective and efficient relations to the key collaborators (Chesbrough 2007). Due to the rapid transformation of the technological and competitive environment, business models require regular monitoring and therefore have themselves become a new subject of innovation (Osterwalder and Pigneur 2010)

Even firms in traditional industries, such as the German energy sector, are realizing the disruptive potential of digital innovations. For instance, the Fraunhofer Institute for Integrated Circuits is currently offering its OGEMA 2.0 (Open Gateway Energy Management 2.0) open source framework, enabling the development and implementation of all kinds of systems, components and apps for energy and facility management. Moreover, start-ups like Next Kraftwerke leverage digital technology to create and implement new business models for VPPs.

However, as digitization redefines all elements of doing business such as customer interactions, deployment of resources and economic modes (Jong and van Dijk 2015), it also gives birth to new risks for the actors in novel ecosystems.

### 4.3.3. How the IoT Transforms the Energy Industry

In the past, the success of German energy supply companies resulted from the ownership of big centralized power plants that mass-produced electricity for many households and industrial customers. In this business environment, energy providers were able to gain competitive advantages particularly by building on the economies of scale. This business model, however, comes increasingly under siege from the shift towards decentralization of production, ecological consciousness of customers (e.g. de-carbonization and the so-called *"German nuclear exit"*) as well as the liberalization of energy markets. Realizing this dramatic change and addressing it by the innovation strategy becomes therefore crucial for firms competing in energy markets. Especially the municipal utility companies that used to rely heavily on conventional (fossil) power plants are currently facing significant disruption through increasing capacity additions of renewable energies. For these companies, an innovative response aimed to compensate the loss of market share is the offering of consultancy services and new storage solutions for the fluctuating renewable power supply. This change means much more than merely new business activities: It leads to a fundamental transformation of the underlying business model. To describe and highlight the relevance of digitization (and the resulting new business models) for the energy industry lessons from other sectors can be drawn. Therefore, many other industries deliver cautionary examples for underestimating the impact of digitization on existing business models like the print industry, the music industry, or streaming services. In consequence, many digitization-driven innovations evolve in the energy industry over the last years such as Smart Home and Smart Grid solutions. The chance to use and exploit Big Data offers many new business opportunities for energy providers since the data that is anyway available can be transformed to develop new products and services for customers (e.g. weather forecasts, energy consumption data, and optimization of connected electricity flows).

## 4.3.4. VPP as Entrepreneurial Action in the Energy Sector

Caused by highly fluctuating feed-in times of renewable energies, a stable energy supply, a reliable base load that enables companies to ensure their power supply, has become much more important than in previous years. VPPs provide an innovative solution to this problem as they integrate several small, decentralized power-generating units – foremost renewable ones such as photovoltaic, wind farms and biogas plants.

This business model design in the energy sector is enabled especially by recent technological innovations in the IT area, particularly IoT. Moreover, the VPP business model transcends industrial borders as it relies on orchestrated activities of various actors from different industries – such as IT vendors, energy producers, hardware developers and marketer service providers. In this new ecosystem, digitization facilitates organizational learning across borders and co-innovation that builds on creative combinations of knowledge from diverse technology and application fields.

Hence, digitally enabled business model design fundamentally transforms the value chain from the conventional unidirectional, i.e. vertical design towards a network-centric approach. For instance, electricity customers (factories, farmers etc.) in VPPs play a dual role both as consumers and as producers of electricity. The involvement of these *"prosumers"* in the ecosystem fundamentally modifies the key element of any business model – the fundamental logic of how value is produced and delivered to customers at an appropriate cost.

The basic principle of a VPP is as follows: Several operators of energy generating units (such as biomass plants, photovoltaics, wind farms etc.) which are mostly characterized by erratic feeds to the grid are virtually combined by the VPP-operator into one unit. The advantage is that other participants of this business model equalize the inconstant generation on a broader scale. Thus, the VPP-operator can offer stable energy deliverance. In addition, in the pooling effort weather forecasts as well as grid requirements are assessed to optimize the orchestrated sale of the generated power. The target customers in the commercialization phase may be the own power producers themselves (upgrade their high unstable generation towards a fitting stable one) or energy markets in times of high revenues. This is also assessed by the VPP-operator who can also be described as the business model innovator.

Consequently, VPPs provide a very current and rich real-life example to explore risks that are associated with the co-innovation of a digital business model – that is, the design of a new IT-enabled ecosystem that consists of many diverse and interdependent actors integrated into value creation and capture processes.

### 4.3.5. The Role of Risks Management in Entrepreneurial Actions

Traditionally managers focus on risk management at the operational level, while its strategic role in the multi-partner business model design remains under-investigated (Calandro 2015). Traditional risk management techniques like VaR (Value at Risk) rely on quantitative and historic data and on predicting and controlling specific risk events. Thus, they provide little help for digital business model designs. As companies design their digital business models together with external partners, the accelerating interdependence on these partners makes new strategic approaches of managing risk indispensable.

New threats in digitally enabled ecosystems are often beyond the direct unilateral control of the innovator and related to the interdependence of suppliers, vendors of complementary products and services and other relevant actors. To address these risks, new risk management frameworks and tools are required. This is also particularly true for the German energy industry that is currently being dramatically changed by the digitization.

## 4.3.6. Methodology

To address co-innovation risks in digital business models, the paper adopts an exploratory multiple case study research design that particularly allows the research of unexplored topics (Yin 2017). For an in-depth examination of the risks, the authors conducted 22 semi-structured interviews with managers from leading German energy utilities as well as major providers of VPP-technology. The case companies were chosen due to their popularity and their market position. Hereby I deliberately chose pioneers as well as fast and late followers to compare their differing approaches. To provide a holistic view, I included perspectives from all ecosystem actors over the value chain. For instance, leading project managers from the supply side (power operators), the customer side (industry, grid operator), internal project managers as well as leading research institutes for industry standard development have been interviewed. In addition, I collected, clustered, and listed 36 press articles and official documents along with internal documents (partnership agreements, supplier conditions, legal documents etc.). Finally, the observations made by one of the authors who participated in the VPP project in one of the case companies for more than 10 weeks were also very valuable.

## 4.3.7. Framework for Managing Entrepreneurial Risk

From the in-depth analysis of both the best as well as the worst practices and experiences mentioned by the interview partners my research findings suggests a four-step framework for the management of risks associated with the co-innovation of business models with multiple partners particularly in the energy industry (see Figure 18).

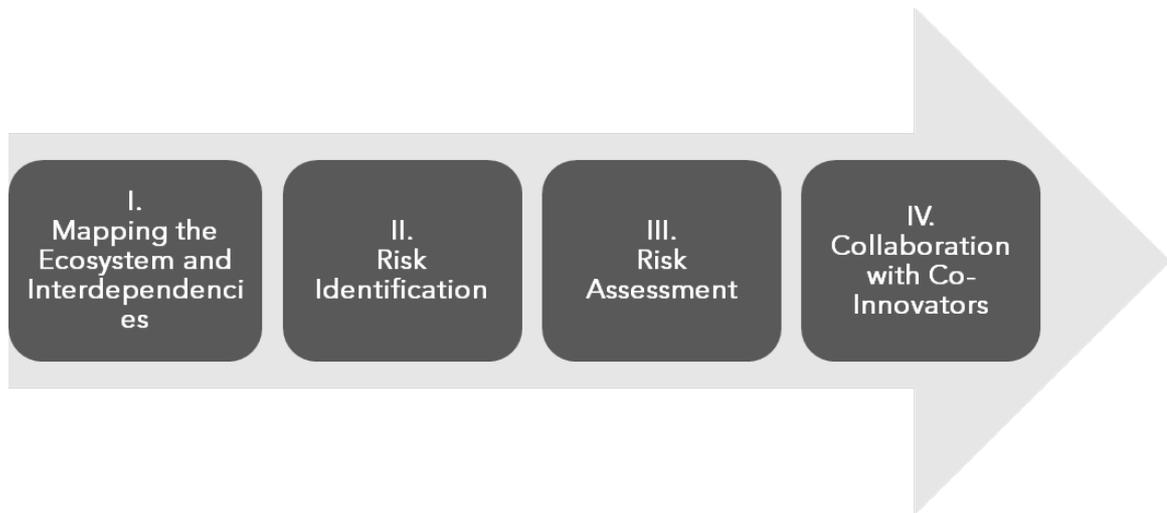

*Risk Management Process for Business Model Design*

## Step I: Mapping the Ecosystem

The major challenge of this step is to realize and assess that the company is manoeuvring through the interplay of several interdependencies. Managers need to identify their ecosystem partners and their roles first. The actors participating in the digital business model design are, for instance, the providers of technical components, complementary products, and services as well as the marketer institutions and the customers.

At this stage, it is crucial to diagnose the interdependencies for each partnership that are relevant for the co-creation and functioning of the new business model. Managers apply the concept of interdependence to consider organizations as entities that rely on an exchange of resources with external organizations such as suppliers, competitors, or regulators (Katila et al. 2008). My research revealed three main sources of interdependence that are particularly salient in the VPP business model design: Regulation-driven interdependence, technological interdependence, and collaborative interdependence.

First, regulatory requirements shape the interdependence of ecosystem relationships, especially in highly regulated industries like the German energy sector where the partners must apply directives for shut-down/response times or grid-operator requirements and certified guarantees of supply origins simultaneously.

Second, the co-innovation of a digitally enabled business model bears critical technological interdependencies: To function appropriately software and hardware components from several providers must be made compatible and technologically integrated. In VPPs, this is especially important for the connectedness of all control devices in the complete system as well as for valid software codes for the linkage to the grid system operators of VPPs.

Third, critical interdependencies become manifest in collaborative agreements that set mutual contractual obligations for actors as well as economic sanctions for failure to fulfil them. Since in the VPPs all customers are also suppliers of energy, the failure of performance caused by one actor may affect the whole ecosystem. For instance, if a vendor of hardware boxes does not deliver on time, the customers cannot be connected to the VPP in a timely manner from which in turn the whole business model suffers.

**Step II: Risk Identification**

Following the insights gained in step one, managers should identify distinct categories of risks that are associated with the innovative business model. First, there are typical risks of internal corporate R&D and product innovation projects (e.g. development of a new component). These risks can be treated with well-known technology and innovation management tools, such as the Stage-Gate® model. Another category of risks is related to the strategic environment of the company and its dynamics (e.g. market changes, appearance of new substitute technologies, changing regulations and governmental interventions, etc.). Those external risks, such as tightening of ecological regulations, can seriously affect the new ecosystem and its actors.

In addition to these two types of risks that are relevant for innovations, the third risk category is directly caused by the fact that multiple partners base the novel ecosystem on co-innovation activities. Such co-innovation risks can be divided into two subcategories: relational and performance risk (Das and Teng 1996). While relational risk refers to the *"will"* dimension of co-innovation, performance risk is primarily related to the *"skill"* dimension. Relational risk is particularly associated with opportunistic behaviour such as distortion of information and fraud. On the contrary, performance risk of co-innovation is particularly related to capability factors: Despite the willingness to co-innovate, firms might not be able to do so due to the lack of skills. For example, in the case of VPP, the developer of important software was not able to deliver the sophisticated and novel software. As a result, the software firm had to be replaced by another provider, which caused an essential delay and opportunity costs borne by all ecosystem actors as well as high transaction costs for search and negotiation borne particularly by the system integrator. My study has shown that managers should distinguish between different types of risks to be able to address them in an effective way.

**Step III: Risk Assessment**

To map and assess risks, many companies deploy a risk response matrix. This popular managerial tool reflects two key risk dimensions – the potential impact, or magnitude, and the likelihood of a certain risk (Aabo et al. 2005). My research shows, however, that especially for the co-innovation of digital business models, such as VPPs, that rely on manifold interdependencies between diverse actors this approach to risk assessment must be expanded.

It must be considered that the risks faced by the given company can also affect other co-innovators who consequently might be hindered or even become unable to provide their specific contributions to the ecosystem. This is true for both internally caused risks (e.g. the risk of R&D failure in the focal company) as well as risks rooted in dyadic relationships with partners (e.g. the relational risk of fraud). Consequently, both risks can be contagious as they might affect not only the given dyadic relationship but seriously damage other interdependencies within the ecosystem and pass problems onto other partners in the value network, such as complementary innovators, intermediaries, or system integrators.

Hence, based on my research findings I assume that in new digital ecosystems an additional dimension for risk assessment must be considered: the outreach. This dimension reflects whether the impact of the given risk is local, dyadic, or systemic and therefore that risk does affect only the focal company, one or many of its dyadic partners or even the entire ecosystem.

Based on the key risk facets mentioned above the following risk radar can be suggested. The novel tool helps visualize and compare risks associated with relationships the company maintains to different co-innovators within the ecosystem. The five-point scale for three interdependencies with the given co-innovator shows their estimated degrees. For the two sorts of co-innovation risks as well as the strategic environmental risk (as far as it affects the relationship with the given co-innovator) the radar also assigns a risk rating on a five-point scale by combining both estimated magnitude of risk and the likelihood of its occurrence (Aabo et al., 2005). The risk outreach goes beyond the bilateral dimension and reflects the levels in the ecosystem that are affected by the risks embedded in the given co-innovation relationship.

**Step IV: Integrate the Ecosystem**

After identifying and assessing the interdependencies and risks, organizational decision-makers must be able to come up with a strategic action plan. For successfully managing the risks of digital business model designs, it is crucial to integrate selected partners in this process. Depending on the allocation of responsibility for mitigating risks and the decision whether those risks are manageable independently or collaboratively, I suggest the following risk matrix that helps draw detailed mitigation plans for specific types of risks.

In sum, the practical use of the proposed framework can be illustrated by the following example. As indicated above, an essential performance risk in setting up a VPP business model is technological complexity. The operator of a VPP needs to ensure the synchronization of the software layer, the hardware devices, and the transmission standards of the respective grid operators. Hereby, her co-innovation partners are the software vendors and the grid operator. The relationship with the grid operator is characterized by a high level of regulatory and technological interdependencies, the relation with the software provider by technological and collaborative interdependencies.

The findings reveal that the likelihood of the technological interdependency risk in setting up VPPs is high. Moreover, the outreach of this risk is systemic, as it affects the entire ecosystem. Without a functional software layer not even, a single power supplier — and thus, not even a single customer — can be connected. In this case, the regulation will not qualify the business model for going online on the grid. Also, the supplier of hardware boxes that allow communication and control of the decentral power plants will be affected, as its hardware is highly interdependent with the software.

For the management of this sort of risk the most suitable approach is *to "help them do it"*. Even if the VPP operator is not directly responsible for managing the risk, it must collaborate with the partners (e.g. cross-organizational teams) to support them mitigating the risk as the performance of all other partners (e.g. customers, power suppliers, software startups) relies on the prevention of this hazard.

## 4.3.8. Conclusion

In sum, this paper reveals that digital business model designs do not only give birth to new business opportunities, but they also give rise to serious new risks. These risks result particularly from manifold interdependencies between the multiple partners who co-innovate the business model. Therefore, executives must identify, assess, and manage these risks in a strategic manner. To make digitally enabled ecosystems both profitable and sustainable, risk management calls for new strategies that transcend the boundaries of a single firm and build on collaboration between interdependent actors for the creation of mutual value. By applying such collaborative approaches to risk management, firms can strengthen the relationships with key partners and gain the ability to manage the complexity of co-innovation in setting up digital business models. Collaborative risk management thus must become an essential part of the new approach to the risk management in technology-driven industries

# Chapter III

## Solution I: Crowd-based Decisional Guidance Prologue

# 5. Solution I: Crowd-based Decisional Guidance Design Paradigms and Design Principles

## Purpose and Findings

The purpose of Chapter III is to examine the cognitive rational of the relevance of integrating stakeholders in entrepreneurial decision-making as well as its role under uncertainty. Therefore, I conceptually explore the cognitive limitations of current forms of decisional guidance and develop the idea of collective intelligence as a design paradigm for offering superior forms of guidance. I then suggest crowdsourcing as a mechanism to access such collective intelligence of the ecosystem and show how crowdsourcing can be applied to guide entrepreneurs (Section 5.1).

Based on these arguments, I highlight limitations in the previous design of crowdsourcing mechanisms and derive requirements for leveraging crowdsourcing as a design paradigm for decisional guidance in the context of entrepreneurial decision-making (Section 5.2).

In Sections 5.3 and 5.4 I propose design principles for CBMV systems and develop several IT artefacts that are required for the design of crowd-based DSS that provides guidance for entrepreneurs under uncertainty and risk.

## Relevance for Dissertation

The findings of this chapter first identify the cognitive constraints of decisional guidance in entrepreneurship and explores the cognitive rational for applying collective intelligence and crowdsourcing as design paradigm. The limitations of previous crowdsourcing mechanisms provide a first step towards developing design principles that are then further used for the design of hybrid intelligence as design paradigm for decisional guidance and the development of the HI-DSS in Section 6.5.

## 5.1. The Application of Crowdsourcing for Guiding Entrepreneurial Decisions

The findings of this chapter were previously presented at the AOM Annual Meeting (Dominik Dellermann et al. 2017) and accepted for publication at the International Journal of Entrepreneurial Venturing (IJEV). This study conceptually develops the idea of integrating the ecosystem in entrepreneurial decision-making through the mechanism of crowdsourcing. Moreover, I examine the cognitive rational of integrating collective intelligence in guiding entrepreneurial decision-making.

### 5.1.1. Introduction

In the era of digital economy, IT is becoming the enabler of novel products, serv might reduce an entrepreneur's chances to receive reasonable feedback and persuade a reasonable number of stakeholders of the viability of the opportunity to gain access to further valuable resources that support the entrepreneur in enacting the opportunity (Alvarez et al. 2013). Furthermore, the demand-side knowledge of potential customers is frequently not accessible (Nambisan and Zahra 2016).

One solution is the use of collective intelligence. This approach enables socially constructed co-creation by providing scalability, diversity, and flexibility beyond the boundaries of an entrepreneur's social network (Jeppesen and Lakhani 2010; Leimeister et al. 2009). While current literature seems to suggest that crowdsourcing as a concrete mechanism for accessing collective intelligence is a powerful tool to discover innovative ideas, I argue that crowdsourcing can also be applied to entrepreneurial opportunity creation. Thus, crowdsourcing might serve entrepreneurs in co-creating opportunities with potential market stakeholders and observing how consumers respond to their actions as well as giving them more flexible access to human resources or financial support (Mollick and Robb 2016). I propose accessing collective intelligence through crowdsourcing as a suitable mechanism to opportunity creation by providing access to social resources, reducing uncertainty about the objective value of an opportunity, and ensuring iterative development, learning, and resource support.

The contribution of my work is threefold. First, I contribute to research on how opportunities emerge from interactions between entrepreneurs and their social environment (e.g. Alvarez and Barney 2007; Alvarez et al. 2013; Tocher et al. 2015) and on the cognitive perspective of opportunity creation and enactment (Grégoire et al. 2011) by highlighting the role of leveraging external heterogeneous social resources in objectifying and enacting an opportunity. I therefore, provide a theoretical rational for why and how crowdsourcing can accelerate entrepreneurial processes. Second, I contribute to the emerging literature stream of digital entrepreneurship (e.g. Nambisan 2017) by showing how the affordances of digital infrastructures make the boundaries and agency of entrepreneurial processes less bound. I therefore introduce a new sub field for digital entrepreneurship research, which requires further consideration due to its enormous potential: crowdsourcing for supporting entrepreneurship by expanding the scope of collective intelligence to entrepreneurship research. Finally, my research contributes to the literature stream of crowdsourcing by emphasizing the potential role of a crowd in entrepreneurship. I put a constructivism lens on the entrepreneurial process by proposing that crowdsourcing cannot only be the source of creative ideas, as in previous studies, but also serve the purpose of sense making between the entrepreneur and the social environment to further develop and construct opportunities via social interaction.

## 5.1.2. A Creation Perspective on Entrepreneurship

The opportunity construct is one of the most pivotal concepts in the field of entrepreneurship (Davidsson 2015; McMullen and Shepherd 2006). In general, an opportunity is defined as a desirable future situation (Stevenson and Jarillo, 1990). Researchers in the academic field of entrepreneurship, however, have different notions of the nature of such opportunities. The literature distinguishes between two perspectives, the discovery view (Shane and Venkataraman 2000) and the creation view (Alvarez et al. 2013; Alvarez and Barney 2007; Alvarez et al. 2014) on opportunities.

The discovery perspective uses a critically realistic view to perceive opportunities as objective and formed by exogenous shocks to existing markets and industries (Shane and Venkataraman 2000). Opportunities are therefore discovered by the alert entrepreneur who aims at creating wealth (Kirzner 1979; Kauppinen and Puhakka 2010). From such a perspective, decision-making is risky. This means that both, outcomes, and their probabilities can be derived from the information that objectively exists in the environment, for example through customer surveys (Alvarez et al. 2014). For instance, approaches such as idea sourcing (e.g. Leimeister et al. 2009) help the entrepreneur in revealing an opportunity that is "waiting to be recognized" and tools such as customer surveys support the assessment of the probability of an opportunity's success. Outside actors and the environment therefore function as a source for novel and creative ideas.

On the other hand, opportunity creation theory (OCT) (Alvarez and Barney 2007/2010; Alvarez et al. 2013) applies an evolutionary realism lens and is based on the view that reality is socially constructed (Weick 1993). This perspective implies that opportunities are not existing independently of the entrepreneur but emerge from the iterative actions undertaken to create novel ways to achieve wealth (Sarasvathy 2001). Market disruptions are therefore not caused by exogenous changes but created endogenously by the actions of entrepreneurs (Wood and McKinley 2010). The opportunity creation perspective proposes that entrepreneurs should follow multiple and iterative developmental stages to fully enact an opportunity (Haynie et al. 2009). First, during the opportunity objectification stage, entrepreneurs start a sense-making process to validate the viability of their conceptualized idea by gaining feedback (Wood and McKinley 2010). Second, in the opportunity enactment stage, the entrepreneur builds stakeholder support by signaling the value of the opportunity and persuading the social environment of the value of the opportunity (Alvarez and Barney 2007; Tocher et al. 2015). Entrepreneurs create opportunities based on their individual beliefs and perceptions, imagination, and social interaction with the environment (Alvarez and Barney 2014). Contrary to the discovery view, the decision-making context is highly uncertain and requires incremental and intuitive decision-making as entrepreneurs create context-specific knowledge where none previously existed (Alvarez et al. 2013). The probability of future success is unknown as neither information about supply nor demand exists before the opportunity is enacted (Sarasvathy et al. 2003). Thus, opportunities are emerging as entrepreneurial actors wait for a response from their actions (e.g. testing it in the market) and then adjust their beliefs accordingly. Therefore, they are co-created by the entrepreneur, customers, and other stakeholders (Alvarez et al. 2013).

### 5.1.3. The Process of Entrepreneurial Decisions

The opportunity creation process thereby starts when an entrepreneur conceptualizes a potential future business idea based on individual social experiences and the formation of its cognitive evaluation of such reality (Wood and McKinley 2010).

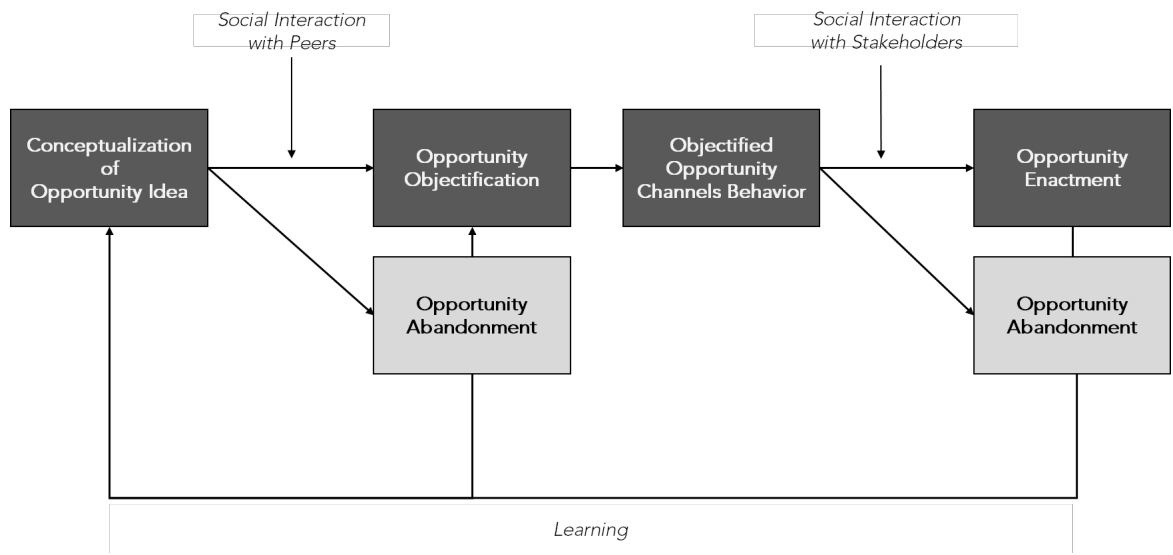

**Opportunity Creation Process**

After the entrepreneur imagined an opportunity idea, the objectification as an act of sense making starts by interacting with the social environment to verify initial beliefs (Weick 1993). In this early stage of opportunity creation, the entrepreneur is confronted with a high level of uncertainty regarding the prospect of a business idea. To reduce such uncertainty, the entrepreneur aims at validating her or her beliefs by enacting with the social environment. During this phase, the entrepreneur interacts with peers to test the viability of the idea and reduce uncertainty. Typically, entrepreneurial actors rely on peers such as friends, family members, or other contacts within their direct social network due to their instant availability. However, the value of feedback from peers that is provided in the process of sense making is highly dependent on their experience in this field, industry, or entrepreneurial practice in general (Dubini and Aldrich 1991). Feedback from peers may be both informal, for instance when provided in conversations, or formal by using meetings. Independent from the form of feedback, entrepreneurs attempt to create consensus within these social interactions to gather information about whether their initial idea and beliefs represent a real and viable opportunity. This process transforms an idea that was previously formed in the mind of the entrepreneur into an objectified opportunity or abandons it if consensus cannot be achieved. Thereby, the objectification of an opportunity reduces an entrepreneur's perceived uncertainty (Wood and McKinley 2010).

Once the opportunity is objectified, the entrepreneurs minimize their individual uncertainty regarding the value of the business idea. Therefore, their beliefs and actions will become guided by the opportunity idea. In the next step, the entrepreneur actively explores and leverages ways to capitalize on the opportunity (Alvarez et al. 2013). To this end, the potential founder needs to engage and gain solid traction among stakeholders. In doing so, they expand their scope beyond the directly related peer group and obtains access to further resources, for instance financial or human capital that allow them to fully exploit the envisioned opportunity and are critical for the opportunity creation process (Wood and McKinley 2010).

At the heart of opportunity enactment, an entrepreneur needs to create a shared understanding of the future idea among all involved stakeholders. This can take on several forms such as negotiations with investors, contacting employees, surveying potential customers, or searching for new technologies that might help to fulfil the opportunity. In this process, the entrepreneur needs to convince stakeholders of the potential idea, thereby increasing the odds of opportunity enactment (Alvarez and Barney 2010). Thus, the value of an opportunity can only be observed and understood after the entrepreneur has acted and thereby stimulated reaction from the market and gained validation from the market (Alvarez et al. 2013). In doing so, an entrepreneur can observe customer responses to products and services, which allows him to identify a divergence between an idea and actual customer perceptions and needs (Alvarez and Barney 2007). If an entrepreneur finds significant divergence, she may change the idea in a process of iterative actions and reactions until she receives a market fit or she might abandon the idea altogether. Furthermore, entrepreneurs draw on their social contacts to gain access to key resources such as potential employees or investors that support their idea. Such resources are crucial to fully turn the opportunity idea into a new venture (Wood and McKinley 2010).

### 5.1.4. Entrepreneurial Actions for Making Decisions

The central concept within the opportunity creation process is uncertainty. Uncertainty in this context regards the objective value of an idea, the needs of stakeholders, and the outcome of this iterative process (Alvarez et al. 2013). Contrary to risks, where decision makers can estimate the outcomes and the probability of such outcomes associated with a decision, uncertainty implies neither the outcomes associated with a decision nor their probability to occur (March and Zur Shapira 1987; Knight, Frank, H. 1921). Uncertainty has a dual role. For instance, the entrepreneur has only insufficient information about responses from the market or other stakeholders regarding a novel technology-based value proposition. On the other hand, stakeholders such as potential investors perceive uncertainty or doubts about the actual value of the idea (McMullen and Shepherd 2006). For a successful opportunity creation process, entrepreneurs should reduce both their individual uncertainty to objectify an opportunity and the uncertainty of their stakeholders to further develop the initial idea and get potential stakeholders on board. Reducing the uncertainty of the environment enables the creation of a potential market as well as provides access to further resources, for instance human capital or investments (Haynie et al. 2009).

OCT indicates three central concepts to reduce uncertainty in the entrepreneurial process: social interaction, iterative development, and learning. The starting point of each creation process is uncertainty. Therefore, entrepreneurs use social interaction with their peers, customers, and other stakeholders to reduce such uncertainty by gathering feedback. The uncertainty about their opportunity is reduced until opportunities can be objectified (Wood and McKinley 2010). Next, these social interactions lead to iterative changes in the beliefs and mental models about the initial opportunity and finally enable the entrepreneur to create wealth. Therefore, the opportunity emerges, and ideas, products, or entire business models are continuously reassessed, pivoted, or even abandoned (Ojala 2016). Creation theory assumes that the entrepreneur should rely on experiments, gathering feedback, remaining flexible, and learning rather than focus on pre-existing knowledge (Mintzberg, 1994). In the context of the opportunity creation process, learning from feedback and the iterative development of the idea is more important than strategic planning. Therefore, tacit learning in a path-dependent process becomes the major source of competitive advantages for entrepreneurs (Alvarez and Barney 2007).

To leverage these approaches to reducing uncertainty, entrepreneurs engage in several entrepreneurial actions (Wood and McKinley 2010). During the opportunity objectification stage, entrepreneurs focus on their individual sense making of the viability of their idea and the iterative development based on the responses from their actions, usually from the market. In the next step, entrepreneurs persuade interested stakeholders of the viability of their idea and mobilize resources to enact an opportunity (Alvarez and Barney 2007).

Previous research in the context of opportunity creation emphasized the value of an entrepreneur's social resources to objectify and enact an opportunity (Wood and McKinley 2010). The actions of an entrepreneur are therefore heavily influenced by the creativity and judgment gathered through social interaction (Foss et al. 2008).

## 5.1.5. Limitations of Entrepreneurial Decision-Making

Leveraging the entrepreneur's individual social capital to fully exploit the value of social interaction has its limits for several reasons that can be explained through cognitive bounds and cognitive constraints. Most obviously, a lack of social capital or competence, which previous research proposes as crucial resource for opportunity creation (e.g. Tocher et al. 2015), is therefore a common threat for entrepreneurs that prevents them from making sense of the viability of an opportunity or gaining access to external resources that are required to transform an opportunity idea into a novel venture. Following this logic, it would be impossible to objectify and enact an opportunity if entrepreneurs lack social competence or capital.

Second, if entrepreneurs explain their ideas to their related peers and ask for feedback on the value of the opportunity, they will face several cognitive traps. For instance, the entrepreneurs might encounter a self-selection bias by choosing peers that are very likely to support their thoughts and beliefs. Moreover, direct associates will more likely tend to overestimate the viability of an opportunity and therefore lead to biased results of the feedback process (Burmeister and Schade 2007). This fact can create a misleading sense of security that might result in the threat of wrong market moves (Lechner et al. 2006). On the other hand, closely related stakeholders might also face severe biases in the phase of enactment. Previous studies showed that, for instance, venture capitalists tend to evaluate start-ups with a high level of similarity regarding their industry, educational background, or personal characteristics more favourably, bias can potentially lead to disastrous funding decisions (Franke et al. 2008, 2006).

Third, during the objectification process, entrepreneurs need access to experienced experts who are also capable of further evaluating and developing initial ideas (Foss et al. 2008). Therefore, an entrepreneur needs social ties to experts who support the process of confirming if a conceptualized idea is viable to adopt it to a potentially viable idea or even completely reject the opportunity (Wood and McKinley 2010). The major constraint that entrepreneurs face here is the fact that they frequently only have social capital. Moreover, the peers within their direct networks might not necessarily be experts in the required field. For instance, they might not have enough business knowledge, technological expertise, or simply not enough domain experience. This problem is particularly important if the entrepreneurs attempt to converge industry boundaries with their ideas and therefore require experts from various branches (Tocher et al. 2015). Without access to such social resources, an entrepreneur has only little chances to reduce uncertainty and finally objectify the idea (Haynie et al. 2009). However, even if they have access to a small network of social contacts, they might face representativeness bias by relying on and generalizing from small samples rather than comprehensively surveying a huge number of experts (Fischhoff et al. 1977). Limited access to social resources can further have crucial effects on the success of the opportunity enactment as entrepreneurs tend to recruit employees or obtain funding from their individual social network (Mikkola and Gassmann 2003; Hsu 2004).

Fourth and directly related to this fact is the problem of strong ties in the entrepreneur's network, which might lead to a limited heterogeneity of knowledge (Burt 2004; Granovetter 1985). To successfully enact an opportunity, the deep prior experience within one field needs to be balanced with heterogeneous knowledge and insights to enable valuable feedback and learning (Alvarez et al. 2013; Weick 1993). In creating opportunities, closely relying on knowledge and experts from directly related industries or markets may make it difficult to gather valuable feedback. For instance, novel ideas that diminish traditional industry boundaries or disrupt markets require evaluation/information from heterogeneous sources and therefore social interaction with experts from various fields (DiMaggio 2012). However, previous research provides strong evidence that entrepreneurs tend toward interacting with contacts from closed networks that often provide only little additional information to the entrepreneur's beliefs during the objectification of an idea (Ruef et al. 2003). Therefore, information about customers' needs and desires is frequently not accessible as well (Nambisan and Zahra 2016).

Finally, the flexibility of required resources represents a certain issue in the creation context of fully enacting an opportunity (Alvarez and Barney 2007). Such a flexibility of resources is particularly manifested in human resource practices and financing. First, entrepreneurs frequently do not know which skills they finally require for exploiting their opportunity as the outcome of the process is highly blind or myopic (Campbell 1960). Therefore, hiring individuals becomes challenging as the requirements can expand or change in a short time exceeding the human capital of employees (Alvarez and Barney, 2007). Second, entrepreneurs must obtain financial resources to realize an opportunity. However, the context of creating opportunities is highly uncertain due to the lack of information. Therefore, it is difficult to explain the nature and value of the opportunity that is being exploited to traditional sources of capital such as banks and venture capital firms (Bhide 1992). Consequently, using peers and potential stakeholders within an entrepreneur's social network might be insufficient in providing the required flexibility of resources for creating an opportunity.

Therefore, this approach provides only limited support for reducing uncertainty and socially constructing an idea during opportunity objectification and enactment. Lacking proper social resources during the opportunity creation process therefore represents the major reason many entrepreneurial efforts fail (Tocher et al. 2015).

| Phase | Entrepreneurial Action | Limitations of Previous Approaches |
|---|---|---|
| **Objectification** | *Sense making* | ▪ Limited social resources<br>▪ Biased social resources<br>▪ Homogeneity of social resources |
| | *Iterative development* | ▪ Lack of demand-side knowledge |
| **Enactment** | *Persuasion of stakeholder* | ▪ Limited social resources |
| | *Resource mobilisation* | ▪ Limited access to external resources<br>▪ Lack of flexibility |

*Limitations of Previous Approaches in Opportunity Creation*

To overcome the limitations of previous approaches the concept of collective intelligence and the mechanism of IT-mediated crowdsourcing offers tremendous possibilities to enable interaction with potential customers, experts, and other stakeholders by minimizing transaction costs and providing broad access to heterogeneous social resources, thus, reducing cognitive constraints and bounded rationality. I therefore propose that crowdsourcing, which has proven to be a valuable concept in other contexts, is a valuable approach for entrepreneurs to reduce uncertainty and interact with their social environment as it provides access to heterogeneous knowledge from diverse sources and flexible resources.

## 5.1.6. Previous Work on Crowdsourcing

One special instantiation of interacting with a firm's environment during the process of developing new products and services is crowdsourcing (Priem 2007). Crowds define an anonymous group of individuals, ideally domain experts or customers, which can provide dispersed knowledge (e.g. demand-side knowledge, supply-side knowledge etc.). Crowdsourcing was originally considered as a new form of organizing work and denotes the act of taking a task once performed inside an organization and broadcasting it via an open call to individuals outside the organization (Howe 2008). More recent research suggests that crowdsourcing contains much more than outsourcing single tasks to a broad and unknown group of people. Thus, on a broader level crowdsourcing can be considered as a mechanism that allows a firm to attain previously unattainable resources to build a competitive advantage. Since this notion is based on the theoretical considerations of the resourced based view, the crowd is viewed as a valuable resource so called crowd capital, that must be leveraged for resource creation purposes. For an entrepreneur to be able to efficiently utilize this crowd capital she must develop so called crowd capabilities. These capabilities include developing an adequate understanding about the contributions she is seeking (e.g. money, ideas etc.), the IT structure best suited to obtain these contributions (episodic vs. collaborative) as well as strategies on how to process/evaluate these contributions. Depending on these considerations, an entrepreneur can make use of different types of crowdsourcing such as —crowd voting, micro-task, idea, and solution crowdsourcing (Prpic and Shukla 2013, 2014).

The value of crowdsourcing as a mechanism can be explained through gaining access to collective intelligence. Such collective intelligence underlies two basic principles: error reduction and resource/knowledge aggregation (Mannes et al. 2012; Mannes et al. 2014). Error reduction is achieved as although an individual decision maker might be prone to biases and errors (such as individual entrepreneurs or mentors in my context), the principle of statistical aggregation minimizes such errors by combining multiple perspectives (Armstrong 2001). Second, resource aggregation describes the diversity of external resources that can be integrated. For instance, supply-side knowledge that can be aggregated by combining multiple decision makers and enables a user to capture a fuller understanding of a certain context (Keuschnigg and Ganser 2016). Moreover, demand-side (i.e. market) knowledge can be accessed and is most frequently applied for user innovation (Priem 2007; Soukhoroukova et al. 2012).

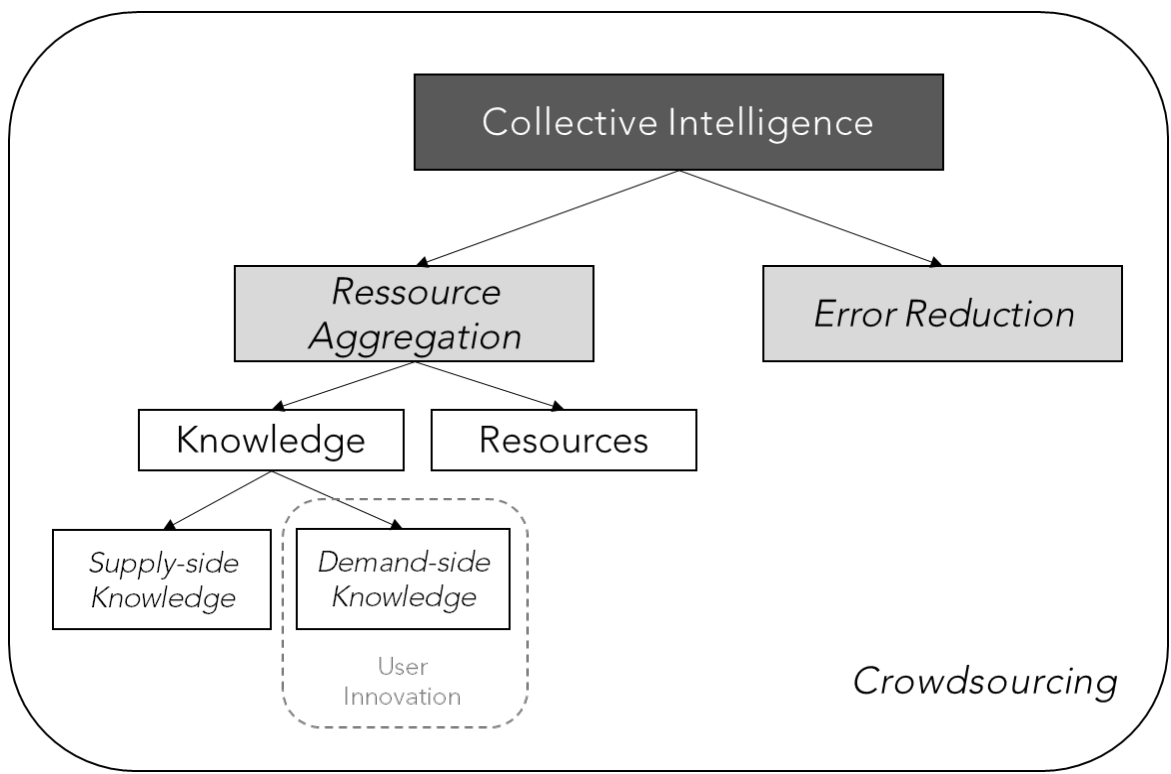

*Relation of Crowdsourcing Concepts for this Study*

Firms that apply crowdsourcing benefit from the heterogeneous and diverse crowd, which can provide the ability to discover creative solutions or solve problems. Interaction with a crowd enables firms to discover novel customer requirements and user input for ideas, thereby representing a "voice of the customer" (Dahan and Hauser 2002; Griffin and Hauser 1993). Therefore, crowdsourcing provides both need-based information (i.e. what is the problem?) as well as solution-based information that guides companies in finding out what a potential new product or service should do (Afuah and Tucci 2003, 2012; Terwiesch and Ulrich 2009). On the other hand, a crowd can be used to gain access to external resources, such as human capital, to recruit freelancers with a specific expertise (e.g. expertise in PHP or Java) to fulfil a certain job (e.g. programming a webpage) or to finance products, investment projects, or entire companies (Mollick 2014).

However, research so far has dealt with crowdsourcing at a surface level. Thus, crowdsourcing has been considered as a tool for ideation, commenting, and voting. One aspect that has been frequently ignored is that crowdsourcing can also be used as an idea development tool that can effectively support idea evolution through various stages of entrepreneurial maturity (i.e. from idea to prototypes and business models) by offering collective intelligence. I argue that applying crowdsourcing to entrepreneurial challenges requires an entirely different perspective that can do more than just helping companies with problems at the fuzzy front end of innovation (i.e. for example ideation). Therefore, a crowd should ideally consist of domain experts and customers that possess the required demand and supply-side knowledge.

### 5.1.7. Crowdsourcing for Entrepreneurial Actions

Based on this argumentation, I build on the process model of opportunity creation by Wood and McKinley (2010). This process includes the stages opportunity conceptualization, opportunity objectification, and opportunity enactment. During the conceptualization of an opportunity, entrepreneurs rely on their individual beliefs and experiences (Wood and McKinley 2010). The term "search" as applied in idea communities has little or no meaning in OCT as the agency of an individual entrepreneur is essential during this phase (Alvarez and Barney 2007). This contrasts with a discovery view on crowdsourcing that might obviously leverage a crowd for discovering novel ideas (e.g. Afuah and Tucci 2012).

Building on previous work on the role of social resources in this process (e.g. Tocher et al. 2015), I argue that crowdsourcing facilitates opportunity objectification by providing entrepreneurs with social resources to engage in a sense-making process. I show that such heterogeneous feedback provides several benefits compared to the knowledge of peers and facilitates the iterative development of an opportunity. After the opportunity is objectified, I argue that crowdsourcing supports the opportunity enactment by signaling the market viability of an opportunity, therefore reducing stakeholders' opportunity-related uncertainty, which arises in the consensus-building stage. Finally, I posit that crowdsourcing facilitates extended access to resources such as human capital or funding to fully enact an opportunity.

## Sense Making

The objectification of an opportunity is a sense making process through which entrepreneurs validate the value of an imagined business idea by interacting with knowledgeable peers (Tocher et al. 2015; Wood and McKinley 2010). This sense making activity supports an entrepreneur in developing an initially vague idea into an articulable vision (Weick 1993). Therefore, social interaction provides feedback from peers that might either confirm the viability of the idea, help the entrepreneur to adapt it, or even reject the envisioned opportunity (Alvarez et al. 2013; Ojala 2016). Thus, the opportunity objectification process highly depends on an entrepreneur's access to social resources such as a group of experienced peers who provide feedback (Foss et al. 2008). If access to such social resources is missing, entrepreneurs lack criticism and advice from their social environment that would reduce individual uncertainty and objectify an opportunity (Haynie et al. 2009). The limited social capital or social competence of entrepreneurs thus makes it difficult to objectify an opportunity idea (Tocher et al. 2015). In this case, using a crowd provides several benefits for entrepreneurs. Crowdsourcing provides access to social resources through scalable IT infrastructures such as platforms (Howe 2008). Therefore, using crowdsourcing for the sense making process of objectification offers a cost-efficient and rapid way to gain access to the social environment while the anonymity of a crowd supports the entrepreneurs in enlarging their social contacts without any high demand for the ability to effectively interact with others (Baron and Markman 2003). Crowdsourcing enables entrepreneurs to test assumptions about their idea with a potential market and therefore enables gathering feedback on the viability of an opportunity (Poetz and Schreier 2012). In this context, crowdsourcing platforms (e.g. Amazon Mechanical Turk) can be leveraged to extend an entrepreneur's social network, which can be used for sense making and idea objectification. Thus, I argue:

**Proposition 1:** Crowdsourcing provides access to social resources to engage in sense making, thus enhancing opportunity objectification.

To objectify their opportunity idea during the sense making process, entrepreneurs strongly rely on directly related peers (Ruef et al. 2003; Stam and Elfring 2008). Previous research shows that entrepreneurs tend to interact with social networks consisting of bonding ties to family members or friends (Tocher et al. 2015). Trust and common norms within such closed networks frequently lead to biased decision-making (Carolis and Saparito 2006). Furthermore, feedback by such homogenous networks provides only limited additional insights that help entrepreneurs to validate their assumptions (Lechner et al. 2006). Therefore, entrepreneurs need access to so-called bridging ties (Putnam 2001), which provide heterogeneous knowledge and feedback (Tocher et al. 2015). In this context, crowdsourcing provides a suitable way to bridge the interface of an entrepreneur's existing social network for accessing heterogeneous valuable knowledge (Howe 2008; Leimeister et al. 2009). Integrating a heterogeneous crowd into the entrepreneurship process therefore provides access to social resources that are characterized by both strong heterogeneity and anonymity. This enables the entrepreneurs to create their opportunity by using the "wisdom of crowds" and related benefits (Surowiecki 2004). The access to such social resources via crowdsourcing provides potential support for the opportunity creation process by enabling evaluation and feedback from potential customers and other stakeholders, therefore reducing uncertainty. Previous research in the field of new product development shows users' appropriateness as "raters" for new product and service ideas (Magnusson et al. 2016; Magnusson 2009). Therefore, one highly important benefit of crowdsourcing is a crowd's ability to provide both user needs (i.e. demand-side knowledge) and product trends (i.e. supply-side knowledge) (Ozer 2009), which is central for opportunity creation (Nambisan and Zahra 2016).

Therefore, a crowd does not only provide access to social resources that are not limited to the entrepreneur's peers but also to the benefits of the heterogeneity of a crowd's knowledge (Jeppesen and Lakhani 2010). Apart from access to further social resources that extend the entrepreneur's social network, using a heterogeneous and anonymous crowd instead of peers provides further valuable benefits for the opportunity creation process (Poetz and Schreier 2012). A crowd is more suitable in preventing self-selection biases as their anonymity ensures more valid and objective feedback on the opportunity idea compared to peers, individual experts, or start-up consultants within closed social networks. The feedback of the heterogeneous crowd represents the "voice" of a potential market and therefore results in a higher level of validity that reduces the threat of an entrepreneur's overestimation of the value of an idea (Magnusson et al. 2016). Approaches such as crowd voting or crowd testing on IT platforms enable entrepreneurs to gather feedback on the viability of an idea by leveraging bridging ties. While individual entrepreneurs have limited social capital, leveraging IT platforms allows interaction with reduced transactions costs and enlarge oneś network. Furthermore, social resources from a crowd support an entrepreneur's sense-making process by reducing representativeness biases (Burmeister and Schade 2007). By challenging their assumptions and beliefs with potential users, the entrepreneurs gather information about the value of their opportunity, the demand-side, and the level of the product-market fit, therefore reducing their individual uncertainty. As crowdsourcing enables the entrepreneurs to use feedback from a huge number of people, the threat that they must generalize and make decisions based on small samples is minimized. Thus, the access to social resources from a heterogeneous and anonymous crowd has tremendous potential to support the opportunity creation process, thus enhancing opportunity objectification. Therefore, I argue:

**Proposition 2:** Crowdsourcing provides access to heterogenous and diverse knowledge to engage in sense making, thus enhancing opportunity objectification.

## Iterative Development

Apart from validating an entrepreneur's individual beliefs, the objectification process requires the continuous modification of the opportunity idea until consensus on the viability of an opportunity is achieved (Alvarez et al. 2013). To obtain consensus among the social environment, entrepreneurs must adapt their initial opportunity idea based on the feedback gathered from social interactions (Dimov 2011; Haynie et al. 2009). Although the entrepreneur might be confident that the idea is valuable, this belief might not be shared by the social environment, leading to iterative changes and adaptions of the idea based on the input from the social environment. This iterative and evolving process of idea feedback, adjustment, and rejection or adoption continues until consensus is achieved and the opportunity idea is finally objectified during complex social interactions between entrepreneurs and their social environment (Dimov 2011; Wood and McKinley 2010). Thereby, crowdsourcing provides support for further developing an idea by testing the opportunity idea in the market and iteratively co-creating it with a crowd. Testing allows the entrepreneurs to gather information about the *"voice of the customer"* (e.g. Dahan and Hauser 2002; Griffin and Hauser 1993). In the context of opportunity creation, it provides a rapid and cost-efficient way to aggregate data about the reactions of the market, feedback of functionality, or the customers' perception of a solution. This allows the entrepreneurs to integrate feedback and further develop an initial idea or prototype to fully enact the opportunity (Breland et al. 2007). Moreover, entrepreneurs might actively engage a crowd of potential stakeholders in the co-creation process of the opportunity. Contrary to previous approaches that focus on the initiation of innovation efforts by a crowd and a linear flow back to the firm (e.g. Leimeister et al. 2009), the starting point of crowdsourcing for entrepreneurship lies with the entrepreneur and leads to an iterative exchange to further develop the opportunity together with a crowd. Interaction with potential stakeholders might relate to solution-based information and enable the entrepreneur to understand what a potential new product or service should do and therefore complements the entrepreneur's technological knowledge (Hippel 2005). For the context of entrepreneurship, it is particularly important that such co-creation does not only lead to creative solutions but to achieving a high level of market fit and viability to create a successful new venture. Thus, during the selection of a crowd for feedback and co-creation, the entrepreneur should balance expert knowledge that can assess the feasibility of an opportunity and supply-side knowledge to provide a high level of customer benefits (Poetz and Schreier 2012). Unlike traditional crowdsourcing efforts that discover novel product ideas, the testing and co-creation of an opportunity is not limited to early-stage ideas or even products. Rather, the development of an opportunity includes different stages ranging from an initial idea to a minimum viable product, a business model, and finally a novel venture (Ojala 2016). Thus, the integration of crowdsourcing is required during various phases of this process. In this context, previous research showed that apart from the co-creation potential for ideas or prototypes, a crowd is also capable of designing and developing novel business models (Ebel et al. 2016). Consequently, crowdsourcing is a valuable approach to validate, co-create, and iteratively develop an opportunity through multiple stages of the process. I thus assume:

**Proposition 3:** Crowdsourcing facilitates an entrepreneurs' ability to maintain continuous dialogue with social resources to co-create the opportunity through multiple iterations of development.

## Stakeholder Persuasion

Once an entrepreneur's opportunity-related uncertainty is reduced during the sense-making process, the entrepreneur will start to obtain resources that are required to enact the opportunity (e.g. Alvarez and Barney 2007; Tocher et al. 2015). During the opportunity enactment, the entrepreneur aims at building stakeholder support to transform the objectified business idea into a new venture (Wood and McKinley 2010). Therefore, it is important for the entrepreneur to reduce the uncertainty of potential stakeholders (e.g. customers, investors, suppliers) regarding the viability of the opportunity to gain solid traction and achieve resource commitment (Im Jawahar and McLaughlin 2001). To reduce such stakeholder uncertainty, entrepreneurs share the knowledge about their opportunity, signal the value of their proposed idea, and create a shared understanding among their environment (Alvarez and Barney 2014; Alvarez et al. 2013). In this context, leveraging a crowd provides several benefits to enact an opportunity. Apart from reducing the entrepreneurs' uncertainty about the value of their beliefs, crowdsourcing enables minimizing potential stakeholders' uncertainty and persuades them of the true value of an opportunity idea. The feedback of a crowd functions as a "voice" from the potential market (Dahan and Hauser 2002). Therefore, mechanisms such as crowd voting signal the responses and thoughts of potential customers and reduce the stakeholders' uncertainty if the idea is objectively valuable (Magnusson et al. 2016). In this context, crowdfunding has proven to be a common mechanism applied in the entrepreneurship context. From the perspective of OCT, however, crowdfunding grants benefits beyond access to financial resources (Lipusch et al. 2018). The funding behaviour of investors, in this case a crowd, may function as a gatekeeper that provides an early evaluation of the opportunity idea.

For instance, (Mollick and Nanda 2015) showed that the funding of democratic individuals is equal to the expert evaluation of ideas and therefore provides valuable insights to reduce stakeholders' uncertainty regarding an opportunity. Crowdsourcing supports the opportunity enactment process by reducing potential stakeholders' uncertainty about the viability of an opportunity, building a shared understanding within a potential market environment, and thus gaining traction among stakeholders by signaling the value of the opportunity. I therefore propose:

**Proposition 4:** Crowdsourcing supports the entrepreneur to reduce stakeholder uncertainty and persuade stakeholders to support by signaling the value of the opportunity.

### Resource Mobilization

As I noted before, opportunity enactment calls for social resources that are larger in size and more diverse than the peers who helped the entrepreneur objectify the idea (Wood and McKinley 2010). Thus, for an entrepreneur to successfully realize an idea, access to a wide and varying base of actors is of crucial importance (Tocher et al. 2012).

In a creation context, entrepreneurs are often confronted with a very dynamic environment that makes the exploitation of opportunities difficult to accomplish. For example, due to the high uncertainty inherent to opportunities, entrepreneurs often need to adapt their product and service offers at short notice in line with dynamic market developments. This presents entrepreneurs with the challenge to hire certain employees flexibly and for short periods of time. Because of this, entrepreneurs need new organizations of work that allow them to hire people with a special expertise periodically and flexibly. This becomes even more important if ventures face monetary constraints, as it is typical for start-ups. One way to address entrepreneurs' needs for more flexible and short-term employment relationships is crowd work (Durward et al. 2016). Thus, platforms such as Freelancer allow to look for a whole variety of skills without incurring the high costs that are associated with the rigidity of long-term employment relationships.

Similarly, dynamic environments, as usually encountered during opportunity creation, are usually associated with high risks. However, in such situations of high risks, traditional external sources of capital—including banks and venture capital firms—are unlikely to provide financing for entrepreneurs (Bhide 1991). Under these conditions, the problem of finding sources of capital is not information asymmetries, it is simply the lack of information. Thus, entrepreneurs in such situations are not capable of reliably presenting economic facts, such as the risks associated with an opportunity, which are required by external capital providers to assess the viability of a new business and therefore a start-up's probability to repay its debts.

One way to address these capital shortages that entrepreneurs inadvertently face is crowdfunding. Crowdfunding is thereby a very versatile tool that through the distributed collection of small sums among many funders can amount to large investment sums granted to the entrepreneur. Even more, crowdfunding is a form of financing that is characterized by a low degree of informational requirements, which makes financing accessible even to entrepreneurs who undergo opportunity creation. In addition to that, crowdfunding provides several other advantages. Thus, it can be used as a method of market research to validate consumer demands as well as a method to gather valuable user feedback to align the product with existing market demands (Lipusch et al. 2018).

**Proposition 5:** Crowdsourcing facilitates access to a diverse base of prospective stakeholders to mobilize resources.

Crowdsourcing provides a rapid and cost-efficient way to aggregate data about the reactions of the market, feedback of functionality, or the customers' perception of a solution (Ries 2011; Blank 2013). Thereby, crowdsourcing helps the entrepreneur overcome limitations such as limited access to or homogeneity of social resources. Furthermore, crowdsourcing provides valuable potentials for the iterative development of the opportunity by offering access to flexible resources (see Table 7).

| Phase | Entrepreneurial Action | Benefits of Crowdsourcing |
|---|---|---|
| Objectification | *Sense making* | <ul><li>Heterogenous knowledge</li><li>Demand-side knowledge</li><li>Representativeness of feedback</li></ul> |
| Objectification | *Iterative development* | <ul><li>Demand-side knowledge</li><li>Supply-side knowledge</li><li>Rapid user feedback</li></ul> |
| Enactment | *Persuasion of stakeholder* | <ul><li>Signaling of market viability</li><li>Integration of stakeholders</li></ul> |
| Enactment | *Resource mobilisation* | <ul><li>Access to human resources</li><li>Access to flexible financing</li><li>Supply-side knowledge</li></ul> |

*Advantages of Crowdsourcing for Opportunity Creation*

## 5.1.8. Directions for Further Research

My discussion shows that crowdsourcing and the opportunity creation perspective in entrepreneurship research can be aligned to support the continuing dialogue between entrepreneurs and their social environment to engage in an interactive learning process, thus objectifying and enacting an opportunity in a co-creative manner (Alvarez et al. 2013). The ideas advanced in this article highlight several interesting RQs for interdisciplinary fields of research in entrepreneurship.

**Research Theme 1:** *The Role of External Knowledge and Resources in Strategic Entrepreneurship*

One related topic in the field of strategic entrepreneurship is the potential role external resources might play in supporting and enabling entrepreneurial efforts. Recent technological trends such as the emergence of digital platforms require open approaches for entrepreneurs to create opportunities (Nambisan and Zahra 2016; Nambisan 2017). Thus, a more co-creative and open orientation of entrepreneurship is required (Alvarez et al. 2015). Therefore, my argumentation about the role of crowdsourcing for opportunity creation reveals two interesting directions for further research: mobilizing and leveraging external resources and the role of customer centrism. First, previous research showed that resource constraints are one of the central issues of entrepreneurial failure (Shane 2003). Thus, research on open approaches in entrepreneurship might examine how entrepreneurs can leverage external resources (e.g. human resources, external knowledge) to start learning how to acquire all required resources through open innovation (Blank 2013). The second direction for further research lies in the need for enhanced customer centrism (e.g. Demil et al. 2015; Nambisan and Zahra 2016; Priem 2007). Thus, future research should focus on deepening the understanding of the role of customer integration into the early phases of entrepreneurial efforts.

**Research Theme 2**: *The Role of Entrepreneurial Agency in Crowdsourcing*

Another interesting field for further research is the role of a fluid entrepreneurial agency in less predefined and distributed entrepreneurial actions among a heterogeneous set of participants, as in the case of crowdsourcing (Nambisan 2017). The process of opportunity creation with a distributed entrepreneurial agency among the entrepreneur and a crowd might lead to emergent roles (Faraj et al.

2011). For instance, during different phases of the creation process, either the entrepreneur or a crowd might start sense making or adapting the opportunity idea based on feedback. Further research might therefore focus on the dynamics of entrepreneurial agencies in crowdsourcing for opportunity creation. Moreover, the nature of an opportunity and its relation to entrepreneurial agency are a valuable starting point for research. As previous literature on innovation in online communities showed, ideas can become disembodied (i.e. independent) from their authors and the context in which they were created (Faraj et al. 2011). This might lead to generative outcomes that may lie beyond the entrepreneur's control and aspiration, leading to debates on how to proceed or the right way to enact the opportunity. Thus, further research should examine the role of such tensions and how to deal with them by focusing on heterogeneous opinions and goals in distributed entrepreneurial agencies. Furthermore, a valuable pathway for entrepreneurship research is providing a deeper understanding of how entrepreneurial cognition and decision-making occurs when collectives are involved (Nambisan 2017). How do entrepreneurial agents select, assume quality, or integrate heterogeneous feedback from a crowd, especially when such feedback is diverging from an individual entrepreneur's assumptions?

**Research Theme 3:** *Participation Architectures for Crowdsourcing for Opportunity Creation*

The next theme for further research is the exploration and design of participation architectures that enable the iterative co-creation of an opportunity and the evolution of an entrepreneurial idea over time, as previous research focused on the generation and discovery of novel ideas than the evolutionary co-creation of an entrepreneurial opportunity (Faraj et al. 2011; Majchrzak and Malhotra 2013). Therefore, it is important to explore the participants, governance, and technological affordances of crowdsourcing for opportunity creation. Empirical examples such as the crowdfunding platform JumpStart

Fund or Quirky point towards such directions. First, further research should examine who participates in such entrepreneurial co-creation processes. Which new participant roles emerge when agents such as customers or investors (e.g. in crowdfunding) are empowered to be co-creators? And what are their intentions, motives, and goals to participate? Second, research on governance in crowdsourcing should explore what determines the nature and structure of the participation and contribution of collective entrepreneurial agents (Nambisan 2017). How can fluid boundaries of communities be managed? Or is traditional community management even relevant in such settings when entrepreneurial agents want to verify their assumptions and beliefs through potential customers? Moreover, it is important to research appropriate incentive mechanisms to share, for instance, future revenue or reward contributions. Finally, future research regarding this theme might explore the role of technological affordances such as experimentation, reviewability, and re-combinability of entrepreneurial opportunities. Thus, it is for example crucial to understand adequate mechanisms and tools to provide feedback. Furthermore, the optimal amount of feedback, the heterogeneity of knowledge, or the number of iterations is a promising field for future research efforts.

**Research Theme 4:** *Design Oriented Research on Crowdsourcing for Opportunity Co-Creation*

From a methodological perspective, interdisciplinary research on the topic of crowdsourcing for opportunity creation might consider design-oriented research approaches (e.g. Hevner et al., 2004; Peffers et al., 2007). This research paradigm has been studied well in the information systems literature and provides enormous potential for the field of entrepreneurship (Nambisan, 2016; Venkatraman et al., 2012). Design science offers a deeper understanding of the design process of an artefact (i.e. the technological solution for a problem) and provides the possibility to create systems that do not yet exist, such in the case of crowdsourcing for opportunity creation. Therefore, this approach proposes the possibility to design and build systems that enable the integration of crowdsourcing into the process of opportunity creation. Such possibilities might be, for instance, the development of tools to validate entrepreneurial assumptions and business models (Ries, 2011) or systems to enable online co-creation between entrepreneurs and a crowd. Thus, such research can inform practical orientation while maintaining theoretical rigor (Gregor and Hevner, 2013). Thus, the methodological approach of design science provides a valuable direction for further research (Dimov, 2015) that is particularly relevant for the topic of crowdsourcing for opportunity creation.

### 5.1.9. Conclusion

In this paper, I argue that integrating crowdsourcing into the entrepreneurial opportunity creation process provides access to heterogeneous social resources. Crowdsourcing therefore offers entrepreneurs the possibility to gain access to collective intelligence and leverage supply-side knowledge (Ozer 2009), demand-side knowledge about customer needs and desires (Nambisan and Zahra 2016), flexible external resources (Howe 2008), and reduce the cognitive bounds and constraints of individual decision makers. Entrepreneurs are thus able to objectify their opportunity idea by starting a sense-making process and iteratively developing their opportunity. Furthermore, entrepreneurial agents can enact the opportunity by applying crowdsourcing to persuade interested stakeholders and mobilize external resources.

The aim of my research is not to discriminate the discovery view of opportunities (e.g. Shane and Venkataraman 2000; Shane 2003) or the potential role crowdsourcing might play in identifying market imperfections through idea sourcing (e.g. Leimeister et al. 2009). Rather, I argue that the emergence of new digital infrastructures (Nambisan 2016) provides a promising approach to opening the boundaries of entrepreneurial processes and integrating the social environment into the iterative and evolutionary creation process of emergent opportunities (Garud and Karnoe 2003; Alvarez and Barney 2007; Alvarez et al. 2013). Therefore, my discussion shows applications of crowdsourcing during different stages as well as entrepreneurial actions to support the creation process and points toward interesting themes for further research.

My contribution is noteworthy for several reasons. First, I contribute to the discourse in entrepreneurship how opportunities emerge from the interactions between entrepreneurs and their social environment (e.g. Alvarez and Barney 2007; Alvarez et al. 2013; Tocher et al. 2015; Kauppinen and Puhakka 2010). I also contribute to the cognitive perspective of opportunity creation and enactment (e.g. Gregoire et al. 2011) by highlighting the role of leveraging external heterogeneous social resources in objectifying and enacting an opportunity. I, therefore, provide a theoretical rational for the value of collective intelligence in the cognitive processes of entrepreneurial agents. For this purpose, I show how crowdsourcing may overcome the cognitive constrains and bounds of previous approaches, such as interacting with peers to open the boundaries of entrepreneurs' existing social networks or integrating demand-side knowledge (e.g. Nambisan and Zahra 2016) into the creation of entrepreneurial opportunities and provide applications during different stages of the creation process.

Second, I introduce the topic of crowdsourcing for opportunity creation as a promising field for further research in the field of digital entrepreneurship (e.g. Nambisan 2016) and propose a research agenda that may guide future efforts. I particularly argue for interdisciplinary research that might include the fields of strategy, information system, as well as cognitive entrepreneurship scholars and suggest design-oriented research (e.g. Hevner et al. 2004) on crowdsourcing for opportunity creation. Such design-oriented research might be especially interesting to ensure the practical relevance of the entrepreneurship discourse.

Finally, I provide very practical applications of crowdsourcing in the creation process that may guide both institutions that support entrepreneurial talent (e.g. incubators, accelerators) and entrepreneurs themselves to use novel, customer-centric, and cost-efficient approaches across the interface to gather rapid feedback and flexible skills to create wealth.

## 5.2. The Requirements of Crowdsourcing for Guiding Entrepreneurial Decision-Making

The findings of this chapter were previously published as Dellermann et al. (2017). This study builds on the previous Section 5.1 and conceptually examines the limitations of the previous forms of crowdsourcing to support entrepreneurial decision-making. Furthermore, I suggest crowdsourcing for entrepreneurial opportunity creation as novel application of the mechanism and identify requirements for this approach, therefore, suggesting a research agenda on crowdsourcing in IS research.

### 5.2.1. Introduction

As I showed in the previous chapters, the consensus building among the entrepreneur and stakeholders, for instance potential customers or investors, leads to a common understanding of value of the proposed entrepreneurial venture (Alvarez et al. 2013).

One possible way to reduce these limitations can be found in the literature on collective intelligence and crowdsourcing. Research on crowdsourcing in the context of innovation extensively showed the potential of integrating the *"voice of the market"* by using collective intelligence for sourcing and evaluating novel ideas and customer co-creation (Blohm et al. 2016; Leimeister et al. 2009; Schlagwein and Bjørn-Andersen 2014) and provides evidence for the value of integrating the social resources of a heterogeneous crowd into different innovation activities.

While previous work on crowdsourcing in IS research has focused on discovering solutions for problems via distant search, and how to design web-based platforms and participation architectures for this context, the creation view on opportunities requires novel perspective on how crowdsourcing should be conducted. Previous research frequently focused on the generation and discovery of novel ideas than the evolution of an entrepreneurial opportunity. In this vein, collaboration among participants and feedback-based idea evolution remains minimal. Additionally, participation architectures are designed for incentivizing the post of novel ideas than co-creating and refining an existing idea (Majchrzak and Malhotra 2013). Furthermore, entrepreneurial opportunities do not merely include a single product but the development of an entire firm (Ojala 2016). This contrasts with conventional crowdsourcing efforts that consist of an open call for a modular, self-contained, and closed problem (Terwiesch and Xu 2008).

Given the unique characteristics of opportunity creation as an emergent and uncertain process of iterative development fostered by interaction with the market and the important role of the crowd in the context of innovation, I conjecture that crowdsourcing is suitable to support the entrepreneurial opportunity creation process. I argue that interaction with a heterogeneous crowd allows entrepreneurs to reduce uncertainty about the objective value of an opportunity and thereby promotes the iterative development of an idea and entrepreneurial learning. This is grounded in the general logic of collective intelligence that allows to aggregate knowledge while reducing individual biases (such as overestimation). Thus, I suggest crowdsourcing for entrepreneurial opportunity creation as a noteworthy field for further research to successfully develop participation architectures, i.e. the socio-technological framework that shapes interaction and exchange, which enable an integration of the crowd with the aim of supporting the evolution of an entrepreneurial opportunity.

I seek to make four main contributions to research and practice. First, I extend previous work on theories of entrepreneurial action by showing limitations of previous approaches in the opportunity creation process (Alvarez and Barney 2007; Venkataraman 2003). Second, I expand research of crowdsourcing to the field of entrepreneurship by extending the principles of crowdsourcing for innovation for entrepreneurial opportunity creation. Third, I develop a research agenda to start a discourse for enhancing existing literature on the application of the crowd. Finally, I propose crowdsourcing as a practical way for entrepreneurs to apply user-centric entrepreneurship principles (Blank 2013; Ries 2011).

## 5.2.2. Opportunity Creation and Entrepreneurial Decision-Making

Based on my discussion of entrepreneurial action, OCT comprises four concepts that are central to entrepreneurial action taking: uncertainty, social interaction, iterative development, and learning. First, the underlying core assumption of each creation process is uncertainty. Uncertainty in this context regards the objective value of an idea, the needs of stakeholders, and the outcome of this iterative process (Alvarez et al. 2013). Contrary to the concept of risk, where decision makers can estimate the outcomes and the probability of such outcomes associated with a decision, uncertainty neither implies the outcomes associated with a decision nor is their probability known (March and Shapira 1987).

For opportunity creation, the concept of uncertainty has a dual role. On the one side, the entrepreneur has only insufficient information about the responses from the market or other stakeholders regarding a novel technology-based value proposition. On the other side, stakeholders, for example potential investors, perceive uncertainty or doubts about the actual value of the idea (McMullen and Shepherd 2006). For a successful opportunity creation process, entrepreneurs should reduce both their individual uncertainty to objectify an opportunity and the uncertainty of their stakeholders to further develop the initial idea and get potential stakeholders on board (Haynie et al. 2009).

Second, entrepreneurs use social interaction with their peers, customers, and other stakeholders to reduce such an uncertainty by gathering feedback. The uncertainty about their opportunity is reduced until opportunities can be objectified and the enactment can occur (Wood and McKinley 2010). Third, these social interactions lead to iterative changes in the beliefs and mental models concerning the initial opportunity and finally enable the entrepreneur to create wealth. Therefore, the opportunity emerges, and ideas, products, or total business models are continuously reassessed, pivoted, or even abandoned (Ojala 2016). Fourth, directly related to the iterative development, creation theory assumes that the entrepreneur should rely on experiments, feedback, and learning rather than pre-existing knowledge (Mintzberg 1994).

I therefore propose that crowdsourcing, which proved to be a valuable concept in the context of innovation, is a valuable approach for entrepreneurs to reduce uncertainty and interact with the market as it provides access to heterogeneous knowledge from diverse sources.

### 5.2.3. Previous Work on Crowdsourcing for Innovation Decisions

One special instantiation of integrating interaction with a firm's environment into the process of innovation is crowdsourcing. Crowdsourcing has been developing as part of the greater open innovation movement and is thus increasingly used by firms to innovate (e.g. Poetz and Schreier 2012). It denotes the act of a *"[...] participative online activity in which an individual, an institution, a non-profit organization, or company proposes to a group of individuals of varying knowledge, heterogeneity, and number, via a flexible open call, the voluntary undertaking of a task. The undertaking of the task, of variable complexity and modularity, and in which the crowd should participate bringing their work, money, knowledge and/or experience, always entails mutual benefit"* ((Estellés-Arolas and González-Ladrón-De-Guevara 2012).

The underlying rationale suggests that a large diverse crowd of independent strangers performs better on certain types of challenges than a small number of experts (Brabham 2008, 2013; Lakhani and Jeppesen 2007; Lakhani et al. 2013). At the heart of the concept are new information systems that allow to leverage networks and therefore innovate with users outside one's association (Doan et al. 2011; Dodgson et al. 2006; Lindič and Marques da Silva 2011; Trott and Hartmann 2009).

To argue for the application of crowdsourcing for opportunity creation, I extensively reviewed literature on crowdsourcing for innovation to present its current applications, benefits, and organization. For this study, I focus on crowdsourcing in the context of innovation. Although crowdsourcing for innovation is far from being a new concept, it is still a topic of high interest and relevance, especially among innovation scholars (Terwiesch and Xu 2008, Chesbrough et al. 2006).

Prominent applications of crowdsourcing in the innovation context include idea generation (Leimeister et al. 2009), idea evaluation (Blohm et al. 2016), as well as co-creation for new product development (e.g. Poetz and Schreier, 2012; Terwiesch and Ulrich, 2006). Previous research in the field shows the crowd's appropriateness as both source of and *"rater"* for new product and service ideas (Ogawa and Piller 2006). Firms that apply crowdsourcing for innovation benefit from the heterogeneous and diverse crowd, which can provide the ability to discover creative solutions. Interaction with the crowd enables firms to discover novel customer requirements and user input for ideas, representing a *"voice of the customer"* (e.g. Dahan and Hauser 2002; Griffin and Hauser 1993). Therefore, crowdsourcing provides both need-based information (i.e. what is the problem?) as well as solution-based information that guides companies to finding out what a potential new product or service should do (Terwiesch and Ulrich 2009; Van Hippel 2005).

Previous research on crowdsourcing for innovation emerged around finding an innovative solution to a certain problem. Prominent examples include idea communities such as Dell Idea storm, MyLego or Foldit where users can brainstorm and provide solutions to new products of the respective companies (Schlagwein and Bjørn-Andersen, 2014). Therefore, one highly important benefit of crowdsourcing is the crowd's ability to provide both user needs (i.e. demand-side knowledge) and product trends (i.e. supply-side knowledge) (Ozer 2009). Moreover, the concept of crowdsourcing leverages the cognitive principle of collective intelligence, which aggregates heterogeneous knowledge while reducing errors that arise from human biases (Malone et al. 2009). Although it has been acknowledged that crowdsourcing can also be used to solve more complex tasks, predominant applications still address problems that address the fuzzy front end of innovation. Thus, idea communities providing solutions to complex problems (such as Quirky) still seem to be the exception than the rule.

In finding innovative solutions to a certain problem, requestors (i.e. companies) usually call upon the crowd. The crowd is thereby understood as an undefined, heterogeneous mass of people that is expected to differ in their capabilities and knowledge to solve a certain problem. The diversity of the people is believed to increase the likeliness that an innovative and creative solution will be found (Brabham 2013). Apart from that, little is known about how crowds differ in terms of knowledge and skills across different domains and solution spaces. Although research highlights the role of users, most of literature still treats the crowd as an undefined mass of people (Magnusson et al. 2016). Therefore, there still seems to be very little understanding of the adequate composition of crowds and which crowds may be most effective in solving certain types of problems.

In summary, crowdsourcing for innovation uses the creativity, expertise, and knowledge of a heterogeneous crowd to generate novel ideas (Leimeister et al. 2009). The sponsoring firm then discovers such solutions as novel opportunities and uses the crowd to filter the most promising ideas through voting (Magnusson et al. 2016).

Crowdsourcing is thus a linear process of infusing external knowledge to a firm's innovation management. The development of novel products and services is then based on the ideas of the crowd. However, one limitation of previous applications of crowdsourcing for innovation is that the focus is on the generation and discovery of novel ideas rather than the evolution and iterative co-creation of such between a firm and the crowd, which is required to turn ideas into novel value propositions and business models (Cullina et al. 2016; Majchrzak and Malhotra 2013).

### 5.2.4. Crowdsourcing for Guiding Entrepreneurial Decision-Making

Following the findings of the previous sections, the state of the art in research on OCT demands a higher level of heterogeneity of feedback to understand the value of an opportunity and iteratively develop such. On the other hand, the analysis of literature on crowdsourcing for innovation clarifies that crowdsourcing provides exactly those benefits for entrepreneurs and might be used to digitize the entrepreneurial process of opportunity creation.

I therefore focus on crowdsourcing from an opportunity creation perspective which offers a fruitful lens to examine how crowdsourcing can help entrepreneurs to generate value in an iterative and co-creative way. I argue that social interaction with the crowd reduces uncertainty and enables the iterative development of the initial opportunity that triggers entrepreneurial learning. Thus, crowdsourcing could help entrepreneurs to share opportunity ideas with potential customers and stakeholders to iteratively modify and test them over time (Majchrzak and Malhotra 2013).

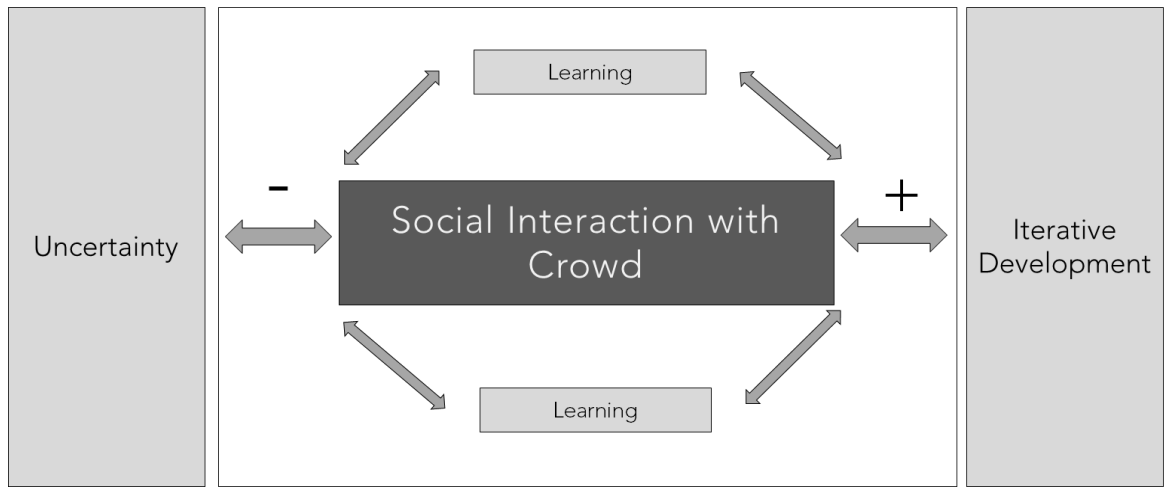

***Conceptual Model of Crowdsourcing from an OCT Perspective***

Integrating a heterogeneous crowd into the entrepreneurial process therefore provides access to social resources that are characterized by both strong heterogeneity and anonymity. Form a holistic perspective, this enables the entrepreneur to create their opportunity by using collective intelligence (Malone et al. 2009) and enabling the entrepreneur to gather data about the *"voice of the customer"* (e.g. Dahan and Hauser 2002; Griffin and Hauser 1993). The feedback of the crowd represents the *"voice"* of a potential market and therefore results in a higher level of validity meanwhile reducing the threat of overestimating the value of an idea and supporting the further development of an initial opportunity idea into a novel venture. It provides a rapid and cost-efficient way to aggregate data about the reactions of the market, feedback of functionality, or the customers' perception of a solution (e.g. Blank 2013; Ries 2011).

By challenging her assumptions and beliefs with potential users, the entrepreneur gathers information about the value of the opportunity and the level of its product-market fit and therefore reduces her individual uncertainty by validating her assumptions (Alvarez and Barney 2007). The feedback of the heterogeneous crowd therefore results in a higher level of validity that reduces the threat of an entrepreneur's overestimation of the value of an idea. As crowdsourcing enables the entrepreneur to use feedback from a huge number of people, the threat that she must generalize and make decisions from small samples is minimized. Vice versa the feedback from the crowd can also function as signaling that the opportunity is desirable for the market (Tocher et al. 2015).

Thereby, crowdsourcing helps the entrepreneur to overcome limitations like limited access to or homogeneity of social resources. Furthermore, crowdsourcing provides valuable potential for the iterative development of the opportunity. By providing feedback, the crowd acts as active co-creator and supports the entrepreneur in further developing the opportunity. One major benefit of crowdsourcing is here the access to knowledge from the market.

Finally, crowdsourcing for opportunity creation fosters entrepreneurial learning during this process by offering new insights on the market and other stakeholder's perception of the opportunity. Thus, it enables the entrepreneur to integrate this knowledge in the further opportunity creation process.

In summary, the different concepts from OCT can be addressed by using the crowd to overcome the limitations of previous approaches, making crowdsourcing a central part of the opportunity creation process compared to single creative campaigns of huge firms. I therefore show how the main concepts of OCT and the limitations can be addressed through crowdsourcing.

### 5.2.5. Idiosyncrasies of Crowdsourcing for Guiding Entrepreneurial Decisions

In the previous section, I argued that crowdsourcing provides several benefits for opportunity creation. The actual architectures of crowdsourcing, however, have previously been examined and tailored to the demands of the innovation management in established firm (e.g. Leimeister et al. 2009).

However, I argue applying crowdsourcing to entrepreneurial opportunity creation requires an entirely different perspective that can do more than just help companies at the fuzzy front end of innovation by enabling iterative co-creation between the entrepreneur and her environment. Although single parts of crowdsourcing mechanisms and participation architectures are suitable for the iterative creation of opportunities, they do not perfectly fit their requirements due to various reasons.

First, in previous studies on crowdsourcing, the crowd represents a source of creative ideas for problem solving that can be objectively discovered through distant search (e.g. Leimeister et al. 2009). Thus, linear, and one-directional social interactions with the crowd constitute an accelerator for recognizing ideas than collaboratively co-creating innovative value propositions. Consequently, the crowd is incentivized for posting new ideas rather than refining an existing one (Majchrzak and Malhotra 2013).

Second, following this argumentation, crowdsourcing in the context of entrepreneurial opportunity creation requires multi-directional interactions with the crowd. From a constructivist's perspective, this is crucial to foster feedback-based idea evolution (Alvarez et al. 2013). Apart from an open call to the crowd, it requires further and intensive exchange between the initiator and the crowd. The crowd is therefore not the source of an initial idea but provides feedback on the correctness of an entrepreneur's assumptions and refines an idea. The initiation of innovation in this process, however, is to the entrepreneur, who starts the interaction with the crowd by showing her beliefs and ideas about an opportunity (Alvarez and Barney 2007). This is a central limitation of previous IS research on crowdsourcing architectures that led to lots of failures in creating solutions that could be implemented by sponsoring firms (Majchrzak and Malhotra 2013).

The third difference between traditional crowdsourcing efforts to foster innovation and the context of entrepreneurial opportunity creation is the level of task complexity. Contrary to previous research that focuses on using the crowd on the fuzzy front end of innovation (e.g. Poetz and Schreier 2012), the support of the opportunity creation represents a more complex task. The development of an opportunity goes far beyond the creation of early-stage ideas or product innovation as it includes the complete process including an initial idea of the entrepreneur, prototypes, and finally the development of a business model and an entire start-up (Ojala 2016). This contrasts with previous IS research that has focused on participation architectures and platforms for modular and closed problems solving tasks and leveraging the crowd for suggesting ideas while leaving the subsequent steps in the innovation process inside the boundaries of the sponsoring firm (Leimeister et al. 2009).

Fourth, identifying a suitable crowd that represents an entrepreneur's potential stakeholders (e.g. investors, customers) is different from crowdsourcing in existing innovation communities that foster the discovery of novel ideas among existing users (e.g. Poetz and Schreier 2012). In this context, the selection of crowd members should balance heterogeneity and expertise in the entrepreneur's technological and industrial domain. However, required application contexts and markets are frequently not known a priori but emerging (Alvarez and Barney 2007). Therefore, the requirements for crowd members' supply- and demand-side knowledge might also change over time and recruiting the crowd from a firm-specific community might be misleading. This contrasts with the widespread principles of an open call.

Finally, opportunity creation is an evolutionary and iterative process to develop an initial idea into a new venture. In the context of crowdsourcing for innovation, however, participation architectures and platforms focus on the contribution of creative ideas while they provide only limited support for the evolution of an idea or the generative co-creation to further develop such ideas into novel value propositions and business models (Majchrzak and Malhotra 2013). In general, there is frequently minimal collaboration among the innovating firm and the crowd. Therefore, the current architectures of crowdsourcing platforms for innovation emphasize the generation of novel ideas over the evolution of opportunities that are suggested by either one member of the community or the innovating firm (Madsen et al. 2012). Such participation architectures, however, are required to integrate the crowd to provide feedback and support the opportunity creation of an entrepreneur and point towards directions for developing IS research on crowdsourcing and online communities.

## 5.2.6. Guiding Entrepreneurial Decisions - A Research Agenda

Based on these holistic differences between crowdsourcing for innovation and crowdsourcing for opportunity creation, I derived a structured description of more detailed requirements for the application of crowdsourcing in this new context. Based on this, I develop a research agenda that motivates and potentially guides future research. To structure my research agenda, the conceptual framework of Pedersen et al. (2013) guided me (see Figure 23). Thereby, I attempt to show how crowdsourcing could be designed to meet the requirements of OCT (Alvarez and Barney 2007; Alvarez et al. 2013) and provide directions for further research, thus, digitizing the entrepreneurial process

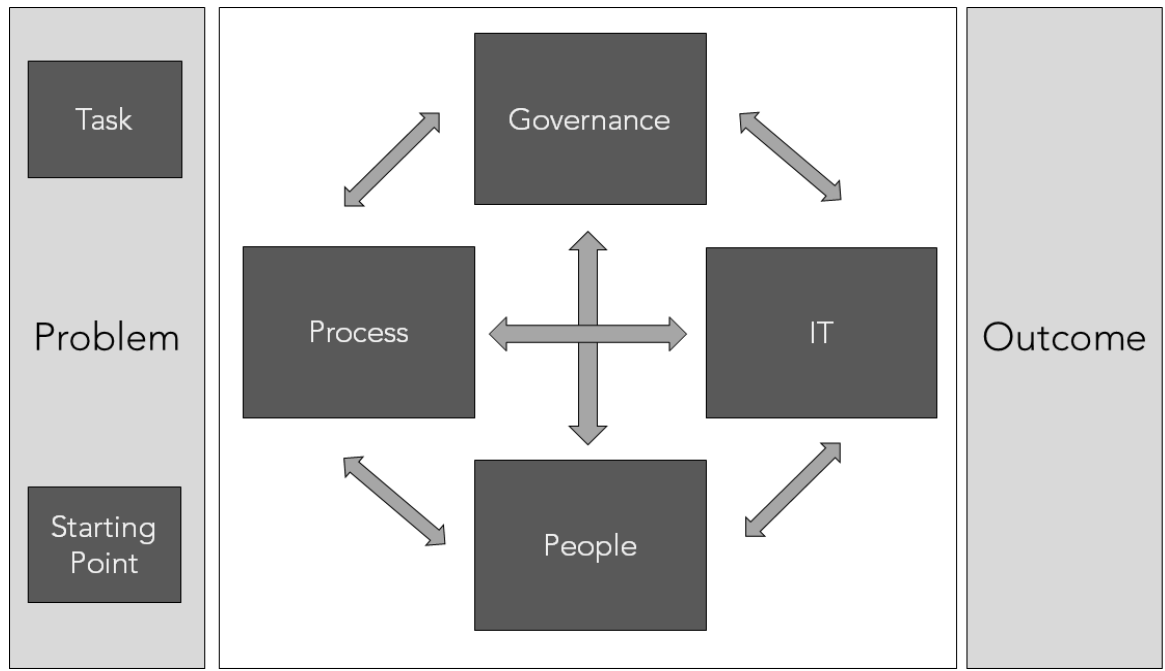

*Conceptual Model of Crowdsourcing*

The starting point in crowdsourcing for opportunity creation is an entrepreneur's initial opportunity. Opportunity creation is an iterative process that includes not just an idea or product but the development of an entire start-up (Alvarez et al. 2013; Ojala 2016). Therefore, it is crucial to understand what stage of opportunity the best starting point is to apply crowdsourcing and the path dependence behind that rationale. More practical research could further focus on suitable representations of the opportunity to provide it to the crowd.

The problem task can start at different stages such as the state of an early idea pitch, a minimally viable product, a prototype, or even a business model. The task of the crowd is providing evaluation and feedback, thus reducing uncertainty (e.g. Mangusson et al. 2014) by gathering information about the *"voice of the customer"* (Griffin and Hauser 1993). The focus of the problem task is therefore on the evolution of an entrepreneur's opportunity rather than the generation and discovery of novel opportunities (e.g. Majchrzak and Malhotra 2013). The role of such opportunity evolution in a co-creative process with the crowd is one central theme for further research to better understand the core of crowdsourcing for opportunity creation. Furthermore, future work should show how this task can be supported.

The people in crowdsourcing for opportunity creation include an anonymous and heterogeneous crowd as *"solvers"* and the individual entrepreneur as requestor. In this context heterogenous covers the aspect of gathering insights from multiple perspectives (e.g. customers, partners, etc.) that are all aligned by a common interest. From a holistic perspective, this enables the entrepreneur to create their opportunity by using the *"wisdom of crowds"* and benefit from heterogeneous knowledge (Surowiecki 2004). Although I refer to the crowd as a heterogeneous and anonymous mass of people, the context of entrepreneurial opportunity creation demands several requirements from a *"suitable"* crowd that has expertise to support the entrepreneur in evolving the opportunity. In this context, I argue that it is particularly important that the crowd represents potential customers and stakeholders (e.g. partners or investors) to assemble people that are interested in the opportunity. The benefit of crowdsourcing therefore lies in balancing supply- and demand-side knowledge (Lüthje 2004; Magnusson 2009). Consequently, further research in this field should focus on how to find a suitable crowd or the role of different expertise in crowdsourcing for opportunity creation. For this purpose, it is central to understand if and what different crowd characteristics are more suitable for different maturity stages of the opportunity creation process and consequently how matching mechanism might support a crowd segmentation process.

The crowdsourcing for opportunity creation process starts with the initial conceptualization of a potential future business idea by the entrepreneur (Wood and McKinley 2010). After the entrepreneur has imagined an opportunity, the sense making starts to verify her initial beliefs (Weick 1993). Therefore, the entrepreneur starts with a call to the crowd. The call should go to a crowd consisting of potential future customers and other stakeholders such as investors or business partners (Alvarez et al. 2013; Tocher et al. 2015). This initiation of the crowdsourcing effort elicits an iterative, dynamic process of evaluation and feedback between the entrepreneur and the crowd to co-create the opportunity. As the entrepreneur uses the interaction with and the feedback from the crowd as sense making, the process is open ended until the entrepreneur has finally objectified and enacted the opportunity (Wood and McKinley 2010). An entrepreneur therefore individually decides about the numbers of iterations and the end of this process. Thereby, it is particularly important to understand what amount of iterations might be ideal to balance feedback and information overload. Moreover, this process requires research on guidance to support the entrepreneur during opportunity creation.

The governance of a crowdsourcing for opportunity creation process requires the entrepreneur to select an appropriate crowd (e.g. Magnusson et al. 2016), to deliberately design the task for the crowd by providing suitable feedback mechanisms (e.g. Blohm et al. 2016; Leimeister et al. 2009), and to decide on representations of the opportunity as well as to ensure an effective incentivization (e.g. Malone et al. 2009). Therefore, it is crucial to provide suitable feedback mechanisms to the crowd to ensure high-quality feedback (Riedl et al. 2013, 2010) as well as representations of the opportunity that help the crowd to understand the content of the entrepreneurial opportunity (e.g. idea pitches, ontologies, videos). An important issue for IS research is therefore the exploration and design of appropriate evaluation and feedback mechanisms that increase feedback quality. Following previous entrepreneurship research, the interaction between an individual and the crowd might be more important than a sense of community and exchange among the crowd members for opportunity creation. However, I argue that this is an interesting field for further research to explore the role of community engineering in crowdsourcing for opportunity creation. Finally, the entrepreneur should provide incentives to the crowd to ensure that it is not only capable but also willing to provide feedback and help to support the further development of the opportunity. Future work should therefore point towards an understanding of motives in opportunity creation and suitable incentivization mechanisms to attract the crowd.

IT, in this context, is an enabler that provides the technical capabilities to implement crowdsourcing for opportunity creation and guides the interaction between the entrepreneur and the crowd. The iterative co-creation process demands extensive tool support, asynchronous capabilities, and collaboration capabilities to form a crowd and facilitate successful task completion (Pedersen et al. 2013). Thus, an interesting entry point for further research can be architectures for integrated and automated platforms, which supports the selection of a suitable crowd as well as IT tool support that facilitates co-creation during different stages of the opportunity creation process.

Finally, the outcome of crowdsourcing for opportunity creation is the feedback from the crowd, the iterative integration of knowledge into the co-creation process, and finally a fully enacted opportunity. The iterative nature of crowdsourcing for opportunity co-creation require also an iterative evolvement of outcomes from simple feedback to fully co-created value propositions. The access to such social resources through crowdsourcing results in evaluations and feedback from potential customers and other stakeholders and reduces an entrepreneur's uncertainty. Thus, crowd feedback signals the response and thoughts of potential customers and reduces uncertainty if the idea is objectively valuable. Furthermore, crowdsourcing for opportunity creation can help the entrepreneur to create an early sense of urgency for her opportunity idea and create awareness as well as commitment among potential customers. To benefit from the outcomes of the process, the entrepreneur should accept and integrate the information acquired in her future actions to facilitate learning (Alvarez et al. 2013). Hence, it is crucial to understand how outcomes should be structured and presented to the crowd and what are integration mechanisms to support entrepreneurial learning.

Based on this discussion, I propose a research agenda for crowdsourcing for opportunity creation that describes how crowdsourcing should be designed to meet the requirements of OCT.

| | Crowdsourcing for Opportunity Creation | Potential Research Issues | Exemplary Contributions |
|---|---|---|---|
| *Starting Point* | ▪ Initial opportunity idea at different stages (e.g., idea pitch, MVP, prototype, business model) | 1) *What are suitable ontologies for representation of the starting point?*<br>2) *What stage is the best starting point?*<br>3) *What are process path dependencies?* | ▪ Business model or prototype ontologies for IT artefacts<br>▪ Reference Process Models |
| *Tasks* | ▪ Evaluation and feedback<br>▪ Idea evolution instead of generation<br>▪ Should address problems that are highly contextual and require experts | 4) *What is the role of evolution in entrepreneurship?*<br>5) *How can the task be supported?* | ▪ Platform and task design for complex problem solving |
| **People** | ▪ Crowd: volunteering potential stakeholders, potential customers, investors, and experts with specific knowledge<br>▪ Requestor: Individual entrepreneurs seeking to validate and create their opportunity ideas | *6) What is the role of different expertise?*<br>*7) Are different crowd characteristics for different tasks more suitable?* | ▪ Expertise requirements in crowdsourcing |
| **Process** | ▪ Initial opportunity creation by the entrepreneur<br>▪ Call to "suitable" /"select" crowd<br>▪ Iterative exchange Open-ended process<br>▪ Evaluation and co-creation between crowd and entrepreneur | *8) How to support process guidance?*<br>*9) What are appropriate feedback mechanisms?*<br>*10) What is the best amount of iterations?* | ▪ Experimental findings on the effect of collaboration on crowdsourcing outcomes |
| **Governance** | ▪ Less requirements for community engineering<br>▪ Immediate incentivization needed<br>▪ Feedback mechanisms<br>▪ Quality management | *11) What is the role of community engineering in crowdsourcing for opportunity creation?*<br>*12) What are suitable incentivization mechanisms?* | ▪ Activation supporting components and participation architectures for platform design |
| **Role of IT** | ▪ IT as enabler<br>▪ Extensive tool support for opportunity creation required<br>▪ Need for integrated platform | *13) How can tool support be designed?*<br>*14) What are appropriate platform architectures?*<br>*15) How can matching mechanisms help to find suitable crowd members?* | ▪ Novel platform design principles<br>▪ Recommender based on crowdsourcing contributions |
| **Outcomes** | ▪ Broad solution space<br>▪ Signalling<br>▪ Feedback and validation<br>▪ Fully enacted opportunity<br>▪ Acceptance of external feedback | *16) How can outcomes be aggregated, structured, and presented in IS tools?*<br>*17) What are integration mechanisms to support entrepreneurial learning?* | ▪ Visualization of decisional guidance<br>▪ Design and development of decision support systems based on crowdsourcing |

*Research Agenda for Crowdsourcing for Opportunity Creation*

## 5.2.7. Conclusion

On a broader level, I have proposed crowdsourcing for opportunity creation as a new field of further research in both IS and entrepreneurship. I therefore took an opportunity creation perspective on entrepreneurship and highlighted the limitations of previous approaches in entrepreneurial interaction with the social environment to validate the beliefs and assumptions about an opportunity, thus, reducing uncertainty. I then conceptually developed the idea that crowdsourcing, which was previously applied in the context of innovation management in established firms is a suitable way to overcome these limitations by using the feedback from a heterogeneous crowd to reduce uncertainty and iteratively develop an opportunity into a new venture.

My further discussion shows that crowdsourcing in its current form is tailored for the application in established firms than the opportunity creation context. Thus, I revealed differences between both approaches and based on this developed an agenda for further research to point towards research that explores the adaption of previous crowdsourcing mechanisms in the field of innovation for the special context of entrepreneurial opportunity creation. This is crucial for IS research to design novel IT and platform architectures that enable iterative interaction between entrepreneurs and the crowd. From a methodological perspective, interdisciplinary research on the topic of crowdsourcing for opportunity creation might consider design-oriented research approaches (e.g. Hevner et al. 2004; Peffers et al. 2007). Such possibilities might be, for instance, the development of tools to validate entrepreneurial assumptions and business models (Ries 2011) or systems to enable online co-creation between entrepreneurs and the crowd. Thus, such research can inform practical orientation while maintaining theoretical rigor (Gregor and Hevner 2013). Moreover, exploratory research might empirically examine recent innovative platforms such as JumpStart Fund (e.g. (Dellermann et al. 2017b) or Quirky to provide a deeper understanding on how the interplay of openness and IT should function for supporting entrepreneurial opportunity creation.

My theoretical contribution is therefore three-fold. First, I contribute the OCT (Alvarez and Barney 2007; Alvarez et al. 2014) by revealing limitations of previous approaches that entrepreneurs use to interact with the social environment to reduce uncertainty. Thereby, I showed various reasons why the social interaction with peers is insufficient to gather feedback. Second, I contribute to research on crowdsourcing in IS by extending the theoretical scope to a new field of application. Third, by highlighting the requirements of crowdsourcing for opportunity creation, I point towards potential future research issues. Such research should examine novel participation architectures that enable the iterative co-creation of an opportunity through different maturity stages, thereby overcoming the limitations of previous crowdsourcing efforts that focus on the generation of novel ideas than its evolution (Majchrzak and Malhotra 2013). The crowdsourcing for opportunity creation research agenda proposed here rests on these premises. I therefore aim at making a first step towards this direction. The potential issues for future research outlined here would hopefully not only motivate but also guide future research efforts in the field of entrepreneurship and crowdsourcing in IS.

As a practical contribution of my research I propose crowdsourcing as a practical way for entrepreneurs to validate their assumptions about the objective value of their opportunity. Therefore, crowdsourcing might offer tremendous possibilities to test ideas in the market, achieve fast and early product-market fit and apply customer-centric principles to entrepreneurship (Blank 2013; Ries 2011). Entrepreneurs might consider applying such mechanisms for instance during crowdfunding campaigns (e.g. Lipusch et al. 2018) or use existing platforms such as JumpStart Fund (Dellermann et al. 2017) or Quirky that became recently popular due to the hyperloop project. This allows entrepreneurs to validate and refine their ideas early and iterative while reducing the risk of missing customer needs.

## 5.3. Designing Crowd-based Guidance for Entrepreneurial Decision-Making

The findings of this chapter were previously published as developing (Dellermann et al. 2017c). This study leverages the idea of using crowdsourcing for guiding entrepreneurial decision-making to design an IT tool by conducting a DSR approach. The resulting DR are then further used for the design in Section 6.5.

### 5.3.1. Introduction

The rapid digital transformation of businesses and society creates tremendous possibilities for novel business models to create and capture value. Many Internet start-ups such as Hybris, Snapchat, and Facebook are achieving major successes and quickly disrupting whole industries. Yet, many digital ventures fail. One reason for this is that entrepreneurs face high uncertainties when creating their business models. Consequently, entrepreneurs must constantly re-evaluate and continuously adapt their business models to succeed (Andries and Debackere 2007).

One way to deal with uncertainty during the development of business models is the validation of the entrepreneur's assumptions by testing them in the market or with other stakeholders such as suppliers or complementors (Blank 2013). Such a validation allows the entrepreneur to gather feedback to test the viability of the current perception of a business model and adapt it, if necessary, before potentially wasting money. For this purpose crowdsourcing has proven to be a valuable mechanism (Ebel et al. 2016b; Ebel et al. 2016a) in other contexts.

Literature on business models provides a rich body of knowledge about different components or the initial design (Al-Debei and Avison 2010; Osterwalder and Pigneur 2013), however, they do not provide any information systems that support such processes and enable the integration of the diverse voices of stakeholders (Veit et al. 2014). Thus, service institutions that create a supportive environment for start-ups, so-called incubators, function as intermediaries that connect different actors such as consultants, business angels, or venture capitalists with entrepreneurs for the exchange of services. Although business model validation services are a repetitive activity of incubators, systematic and scalable solutions to enable interaction to validate business models do not exist. In this context, IT creates opportunities to design systems that support the entrepreneur in business model validation.

Therefore, the aim of this paper is to develop tentative DPs for crowd-based business model validation (CBMV) systems. Such information systems support entrepreneurs in learning and reducing the uncertainty about the validity of their assumptions. With this aim in view, I develop DPs that guide the design of prototypes for CBMV systems. I refer to DPs as the tentative properties of a generic solution drawn from literature that address the potential solution space of such artifacts. The purpose of this paper is thus to develop DPs for information systems that feature CBMV.

To derive my DPs, I follow a design science approach (Hevner 2007) guided by the process of Vaishnavi and Kuechler (2015). This paper follows a theory-driven design approach based on knowledge drawn from literature and complemented by empirical insights. For developing CBMV systems, I combine the concept of crowdsourcing with findings from research on DSSs to propose tentative DPs. The identified DPs describe the core of a solution to a problem that previous research proved as viable. I therefore ensure theoretical rigor while developing a system to solve a real-world business problem.

## 5.3.2. Business Model Validation in Early-Stage Start-ups

To formulate the problem for my design research approach, I reviewed current literature on business model development. The concept of business models has gathered substantial attention from both academics and practitioners in recent years (Veit et al. 2014). In general, it describes the logic of a firm to create and capture value (Osterwalder and Pigneur 2013; Amit and Zott 2001). Although there is no commonly accepted definition of the term, this concept provides a comprehensive approach toward describing how value is created for all engaged stakeholders, the allocation of activities among them, and the role of information technology (Bharadwaj et al. 2013). Following Teece (2010), a business model reflects the assumptions of an entrepreneur and can therefore be considered as a set of hypotheses about the ecosystem.

In the context of early-stage start-ups, business models become particularly relevant as entrepreneurs define their ideas more precisely in terms of how market needs might be served. In addition to that, it helps the entrepreneur to examine which kind of resources have to be deployed to create value and how that value might be distributed among the stakeholders (Demil et al. 2015). Such early conceptualizations of a start-up's business model represent an entrepreneur's assumptions about what might be viable and feasible but are mostly myopic in terms of the outcome as entrepreneurs are acting under high levels of uncertainty (Alvarez et al. 2013). Since entrepreneurs are operating under high levels of uncertainty, they start a sense-making process in which they test their initial beliefs about the market through iterative experimentations and learning from successful or failed actions (Alvarez and Barney 2007). When the entrepreneurs' assumptions contradict with the reaction of the market, this might lead to a rejection of erroneous hypotheses. This will require a reassessment of the business model to test the market perceptions again. Thus, the business model evolves toward the needs of the market and changes the assumptions of entrepreneurs (Alvarez et al. 2014). The success of start-ups thus heavily depends on the entrepreneurs' ability to develop and continuously adapt their business models to the reactions of the environment.

### 5.3.3. Previous Work on Crowd-Based Guidance

Practitioner literature recognizes that many business models fail due to wasting resources before validation (Blank 2013). Consequently, entrepreneurs should test the assumptions about their business model with customers, partners, complementors, and suppliers to gather feedback and validate the current version before continuing and wasting money. The feedback from external actors enables entrepreneurs to reflect on the current version. Thus, entrepreneurs may start thinking about the drawbacks of their hypothesized business model and exert effort on resolving these by reassessing, pivoting, or even abandoning elements (Ojala 2016).

One mechanism that has proven to be valuable to gain access to such feedback is crowdsourcing (Blohm et al. 2011; Blohm et al. 2013). Research on crowdsourcing shows the value of integrating customers and other stakeholders into the evaluation process to support decision-making during the development of new products. For instance, crowd voting provides extensive evidence for the suitability of a crowd in evaluation tasks as it is equally capable of identifying viable ideas (Klein and Garcia 2015; Toubia and Florès 2007). Therefore, many companies have started to use the collective intelligence of a heterogeneous crowd to evaluate ideas (Kornish and Ulrich 2014). Thereby, a heterogeneous crowd, most commonly end users of a certain product, rates certain product ideas. Crowd-based online validation of innovation is particularly beneficial compared to industry expert evaluation due to time and cost efficiency reasons (Toubia and Florès 2007), the reduction of individual biases through averaging the results (Mannes et al. 2012), and the possibility to focus on the demand side perspective of innovation (Di Gangi and Wasko 2009) including a much higher number of raters compared to offline approaches. This assessment constitutes a proxy to distinguish between high- and low-quality ideas and the feedback of the crowd is then used as decision support on how to proceed. The appropriateness for using a crowd has also been shown for business models (Ebel et al. 2016a). I thus argue that crowd-based validation is also suitable for the highly uncertain context of start-up business models and provides a superior approach compared to consultancy feedback or offline approaches such as design thinking, which might force the entrepreneur to follow biased individual feedback or to draw conclusions from small samples.

## 5.3.4. Methodology

For developing DPs for a CBMV system, I conducted a DSR project (Peffers et al. 2007) in the broader context of a research project that attempts to provide crowd-based services for incubators to design a new and innovative artefact grounded in theoretical rigor that helps to solve a real-world problem. Therefore, I followed the design research cycle methodology as introduced by Vaishnavi and Kuechler (2015).

First, I conducted a literature analysis as well as exploratory qualitative interviews. I contacted executives of German business incubators (n=17) that provide business model validation services and decision makers in start-ups (n=28) to analyse the status quo of business model validation, the limitation of those, and requirements for a solution. For this procedure, I used a semi-structured interview guideline, which followed the theoretical concepts of OCT. This theory-guided approach provides two benefits. First, I could justify the DRs derived from theory. Second, I obtained a deeper understanding of the requirements from the practical problem domain. The requirements identified through the interviews were aggregated and coded. Thus, I could derive four additional DRs. The interviews lasted between 30 and 45 minutes and were coded by two of the authors. A cross case analysis was conducted to identify common themes. To develop suggestions for a solution, I applied a theory-driven design approach and OCT (Alvarez et al. 2007), which explains how business models are co-created, as general scientific knowledge base that provides theoretical abstraction of the cause and effect of the problem space and informs my design (Briggs 2006). From this kernel theory, I derived DRs that were validated and complemented with findings from the interviews. I then used previous work on crowdsourcing evaluation as well as DSSs as relevant knowledge base that provides me with guidance in the development of the DPs for the CBMV system. Such DPs drawn from literature are tentative properties that may inform the design of a first prototype. Through an expert workshop (n=7) I evaluated the validity of my conceptual tentative DPs. These DPs will then be instantiated into an IT artefact and finally evaluated in an experimental setting of a business model competition. Applying this approach allows me to use theoretical rigorous knowledge for developing an innovative IT artefact, which helps to solve a real-world problem, thus ensuring practical relevance.

## 5.3.5. Awareness of the Problem

The DSR project is motivated by both a gap in IS research on systems that support business model validation services and practical problems of entrepreneurs and incubators. Therefore, I conducted exploratory interviews with incubators (n=17) as well as entrepreneurs (n=28) to include a two-sided perspective on the problem and to create awareness. The interviews were guided by the central question of how incubators as service providers typically conduct the validation of entrepreneurs' business models and the perceived limitations of these approaches. By analysing the interviews, I gained a deeper understanding of practical business model validation for start-ups and discovered four key problems:

**Problem 1:** *Incubators do not use structured processes to conduct business model validation services, which represent a repetitive task*.

**Problem 2:** *Both incubators and entrepreneurs have only limited access to expertise. Access to demand-side knowledge is especially scarce.*

**Problem 3:** *The feedback of consultancy services is frequently perceived as subjective, industry bound, and thus misleading.*

**Problem 4:** *Resource constraints make scalable and iterative validations of business models impossible.*

Although the validation of business models is one of the most pivotal parts of business model creation (Ojala 2016), to the best of my knowledge, there are no systems that support this service.

## Theory-Driven Design for CBMV Systems

To define the objectives of the solution for my design science approach, I zoomed in on the entrepreneurial process and identified OCT as a kernel theory that informs me about the requirements of a CBMV. OCT is a theoretical lens to examine business co-creation under uncertainty and risk (Ojala 2016) This perspective implies that opportunities emerge from the iterative actions undertaken with the social environment. Entrepreneurs create business models based on their individual beliefs and perceptions, imagination, and social interaction with the environment (Alvarez and Barny 2007). Entrepreneurial actors then wait for responses from testing their models in the market to understand the perceptions of customers and other stakeholders and then adjust their beliefs accordingly to adapt their business models (Wood and McKinley 2010). During the validation of the entrepreneur's assumptions, a mismatch between the entrepreneurial idea and the opinion of the social environment may become evident. The entrepreneur will therefore need to reassess her assumptions and adapt the business model to the feedback of the market. This integration of customers, suppliers, and other stakeholders into the evolvement of a business model enables the entrepreneur to learn and further develop the initial version of the business model; it also reduces uncertainty about the validity of her assumptions (Ojala 2016).

## DRs from OCT

This entrepreneurship theory perfectly fits the context of my research as it explains how entrepreneurs create their businesses under uncertainty and helps to understand the problem domain of business model validation. Using this kernel theory, I developed the DRs for my artefact.

During the process of business model creation, entrepreneurs should validate their assumptions to validate the initial form of the business model and reassess parts of it if needed (Ojala 2016). To support this validation process, the CBMV system should consequently be able to support the entrepreneur in engaging in social interaction with potential customers or other stakeholders to validate the assumptions about the business model with the broader environment and make sense of it.

**DR1:** *Business model validation should be supported by systems that enable social interaction with potential customers or other stakeholders to test an entrepreneur's assumptions and support the sense-making process.*

To capitalize from social interaction, entrepreneurs gather external feedback on the viability of their business model hypothesis to make sense of their assumptions (Alvarez et al. 2013). Therefore, the feedback providers require suitable mechanisms to provide adequate responses (Blohm et al. 2016). Following this argumentation, CBMV systems should support the entrepreneur in gathering feedback through social interaction and, on the other hand, enable the crowd to provide such.

**DR2:** *Business model validation should be supported by systems that enable providing and receiving feedback to test an entrepreneur's assumptions and support the sense-making process.*

The creation of an initial version of the business model represents an entrepreneur's individual assumptions and beliefs (Gioia and Chittipeddi 1991). To start a sense-making process by interacting with external actors who provide feedback, entrepreneurs must translate their mental model of what is viable into a transferable format to communicate the imagined business model to others (Wood and McKinley 2010). Thus, entrepreneurs need to turn their assumptions regarding their business model into a transferable format to create a shared understanding between themselves and the external environment, which should provide feedback.

**DR3:** *Business model validation should be supported by systems that enable the entrepreneur to transfer their mental representation of a business model to the external environment for creating a shared understanding.*

Such mental representations of business models are not static but rather emergent assumptions that evolve through the process of social interaction and feedback (Eggers and Kaplan 2013). Thus, the creation process of a business model is highly iterative as entrepreneurs should start a sense-making process again when their assumptions about a desired business model change (Eggers and Kaplan 2009). To reduce incongruities in the assumptions of the business model, entrepreneurs incorporate the feedback from external actors (Tocher et al. 2015). Validating a business model might therefore need multiple iterations. Thus, systems that support business model validation should provide two affordances to enhance the iterative development of an entrepreneur's business model. First, such systems should easily allow for the adaption of the business model representation (see **DR3**); and second, they should enable the entrepreneur to iterate the process of gathering feedback and adapting the business model.

**DR4:** *Business model validation should be supported by systems that enable the iterative development and adaption of the business model representation during the sense-making process.*

Finally, entrepreneurs need to learn from the feedback and integrate the learning into the reassessment of their business model (Ojala 2016). The feedback that actors provide will include specific knowledge or expertise (Zott and Huy 2007) and thus change the information that is available for the entrepreneur during this emergent process (Alvarez and Barney 2010). Such feedback serves as a form of formative assessment that alters an entrepreneur's assumptions and accelerates learning (Nambisan and Zahra 2016). Thus, feedback-based learning might create a mental shift that orients the entrepreneur toward a specific direction. However, to facilitate the process of learning from the supply of extra knowledge through feedback from the social environment, entrepreneurs need guidance on what to do and how to derive actions based on this (Nguyen Huy 2001). Systems for business model validation should therefore support entrepreneurial learning through guidance on how to leverage feedback for the interpretation and update of an entrepreneur's assumptions and finally improve future versions of the business model (Ojala 2016).

**DR5:** *Business model validation should be supported by systems that enable the entrepreneur to learn from the results of the sense-making process through guidance that instructs future entrepreneurial actions.*

## Practical Requirements

To complement the theoretical DRs, I gathered practical requirements from the problem domain to balance the artefact's grounding in both theoretical rigor as well as practical relevance. I therefore derived additional DRs from the qualitative interviews with executives of incubators (n=17) and entrepreneurs (N=28) following the data collection approach stated in Section 5.3.4.

As resource constraints are one of the major problems for early-stage start-ups, the interviewees agreed on the theme of time and money as the crucial requirements for the usefulness of a CBMV systems. The dynamic and fast-changing environment as well as the limited time that entrepreneurs typically spend within incubators require the collection of feedback as fast as possible. Such rapid feedback was identified as particularly important to reduce the amount of time for each validation iteration.

**DR6:** *Business model validation should be supported by systems that enable the entrepreneur to obtain rapid feedback.*

Furthermore, limited financial resources are a main reason that hinders entrepreneurs to validate their assumptions as they are typically not able to afford multiple rounds of consultancy, conducting workshops with potential customers, or building a community around their business idea.

**DR7:** *Business model validation should be supported by systems that enable the entrepreneur to obtain cost-efficient feedback.*

Apart from resource constraints, entrepreneurs are concerned about the competency of their feedback providers. They demand to obtain feedback from multiple sources (e.g. customers, investors, consultants) rather than from a single person who might be biased due to subjective perceptions of the entrepreneur's business model.

**DR8:** *Business model validation should be supported by systems that enable access to multiple feedback sources to enhance objectivity.*

Finally, one additional requirement derived from the interviews is the heterogeneity of knowledge among the feedback providers. The interviewees agreed that the convergence of traditionally separated industries (e.g. manufacturing and IT) requires novel types of business models that might blur traditional industry standards. CBMV systems should therefore provide access to heterogeneous knowledge to obtain adequate feedback.

**DR9:** *Business model validation should be supported by systems that enable access to heterogeneous knowledge to enhance the feedback quality.*

## 5.3.6. Translating DRs into Tentative DPs

Based on the nine DRs derived from OCT and the qualitative interviews, I continued my research by identifying tentative DPs for a CBMV. First, I identified DPs by analysing literature to identify design-relevant knowledge from previous work, which helped me to address the identified DRs. Second, to ground my artefact in practical relevance, I conducted an expert workshop (n=7) to justify the tentative DPs derived from the literature. The participants in the workshop had both expertise in software engineering to evaluate the usability of the DPs to be implemented in an IT artefact as well as knowledge of the problem domain (i.e. business model validation) to assess the efficiency of the derived principles to solve the practical problem.

To gain access to social resources that might be used to validate the entrepreneur's assumptions quickly and iteratively, using a crowdsourcing platform constitutes a suitable approach (John 2016).

This approach is based on the findings of previous studies, which showed that a heterogeneous crowd can assess the value of creative solutions, such as an entrepreneur's business model, at a level comparable to that of experts, but at substantially lower costs (Magnusson et al. 2016). As neither incubators nor entrepreneurs have so far been able to build a community around their efforts, using existing crowd platforms can be leveraged through APIs (e.g. Amazon Mechanical Turk) to gain access to hundreds of thousands of problem solvers (John 2016). Thus, CBMV systems allow access to huge crowds to validate an entrepreneur's business model. This DP is suitable due to various reasons. First, it provides a scalable and cost-efficient way for tapping social resources to obtain feedback. Second, it enables the entrepreneur to provide monetary incentives to ensure participation (Klein and Garcia 2015). Third, creating tasks and retrieving validation results from individual participators, whose previous ratings by other users cannot be seen, avoids information cascades (Riedl et al. 2013). Thus, I suggest:

**DP1:** *Provide the CBMV systems with access to existing crowdsourcing platforms to provide the entrepreneur access to social resources.*

This procedure continues at least until the crowd has the necessary knowledge of the context in which they validate a business model. Past literature shows that a judge who is qualified for validating a business model is also an expert in the respective context (Ozer 2009). Such appropriateness then results in a higher ability to provide valuable feedback. This enables the prediction of the potential future success of a business model even in highly dynamic contexts. Therefore, a participant in the crowd should have two types of expertise to be suitable as a judge and provide more accurate predictions (Ozer 2009): demand- and supply-side knowledge. While the first type is necessary to understand users' needs and wants explaining the desirability of a business model, the latter one consists of knowledge on feasibility (Lüthje 2004). Both are necessary for the crowd to accurately validate an entrepreneur's business model, which represents the problem-solution fit. For this purpose, recommender systems that ensure to find a fit between the expertise requirements for being suitable as a judge and the validation task have proven to be a suitable approach in crowdsourcing (Geiger and Schader 2014). In particular, expertise retrieval, which suggests people with relevant expertise for the topic of interest, can be leveraged to find suitable judges on existing crowd platforms (Deng et al. 2012).

**DP2:** *Provide the CBMV systems with a recommender system in order that the entrepreneur obtains access to expertise.*

To apply CBMV, entrepreneurs must transfer their implicit assumptions to the crowd participants for creating a shared understanding. Business models are mental representations of an entrepreneur's individual beliefs that should be made explicit by transferring them into a digital object (Bailey et al. 2012; Carlile 2002). In particular, approaches to transfer such knowledge into a common syntax are required (Nonaka and Krogh). Therefore, ontologies can be used to leverage knowledge sharing through a system of vocabularies, which is the gold standard in the context of business models (Osterwalder 2004). Previous work on human cognition showed that the representation of knowledge in such an object (i.e. digital representation of the business model) should fit the corresponding task (i.e. judging the business model) to enhance the quality of the crowd's feedback (Khatri et al. 2006). Due to the fact that judging a business model is a complex task, a visual representation is most suitable as it facilitates cognitive procedures to maximize the decision quality (Speier and Morris 2003).

**DP3:** *Provide the CBMV systems with an ontology-based, visual business model representation to transfer an entrepreneur's assumptions and create a shared understanding among the crowd and the entrepreneur.*

To validate an entrepreneur's business model, the crowd needs adequate feedback mechanisms to evaluate the assumptions (Blohm et al. 2016). From the perspective of behavioural decision-making, this feedback can be categorized as a judgment task in which a finite set of alternatives (i.e. business models) is evaluated by applying a defined set of criteria by which each alternative is individually assessed by using rating scales (Dean et al. 2006; Riedl et al. 2013). In the context of crowd validation, individual ratings can be aggregated to group decisions (Zhao and Zhu 2014). Using rating scales for judging and thus validating an entrepreneur's business model is therefore most suitable for improving the quality of crowd evaluations (Di Gangi and Wasko 2009). In particular, elaborated rating scales with multiple response criteria lead to more consistent results of crowd-based validations. These multi-criteria rating scales should thus cover the viability and probability of success of a business model by assessing dimensions, which are strong predictors for the future success, such as the market, the business opportunity, the entrepreneurial team, and the resources (Song et al. 2008).

**DP4:** *Provide the CBMV systems with an elaborated feedback mechanism to enable the crowd to provide adequate feedback.*

As business model validation is an iterative process of adapting the current version of the business model and validating it again, CBMV systems should aggregate the results of each validation round to transient domain knowledge to show how the crowd feedback changes an entrepreneur's assumptions and how such changes are again evaluated by the crowd. The accumulation of such knowledge can trigger cognitive processes that restructure the entrepreneur's understanding of the domain (Wooten and Ulrich 2017). Learning can occur when entrepreneurs add new information from the feedback to their existing knowledge and cognitive schemas (Gönül et al. 2006).

**DP5:** *Provide the CBMV systems with an accumulation of domain knowledge by aggregating the results of the iterative feedback rounds so that the entrepreneur can learn.*

The feedback from the crowd provides extra knowledge about the validity of an entrepreneur's assumptions. To support entrepreneurs in reducing uncertainty and executing their task of adapting and further developing their business model, the CBMV systems should provide guidance to facilitate learning from the system (Silver 1991). Such decisional guidance, often studied in the context of DSSs (Silver 2006), is a DP that intends to reduce an entrepreneur's uncertainty and directs an entrepreneur's future actions by structuring decision-making processes under uncertainty (Mahoney et al. 2003). Decisional guidance can either be suggestive (i.e. explicitly recommending what to do) or informative, providing pertinent information that enlightens the user's choice without suggesting or implying how to act (Morana et al. 2017). This type of guidance provides information that supports the entrepreneur in reaching a conclusion of what to do. As the aim of the guidance of a CBMV system is fostering entrepreneurial learning, informative guidance is most suitable, especially for complex tasks such as adapting business models (Montazemi et al. 1996). Informative guidance outputs are the result of the crowd's judgment and support the entrepreneurs in learning from this additional information by enlightening the understanding of the social environment's reaction to their assumptions, especially when this feedback adds new perspectives, and lead to more reflective and deliberate thinking. Such learning may therefore increase the confidence of the entrepreneurs and develop a greater understanding of the problem domain. The mode of guidance is dynamic as the system should *"learn"* from the input of the judgment by the crowd and provide the guidance on demand when the entrepreneur decides to iterate the validation process. This mode is particularly effective for improving the decision quality, the entrepreneurial learning, and the decision performance (Parikh et al. 2001).

**DP6:** *Provide the CBMV systems with dynamic informative guidance so that the entrepreneur can guide the reactions to the provided feedback and learn.*

## 5.3.7. Conclusion

In this paper, I investigated tentative DPs for a CBMV systems that supports business model validation services to provide concrete principles that may guide the development of an IT artefact to solve a real-world problem. Therefore, I identified OCT as kernel theory to explain business model creation under uncertainty and derive five DRs from this theory. These are complemented by four additional requirements identified during interviews. Based on findings from literature, I develop six DPs that match my derived requirements for a CBMV systems and were validated within an expert workshop.The tentative DPs drawn from literature manifest a potential solution space of tentative properties that may inform the design of a first prototype.

My findings provide several contributions. First, I contribute to the body of knowledge on crowdsourcing and crowd evaluation by extending these mechanisms from the evaluation of creative ideas to the uncertain and complex context of start-ups business models, where I intend to show that the crowd is also able to assess the desirability and feasibility of entrepreneurial opportunities. Second, I provide a design for DSSs based on collective intelligence. I show that using this approach enables academia and practice to extend decision support services to the context of entrepreneurship and innovation. Finally, my tentative DPs provide practical guidance for providers of business model validation services, such as incubators, to develop information systems as well as a novel, crowd-based approach to conduct such services.

## 5.4. Expertise Requirements for Crowdsourcing in Guiding Entrepreneurial Decisions

The findings of this chapter were previously published as (Dellermann et al. 2018c). Based on the requirements for expertise requirements in crowdsourcing for entrepreneurial opportunity creation in Section 5.2 and the conceptually developed DP in Section 5.3, this study uses a DSR approach to design and evaluate an innovative approach for using expertise requirements in the crowd-based evaluation of business models. This approach combines text mining techniques to identify topics of business models and matches their contributors with novel business models that require evaluation. The findings of this Section are then used for the final design in Section 6.5.

### 5.4.1. Introduction

Firms increasingly engage in open innovation efforts to leverage the creative potential of a huge and diverse crowd of contributors (Leimeister et al. 2009). Therefore, one popular approach is to solve innovative problems by starting an open call to a crowd with heterogeneous knowledge and diverse experience via a web-based innovation platform (e.g. Bright Idea, Salesforce, and Ideascale). Individual members of the crowd then contribute creative opportunities to solve such problems and the firm rewards the best contribution in a contest approach (Lakhani and Jeppesen 2007). This novel way to solicit opportunities from online communities is a powerful mechanism to utilize open innovation.

However, the creative potential that arises from the innovative contributions of the crowd constitutes some critical challenges. The quantity of contributions and the demands on expertise to identify valuable opportunities is high and remains challenging for firms that apply crowdsourcing. Famous examples illustrate these novel phenomena. For instance, during the IBM "Innovation Jam" in 2006 more than 150,000 users from 104 countries generated 46,000 product opportunities for the company. Moreover, Google launched a crowd-innovation challenge in 2008 to ask the crowd opportunities that have the potential to change the world in their "Project 10^100". After receiving over 150,000 submissions, thousands of Google employees reviewed the ideas to pick a winner, which took nearly two years and tens of thousands of dollars (Bayus 2013). As previous research suggests only about 10–30% of the entrepreneurial opportunities from crowdsourcing engagements are considered valuable. Furthermore, screening this vast amount of contributions to identify the most promising opportunities is one of the toughest challenges of crowdsourcing to date (Blohm et al. 2016).

To solve these problems, different streams of research emerged that attempt to filter entrepreneurial opportunities (Klein and Garcia 2015). First, expert evaluations, which use executives within the firm to screen opportunities, were identified as costly and time consuming (Blohm et al. 2016). Second, research on algorithmic approaches proved to be a valuable way by identifying metrics to distinguish between high- and low-quality opportunities (Walter and Back 2013; Westerski et al. 2013; Rhyn and Blohm 2017). However, such filtering approaches always risk missing promising opportunities by identifying "false negatives" (classifying good opportunities as bad ones) and are rather capable to cull low quality opportunities than identifying valuable ones, which is a task that demands human decision makers. In response to this, the third approach to screen opportunities is crowd-based evaluation (Klein and Garcia 2015; Blohm et al. 2013; Riedl et al. 2013). Organizations have turned to the crowd to not just for generating opportunities but also to evaluate them to filter high quality contributions. This way has in fact shown to be of same accuracy such as expert ratings if the members of the crowd have suitable domain knowledge (Magnusson et al. 2016). However, this approach frequently fails in practice, when facing huge amounts of opportunities. Crowd-based filtering approaches tend to perform poorly as they make unrealistic demands on the crowd regarding their expertise, time, and cognitive effort (Klein and Garcia 2015).

By combining algorithmic ML approaches with human evaluation to adaptively assign crowd members that have the required domain knowledge to entrepreneurial opportunities, I propose a semi-automatic approach that leverages the benefits of both approaches and overcomes limitations of previous research. I thus propose that a hybrid approach is superior to sole crowd-based and computational evaluation for two reasons: First, various research suggests that computational models (or machines) are better at tasks such as information processing and provide valid results (Nagar et al. 2016), while human decision makers are cognitively constrained or biased (Kahneman and Tversky 2013). Additionally, previous research shows the importance of human decision makers in the context of innovation (Kornish and Ulrich 2014). In this highly uncertain and creative context, decision makers can rely on their intuition or gut feeling (Huang 2016).

Following a design science approach, I identified awareness of real-world problems in the context of filtering crowdsourcing contributions and derived DPs for such systems, which I evaluated with experts on crowdsourcing and requirement engineering.

I, therefore, intend to extend previous research on idea filtering in crowdsourcing engagements through combining algorithmic and crowd-based evaluation. This research therefore will contribute to both descriptive and prescriptive knowledge, which may guide the development of similar solutions in the future.

## 5.4.2. Entrepreneurial Contributions of the Crowd

In general, crowdsourcing denotes a mechanism that allows individuals or companies, who face a problem to openly call upon a mass of people over the web to provide potentially valuable solutions. One instantiation of crowdsourcing that is particularly interesting from both a practical and a research perspective are idea contests (Blohm et al. 2016). Idea contests are usually conducted via platforms that allow companies to collect opportunities from outside the organization. The output (i.e. the opportunities) of such contests are usually artefact opportunities that can take on different forms such as plain text, plans, designs and predictions from both experts and lay crowds (Riedl et al. 2013). The basic idea behind idea contests is thereby for companies to expand the solution space to a problem and thereby increasing the probability to obtain creative solutions to said problem (Klein and Garcia 2015). The effectiveness of idea contests is also underpinned by research showing that only under certain conditions users are willing, as well as capable to come up with innovative opportunities (Magnusson et al. 2016). Thus, by providing various incentives such as monetary rewards, firms increase the number of contributions and the probability to receive a creative submission. In simple terms attracting larger crowds leads to a more diverse set of solutions (Afuah and Tucci 2012).

### 5.4.3. Previous Approaches to Identify Valuable Opportunities

Such idea contests lead to a high number of opportunities that cannot be efficiently processed by current approaches. Thus, successful idea contests often lead to a flood of contributions that must be screened and evaluated before they can be moved to the next stage and further developed (Blohm et al. 2016). To identify valuable contributions that are worth implementing, one important task is filtering the textual contributions in such idea contests. Existing filtering approaches to separate valuable from bad contributions in crowdsourcing apply two content-based filtering approaches to evaluate the creative potential of opportunities: computational, algorithmic evaluation approaches and crowd-based evaluation approaches.

## Computational Evaluation Approaches

One current approach to evaluate textual contributions in the context of crowdsourcing is computational evaluation, wherein algorithms are used to filter opportunities based on metrics for idea quality such as word frequency statistics (Nagar et al. 2016). Within the approaches for computational evaluation, two dominant approaches are emerging to support the decision-making of the jury, which reviews the opportunities to identify the most valuable ones.

First, clustering procedures examine how the vast amount of textual data from crowdsourcing contributions can be organized based on topics (Walter and Back 2013) or domain-independent taxonomy for idea annotation (Westerski et al. 2013). Second, ML approaches can be used to filter opportunities based on rules that determine the value of the content (Rhyn and Blohm 2017). This approach is particularly useful if training data sets are available. Previous research in this context uses variables for contextual (e.g. length, specificity, completeness, writing style) or representational (e.g. readability, spelling mistakes) characteristics as well as crowd activity (e.g. likes, page views, comments), and behavior of the contributor of the idea (e.g. date of submission, number of updates) to determine the value of crowdsourcing contributions.

## Crowd-based Evaluation Approaches

The second approach to evaluate crowdsourcing contributions is applying crowd-based evaluation approaches. In this context, members of the crowd evaluate contributions individually and the results are aggregated (Klein and Garcia 2015). Such users might include other users of the contest, or even paid crowds on crowd work platforms (John 2016) that are asked to evaluate opportunities from the crowdsourcing engagement.

Previous research on crowd-based evaluation examined the applicability of one or multiple criteria in voting mechanism (where users vote for valuable opportunities), ranking approaches (where members of the crowd rank submissions), and rating mechanisms (where the crowd score opportunities) (Salganik and Levy 2015; Soukhoroukova et al. 2012; Bao et al. 2011). Moreover, prediction markets can be used where users trade opportunities by buying and selling stocks to identify the most valuable idea by aggregating these trades as a stock price (Blohm et al. 2016). Depending on the context of evaluation settings, these approaches proved to be equally accurate compared to the evaluation of experts (Magnusson et al. 2016).

## 5.4.4. Methodology

For resolving the above-mentioned limitations, I conduct a DSR project (Gregor and Hevner 2013) to design a new and innovative artefact that helps to solve a real-world problem. To combine both relevance and rigor I use inputs from the practical problem domain and the existing body of knowledge (rigor) for my research project (Gregor and Jones 2007). Abstract theoretical knowledge thus has a dual role. First, it guides the suggestions for a potential solution. Second, the abstract learnings from my design serve as prescriptive knowledge to develop other artefacts that address similar problems in the future (Hevner 2007).

So far, I analysed the body of knowledge on collective intelligence, idea contests, and crowd-based evaluation as well as computational filtering approaches and identified five theory-driven problems of current idea filtering approaches that adversely affects evaluation accuracy. These problems represent the starting point for my solution design. Based on deductive reasoning, I derived five DPs for a potential solution that I evaluated in an ex-ante criteria-based evaluation with experts in the field of community- and service -engineering. In the next steps, I will develop a prototype version of the novel filtering technique and implement it within the context of an idea contest. By conducting an A/B-test to compare the accuracy of my filtering approach against current filtering approaches, I intend to evaluate my proposed design. This also constitutes my summative design evaluation. I will, therefore, use a consensual assessment of experts as baseline (Blohm et al. 2016). Finally, the abstract learning from my design will provide prescriptive knowledge in the form of principles of form and function for building similar artefacts in the future.

### 5.4.5. Awareness of Limitations of Computational and Crowd Approaches

One solution that is currently employed in idea contests is shortlisting. Shortlisting can be considered as an algorithmic solution with the aim to shortlist the best opportunities. In doing so shortlisting algorithms often face a trade-off between specificity and sensitivity. Thus, if such algorithms are not balanced out (i.e. they are too specific, or they are too sensitive) this may lead to opportunities being shortlisted that are not innovative (i.e. the algorithm might include false positives) or to promising opportunities not being shortlisted (i.e. the algorithm might favour false negatives). In both cases this might lead to unfavourable results such as opportunities that are labelled as innovative when in fact they are not truly innovative opportunities (**Problem 1**).

One limitation of previous crowd-based evaluation approaches is the cognitive load associated with the volume and variety of idea contributions in crowdsourcing (Blohm et al. 2016; Nagar et al. 2016). As cognitive load increases, users in the crowd may become frustrated make low quality decisions or simply deny evaluating opportunities (Nagar et al. 2016) . Such load may arise due to the complexity of the evaluation mechanism itself (e.g. prediction markets) and the increasing time and cognitive complexity demands for the raters. Moreover, the information overload in which cognitive processing capacity is exceeded by the volume and diversity of the crowdsourcing contributions makes it difficult for the crowd to evaluate each idea especially when the proposals are complex, such as in the context of innovation problems. Thus, users need to judge manifold, diverse, maybe even paradox opportunities with a high degree of novelty. This cognitive load renders previous approaches of crowd-based evaluation problematic for use in practice, where the number of contributions is large (**Problem 2**).

Furthermore, contributions will vary in their textual representation such as writing style, schema, or language which accelerates the cognitive demands on the crowd. Consequently, in practice only a small number of contributions are evaluated. These contributions and their (positive) evaluations then create an anchoring effect (Kahneman 2011) and will socially influence other decision makers in the crowd (Deutsch and Gerard 1955). Generally, the ones that are presented on the top of the page and have been positively evaluated by peers a priori, which creates (potentially negative) information cascades (Klein and Garcia 2015) (**Problem 3**).

Another major problem in crowd-based evaluation methods so far is that not all users in an idea contest are necessarily capable to evaluate opportunities. Therefore, the crowd-based evaluation results might not be a proxy for expert ratings, if users do not have the required expertise for being a *"judge"* (Magnusson et al. 2016; Ozer 2009). This is particularly problematic when crowdsourcing contributions are complex and diverse. Although previous research highlighted the requirements on the crowd for evaluating opportunities, the bottleneck of domain expertise is almost neglected in both theory and practice. To be appropriate for identifying valuable opportunities and improving decision quality and predictions in idea filtering, a user should also be an expert in the field (Keuschnig and Ganser 2016). Therefore, the crowd should combine both problem knowledge as well as solution knowledge (Hippel 1994) , which are crucial in the evaluation of innovation. While knowledge about the problem domain might be assumed for users that contribute an idea to a specific problem call, the variety of submitted solutions might be enormous as each diverse solver within the crowd deeply know different parts of the potential solution landscape (Faraj et al. 2016). Therefore, not every user in the crowd is equally appropriate to evaluate a certain idea due to limited domain knowledge of each part of the solution space submitted, which represents a major weakness of previous approaches in crowd-based evaluation (**Problem 4**).

## 5.4.6. Suggestion and Development of DPs for a Hybrid Filtering Approach

To overcome the limitations of previous approaches and to define objectives for a potential solution, I combine algorithmic approaches from ML with crowd-based evaluation approaches rather than treat them as substitutes. This approach enables my solution to support the human judge by using ML algorithms that identify the expertise of a crowd user, the expertise requirements for evaluating a specific crowdsourcing contribution, and match both to gather more reliable results in identifying valuable contributions. My proposed DPs mainly focus on improving the idea evaluation phase in innovation contests (see Figure 28).

First, the expertise requirements for each textual contribution needs to be identified to match it with suitable members of the crowd (Ozer 2009). Therefore, the hybrid filtering approach should extract topical features (i.e. latent semantics) to identify the knowledge requirements for potential judges. Thus, I propose:

**DP1:** *Filtering crowdsourcing contributions should be supported by approaches that extract solution knowledge requirements from textual idea contributions within an idea contest by identifying relevant themes.*

In the next step, the hybrid filtering approach needs to consider the expertise of a crowd participant (Kulkarni et al. 2014). One source of such expertise description is the user profile, which includes the self-selected proficiency of a participant. Thus, I propose:

**DP2:** *Filtering crowdsourcing contributions should be supported by approaches that screen user profiles to extract expertise.*

Apart from the expertise description in the users´ profile (i.e. static), crowd participants gain ability through their activity (i.e. dynamic) in idea contests over time. Users constantly learn through their own contributions (Yuan et al. 2016a). This needs to be additionally considered for the hybrid filtering approach. Moreover, this offers the possibility to ensure that users have really expertise in a topic as they proved it by making contributions. In contrast, expertise descriptions in user profiles might be biased due to overconfidence. Thus, I propose:

**DP3:** *Filtering crowdsourcing contributions should be supported by approaches that extract solution expertise from users´ prior textual idea contributions across idea contests by identifying relevant themes.*

Idea contest are highly dynamic (Blohm et al. 2013). To match crowd participants with suitable opportunities for evaluation, the expertise profiles of each user need to be dynamic (Yuan et al. 2016a). This means it should constantly update the expertise of a user through dynamically updating the abstract user profile based on the input and contributions of a crowd participant. Contributions include both past idea proposals, as well an idea quality indicator (i.e. the corresponding idea rating) Thus, I propose:

**DP4:** *Filtering crowdsourcing contributions should be supported by approaches that create adaptive user profiles containing expertise extracted from the user profile and prior contributions.*

As the evaluation quality of the crowd is highly dependent on the ability of each individual member of the crowd (Mannes et al. 2014), in the last step the hybrid filtering approach needs to match crowdsourcing contributions with suitable users. Previous work on such select crowd strategies in the field of psychology suggests that approximately five to ten humans are required to benefit from the aggregated results of evaluation (Keuschnigg and Ganser 2016). This sample size is most suitable for leveraging the error reduction of individual biases as well as the aggregation of diverse knowledge. Thus, I propose:

**DP5:** *Filtering crowdsourcing contributions should be supported by approaches that match solutions with users that have the required expertise and assign textual contributions to this user for evaluation.*

Figure 28 illustrates how the DPs relate to each other. Same topics are represented by the same colour codes. The solution is designed in a way that it allows to match topics (i.e. expertise) that are extracted from a static user profile and a dynamic user profile (i.e. past idea proposals). The adaptive profile thus includes both the self-reported topics of their expertise, as well as expertise that individuals acquired in past idea proposals. These extracted topics are then matched with topics of the current idea proposals.

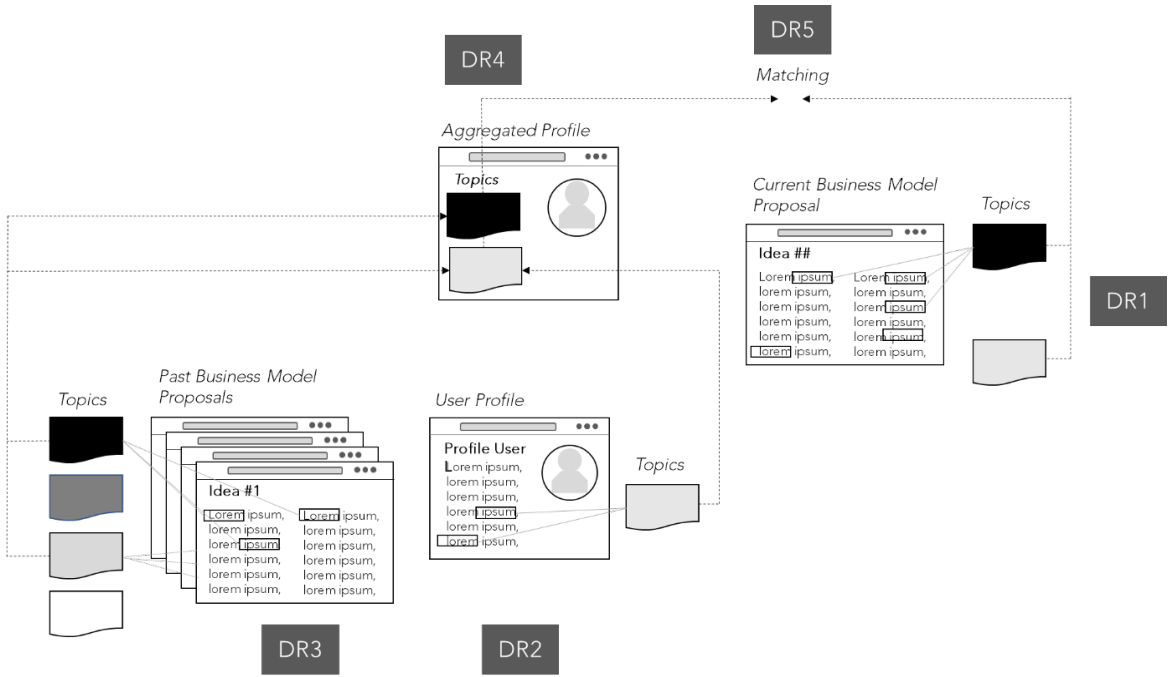

*Crowd Expertise Matching Approach*

## 5.4.8. Conclusion

My research introduces a novel filtering approach that combines the strengths of both machines and humans in evaluating creative opportunities by using ML approaches to assign the right user with the required solution knowledge to a corresponding idea. To this end, I propose tentative DPs that I validated in the field with experts on crowdsourcing and system engineering. To the best of my knowledge, this is the first study that takes this topic into account. My research offers a novel and innovative solution for a real-world problem and contribute to the body of knowledge on idea filtering for open innovation systems by considering the required expertise of crowd evaluations (Bayus 2013; Klein and Garcia 2015; Keuschnigg and Ganser 2016). I, therefore, intend to extend previous research on idea filtering in crowdsourcing engagements through combining algorithmic and crowd-based evaluation.

# Chapter IV

## Solution II: Hybrid Intelligence Decisional Guidance

# 6. Solution II: Hybrid Intelligence Decisional Guidance Design Paradigms and Design Principles

## Purpose and Findings

The purpose of Chapter IV is to propose a novel and innovative design paradigm for decisional guidance in contexts of uncertainty and risk: hybrid intelligence. This new paradigm has the potential to augment human provided decisional guidance and offers a first step towards solving wicked AI-complete problems for DSSs. Therefore, I start with conceptually developing this new paradigm and providing a rational for its applicability in such contexts (Section 6.1).

In the next step, I develop a structured taxonomy as classification schema for existing design knowledge that can be leveraged to build hybrid intelligence systems in general. This supported the design of the hybrid intelligence prediction method in Section 6.4 and the HI-DSS in Section 6.5.

To apply such hybrid intelligence and especially ML techniques for the context of entrepreneurial decision-making, I once again used the business model design as central decision problem and investigated design pattern that allow to provide data driven guidance for entrepreneurs.

In Section 6.4 I develop design principles for a generic method for making predictions under uncertainty by combining human intelligence and AI. Therefore, I propose a method for combining both predictions to achieve superior outcomes. This is then used as the hybrid intelligence core for the HI-DSS.

Section 6.5. then uses all the knowledge from the previous Chapters of this thesis to develop design principles that are used for building a DSS based on the novel design paradigm of hybrid intelligence.

**Relevance for Dissertation**

The findings of this Chapter IV are an improved version of design paradigms and design principles for decisional guidance in contexts under uncertainty and risk. By combining the complementary capabilities of human intelligence and AI, I propose novel design knowledge on the development of decisional guidance and DSS for complex decision-making problems. This Chapter, thus, contains the main findings of this thesis.

## 6.1. Conceptualizing Hybrid Intelligence for Decisional Guidance

The findings of this chapter are accepted for publication at Business & Information Systems Engineering (BISE). This part of the dissertation conceptually develops the idea of hybrid intelligence, an approach that combines human (collective) and AI to provide guidance in contexts of extreme uncertainty such as entrepreneurial decision-making. The findings of this study build the conceptual core for the design of the HI-DSS in Section 6.5.

### 6.1.1. Introduction

Research has a long history of discussing what is superior in predicting certain outcomes: statistical methods or the human brain. This debate has come up again and again due to the remarkable technological advances in the field of artificial intelligence (AI), such as solving tasks like object and speech recognition, achieving significant improvements in accuracy through deep-learning algorithms (Goodfellow et al. 2016) or combining various methods of computational intelligence, such as fuzzy logic, genetic algorithms, and case-based reasoning (Medsker 2012). One of the implicit promises that underlie these advancements is that machines will one day can perform complex tasks or may even supersede humans in performing these tasks. This heats up new debates of when machines will ultimately replace humans (McAfee and Brynjolfsson 2017). While previous research proves that AI performs well in some clearly defined tasks such as playing chess, Go or identifying objects on images, artificial general intelligence (AGI), however, which is able to solve multiple tasks at the same time, is doubted to be achieved in the near future (e.g. Russel and Norvig 2016). Moreover, using AI to solve complex business problems in organizational contexts occurs rather scarcely and applications for AI that solve complex problems remain mainly in laboratory settings than in the "wild."

Since the road towards AGI is still a long one, I argue that the most likely paradigm for the division of labour between humans and machines in the next year or probably decades is, thus, Hybrid Intelligence. This concept aims at using the complementary strengths of human intelligence and AI, so they behave more intelligently than each of the two could separately (e.g. Kamar 2016).

## 6.1.2. Conceptual Integration of Hybrid Intelligence

Before focusing on Hybrid Intelligence in detail, I first want to delineate the concept from related but still different forms of intelligence in this context.

### Intelligence

Various definitions and dimensions (e.g. social, logical, spatial, musical etc.) of the term intelligence exist in multiple research disciplines, such as psychology, cognitive science, neuro science, human behavior, education, or computer science. To my research, I use an inclusive and generic definition of describing general intelligence. It is the ability to accomplish complex goals, learn, reason, and adaptively perform effective actions within an environment. This can be subsumed with the capacity to both acquire and apply knowledge (Gottfredson 1997). While intelligence is most commonly used in the context of humans (and more recently intelligent artificial agents), it also applies to intelligent, goal-directed behavior of animals.

### Human Intelligence

The sub-dimension of intelligence that is related to the human species defines the mental capabilities of human beings. On the most holistic level, it covers the capacity to learn, reason, and adaptively perform effective actions within an environment, based on their knowledge. This allows humans to adapt to changing environments and act towards achieving their goals.

While one assumption of intelligence is the existence of a so-called *"g-factor"*, which indicates a measure for general intelligence (Brand 1996), other research in the field of cognitive science explores intelligence in relation to the evolutionary experience of individuals. This means that, rather than having a general form of intelligence, humans become much more effective in solving problems that occur in the context of familiar situations (Wechsler 1964).

Another view on intelligence supposes that general human intelligence can be subdivided into specialized intelligence components, such as linguistic, logical-mathematical, musical, kinesthetics, spatial, social, or existential intelligence (Gardner 2000).

Synthesizing those perspectives on human intelligence, Sternberg (1985) proposes three distinctive dimensions of intelligence: componential, contextual, and experiential. The componential dimension of intelligence refers to some kind of individual (general) skill set of humans. Experiential intelligence refers to oneś ability to learn and adapt through evolutionary experience. Finally, contextual intelligence defines the capacity of the mind to inductively understand and act in specific situations as well as the ability to make choices and modify those contexts.

**Collective Intelligence**

The second related concept is collective intelligence. According to Malone and Bernstein (2015:3), collective intelligence refers to *"[...] groups of individuals acting collectively in ways that seem intelligent [...]"*. Even though the term *"individuals"* leaves room for interpretation, researchers in this domain usually refer to the concept of wisdom of crowds and, thus, a joint intelligence of individual human agents (Woolley et al. 2010). This concept describes that, in certain conditions, a group of average people can outperform any individual of the group or even a single expert (Leimeister 2010). Other well-known examples of collective intelligence are phenomena found in biology, where, for example, a school of fish swerves to increase protection against predators (Berdahl et al. 2013). These examples show that collective intelligence typically refers to large groups of homogenous individuals (i.e. humans or animals), while Hybrid Intelligence combines the complementary intelligence of heterogeneous agents (i.e. humans and AI).

## Artificial Intelligence

The subfield of intelligence that relates to machines is called artificial intelligence (AI). With this term, I mean systems that perform *"[…] activities that I associate with human thinking, activities such as decision-making, problem solving, learning […]"* (Bellman 1978:3). It covers the idea of creating machines that can accomplish complex goals. This includes facets, such as natural language processing, perceiving objects, storing of knowledge, applying this knowledge for solving problems, machine learning to adapt to new circumstances and act in its environment (Russel and Norvig 2016). Other definitions in this domain focus on AI as the field of research about the *"[…] synthesis and analysis of computational agents that act intelligently […]"* (Poole and Mackworth 2017:3). Moreover, AI can be defined as having the general goal to replicate the human mind by defining it as *"[…] the art of creating machines that perform functions that require intelligence when performed by people […]"* (Kurzweil 1990: 117). The performance of AI in achieving human-level intelligence can then be measured by, for instance, the Turing test. This task asks an AI program to simulate a human in a text-based conversation. Such capabilities can be seen as a sufficient but not necessary criterion for artificial general intelligence (Searle 1980).

Synthesizing those various definitions in the field, AI includes elements such as the human-level ability to solve domain-independent problems, the capability to combine highly task-specialized and more generalized intelligence, the ability to learn from its environment and interaction with other intelligent systems, or human teachers, which allows intelligent agents to improve in problem solving through experience.

To create such a kind of AI in intelligent agents, various approaches exist that are more or less associated with the understanding and replication of intelligence. For instance, the field of cognitive computing *"[...] aims to develop a coherent, unified, universal mechanism inspired by the mind's capabilities. [...] I seek to implement a unified computational theory of the mind [...]"* (Modha et al. 2011:60). Therefore, interdisciplinary researcher teams rely on the reverse-engineering of human learning to create machines that *"[...] learn and think like people [...]"* (Lake et al. 2017:1).

### 6.1.3. The Complementary Benefits of Humans and AI

The general rationale behind the idea of Hybrid Intelligence is that humans and computers have complementary capabilities that can be combined to augment each other. The tasks that can be easily done by artificial and human intelligence are quite divergent. This fact is known as Moravecś paradox (1988:15), which describes the fact that *"[...] it is comparatively easy to make computers exhibit adult level performance on intelligence tests or playing checkers, and difficult or impossible to give them the skills of a one-year-old when it comes to perception and mobility [...]."* This is especially represented by the human common sense that is challenging to achieve in AI (Lake et al. 2017).

This can be explained by the separation of two distinctive types of cognitive procedures (Kahneman 2011). The first, system 1, is fast, automatic, affective, emotional, stereotypic, subconscious and it capitalizes on what one might call human intuition. The second one, system 2 reasoning, is rather effortful, logical, and conscious and is ideally following strict rational rules of probability theory. In the context of complementary capabilities of human and AI, humans proved to be superior in various settings that need system 1 thinking. Humans are flexible, creative, empathic, and can adapt to various settings. This allows, for instance, human domain experts to deal with so called *"broken-leg"* predictions that deviate from the currently known probability distribution. However, they are restricted by bound rationality that prevents them from aggregating information perfectly and drawing conclusions from that. On the other hand, machines are particularly good at solving repetitive tasks that require fast processing of huge amounts of data, recognizing complex patterns, or weighing multiple factors following consistent rules of probability theory. This has been proven by a long-standing tradition of research that shows the superiority. Even in very simple actuarial models, they outperform human experts in making predictions under uncertainty (Meehl 1954).

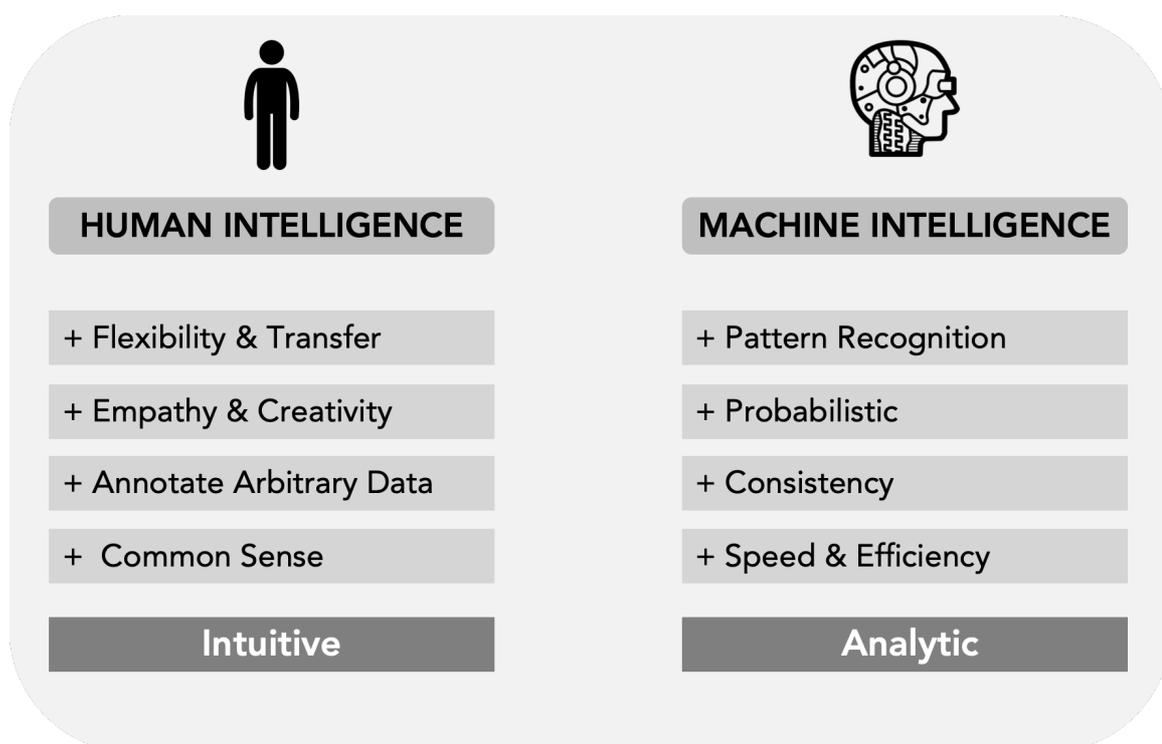

*Complementary Strengths of Humans and Machines*

These complementary benefits of humans and machines has since led to the application of both, that is, AI is in the loop of human intelligence, which improves human decisions by providing predictions, and humans are in the loop of AI, which is frequently applied to train machine learning models.

## AI in the Loop of Human Intelligence

Currently, in typical business contexts, AI is applied in two areas. First, they are used in automating tasks that can be solved by machines alone. While this is often associated with the fear of machines taking over jobs and making humans obsolete in the future, it might also allow humans to solve tasks they do not want to do. Second, AI is applied to provide humans with decision support by offering predictions. This ranges from structuring data, making forecasts, for example, in financial markets, or even predicting the best set of hyperparameters to train new machine learning models (e.g. AutoML). As humans often act non-Bayesian by violating probabilistic rule and thus making inconsistent decisions, AI has proven to be a valuable tool to help humans in making better decisions (Agrawal et al. 2018). The goal in this context is to improve human decision effectiveness and efficiency.

For instance, in settings where AI provides the human with input that is then evaluated to decide, humans and machines act as teammates. For instance, AI can help human physicians by processing patient data (e.g. CT scans) to make predictions on diseases such as cancer, empowering the doctor to learn from the added guidance. In this context, the Hybrid Intelligence approach allows human experts to use the predictive power of AI while using their own intuition and empathy to make a choice from the predictions of the AI.

## The Human in the Loop of AI

On the other hand, human intelligence also has a crucial role in the loop of machine learning and AI. Humans aid in several parts of the machine learning process to support AI in tasks that it cannot (yet) solve alone. Here, humans are most commonly used for the generation of algorithms (e.g. hyperparameter setting/tuning), training or debugging models and making sense of unsupervised approaches such as data clustering.

AI systems can help and learn from human input. This approach allows to integrate human domain knowledge in the AI to design, complement and evaluate the capabilities of AI (Mnih et al. 2015). Many of these applications are based on supervised and interactive learning approaches and need an enormous amount of labelled data, provided by humans (Amershi et al. 2014). The basic rational behind this approach is that humans act as teachers who train an AI. The same machine teaching approach can also be found in the area of reinforcement learning that used, for instance, human game play as input to initially train robots. In this context, human intelligence functions as a teacher, augmenting the AI. Hybrid Intelligence allows to distribute computational tasks to human intelligence on demand (e.g. through crowdsourcing) to minimize shortcomings of current AI systems. Such human-in-the-loop approaches are particularly valuable when only little data is available to date, pre-trained models need to be adapted for specific domains, the stakes are high, there is a high level of class imbalance, or in contexts where human annotations are already used.

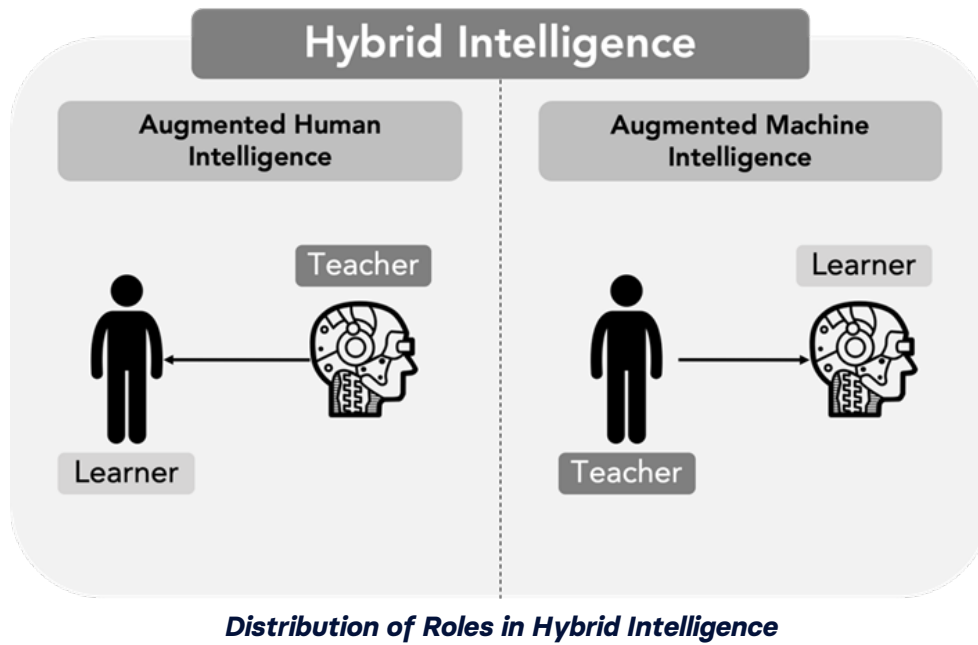

*Distribution of Roles in Hybrid Intelligence*

As humans in the loop of AI am most frequently applied in settings where models are initially set up or in the field of research, the goal is to make AI more effective.

## 6.1.4. Defining Hybrid Intelligence

One other approach beyond trying to replicate human level intelligence and related learning mechanisms is to combine human and artificial intelligence. The basic rational behind this is the combination of complementary heterogeneous intelligences (i.e. human and artificial agents) into a socio-technological ensemble that is able to overcome the current limitations of artificial intelligence. This approach is neither focusing on the human in the loop of AI nor automating simple tasks through machine learning but on solving complex problems using the deliberate allocation of tasks among different heterogeneous algorithmic and human agents. Both the human and the artificial agents of such systems can then co-evolve by learning and achieve a superior outcome on the system level.

I call this concept Hybrid Intelligence, which is defined as *"[...] the ability to accomplish complex goals by combining human and artificial intelligence to collectively achieve superior results than each of them could have done in separation and continuously improve by learning from each other [...]"* (Dellermann et al. 2019:3). Several core concepts of this definition are noteworthy:

(1) **Collectively:** Hybrid Intelligence covers the fact that tasks are performed collectively. Consequently, activities conducted by each agent are conditionally dependent. However, their goals are not necessarily always aligned to achieve the common goal such as when humans are teaching an AI adversarial tasks in playing games.

(2) **Superior results:** defines the idiosyncratic fact that the socio-technical system achieves a performance in a specific task that none of the involved agents, whether they are human or artificial, could have achieved without the other. The aim is, therefore, to make the outcome (e.g. a prediction) both more efficient and effective on the level of the socio-technical system by achieving goals that could not have been solved before. This contrasts Hybrid Intelligence with the most common applications of human-in-the-loop machine learning.

(3) **Continuous learning:** a central aspect of Hybrid Intelligence is that over time this socio-technological system improves both as a whole and as each single component (i.e. humans and machine agents). This facet defines that they learn from each other through experience. The performance of Hybrid Intelligence systems can, thus, not only be isolated, measured by the superior outcome of the whole socio-technical system, but also by the learning (i.e. performance increase) of human and machine agents that are parts of the system.

One recent example that provides an astonishing indicator for the potential of Hybrid Intelligence is DeepMind[1]ś AlphaGo. For training the game -playing AI, a supervised learning approach was used that learned from expert human moves and, thus, augmented the AI through human input, which allowed AlphaGo to achieve superhuman performance over time. During its games against various human world-class players, AlphaGo played several highly inventive moves that previously were beyond human players´ imagination. Consequently, AlphaGo was able to augment human intelligence as well and somehow taught expert players completely new knowledge in a game that is one of the longest studied in human history (Silver et al. 2015).

*"I believe players more or less have all been affected by Professor Alpha. AlphaGo's play makes us feel freer and no move is impossible to play anymore. Now everyone is trying to play in a style that hasn't been tried before."* – **Zhou Ruiyang, 9 Dan Professional, World Champion**

Solving problems through Hybrid Intelligence offers the possibility to allocate a task between humans and intelligent agents and deliberately achieve a superior outcome on the socio-technical system level by aggregating the output of its parts. Moreover, such systems can improve over time by learning from each other through various mechanisms, such as labelling, demonstrating, teaching adversarial moves, criticizing, rewarding and so on. This will allow us to augment both the human mind and the AI and extend applications when men and machines can learn from each other in much more complex tasks than games: for instance, strategic decision-making, managerial, political, or military decisions, science and even AI development leading to AI reproducing itself in the future. Hybrid Intelligence, therefore, offers the opportunity to achieve super-human levels of performance in tasks that so far seem to be the core of human intellect.

---

[1] https://deepmind.com

### 6.1.5. The Advantages of Hybrid Intelligence

This hybrid approach provides various advantages for humans in the era of AI such as generating new knowledge in complex domains that allow humans to learn from AI and transfer implicit knowledge from experienced experts to novices without some kind of social interaction. On the other hand, the human teaching approach allows to control the learning process by ensuring that the AI makes inferences based on humanly interpretable criteria – a fact that is crucial for AI adoption in many real-world applications and AI safety and that allows to exclude biases such as racism and so forth (Bostrom 2017). Moreover, such hybrid approaches might allow for a better customization of AI, based on learning the preferences of humans during interaction. Finally, I argue that the co-creation of Hybrid Intelligence services between humans and intelligent agents might create a sense of psychological ownership and, thus, increasing acceptance and trust.

### 6.1.6. Future Research Directions in the Field of Hybrid Intelligence

As technological advances further, the focus of machine learning and Hybrid Intelligence is shifting towards applications in real-world business contexts, solving complex problems will become the next frontier. Such complex problems in managerial settings are typically time variant, dynamic, require much domain knowledge and have no specific ground truth. These highly uncertain contexts require intuitive and analytic abilities and further human strengths such as creativity and empathy. Consequently, I propose three specifics but also interrelated directions for further development of the concept that are focused on socio-technical system design.

A core requirement for integrating human input into an AI system is interaction design. For instance, semi-autonomous driving requires the AI to sense the human state to distribute tasks between itself and the human driver. Furthermore, it requires examining human-centred AI architectures that balance, for instance, transparency of the underlying model and its performance, or create trust among users. However, domain specific design guidelines for developing user-interfaces that allow humans to understand and process the needs of an artificial system are still missing. I, therefore, believe that more research is needed to develop suitable human-AI interfaces as well as to investigate possible task and interface designs that allow human helpers to teach an AI system (e.g. Simard et al. 2017). Interpretability and transparency of machine learning models while maintaining accuracy is one of the most crucial challenges in research on Hybrid Intelligence. This was most recently covered by the launch of the People + AI Research (PAIR)2 group at Google.

Second, research in the field of Hybrid Intelligence might investigate how mechanisms of traditional crowdsourcing strategies can be used to train and maintain Hybrid Intelligence systems. Such tasks frequently require domain expertise (e.g. health care) and, thus, crowdsourcing needs to focus on explicitly matching experts with tasks, aggregating their input and assuring quality standards. I, therefore, argue that it might be a fruitful area of research to further investigate how current forms of crowdsourcing and platforms should evolve. Moreover, human teachers may have different motivations to contribute to the system. Consequently, research in the field tries to shed light on the question of how to design the best incentive structure for a predefined task. Especially, when highly educated and skilled experts are required to augment AI systems, the question arises if traditional incentives of micro-tasking platforms (e.g. monetary reward) or online communities (e.g. social rewards) are sufficient.

---

2 https://ai.google/research/teams/brain/pair

A third avenue for future research is related to digital work. The rise of AI is now changing the capabilities of IS and the potential distribution of tasks between human and IS dramatically and, hence, affects the core of my discipline. Those changes create novel qualification demands and skill sets from employees and, consequently, provide promising directions for IS education. Such research might examine the educational requirements for democratizing the use of AI in future workspaces. Finally, Hybrid Intelligence also offers great possibilities for novel forms of digital work such as internal crowd work to leverage the collective knowledge of individual experts that resides within a company across functional silos.

## 6.2. Deriving Design Knowledge for Hybrid Intelligence Systems

The findings of this chapter have been previously published as Dellermann et al. (2019). This part of the dissertation derives design knowledge on hybrid intelligence information systems in general. By combining a literature review and a taxonomy development method, I structured existing interdisciplinary knowledge that can be developed for the design of such systems. The findings of this study provide knowledge for the design of the HI-DSS in Section 6.5.

### 6.2.1. Introduction

Recent technological advances, especially in the field of deep learning, provide astonishing progress on the road towards AGI (Goertzel and Pennachin 2007; Kurzweil 2010). AI is progressively achieving (super-) human level performance in various tasks, such as autonomous driving , cancer detection, or playing complex games (Mnih et al. 2015; Silver et al. 2016). Therefore, more and more business applications that are based on AI technologies arise. Both research and practice are wondering when AI will be capable of solving complex tasks in real-world business applications apart from laboratory settings in research.

However, those advances provide a rather one-sided picture on AI, denying the fact that although AI is capable to solve certain tasks with quite impressive performance, AGI is far away from being achieved. There are lots of problems that machines cannot solve alone yet (Kamar 2016), such as applying expertise to decision-making, planning, or creative tasks, just to name a few. ML systems in the wild have major difficulties with being adaptive to dynamic environments and self-adjusting (Müller-Schloer and Tomforde 2017), and the lack of what humans call common sense. This makes them highly vulnerable for adversarial examples (Kurakin et al. 2016). Moreover, AGI needs massive amounts of training data compared to humans, who can learn from only few examples (Lake et al. 2017) and fails to work with certain data types (e.g. soft data). Nevertheless, a lack of control of the learning process might lead to unintended consequences (e.g. racism biases) and limit interpretability, which is crucial for critical domains such as medicine (Doshi-Velez and Kim 2017). Therefore, humans are still required at various positions in the loop of the ML process. While a lot of work has been done in creating training sets with human labellers, more recent research points towards end user involvement (Amershi et al. 2014) and teaching of such machines (Mnih et al. 2015), thus, combining humans and machines in hybrid intelligence systems.

The main idea of hybrid intelligence systems is, thus, that socio-technical ensembles and its human and AI parts can co-evolve to improve over time. The purpose of this paper is to point towards such hybrid intelligence systems. Thereby, I aim to conceptualize the idea of hybrid intelligence systems and provide an initial taxonomy of design knowledge for developing such socio-technical ensembles. By following a taxonomy development method (Nickerson et al. 2013), I reviewed various literature in interdisciplinary fields and combine those findings with an empirical examination of practical business applications in the context of hybrid intelligence.

The contribution of this paper is threefold. First, I provide a structured overview of interdisciplinary research on the role of humans in the ML pipeline. Second, I offer an initial conceptualization of the term hybrid intelligence systems and relevant dimensions for system design. Third, I intend to provide useful guidance for system developers during the implementation of hybrid intelligence systems in real-world applications. Towards this end, I propose an initial taxonomy of hybrid intelligence systems.

## 6.2.2. ML and AI

The subfield of intelligence that relates to machines is called AI. With this term I mean systems that perform*" [. . .] activities that I associate with human thinking, activities such as decision-making, problem solving, learning [. . .]"* (Bellman 1978). Although various definitions exist for AI, this term generally covers facets, such as creating machines that can accomplish complex goals. This includes facets such as natural language processing, perceiving objects, storing of knowledge, and applying it for solving problems, and ML to adapt to new circumstances and act in its environment (Russell and Norvig 2016).

A subset of techniques that is required to achieve AI is machine learning (ML). (Mitchell 1997)defines it the following way: *"[...] A computer program is said to learn from experience E with respect to some class of tasks T and performance measure P, if its performance at tasks in T, as measured by P, improves with experience E [...]."* A popular approach that drives current progress in both paradigms is deep learning (Kurakin et al. 2016). Deep-learning constitutes a representation learning method that includes multiple levels of representation, obtained by combining simpler but non-linear models. Each of those models transform the representation of one level (starting with the input data) into a representation at more abstract level (LeCun et al. 2015). Deep learning is a special ML technique. Finally, human-in-the-loop learning describes ML approaches (both deep and other) that use the human in some part of the pipeline. Such approaches contrast with research on most knowledge-based systems in IS that use rather static knowledge repositories. I will focus on this in the following chapter.

### 6.2.3. The Role of Humans-in-the-Loop of ML

Although, the terms of AI and ML give the impression that humans become to some extent obsolete, the ML pipeline still requires lot of human interaction such as for feature engineering, parameter tuning, or training. While deep learning has decreased the effort for manual feature engineering and some automation approaches (e.g. AutoML (Feurer et al. 2015)) support human experts in tuning models, the human is still heavily in the loop for sense-making and training. For instance, unsupervised learning requires humans to make sense of clusters that are identified as patterns in data to create knowledge (Gomes et al. 2011). More obviously, human input is required to train models in supervised ML approaches, especially for creating training data, debug models, or train algorithms such as in reinforcement learning (Mnih et al. 2015). This is especially relevant when divergences of real-life and ML problem formulations emerge. This is, for instance, the case when static (offline) training datasets are not perfectly representative of realist and dynamic environments (Kleinberg et al. 2017). Moreover, human input is crucial when models need to learn from human preferences (e.g. recommender systems) and adapt to users or when security concerns require both control and interpretability of the learning process and the output (Doshi-Velez and Kim 2017). Therefore, more recent research has focused on interactive forms of ML (Holzinger 2016) and machine teaching (Simard et al.

2017). Those approaches make active use of human input (Settles 2014) and, thus, learn from human intelligence. This allows machines to learn tasks that they cannot yet achieve alone (Kamar 2016), adapt to environmental dynamics, and deal with unknown situations (Attenberg et al. 2015).

## 6.2.4. Hybrid Intelligence

Rather than using the human just in certain parts and time during the process of creating ML models, applications that can deal with real-world problems require a continuously collaborating socio-technological ensemble integrating humans and machines, which is contrast to previous research on decision support and expert systems (Holzinger 2016).

Therefore, I argue that the most likely paradigm for the division of labour between humans and machines in the next years, or probably decades, is hybrid intelligence. This concept aims at using the complementary strengths of human intelligence and AI to behave more intelligently than each of the two could be in separation (Kamar 2016). The basic rational is to try to combine the complementary strengths of heterogeneous intelligences (i.e. human, and artificial agents) into a socio-technological ensemble.

I envision hybrid intelligence systems, which are defined as systems that can accomplish complex goals by combining human and AI to collectively achieve superior results than each of them could have done in separation and continuously improve by learning from each other.

The idea of hybrid intelligence systems is thus that socio-technical ensembles and its human and AI parts can co-evolve to improve over time. The central questions are, therefore, which and how certain design decisions should be made for implementing such hybrid systems rather than focusing.

## 6.2.5. Taxonomy Development Method

For developing my proposed taxonomy, I followed the methodological procedures of Nickerson et al. (2013). In general, a taxonomy is defined as a *"fundamental mechanism for organizing knowledge"* and the term is considered as a synonym to *"classification"* and *"typology"* (Nickerson et al., 2013). The method follows an iterative process consisting of the following steps:

(1) *defining a meta-characteristic;*
(2) *determining stopping conditions;*
(3) *selecting an empirical-to-conceptual or conceptual-to-empirical approach;*
(4) *iteratively following this approach, until the stopping conditions are met.*

The process of the taxonomy development starts with defining a set of meta-characteristics. This step limits the odds of *"naive empiricism"* where many characteristics are defined in search for random pattern and reflects the expected application of the taxonomy (Nickerson et al. 2013). For this purpose, I define those meta-characteristic as generic design dimensions that are required for developing hybrid intelligence systems. Based on my classification from literature, I choose four dimensions: task characteristics, learning paradigm, human-AI interaction, and AI-human interaction. In the second step, I selected both objective and subjective conditions to conclude the iterative process.

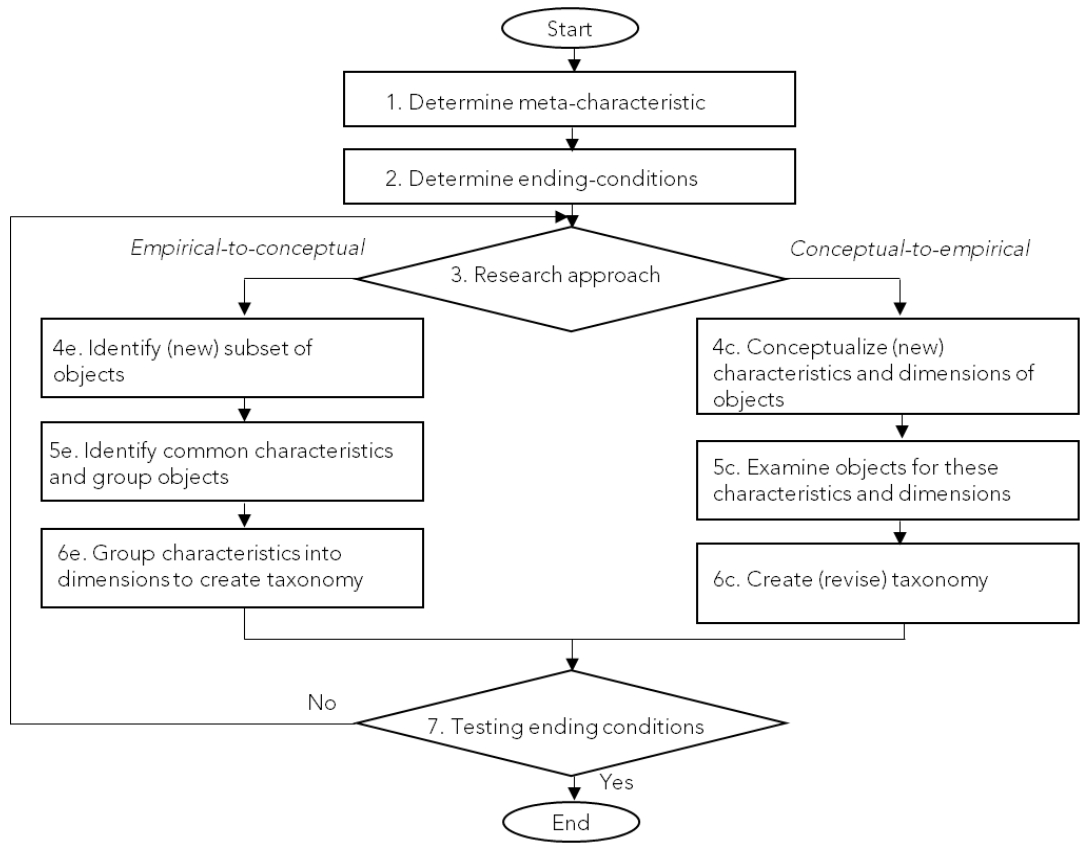

***Taxonomy Development Method***

The following conditions, adapted from Nickerson et al. (Nickerson et al. 2013), were selected: First, I applied the following objective conditions:

(5)     *All papers from the sample of the literature review and empirical cases are examined;*

(6)     *Then, at least one object is classified under every characteristic of every        dimension;*

(7)     *While performing the last iteration, no new dimension or characteristics        are added;*

(8)     *I treated every dimension as unique;*

(9)     *Lastly, every characteristic is unique within its dimension.*

The following subjective conditions were considered: conciseness, robustness, comprehensiveness, extensibility, explanatory, and information availability. I included no unnecessary dimension or characteristic (conciseness), whereas there are enough dimensions and characteristics to differentiate (robustness). At this point, all design decisions can be classified in the taxonomy (comprehensiveness), while still allowing for new dimensions and characteristics to be subsequently added (extensible). Furthermore, the information is valuable for guiding hybrid intelligence systems design decisions (explanatory) and is typically available or easily interpretable (information availability).

I conducted a total of three iterations so far. The first iteration used a conceptual-to-empirical approach, where I used extant theoretical knowledge from literature in various fields such as computer science, HCI, information systems, and neuro science to guide the initial dimensions and characteristics of the taxonomy.

Based on the identified dimensions of hybrid intelligence systems, I sampled seven real-world applications that make use of human and AI combinations. The second iteration used the empirical-to conceptual approach focuses on creating characteristics and dimensions based on the identification of common characteristics from a sample of AI applications in practice. The third iteration then used the conceptual-to-empirical approach, based on an extended literature review including newly identified search termini.

## 6.2.6. Data Sources and Sample

**Literature Review**

For conducting my literature review, I followed the methodological procedures of (Webster and Watson 2002; vom Brocke et al. 2009). The literature search was conducted from April to June 2018. A prior informal literature search revealed keywords for the database searches resulting in the search string *(" hybrid intelligence" OR" human-in-the-loop" OR" interactive machine learning" OR" machine teaching" OR" machine learning AND crowdsourcing" OR" human supervision" OR"* *human understandable machine learning" OR" human concept learning"*).

During this initial phase I decided to exclude research on knowledge-based systems such as expert systems or DSSs in IS (Kayande et al. 2009; Gregor 2001), as the studies either do not focus on the continuous learning of the knowledge repository or do not use ML techniques at all. Moreover, the purpose of this study is to identify and classify relevant (socio-) technical design knowledge for hybrid intelligence systems, which is also not included in those studies.

The database search was constrained to title, abstract, keywords and not limited to a certain publication. Databases include **AISeL, IEEE Xplore, ACM DL, AAAI DL,** and **arXiv** to identify relevant interdisciplinary literature from the fields of IS, HCI, bio-informatics, and computer science. The search resulted in a total of 2505 hits. Titles, abstracts, and keywords were screened for potential fit to the purpose of my study. Screening was conducted by three researchers independently and resulted in 85 articles that were reviewed in detail so far. A backward and forward search ensured the extensiveness of my results.

### Empirical Cases

| Application | Domain | Reference |
|---|---|---|
| **Teachable Machine** | Image Recognition | www.teachablemachine.withgoogle.com |
| **Cindicator** | Asset Management | www.cindicator.com |
| **vencortex** | Innovation Decisions | www.vencortex.com |
| **Cobi** | Conference Scheduling | www.projectcobi.com |
| **Stitch Fix** | Fashion | www.stitchfix.com |

| Alpha Go | Games | www.deepmind.com/research/alphago |
|---|---|---|
| **Custom Decision Service** | E-Commerce | www.portal.ds.microsoft.com |

*Empirical Cases for Taxonomy Development*

To extend my findings from literature and provide empirical evidence from recent (business) applications of hybrid intelligence systems, I include an initial set of seven empirical applications that was analysed for enhancing my taxonomy.

## 6.2.7. Design Knowledge on Hybrid Intelligence Systems

My taxonomy of hybrid intelligence systems is organized along the four meta-dimensions task characteristics, learning paradigm, human-AI interaction, and AI-human interaction. Moreover, I identified 16 sub-dimensions and a total of 50 categories for the proposed taxonomy. For organizing the dimensions of the taxonomy, I followed a hierarchical approach following the sequence of the design decisions that are necessary to develop such systems. The design knowledge is displayed in Figures 32-35.

### Task Characteristics

The goal of hybrid intelligence is to create superior results through a collaboration between humans and machines. The central component that drives design decisions for hybrid intelligence systems is the task, that humans and machines solve collaboratively. Task characteristics focus on how the task itself is carried out (Reynolds and Miller 2003). In context of hybrid intelligence systems, I identify the following four important tasks characteristics.

**Type of Task:** The task to be solved is the first dimension that must be defined for developing hybrid intelligence systems. In this context, I identified four generic categories of tasks: recognition, prediction, reasoning, and action. First, recognition defines tasks that recognize for instance objects (LeCun et al. 2015), images, or natural language (Hinton et al. 2012). On an application level such tasks are used for autonomous driving or smart assistants such as Alexa, Siri, or Duplex. Second, prediction tasks aim at predicting future events based on previous data such as stock prices or market dynamics (Choudhary et al. 2018). The third type of task, reasoning, focuses on understanding data by for instance inductively building (mental) models of a certain phenomenon and therefore make it possible to solve complex problems with small amount of data (Lake et al. 2017). Finally, action tasks are characterized as such that require an agent (human or machine) to conduct a certain kind of action (Mao et al. 2016).

**Goals:** The two involved agents, the human and the AI, may have a common" goal" like solving a problem through the combination of the knowledge and abilities of both. An example for such common goals is research on recommender systems (e.g. Netflix (Gomez-Uribe and Hunt 2016)), which learn a user's decision model to offer suggestions. In other contexts, the agents' goals may also be adversarial. For instance, in settings where AIs try to beat human in games such as IBMs Watson in the game of Jeopardy! (Ferrucci et al. 2010). In many other cases the goal of the human and the AI may also be independent for example when humans train image classifiers without being involved in the end solution.

**Shared Data Representation:** The shared data representation is what is the data that is shown to both the human and the machine before executing their tasks. The data can be represented in different levels of granularity and abstraction to create a shared understanding between humans and machines (Feldman 2003; Simard et al. 2017). Features describe phenomena in different kinds of dimensions like height and weight of a human being. Instances are examples of a phenomena which are specified by features. Concepts on the other hand are multiple instances that belong to one common theme, e.g. pictures of different humans. Schemas finally illustrate relations between different concepts (Gentner and Smith 2012).

**Timing in ML Pipeline:** The last sub-dimension describes the timing in the ML pipeline that focuses on hybrid intelligence. For this dimension I identified three characteristics: feature engineering, parameter tuning, and training. First, feature engineering allows the integration of domain knowledge in ML models. While more recent advances make it possible to fully automatically (i.e. machine only) learn features through deep learning, human input can be combined for creating and enlarging features such in the case of artist identification on images and quality classification of Wikipedia articles (Cheng and Bernstein 2015). Second, parameter tuning is applied to optimize models. Here ML experts typically use their deep understanding of statistical models to tune hyper-parameters or select models. Such human only parameter tuning can be augmented with approaches such as AutoML (Feurer et al. 2015) or neural architecture search (Real et al. 2018; Real et al. 2017) automate the design of ML models, thus, making it much more accessible for non-experts. Finally, human input is crucial for training ML models in many domains. For instance, large dataset such as ImageNet or the lung cancer dataset LUNA16 rely on human annotations. Moreover, recommender systems heavily rely on input of human usage behavior to adapt to specific preferences (e.g. (Amershi et al. 2014)) and robotic applications are trained by human examples (Mao et al. 2016).

**Learning Paradigm**

**Augmentation:** In general, hybrid intelligence systems allow three different forms of augmentation: human, machine, and hybrid augmentation. The augmentation of human intelligence is focused on typically applications that enable humans to solve tasks through the predictions of an algorithm such as in financial forecasting or solving complex problems (Doroudi et al. 2016). Contrary, most research in the field of ML focuses on leveraging human input for training to augment machines for solving tasks that they cannot yet solve alone (Kamar 2016). Finally, more recent work identified the great potential for simultaneously augmenting both at the same time through hybrid augmentation (Milli et al. 2017; Carter and Nielsen 2017) or the example of Alpha Go that started by learning from human game moves (i.e. machine augmentation) and finally offered hybrid augmentation by inventing creative moves that taught even mature players novel strategies (Baker et al. 2009; Silver et al. 2016).

**Machine Learning Paradigm:** The ML paradigm that is applied in hybrid intelligence systems can be categorized into four relevant subfields: supervised, unsupervised, semi-supervised, and reinforcement learning (Murphy 2012). In supervised learning, the goal is to learn a function that maps the input data x to a certain output data y, given a labelled set of input-output pairs. In unsupervised learning, such output y does not exist, and the learner tries to identify pattern in the input data x (Mitchell et al. 1997). Further forms of learning such as reinforcement learning, or semi-supervised learning can be subsumed under those two paradigms. Semi-supervised learning describes a combination of both paradigms, which uses both a small set of labelled and a large set of unlabelled data to solve a certain task (Zhu 2006). Finally, reinforcement learning. An agent interacts with an environment thereby learning to solve a problem through receiving rewards and punishment for a certain action (Mnih et al. 2015; Silver et al. 2016).

**Human Learning Paradigm**: Humans have a mental model of their environment, which gets updated through events. This update is done by finding an explanation for the event (Carter and Nielsen 2017; Milli et al. 2017; Lake et al. 2017). Human learning can therefore be achieved from experience and comparison with previous experiences (Kim et al. 2014; Gentner and Smith 2012) and from description and explanations (Hogarth 2011).

## Human-AI Interaction

**Machine Teaching:** defines how humans provide input. First, humans can demonstrate actions that the machine learns to imitate (Mao et al. 2016). Second, humans can annotate data for training a model for instance through crowdsourcing (Snow et al. 2008; Raykar et al. 2010). I designate that as a labelling. Third, human intelligence can be used to actively identify a misspecification of the learner and debug the model, which I define as troubleshooting (Nushi et al. 2017; Attenberg et al. 2015). Moreover, human teaching can take the form of verification whereby humans verify or falsify machine output (Pei et al. 2017).

**Teaching Interaction:** The input provided through human teaching, can be both explicit and implicit. While explicit teaching leverages active input of the user such as for instance labelling tasks such as image or text annotation (Li et al. 2017), implicit teaching learns from observing the actions of the user and thus adapts to their demands. For instance, Microsoft uses contextual bandit algorithms to suggest users content, using the actions of the user as implicit teaching interaction.

**Expertise Requirements:** Hybrid intelligence systems can have certain requirements for the expertise of humans that provides input for systems. While by now both most research and practical applications focus on human input from an ML expert (Chakarov et al. 2016; Kulesza et al. 2010; Patel et al. 2010), thus, requiring deep expertise in the field of AI. Moreover, end users can provide the system with input for product recommendations and e-commerce or input from human non-experts accessed through crowd work platforms (Chang et al. 2017; Chang et al. 2016; Nushi et al. 2017). More recent endeavours, however, focus on the integration of domain experts in hybrid intelligence architectures that leverage the profound understanding of the semantics of a problem domain to teach a machine, while not requiring any ML expertise (Simard et al. 2017).

**Amount of Human Input:** The amount of human input can vary between those of individual humans and aggregated input from several humans. Individual human input is for instance applied in recommender systems for individualization or due to cost efficiency reasons (Li et al. 2017). On the other hand, collective human input combines the input of several individual humans by leveraging mechanisms of human computation (Quinn and Bederson 2011). This approach allows to reduce errors and biases of individual humans and the aggregation of heterogeneous knowledge (Cheng and Bernstein 2015; Cheng et al. 2015; Zou et al. 2015).

**Aggregation:** When human input is aggregated from a collective of individual humans, different aggregation mechanisms can be leveraged to maximize the quality of teaching. First, unweighted methods can be used that use averaging or majority voting to aggregate results (Li et al. 2017)). Additionally, aggregation can be achieved by modelling the context of teaching through algorithmic approach such as expectation maximization, graphical models, entropy minimization, or discriminative training. Therefore, the aggregation can be human dependent focusing on the characteristics of an individual human (Kamar et al. 2012; Dawid and Skene 1979), or human-task dependent adjusting to the teaching task (Kosinski et al. 2014; Whitehill et al. 2009).

**Incentives:** Humans need to be incentivized to provide input in hybrid intelligence systems. Incentives can be monetary rewards such in the case of crowd work on platforms (e.g. Amazon Mechanical Turk), intrinsic rewards such as intellectual exchange in citizen science (Segal et al. 2018), fun in games with a purpose (Ahn and Dabbish 2008), learning (Vaughan 2017). Another incentive for human input is customization, which allows to increase individualized service quality for users that provide a higher amount of input to the learner (Bernardo et al. 2017; Amershi et al. 2014).

## AI-Human Interaction

This sub-dimension describes the machine part of the Interaction, the AI-human interaction. At first, which query strategy the algorithm used to learn. Second, I describe the feedback of the machine to humans. Third, I carry out a short explanation of interpretability to show the influence for hybrid intelligence.

**Query Strategy:** Offline query strategies require the human to finish her task completely before her actions are applied as input to the AI (Lin et al. 2014; Sheng et al. 2008). Handling a typical labelling task, the human would first need to go through all the data and label each instance. Afterwards the labelled Data is fed to a ML algorithm to train a model. In contrast, online query strategies let the human complete subtasks whose are directly fed to an algorithm, so that teaching and learning can be executed almost simultaneously (Chang et al. 2017; Nushi et al. 2017; Kamar et al. 2012). Another possibility is the use of active learning query strategies (Zhao et al. 2014; Settles 2014)Zhao and Zhu 2014). In this case, the human is queried by the machine when more input to give an accurate prediction is required.

**Machine Feedback:** Those four categories describe the feedback that humans receive from the machine. First, humans can get direct suggestions from the machine, which makes explicit recommendations to the user on how to act. For instance, recommender systems such as Netflix or Spotify provide such suggestions for users. Furthermore, systems can make suggestions for describing images (Nushi et al. 2017). Predictions as machine feedback can support humans e.g. to detect lies (Cheng and Bernstein 2015), predict worker behaviours (Kamar et al. 2012), or classify images. Thereby, this form of feedback provides a probabilistic value of a certain outcome (e.g. probability of some data x belonging to a certain class y). The third form of machine feedback is clustering data. Thereby, machines compare data points and put them in an order for instance to prioritize items (Kou et al. 2014), or organize data among identified pattern. Furthermore, another possibility of machine feedback is optimization. Machines enhance humans for instance in making more consistent decisions by optimizing their strategy (Chirkin and Koenig 2016).

**Interpretability:** For AI-Human interaction in hybrid intelligence systems interpretability is crucial to prevent biases (e.g. racism), achieve reliability and robustness, ensure causality of the learning, debugging the learner if necessary and for creating trust especially in the context of AI safety (Doshi-Velez and Kim 2017). Interpretability in hybrid intelligence systems can be achieved through algorithm transparency, which allows to open the black box of an algorithm itself, global model interpretability that focuses on the general interpretability of a ML model, and local prediction interpretability that tries to make more complex models interpretable for a single prediction (Lipton 2018).

## 6.2.8. Discussion

My proposed taxonomy for hybrid intelligence systems extracts interdisciplinary knowledge on human-in-the-loop mechanisms in ML and proposes initial descriptive design knowledge for the development of such systems that might guide developers. My findings reveal the manifold applications, mechanisms, and benefits of hybrid systems that might probably become of increasing interest in real-world applications in the future. My taxonomy of design knowledge offers insights on how to leverage the advantages of combining human and machine intelligence.

For instance, this allows to integrate deep domain insights into ML algorithms, continuously adapt a learner to dynamic problems, and enhance trust through interpretability and human control. Vice versa, this approach offers the advantage of improving humans in solving problems by offering feedback on how the task was conducted or the performance of a human during that task and machine feedback to augment human intelligence. Moreover, I assume that the design of such systems might allow to move beyond sole efficiency of solving tasks to combined socio-technical ensembles that can achieve superior results that could no man or machine have achieved so far. Promising fields for such systems are in the field of medicine, science, innovation, and creativity.

## 6.2.9. Conclusion

Within this paper I propose a taxonomy for design knowledge for hybrid intelligence systems, which presents descriptive knowledge structured along the four meta-dimensions task characteristics, learning paradigm, human-AI interaction, and AI-human interaction. Moreover, I identified 16 sub-dimensions and a total of 50 categories for the proposed taxonomy. By following a taxonomy development methodology (Nickerson et al. 2013), I extracted interdisciplinary knowledge on human-in-the-loop approaches in ML and the interaction between human and AI. I extended those findings with an examination of seven empirical applications of hybrid intelligence systems.

Therefore, my contribution is threefold. First, the proposed taxonomy provides a structured overview of interdisciplinary research on the role of humans in the ML pipeline by reviewing interdisciplinary research and extract relevant knowledge for system design. Second, I offer an initial conceptualization of the term hybrid intelligence systems and relevant dimensions for developing applications. Third, I intend to provide useful guidance for system developers during the implementation of hybrid intelligence systems in real-world applications.

Obviously, this paper is not without limitations and provides a first step towards a comprehensive taxonomy of design knowledge on hybrid intelligence systems. First, further research should extend the scope of this research to more practical applications in various domains. By now my empirical case selection is slightly biased on decision problem contexts. Second, as I proceed my research I will further condensate the identified characteristics by aggregating potentially overlapping dimensions in subsequent iterations. Third, my results are overly descriptive so far. As I proceed my research I will therefore focus on providing prescriptive knowledge on what characteristics to choose in a certain situation and thereby propose more specific guidance for developers of hybrid intelligence systems that combine human and machine intelligence to achieve superior goals and driving the future progress of AI. For this purpose, I will identify interdependencies between dimensions and sub-dimensions and evaluate the usefulness of my artefact for designing real-world applications. Finally, further research might focus on integrating the overly design oriented knowledge of this study with research on knowledge-based systems in IS to discuss the findings in the context of those class of systems.

## 6.3. A Data Driven Approach to Business Model Design

The findings of this chapter are currently submitted to California Management Review (CMR). Based on the requirements for domain specific ontologies to communicate the cognitive schema of an entrepreneur to the ecosystem to gather guidance in Section 5.2 and the conceptually developed DP in Section 5.3, this study first develops a domain specific business model representation for the IoT. In the next steps, configurations and archetypes of successful business model design choices are explored by applying ML techniques. Finally, the most important design choices are identified to guide entrepreneurial decision-making through a data driven approach. This approach allows provide data driven guidance for business model design. The findings of this Section are then used for the final design in Section 6.5.

### 6.3.1. Introduction

One key conceptualization highlighted by scholars to understand the mechanisms through which entrepreneurs can successfully commercialize new technologies is the business model concept (Chesbrough and Rosenbloom 2002);(Chesbrough 2007). Among entrepreneurs, the business model has been established as a blueprint to identify and assess opportunities for market exploitation (George and Bock 2011), especially in digital contexts (Nelson and Metaxatos 2016). Business models are systems of interrelated elements (Afuah and Tucci 2003; Massa et al. 2017; Afuah 2014; Baden-Fuller and Morgan 2010). Understanding their nature requires to look beyond the isolated elements towards their configuration (Klang et al. 2014; Lindner et al. 2010; Wirtz 2011). In this perspective firms achieve competitive advantage and superior financial performance when there exists *"fitness"* between these elements (Morris et al. 2005; Afuah and Tucci 2003; Teece 2010; Foss and Saebi 2017, 2018).

However, little is known on the role of configurations of single business model components and their complex interactions in influencing the performance of a firm. Thus, neither a theoretical nor a practical rational for how to make choices in the design of a business models. Although business model literature has arrived at a certain degree of consensus about the generic components, a more-fine grained and domain specific 'design choice'-level as the primary unit of analysis is required when relating business model concept to firm performance or for building business model tools. Such an analysis is crucial for understanding what configuration of design choices distinguishes successful from non-successful business models and providing decisional guidance to entrepreneurs with the design paradigm of hybrid intelligence.

The purpose of this research is therefore to provide a fine-grained, domain specific examination of design choices and investigates how configurations of design choices are related to firm performance. By conducting an inductive multiple case study design, the uses scalable machine learning techniques such as clustering and classification methods for data analysis to overcome methodological limitations of existing studies and providing an in-depth analysis of 188 IoT business models along 108 dimensional characteristics. I initiated my research by a taxonomy development (Nickerson et al. 2013) to identify the relevant components of IoT related business models. This initial step comprised the analysis of 188 IoT ventures and extant theories from management research. In a second step I applied a clustering analysis for identifying certain types of configurations. I then used another clustering over those types to identify successful business model archetypes. Finally, I used an ensemble of classification trees to identify pattern that distinguish successful from un-successful business models and the relevance of certain components in doing so.

For the context of this thesis, this offers several contributions that are required to address the conceptual DP of CBMV systems in Section 5.3 as well as to propose DPs and an implementation of the HI-DSS in Section 6.5.First, I provide a fine-grained and domain specific taxonomy of design choices for business models and their components. This allows to use the taxonomy as a cognitive schema to communicate the entrepreneurial opportunity between the entrepreneur and the crowd. Moreover, it provides a formal conceptual representation that can be used as data ontology for providing ML supported guidance.

Second, I identified four archetypes of successful business models design pattern that are context-specific and are identified based on an array of organizational features. This allows me to examine attributes of firm and their effect on performance, thus, providing decisional guidance based on real attributes of a firm.

Finally, I investigated the relevance of entrepreneurial design choices in determining success. Therefore, I can use this as an input to use the identified success pattern for providing decisional guidance through data-driven approaches.

## 6.3.2. Business Model Design Configurations

For this thesis, I view business models as a configuration of design choices that entrepreneurs make to create and capture value.

Prior research on configuration theory conceptualizes entrepreneurial firms from a systemic perspective understanding them as multidimensional constructs of interrelated design choices (Fiss 2011; Fiss et al. 2013). An organizational configuration describes the commonly observable co-occurrence of conceptually distinct attributes that collectively cause a certain outcome (Ketchen and Shook 1996; Meyer et al. 1993). In contrast to the theoretically computationally intractable number of possible design choice configurations these attributes reveal a tendency to occur in real-world organizations as coherent design patterns that are causal dependent (Meyer et al. 1993). Although, configurational view stresses equifinality, i.e., the possibility of a system to reach the same state through different configurations (Katz and Kahn 1978). I argue that such design pattern can be used to guide entrepreneurial decision making. This is in line with recent research that has focused on the relationships between a business model's elements to explain firm performance and competitive advantage (Klang et al. 2014).

### 6.3.3. Methodology

For my research, I apply a multiple case study approach to inductively build a theory on business model design from empirical cases (Yin 2017; Eisenhardt 1989). Building theory from empirical case evidence allows to generate theoretical constructs and midrange theories (Eisenhardt 1989). In my context, the overly phenomenon-driven research question is grounded in the emerging relevance of business models for digital innovation and the lack of plausible existing theory that provide explanatory justification on the design of business models. Little is known about how to configure characteristics of business models to succeed. Based on its replication logic, the theory building process leverages each individual case as distinctive analytic unit that allow to extend nascent theories in a field (Eisenhardt 1989). This process of creating theory is then achieved by recursive loops between the empirical evidence, related literature in the field, and the emerging theory itself (Eisenhardt and Graebner 2007). In contrast to laboratory experiments, examining cases in a real-world context enables to investigate and reason about the phenomenon in its realistic complexity. Moreover, case studies allow to gather rich, empirical descriptions of a phenomenon in a real-world context as well as the integration of various data sources (Yin 2017). This is especially valuable when the phenomenon is emerging, and theoretical rationales are still nascent. Using multiple cases also ensure the creation of more generalizable and deeper grounding in empirical evidence.

For conducting my multiple case study approach, I followed the principles of the process of building theories from case study research Eisenhardt (1989). Given the limited theory, I relied on a multi-method approach to inductively build theory about organizational configurations of digital business models. To apply scalable principles for data analysis and cross-case pattern identification, I applied several machine learning techniques. Data triangulation, the use of multiple investigators, and the combination of qualitative and quantitative methods enables me to gain deep insights in each of my 188 cases along 108 theoretically specified constructs as well as their relation across cases. Therefore, I conducted the following steps.

First, a priori specification of constructs through taxonomy grounded in related literature (Eisenhardt and Graebner 2007). Although researchers agree on the main components of business models as conceptual models, there exists little consensus about the elements that constitute those components (Massa et al. 2017; Foss and Saebi 2017, 2018). Taxonomies represent an established mechanism to organize knowledge in a field by providing a set of unifying constructs that facilitate its systematic description (Nickerson et al. 2013). These characteristics make them particularly useful to analyse complex domains and hypothesize relationships among concepts. Taxonomies, hence, vitally contribute to theory building, especially in the context of configurational research. Taxonomies organize facts and data into meaningful sets, out of which theories can be developed (Dess et al. 1993).

Second, I gathered in-depth data by conducting a descriptive classification of each case as within case analysis (Yin 2017). For this purpose, three researcher gathered data on each of the 188 independently and classified each business model along 108 characteristics. Therefore, web data, CrunchBase descriptions, news articles etc. were used. The three independent classifications for each business model were then compared and discussed until consensus among all researchers was achieved.

Third, for identifying across case pattern, I applied an unsupervised machine learning approach (i.e. clustering). The clustering of characteristics allowed me to identify configurations of business models and find archetypes.

Finally, I applied an inductive machine learning approach to examine the relationship between configurations and their influence on firm performance. I therefore used a decision tree to discriminate successful and non-successful configurations of business model configurations. This inductive approach allowed me to identify the most important configurations that determine success and distinguish successful from non-successful business models.

## Sampling

I focused on the IoT as domain focus for my study due to its disruptive character among emerging technology domains (Manyika et al. 2015). IoT aims for the consolidation of the digital and the physical sphere by equipping physical things with sensors and communication technology that facilitate the collection and analysis of data on top of which digital services can be created (Yoo et al. 2010). This setting is particularly interesting when studying digital business models as it allows ventures to develop innovative value creation and capture mechanisms that serve as a key source for future competitive advantage (Porter and Heppelmann 2014).

My research primarily focuses on 188 ventures retrieved from the online start-up database CrunchBase (www.crunchbase.com). There, each venture was categorized under the tag 'Internet of Things'. I further used three criteria to identify potential case firms. First, each of the ventures had to meet the definition developed by the European Research Cluster on the Internet of Things referring to the IoT as a *"[...]*

*dynamic global network infrastructure with self-configuring capabilities based on standard and interoperable communication protocols where physical and virtual things have identity, physical attributes and virtual personalities and use intelligent interfaces and are seamlessly integrated into the information network [...]"* (Vermesan and Friess 2014a). Second, the firm was active in terms of their business activities. Third, the venture discloses sufficient information to adequately assess their business model design.

To improve the robustness and generalizability of my results, I aimed for variation in terms of industries, technologies, and market segments across the selected ventures (Eisenhardt and Graebner, 2007). Moreover, I just used ventures that were founded after 2014 as in this phase the business model has the most predictive power for entrepreneurial success.

## Data Collection

For each venture I collected evidence of its business model design and financial performance. I relied on two main data sources: company websites and CrunchBase profiles. I triangulated these data sources with complementary data comprising news articles and official social media profiles (LinkedIn, Twitter, Facebook) to improve the robustness of my findings (Yin 2017).

I continued my data collection by interviewing top level managers from two additional ventures. During these interviews I provided them with a preliminary version of my taxonomy and let them explain their business model sub-layer by sub-layer. These interviews (each lasting around 2 hours) were particularly helpful for refining individual manifestations of the taxonomy.

## Taxonomy Development

For the initial step I relied on the taxonomy development method proposed by Nickerson et al (2012) that comprises inductive and deductive elements, and therefore incorporates both empirical and theoretical evidence.

The first step of the taxonomy development method is to determine a meta-characteristic, that serves as the foundation of subsequent choice of characteristics. I defined the meta-characteristic as the main components of a business model reflected by the four dimensions proposed by Gassmann et al. (2014), *Who? What?, How?, Why?.* This step limits the odds of *"naïve empiricism"*, where many characteristics are defined, hoping that a pattern will emerge, and it reflects the expected use of the taxonomy (Nickerson et al. 2013).

The second step comprises the selection of objective and subjective ending conditions to terminate the iterative process facilitating usefulness of the taxonomy. I selected and adapted the following objective and subjective ending-conditions from Nickerson et al. (2013):

## Objective conditions:

(1) *all the ventures of the sample are examined;*
(2) *at least one object is classified under every characteristic of every dimension;*
(3) *no new dimension or characteristics is added in the last iteration;*
(4) *every dimension is unique and not repeated;*
(5) *every characteristic is unique within its dimension and not repeated.*

## Subjective conditions:

(1)   *conciseness: no unnecessary dimension or characteristic is included;*

(2)   *robustness: there are enough dimensions and characteristics to differentiate between the various ventures;*

(3)   *comprehensiveness: all IoT ventures can be classified in the taxonomy;*

(4)   *extendible: new dimensions and characteristics can be subsequently added;*

(5)   *explanatory: the information is valuable and contributes to characterizing IoT ventures;*

(6)   *information availability: the information is typically available or easily interpretable.*

A key objective of the objective ending conditions is to generate dimensions of mutually exclusive and collectively exhaustive characteristics. While aiming for a high degree of mutual exclusiveness it became obvious that for certain dimensions characteristics may apply. A company can target several customer segments, simultaneously occupy multiple layers of the IoT ecosystem and use multiple approaches to monetize their products or services. This is not a unique property of the chosen unit of analysis and has been considered in ontology development research by the concept of slot cardinality that defines how many characteristics may maximally exist for any venture in a certain dimension. Multiple cardinality is contrary to the recommendation of Nickerson et al. (2013) but applied to accommodate the taxonomic nature of IoT business model components. Furthermore, to limit the cognitive load of the taxonomy user and make it more intuitive to use, I intentionally focused on a generalizable abstraction level. Hence, I kept the dimensions and characteristics to a reasonable number while preserving taxonomic completeness to achieve comprehension, application and ultimately usefulness of the taxonomy (Nickerson et al. 2013).

Following the definition of a meta-characteristic and the ending conditions, I began to derive dimensions and characteristics for the taxonomy. The process allows for two different approaches in each iteration. First, the empirical-to-conceptual approach creates characteristics and dimensions based on the identification of common characteristics from the sample. Second, the conceptual-to-empirical approach relies on extant theory to devise characteristics and dimensions, before validating these dimensions and characteristics on the sample. In total I conducted three iterations.

Following the advice by Nickerson and colleagues (2013) I initiated this research step with a conceptual-to-empirical approach as I felt sufficiently knowledgeable about digital business models and the IoT but lacked data. I then examined a subset of 50 ventures to test the appropriateness of each dimension. Although each dimension aligned to the meta-characteristic, I noticed that the dimensions insufficiently described certain important aspects of the business model, indicating a lack of collective exhaustiveness.

Rather than pursuing one of the two ideal typical approaches the second iteration can be described as a back-and-forth between the study of my sample ventures and theory until I had examined an enlarged subset of 100 firms. More specifically, in line with an empirical-to-conceptual approach I discussed notable business model related differences for the studied cases among the researchers, followed by an investigation of corresponding theoretical concepts. Once I felt that the iterated taxonomy was well suited for my subsample well, I conducted interviews with two ventures for evaluative purposes. The interviews revealed necessary adjustments for the characteristics.

In the third iteration I then applied the empirical-to-conceptual approach comprising a review of the full sample. I terminated the taxonomy development after I noticed that I were able to describe all ventures.

## Cluster Analysis

I used cluster analysis as a valuable data analysis technique as it allows the classification of a huge number of cases (i.e. individual business models) along many characteristics (Ketchen and Shook 1996). For the applied case study research methodology, this approach supports inductively examining an open-ended, large-sample examination of cases on a fine-grained level. Clustering is a statistical technique to group a sample by minimizing the variance among cases grouped together while maximizing between-group variance (Ketchen and Shook 1996).

For clustering my data, I choose the characteristics along the conceptually defined subsets of business model dimensions (Ketchen and Shook 1996). As all the features are categorical, I use one hot encoding for categorical features, which results in a dummy variable for each characteristic, where each feature (i.e. characteristic) can take the values 0 or 1. This means that a business model can theoretically consist of $2^{108}$ unique configurations of characteristics. Those characteristics were then used as input features for clustering.

I applied the kModes clustering algorithm for categorical data with a Python implementation (Huang et al.; Huang 1997, 1998; Huang and Ng 2003). Other clustering algorithms such as kMeans use distance such as Euclidian distance measures between two objects, which is not possible for categorical data. Moreover, these approaches represent cluster centroids as means, which is not possible for nominal data. Moreover, this statistical approach captures the conceptual similarity between business models that can be defined as the degree of coincidence of elements (Rumble and Mangematin 2015).

kModes clustering can be seen an extension to the standard kMeans by applying a simple matching dissimilarity measure, using modes to represent cluster centres and updating modes with the most frequent values in each iteration (Huang 1998). This approach ensures that the iterative process converges to a local minimum.

The dissimilarity (Hamming distance) between two objects (i.e. business models) X and Y described by m characteristics is defined as

d(X, Y)= $\sum_{j=1}^{m}$ = δ (xj,yj) where δ (xj,yj) is $0, xj = yj$ 1, xj ≠ yj.

Thereby, xj and yj describe the values of attribute j in X and Y, where a higher number of mismatches of characteristics between X and Y express a higher dissimilarity. The dissimilarity between an object X and a cluster centre $Z_l$ is then calculated as ϕ$(xj, zj)$ = 1- $\dfrac{n_j^r}{n_l}$ for $xj = zj$ and 1 for $xj \neq zj$ (Ng et al. 2007).

Here $zj$ is the categorical characteristic of attribute j in $zl$, while $n_l$ is the count of objects in cluster l and $n_j^r$ is the number of objects whose attribute characteristic is r.

The kModes clustering represents cluster centroids as the vectors of modes (i.e. the most frequent value) of categorical attributes. This means that a data set of m categorical features has a mode vector Z of m categorical values ($z_1$, $z_2$, ..., $z_m$). This mode vector of a cluster then minimizes the sum of between object distances within a distinctive cluster and the cluster centroid (Huang 1998).

Defining the most appropriate number of clusters is one of the biggest challenges in this approach (Ketchen and Shook, 1996). I combined both a statistical (i.e. Silhouette score) as well conceptual a priori constraints to set the number of clusters (Hair et al. 1992). For this purpose, I used a grid search over the possible search space of hyperparameters (i.e. number of clusters). Therefore, I conceptually constraint the number of possible clusters between 2 and 30. I then calculated the average Silhouette score for each cluster number and choose the local maxima of this value. The Silhouette score is an approach for cluster interpretation and validation of consistency by indicating how well each object lies within its cluster (Rousseeuw 1987). This value indicates the similarity of an object to its individual cluster (cohesion) in relation to all other clusters (separation). The silhouette score ranges from −1 to +1, where values close to 1 indicates that the object is properly matched to the right cluster. When most objects have a high score, the clustering configuration is optimal. Therefore, I calculated the average score across all objects per number of clusters.

## 6.3.4. A Taxonomy of Business Model Design Choices

The first findings of this study reveal a detailed taxonomy of design choices for business models. The taxonomy and the related design choices are displayed in Figure 36.

The taxonomy builds on the four business model layers developed by Gassmann et al. (2014): What does the firm offer to target customers? Who are the target customers? How does the firm produce their offering? Why does the firm generate profit? The taxonomy is structured as follows:

$$Business\ Model\ Layers \left\{ Sub-layers \left[ Dimensions \left( Characteristics \right) \right] \right\}$$

| | Sub-layer | Dimension | Characteristics | | | | | | |
|---|---|---|---|---|---|---|---|---|---|
| **WHAT?** | Solution | Solution Type | Mitigation Tool | | Execution Tool | | Improvement Tool | | Control Tool | |
| | | Solution Form | Goods | | | | Services | | | |
| | | Competitive Strategy | Low Cost | Innovativeness | Performance | Customization | Turnkey | Design | Integration |
| | Eco-system | IoT Layers | Device | Content | Network | Management | Application | Service | Security |
| | | Core Functions | Monitoring | | Controlling | | Optimizing | Autonomy | Sharing & collaboration |
| | | Inter-operability | Open | | Limited Openness | | Closed | | |
| | | Ecosystem Ownership | Own | | | | Existing | | |
| **WHO?** | Market | Application Ecosystem | Smart Environment | | Smart Industry | | Smart Health and Wellbeing | | |
| | | Customer | B2B | | B2C | | B2G | | |
| | | Segment | Segmented | | Niche | Mass | Diversified | | Multi-Sided |
| | Customer Relation | Interaction intensity | Loose | | | Highly Engaged | | | |
| | | Retention | Validate Consumer Choice | | Value | Procedural Switching Costs | Financial Switching Costs | | Industry Standard |
| **HOW?** | Resources | Key Internal Resources | Human Resources | | Physical Resources | | Organizational Resources | | |
| | | Technology Combination | Artificial Intelligence | | Robotics | | 3D Printing | | Blockchain |
| | | Data Source | Product State | | Product Context | | Product Usage | | External Data |
| | | Data Usage | Transactional | | | | Analytical | | |
| | Partners | Partners | Component | Content | Network | Management | Security | Channel | Co-Creation |
| | | Institutional Support | Incubator | Accelerator | Corporate Program | | University | Board of Directors | Board of Advisors |
| | | Investors | Angels | Venture Capitals | | Corporate Venture Capitals | Public Funds | | Crowdfunding |
| | Activities | Operational Focus | Operations | | Marketing and Sales | | Research & Development | | Services |
| | | Customer Acquisition | SEM | | Email | | Social Media | Events | Content |
| | | Customer Service | Personal Assistance | | Customized Channels | | Self Service | Community | Automated Services |
| | | Distribution | Direct Online | | Direct Offline | | Indirect Online | | Indirect Offline |
| **WHY?** | Revenues | Revenue Model | One-Time | | Subscription | | Advertising | Brokerage | Licensing |
| | | Pricing Strategy | Fixed Price | | Feature-Dependent | | Usage-Dependent | | |
| | Cost/Profit | Profit Effect | Margin | | Volume | | Use/Transaction | | |

***Taxonomy of IoT Business Model Design***

# What?

The "what?" layer depicts the content of the value proposition of an IoT company, defined as "*the benefits customers can expect from products and services*" (Osterwalder et al. 2014: 6). Anderson et al. (2006: 4) identified three interpretations of value proposition: *"all the benefits"* for the customers, all the favorable points of differentiation from the competition, and a restricted number of key differentiation points. Accordingly, in the first sublayer of *solution*, we distinguish three dimensions: benefits (*solution type*), format (*solution form*), and differentiation (*competitive strategy*). The second sublayer is the *ecosystem*, which has four dimensions: the levels or *layers* in the IoT reference architecture at which a specific company creates value, its *core function*, the *ownership of the ecosystem*, and the combination possibilities with third-party solutions (*interoperability*).

### Solution

The *solution types* are developed in reaction to the latent or explicit need of customers and users to remove hurdles or mitigate risks. Based on a survey conducted by Zebra Technologies for the Strategic Innovation Symposium at Harvard University, we identify four generic types of problems commonly solved by IoT solutions. The four generic problems can be defined as the *"benefits and outcomes required, desired, expected or unexpected by the customers"* (Osterwalder et al. 2014: 8).

|  | Description | Example | Theoretical Foundation |
|---|---|---|---|
| **Solution Type** | | | |
| **Mitigation Tool** | Solution to reduce risks and uncertainty. | Kyontracker, HAAS Alert, SensrTrx | (Fleisch 2010; Manyika et al. 2015; Osterwalder and Pigneur, 2010; Zebra, 2017) |
| **Execution Tool** | Solution enabling the making and repairing of another good or the performance of an activity. | Seebo, Pycom | |
| **Improvement Tool** | Solution to increase the value and quality of products, services and processes. | Dattus, Agroptima, Senseware | |
| **Control Tool** | Solution to influence actions and behaviors of objects and persons. | Cloudleaf | |
| **Solution Form** | | | |
| **Goods** | Tangible offerings which last over time, exist independently from their owner and for which ownership rights may be established. | Onion | (Parry et al. 2011; Turber et al. 2014) |
| **Service** | Intangible offering existing only via other things. The production cannot be distinguished from consumption. | Agroptima, Flytrex | |

The solutions may be offered under the *form* of goods, services, or a combination of both. Turber et al. (2014) describe these formats as the 'carriers of competences.' We adopt the set of distinguishing features to classify goods and services in the taxonomy.

*Competitive strategies* aim at creating a unique value proposition which differentiates a provider's solutions from those of their competitors. Adapting Porter's generic strategies (1980) – i.e. overall low costs, differentiation, and focus with empirical observations – we identify seven generic competitive strategies. Individual ventures may combine various competitive strategies, so they are not mutually exclusive.

| Competitive Strategy | | | |
|---|---|---|---|
| **Low Cost** | The solution is cheaper then the competition. | GPSDome, Onion | |
| **Innovativeness** | The solution is new or significantly different compared to other products on the market. | Flytrex, Keriton, 75F | |
| **Performance** | The solution is more efficient and/or effective than other solutions on the market. | Cognosos, Wesavy | |
| **Customization** | The solution can be adapted to individual specifications. | Scriptr, Momenzz | (Kans and Ingwald, 2016; Osterwalder and Pigneur 2010; Porter 1980 |
| **Turnkey** | The solution comprises all the necessary components for valuable operation, and can be used out-of-the box. | Senseware, Watty | |
| **Design** | The solution is intuitive, easy to use and functional. It can also relate the aesthetical quality. | Soofa , Include, Kwik | |
| **Integration** | The solution seamlessly integrates with the other solutions used by the customer. | Seebo | |

### *Ecosystem*

The IoT combines multiple technologies in a complete stack that is essential to value creation (Püschel et al. 2016). In the taxonomy, we distinguish between seven *IoT layers* to situate each venture within the larger IoT ecosystem and specify the nature of the offering. An IoT company may be active across multiple layers. Therefore, the characteristics of this dimension are not mutually exclusive.

Based on empirical findings and theories developed by Porter and Heppelmann (2014), we define five *core functions* of IoT products. Each of the first four functions builds on the previous one; for example, the controlling function requires a monitoring function (Porter and Heppelmann 2014).

|  | Description | Examples | Theoretical Foundation |
|---|---|---|---|
| **IoT Layers** | | | |
| **Device Layer** | The "things" in the IoT. Combination of the machinery layer (sensors and actuators) and logical capability layer (software and OS). | Butterfleye, Watty, DSP Concepts | (Aazam et al. 2014; Borgia 2014; Cisco Inc. 2014; Fleisch et al. 2014; IERC 2016; Püschel et al. 2016; Ray 2016; Vatsa and Singh 2015; Yaqoob et al. 2017) |
| **Content Layer** | Texts, images, videos, metadata and other types of data in a digital format. | Terbine, | |
| **Network layer** | Physical transport layer (gateways, cables, radio waves and other hardware) coupled with logical transmission layer (standards and protocols). | Cirrent , Morsemicro | |
| **Management Layer** | Aggregation and storage of data; cloud or edge computing. | ProxToMe, WayLay | |
| **Application layer** | Final presentation of the relevant data, which can be edited, controlled and/or monitored. | Butterfleye, Watty, Wexus | |
| **Service Layer** | Practical usage of the application and content for effective use by the users; communication and collaboration between people. | Watty, Wexus, Flytrex, Wesavy | |
| **Security Layer** | Secure devices, systems and processes between all layers; ensure privacy of users. | Wia, GPSDome, Hideez, Cryptalabs | |
| **Core Functions** | | | |
| **Monitoring** | Function of observing the state and usage of a product, as well as its environment. | Butterfleye, Measurence | |
| **Controlling** | Function of influencing the mode of operation of a product. | Hideez | (Lee and Lee 2015; Porter and Heppelmann 2014) |
| **Optimization** | Function of improving products, processes and business models to the best possible level of performance. | Jooxter | |
| **Automation** | Ability of a product or a system to self-sufficiently perform operations and tasks such as servicing, customization and diagnosis. | Flytrex | |
| **Sharing and collaboration** | Function of exchanging data "between people, between things or between people and things". | Cirrent, WayLay, Terbine | |
| **Interoperability** | | | |
| **Open** | Solution based exclusively on open protocols and standards; seamlessly integrate with other solutions via APIs SDKs, or other outbound links. | DSP Concepts, Wesavy | (Elkhodr et al. 2016; Manyika et al. 2015; Yaqoob et al. 2017) |
| **Limited Openness** | Mix of open and proprietary standards, protocols; limited links with third party solutions (SDKs, APIs). | Senseware | |
| **Closed** | Integration only within proprietary ecosystem. | Watty | |
| **Ecosystem Ownership** | | | |
| **Own** | Development and exclusively use of a proprietary ecosystem with no integration with other products. | Watty, Flytrex | (Empirical) |
| **Existing** | Integrates with solutions from partners. | DSP Concepts, IOPipe | |

*Interoperability,* defined as the ability of solutions from different vendors to communicate and integrate with each other seamlessly, is a key driver of value in IoT. McKinsey estimates that interoperability could be responsible for 40% of the potential value created by the IoT (Manyika et al. 2015). The possibility for devices to communicate and operate together depends largely on the use of widely-accepted interface standards or translation schemes between operating systems and applications (Manyika et al. 2015). However, the competitive pressures and the constant evolution of technologies tend to impede the integration of devices in a homogeneous framework. In the taxonomy, we distinguish between three mutually-exclusive types of ecosystems: open, limited openness, and closed.

Finally, the *ecosystem ownership* dimension differentiates companies developing their *own* ecosystem from companies leveraging *existing* ecosystems from third parties. Companies promoting their own ecosystems integrate the different layers of their offering under one proprietary umbrella. Alternatively, firms may restrict their solution to one or more layers, while collaborating with *existing* platforms or solutions from third parties.

## Who?

The business model layer "*who?*" defines the stakeholders for which value is created and the channels through which they are being reached (Gassmann et al. 2014: 55). This layer is split into two sub-layers: the *market* sub-layer describing the specific categories of targeted consumers, and the *relations* sub-layer which specifies the channels and intensity with which they are addressed.

### Market

The nature of the IoT broadens the traditional frontiers of market sectors. Manyika et al. (2015) argue that focusing on industry verticals provides a limited perspective on the value created by IoT because it fails to describe the interaction between systems which do not belong to the same market sectors. Therefore, they adopt the complementary lenses of the setting, defined as the *"context of the physical* *environment in which systems can be deployed"* (Manyika et al. 2015: 18). Specifically, they define nine IoT settings: human, home, retail, offices, factory, worksite, vehicle, city, and outside. Combining the industry verticals with settings provides for a very large number of potential combinations, especially since neither the market sectors nor the settings are mutually exclusive. We identify three dimensions of market: application ecosystems, customer, and market segment.

| | Description | Examples | Theoretical Foundation |
|---|---|---|---|
| **Application Domain** | | | |
| **Smart Environments** | Use of IoT in objects, buildings and conditions in and with which humans are evolving daily. Comprises smart city, home, workplace and mobility, among others. | Watty, Soofa, Transitscreen, Gluehome, Ween | (Aggarwal et al. 2013; Atzori et al. 2010; Borgia, 2014) |
| **Smart Industry** | Use of IoT solutions in industrial activities. Comprises smart manufacturing, agriculture, logistics and transportation, among others. | Petasense, Agroptima | |
| **Smart Health and Wellbeing** | Use of IoT solutions to improve the health and everyday life of users. Comprises smart medical devices and smart consumer goods, among others | Keriton, Glanceclock, MysteryVibe | |
| **Customer** | | | |
| **B2C** | Focus on solutions for end-consumers. | Watty, Wesavy | (Kotler and Keller) |
| **B2B** | Focus on solutions for business customers. | Butterfleye, DSP Concepts | |
| **B2G** | Focus on solutions for government and public institutions. | Soofa, | |
| **Segment** | | | |
| **Segmented** | Focus on a customer group with relatively homogenous needs. | Wesavy | (Osterwalder and Pigneur 2010) |
| **Niche** | Focus on a solution for the needs of a very specific market segment. | Nohocare | |
| **Mass** | No specific distinction between target groups. | Glanceclock, Meural | |
| **Diversified** | Focus on multiple segments. | DSP Concepts, Measurence | |
| **Multi-sided** | Intermediary between two or more interdependent segments. | Slock.it, Terbine | |

Based on the observations mentioned in the previous paragraph, we identify three generic *application ecosystems*: *smart environments*, *smart industry*, and *smart health & well-being*. They represent the three archetypical combinations of industry verticals and settings and are based on the domains proposed by Borgia (2014).

The *customer* dimension describes the type of customer towards whom the transaction is directed. Business-to-consumer (*B2C*) concerns companies that sell goods and services to the end customers. In business-to-business (*B2B*) markets, firms sell their offerings to other firms. In business-to-government (*B2G*), companies sell their products to governments and other public organizations in public markets.

The *market segment* dimension describes the size of the targeted customer segment. In the taxonomy, we adopt the five types of customer segments suggested by Osterwalder and Pigneur (2010).

## *Customer Relation*

|  | Description | Examples | Theoretical Foundation |
|---|---|---|---|
| **Customer Interaction Intensity** | | | |
| **Loose** | Customer do not interact directly with an employee from the company. (transactional). | Butterfleye, Glanceclock | (Empirical) |
| **Highly engaged** | Direct interaction between customer and employee (relational). Comprises personal assistance, co-creation with the user, communities. | DSP Concepts, Measurence | |
| **Customer Retention** | | | |
| **Validate customer's choice** | Activities to validate the customer's choice of supplier, for example through advertisement, influencers, ensuring positive user experience or post-purchase service activities. | Momenzz | (Nagengast et al. 2014) |
| **Value** | Enhancing the *value* of the exchange, that is increasing the customer utility derived from the use of the solution and by proposing add-ons or complements. | Wesavvy | |
| **Procedural Switching Costs** | Time, capital and effort costs to evaluate, learn to use (learning curve trap) or setup a new product or technology (technological lock-in). | Wia | |
| **Financial Switching Costs** | Costs of terminating long term contracts (contractual lock-in) and loosing fidelity benefits. | Soofa | |
| **Industry Standard Lock In** | Leverage the network effect to enforce publicly accepted standard or built upon an industry standard from a third party. | IoPipe | |

Customer relation includes two dimensions: interaction intensity and retention. Interaction intensity describes the magnitude of the interaction between a specific firm and their customers.

Recruiting new customers generally costs more to companies than selling to existing customers. The dimension of retention describes how the ventures retain their customers and increase the likelihood of repurchase. Based on customer retention and switching costs theories, we identify five general retention strategies.

## How?

In the *How* business model layer, we differentiate the value-creation mechanisms by separate *resources*, *partner*s, and *activities* sub-layers.

*Resources*

We adopt the definition of a firm as "all assets, capabilities, organizational processes, firm attributes, information, knowledge etc. controlled by a firm that enable the firm to conceive of and implement strategies that improve its efficiency and effectiveness" (Barney 1991: 101).

| | Description | Examples | Theoretical Foundation |
|---|---|---|---|
| **Key Internal Resources** | | | |
| **Human Resources** | Employees, training, experience, knowledge, skills and abilities. | DSPConcepts | |
| **Physical Resources** | Plant and equipment, technology, raw materials and machines. | Onion | |
| **Organizational Resources** | Structure of the organization, strategic and planning processes, information systems, intellectual properties such as patents, trademark and copyrights, and databases. | Connectric | (Barney, 1991) |
| **Technological Combination** | | | |
| **Blockchain** | Use of distributed ledger technologies in IoT or vice versa. | Slock.it | (IBM and MIT Technology Review 2018) |
| **Artificial Intelligence** | IoT data processed through machine learning, neural networks, deep learning or other Artificial Intelligence technology. | Xesol, Imagimob | |
| **Robotics** | Machine moving independently and making complex movements. | Ready-Robotics | |
| **3D Printing** | Support computer assisted transformation of material to create a three-dimensional object. | Astroprint | |
| **Data source** | | | |
| **Product state** | From the internal status of the sensors, actuators or other components at a given time. | Podtrackers, ReadyRobotics | (Püschel et al.2016) |
| **Product context** | From the local conditions surrounding the IoT hardware and software at a given time. | Sensibo, Ween | |
| **Product usage** | From the various parameters defining the mode of use which are made of a particular solution, especially in relation with other hardware and software components. | Leantagra | |
| **External data** | From information not generated or shared at another layer in the IoT infrastructure than the device layer. | Slock.it, Scanalytics | |

The resource-based view posits that the *internal resources* of a firm are better contributors to competitive advantage than external factors (Barney 1991), and groups them in three generic categories: *human resources*, *physical resources*, and *organizational resources*. In most instances, companies rely on multiple resources. However, the aim of the distinction we propose is to identify which of these resources is the most critical in the value creation process.

The various technologies commonly used across the different layers of the IoT architecture can be complemented with other technologies which are not systematically related with the IoT. We call this blend of technology the *technology combination* and identify five technologies which are regularly incorporated in the IoT: *Blockchain*, *Artificial intelligence*, *Robotics*, and *3D printing*.

Data has become a key asset for companies in the digital age (Bharadwaj et el. 2013). Therefore, it represents an important component of the business model. In the taxonomy, we extend the dimensions of *data source* and data usage identified by Püschel *et al.* (2016) to the complete IoT stack

The second dimension relating to data is the *data usage;* it describes how data is used. We adopt two characteristics (Püschel et al. 2016): transactional and analytical.

| Data Usage | | | |
|---|---|---|---|
| **Transactional** | Data is leveraged as part of the process for the intended use of the IoT solution or traded (descriptive). | Glanceclock, Meural | (Püschel et al. 2016) |
| **Analytical** | Data is processed to derive particular insights (analytical, prescriptive, predictive). | Slock.it, Robby | |

## *Partners*

| | Description | Examples | Theoretical Foundation |
|---|---|---|---|
| **Partners** | | | |
| **Component supplier** | Supplies hardware and software components. | Flytrex | (Aazam et al. 2014; Fleisch et al., 2014; IERC, 2016; Püschel et al. 2016) |
| **Content partner** | Supplies texts, images, videos, metadata and other types of data in a digital format. | Meural, Depict | |
| **Network provider** | Supplies the physical transport and logical transmission layers. | Watty | |
| **Management partner** | Supplies the tools for aggregating, storing and computing data. | Instapio | |
| **Security provider** | Supplies the hardware and software to secure the IoT solution. | EmbarkTrucks, Calipsa | |
| **Channel partner** | Supplies a mean to communicate and sell an IoT solution to the market. | Lattis | |
| **Value co-creation partner** | Customer applying skills and know-how as a contribution to value creation. | DSPConcepts | |
| **Institutional Support** | | | |
| **Incubator** | Institution providing startups with services such as office space and management training. | MysteryVibe | |
| **Accelerator** | Fixed term programs including trainings, networking possibilities, mentorship and seed investment (equity). | Agroptima, SAMLabs | |

| **Corporate Program** | Accelerator program within a corporation. | Chargifi, CarFit | (Bergek and Norrman 2008; Crunchbase, Osterwalder et al. 2014) |
|---|---|---|---|
| **University** | Support from university or other higher education institution for the development of startups. | CocoonCam | |
| **Board of Directors** | Group of people elected by the stockholders who establish and oversee management policies. | DSP Concept | |
| **Board of Advisors** | Group of people providing non-binding strategic advices to the management of a company. | Silk Labs | |

The creation and capture of value involves an increasingly large number of interactions between firms. Consequently, there is an incentive for firms up and down the value chain to align their interests and enter into explicit or implicit partnerships for value creation (Chesbrough and Schwartz 2007). Specifically, Dyer *et al.* (2018) have identified four key determinants for inter-company value creation: complementary resources and capabilities, relation-specific assets, knowledge sharing routines, and effective governance. In the taxonomy, the factor of resource and capability complementarity is reflected by identifying the *key partners* for the creation of final goods and services, the *investors* who contribute with the financial resources, and the *institutional supports*. However, since the taxonomy is developed exclusively with publicly available information, it does not contain any dimension about the assets and routines involved in the partnership, nor about the governance (i.e. the mechanisms to balance the power relationship) and nature (e.g. joint venture, strategic alliance) of the partnership.

| Investors | | | |
|---|---|---|---|
| **Angels** | Individuals using their personal wealth for start-up and early stage business funding, in return for equity. | Koto, Flytrex | (Crunchbase) |
| **Public Fund** | Non-repayable funding by the government. | Agroptima | |
| **Venture Capitals** | Firms specialized in taking equity in exchange for capital. | Robby, Cognosos | |
| **Corporate Venture Capitals** | Arm of a corporation investing in equity of innovative ventures. | DSP Concepts | |
| **Crowdfunding** | Funding by raising small amounts of money from large number of people. | MysteryVibes | |

To categorize the essential *partners* of a given firm, we distinguish between the different IoT layers at which partners supply hardware and software components to be embedded in IoT solutions. We also distinguish between channel partners and co-creation partners.

Given the limited access to resources and knowledge, the success of new ventures often relies on various *institutional support*. Based on our sample, we identified six types of support: *incubators*, *accelerators*, *corporate programs*, *universities*, *boards of directors*, and *boards of advisors*.

In addition to these institutional supporters, new ventures nurture a close relationship with *investors*. Based on the classification from the startup database *Crunchbase* and on empirical evidences from the sample, we identified five main types of investors in IoT companies.

### Activities

Key activities are the most important actions undertaken by an organization to create and distribute value (Osterwalder and Pigneur 2010). The *operational focus* describes key activities performed by an IoT company. Porter differentiates between primary activities relating directly to the creation of the offering, and secondary activities which consist of all activities to support the primary activities (Porter 1980). The primary activities comprise inbound and outbound logistics, the operations transforming input into outputs, marketing and sales, and services. Secondary activities are listed as procurement, human resource management, research and development, and company infrastructure (a company's support systems). As discussed above, the nature of value creation in the IoT scatters activities traditionally concentrated in one firm to an ecosystem of partner firms. By combining empirical observations with the distinctions proposed by Porter (1980), we retained four operational foci for IoT companies: *operations*, *marketing & sales*, *services*, and *R&D*.

| | Description | Examples | Theoretical Foundation |
|---|---|---|---|
| **Operational Focus** | | | |
| **Operations** | Activity related to transforming inputs (resources) into outputs (goods and services). | Onion, Pycom | (Johnson et al. 2014; Porter 1985; Slack et al. 2010) |
| **Marketing and Sales** | Activity related to raising customer's awareness about the solution, and ensuring he is able to purchase it. | Meural, Glanceclock | |
| **Services** | Activity of enhancing and maintaining the value of a solution. | Parkbob | |
| **Research and Development** | Activity of introducing, developing and enhancing existing products and processes. | DSP Concept | |

| Customer Acquisition Channel | | | |
|---|---|---|---|
| Search Engine Marketing | Optimization of the visibility of commercial web pages and web stores as a result of search engines queries. | Meural | (Kotler and Keller 2012; Reynolds 2018) |
| Email Marketing | Newsletters and customized massage sent via email. | Watty | |
| Social Media Marketing | Promotion of pictures, videos and other contents on social media. | Vinaya | |
| Event Marketing | Participation and organization of events such as sport, charity or cultural events, or professional fairs. | Petasense | |
| Content Marketing | Team of employees engage in prospection, promotion and demonstration of the value of the solution. | Switch Automation | |
| Customer Service Channel | | | |
| Personal Assistance | Direct human interaction during or after sale. | Soofa | (Osterwalder and Pigneur 2010) |
| Customized Channels | Personal assistant dedicated to one specific customer (e.g. account manager). | Wexus | |
| Self Service | No direct relationship with customers but provide tools for customer to serve themselves. | Kwik | |
| Community | Maintenance of online communities to exchange utilization advices and peer to peer troubleshooting, such as online threads, reviews. | Onion | |
| Automated Services | Automated self service optimized with information about the customer, such as chat bot. | Litmus | |
| Distribution | | | |
| **Direct Online** | The company sells directly to the end consumer via its own online platform. | Atmoph | (Kotler and Keller, 2012; Osterwalder and Pigneur 2010) |
| **Direct Offline** | The company has a sales team directly interacting with the customers. | Macrofab | |
| **Indirect Offline** | The company sells via brick and mortar stores belonging to sales partner such as retailers and wholesalers. | MysteryVibe, Include | |
| **Indirect Online** | Company uses partner's online store. | Meural | |

Furthermore, all firms promote their solution with *customer acquisition* activities. They may use online and offline channels to raise product awareness, share technical and usage information, and increase willingness to buy. We propose five types of customer acquisition channels for the taxonomy.

In addition to their offering per-se, companies also compete on the c*ustomer service* they provide. We distinguish between five types of customer services that depict commonly observable manifestations.

In addition to activities related to the creation of the core offering, companies need to deliver the created value through *distribution*, or marketing, channels. Osterwalder and Pigneur (2010) distinguish between types of distribution channels based on two characteristics: whether the company owns and manages its own channels or relies on partners, and whether there is a direct interaction between the company and the customers or not. From our sample, we identified four categories of distribution channels.

# Why?

The "Why?" business model layer refers to the ability of the revenue streams and the cost structure to enable the commercial viability of a business model (Gassmann et al. 2014).

| | Description | Examples | Theoretical Foundation |
|---|---|---|---|
| **Revenue Model** | | | |
| **One Time Sale** | Revenue derived from the sale of ownership right on a good. | Hideez Glanceclock | (Carnegie Mellon University 2015; Dijkman et al. 2015; Osterwalder and Pigneur 2010) |
| **Subscription** | Revenue derived from selling access to a solution for a determined period of time. | Wexus Petasense | |
| **Advertising** | Revenue derived from advertisement fees. | Wesavvy | |
| **Brokerage** | Revenue derived from intermediation service between at least two parties. | Slock.it, Terbine | |
| **Licensing** | Revenue derived from permission to use intellectual property against a licensing fee. | DSP Concepts | |
| **Pricing Mechanism** | | | |
| **Fixed Price** | Price fixed for individual services and products. | Hideez | (Osterwalder and Pigneur 2010) |
| **Feature Dependent** | Price is a function of product features such as quality, customization, and design. | Wexus, WigWag | |
| **Volume/ Usage Dependent** | Price is a function of the quantity purchased. | Remicro | |
| **Profit Effect** | | | |
| **Margin** | Profit stems from the large premium between the cost of goods or service sold and the sales price. Focus is on selling low volume at a high margin. | American-Robotics | (Horngren et al. 2012) |
| **Sales Volume** | Profit stems from a large number of goods or services being sold. Focus is on selling high volume at a lower margin. | Hideez | |
| **Use/Transaction** | Profit is directly related to the number of use or transactions of the offering. | Wexus | |

### Revenue

The *revenue model* describes the type of revenue that is generated by the selling of a solution. Osterwalder and Pigneur (2010) define two types of revenue models: transactional revenues, which involve a unique payment by the buyers, and recurring revenues, which generate multiple incoming payments over the lifecycle of the offering. We identified five IoT revenue models.

The dimension of pricing strategy describes how a price is presented to (potential) customers. We identified three types of mechanisms used by IoT companies to set prices.

### Costs

Two direct components drive profitability: the contribution margin and the volume of units sold. Accordingly, the profit is either driven by large sales volume or by selling the product at a high margin. Additionally, for many services or "as-a-service"-solutions, the profit is dependent of the number of transactions or use.

## 6.3.5. Business Model Design Configurations

My clustering approach results in each business model being characterized along nine component cluster memberships, where the number of types describes how many clusters exist for the component in the data.

| Component Name | Sub-Dimensions (Number of Characteristics) | Number of Types | Average Silhouette Score |
|---|---|---|---|
| Solution | 3 (12) | 10 | 0.177 |
| Ecosystem | 3(17) | 2 | 0.2727 |
| Market | 3(11) | 19 | 0.651 |
| Customer Relation | 2(7) | 18 | 0.775 |
| Resources | 4(13) | 2 | 0.294 |
| Partners | 3(18) | 2 | 0.159 |
| Activities | 4(18) | 2 | 0.274 |
| Revenues | 2(8) | 19 | 0.855 |
| Costs | 1(3) | 7 | 1.0 |

*Components and Number of Configurations*

## 6.3.6. Design Pattern of Successful Business Models

For identifying and describing successful business model design pattern, I split the total data set in a subset of successful business models each characterized along its membership along the 9 dimensions. This approach resulted in a set of 31 ventures. I then used another kModes clustering to identify common pattern among those business models. I searched the possible space of design configurations between 2 and 20 clusters and identified 4 configurations as being most appropriate (Silhouette score = 0.053) (Rousseeuw 1987). Those four design patterns of successful business models constitute configurations of the nine business model dimensions identified in the previous section.

| Pattern | Design Configuration | Frequency | Examples |
|---|---|---|---|
| I. | Solution Type: 1<br>Ecosystem Type: 0<br>Market Type: 15<br>Customer Relation Type: 3<br>Resources Type: 1<br>Partners: 1<br>Activities Type: 1<br>Revenues Type: 10<br>Cost Type: 4 | 13 | Cloudleaf, Zoox, Cognosos |
| II. | Solution Type: 0<br>Ecosystem Type: 0<br>Market Type: 3<br>Customer Relation Type: 1<br>Resources Type: 0<br>Partners Type: 0<br>Activities Type: 1<br>Revenues Type: 9<br>Cost Type: 6 | 6 | Smappee, Embark |
| III. | Solution Type: 1<br>Ecosystem Type: 0<br>Market Type: 3<br>Customer Relation Type: 2<br>Resources Type: 0<br>Partners Type: 1<br>Activities Type: 0<br>Revenues Type: 7<br>Cost Type:6 | 6 | SAMLabs, Meural |
| IV. | Solution Type: 7<br>Ecosystem Type: 0<br>Market Type: 3<br>Customer Relation Type: 10<br>Resources Type: 0<br>Partners Type: 0<br>Activities Type: 0<br>Revenues Type: 10<br>Cost Type:4 | 6 | Seebo, scriptr |

*Design Pattern of Successful Business Models*

## Design Pattern I

Design Pattern I creates value by offering Control solutions, which allow to influence the actions and behaviours of objects and persons. As an example, the start-up Avi-on provides clients a Control tool to manage and control lighting infrastructure. They describe their lightning control system as *"[...] purpose-built from the ground up for commercial markets. [...] [we] provide an easy to install, easy change, easy to maintain lighting experience"*.

Design Pattern I solutions can take the form of both goods and services, a characteristic that is not unusual for IoT companies. For instance, the lightning control solutions of Avi-on comprise hardware and software elements. Another example is The Yield, a company offering smart agricultural services, which sells both the sensors and the software to measure and predict soil conditions. They describe their solutions as follows: *"[...] [The Yield] isn't measuring from a weather station 10, 20 or 100 kilometres down the road and leaving you to calculate the difference. It's recording growing conditions under your feet and converting data into on-farm predictions that can help you plan activities with confidence [...]"*.

Design Pattern I companies complement offered solutions with a competitive strategy that relies on turnkey characteristics and superior performance. Turnkey characteristics refer to solutions that are set up instantaneously and intuitively. Avi-on emphasis the easy-to-use, simplistic nature of its products and bases its competitive strategy on the brand message of *"[...] Easy to Install. Easy to Use. Easy to Change [...]".* Superior performance on the other hand is a cornerstone of Cognosos' differentiation strategy. The asset tracker offers a *"[...] smarter, more productive, way to keep up with all your equipment [...]".* Cognosos' solutions *"[...] improve utilization, reduce labour costs, and decrease turnaround time [...]".* To achieve increased efficiency, Cognosos follows Avi-on's and The Yield's example and provides customers with both the hardware and software to track assets and manage inventories.

Because of offering both goods and services, Design Pattern I companies operate on multiple layers of the IoT. As mentioned before, this is a common trait throughout all the observed design patterns. Design Pattern I core functions on these layers focus on monitoring, controlling, and optimizing processes and events.

In the case of Cognosos, expanding solutions onto the Device layer, is essential to realize cost efficiency, as previously mentioned. On the software-based layers that Cognosos is active on, gathered data is stored, processed, and eventually converted into actionable insights. Processed data is controlled and optimized in Cognosos' proprietary RadioCloud ecosystem.

While the core functions describe the purpose of an IoT product or service, the operational focus defines the types of activities performed by a company. For Design Pattern I companies the operational focus is set on Research and Development (R&D) and Services. R&D is a key activity to companies such as Zoox, which is currently in the testing phase of an autonomous vehicle solution. Services on the other hand, refer to activities which contribute to the value of the provided solution. As an integral part of their offering, The Yield provides installation and maintenance services for its smart agriculture solutions.

To be able to carry out core functions and eventually deliver a solution, companies are dependent on different types of resources. Physical resources play an important role for Design Pattern I companies. This type of resources is needed to create devices like Avi-on's lighting control tools, The Yield's sensors and Zoox's car chassis.

A resource all design patterns rely on is data. However, the sources of data differ throughout the four design patterns. In case of Design Pattern, I data is gathered based on the state, context, and usage of a company's solution. This means that data gets originates from a company's solution itself.

Design Pattern I targets B2B clients from Smart Industry domains. Companies are addressing a diversified portfolio of clients. This focus on Smart Industry is unique to Design Pattern I. An example of this Design Pattern I configuration is Cloudleaf, which offers supply chain visibility solutions that are employed throughout multiple contexts and industries. In the words of CEO Mahesh Veerina: *"[...] Today, I solve many of my in-door problems very well. In-door for me would be something like a large warehouse, distribution centre, a factory, or a warehouse. You are tracking your raw materials, supplies, and tools, so whatever you are tracking you want to have visibility on them. I offer that today [...]."*

Concerning client interactions, Design Pattern I companies are distinguishable by a loose interaction intensity. This can be observed in the case of N.thing, a company which provides smart farming equipment online. N.thing's customer interactions are limited to interactions through a conventional online store. In the aspect of interaction intensity, Design Pattern I is identical to Design Pattern III. However, in contrast to Design Pattern III, Design Pattern I solely utilizes financial switching costs as a customer retention mechanism. This mechanism is effective in cases like N.thing's Planty Cube, a 7.5-ton container farm that enables year-long urban farming. The high acquisition costs and the option to stack modules, deters customers from switching provider.

However, even though physical resources are of high relevance for Design Pattern I companies, production and availability of hardware is dependent on Component partners. They play a major role for companies such as Flytrex, a drone delivery service that engages in partnerships with drone manufacturers such as DJI. The partnerships allow Flytrex to focus on its core service – autonomous drone delivery – while having access to equipment of industry leading drone manufacturers. In addition to Component partners, Network and Security partners are essential for Design Pattern I. They ensure that companies can connect their products and services while providing security for all users. Besides having partners on different levels of the value chain, companies rely on Investors, to secure sufficient funding for their business. Business Angels and Venture Capitalists are the investor categories that predominately invest in Design Pattern I companies. In the case of Flytrex, investments from angel investor Joey Low and venture capital firms Armada Investment AG and VI Partners AG resulted in a total funding amount of USD 3,000,000.

Design Pattern I companies utilize Subscriptions as primary revenue model. Connecterra, a provider of dairy monitoring systems, charges monthly fees per cow equipped with a sensor. Flytrex prices the access to its drone Control Center with monthly rates. The monthly rates are feature dependent, which allows an improved targeting of different customer needs. Connecterra provides *"Standard", "Pro"* and *"Flex"* subscriptions, which differ in monthly fees, start-up fees and additional features. Flytrex utilizes monthly fees that dependent on the type of drone you want to access via the Control Center.

**Design Pattern II**

In terms of offered solution, Design Pattern II exhibits multiple similarities to Design Pattern I and Design Pattern III. Remarkedly, all three design patterns provide control tools, which take the forms of both goods and services. Smappee exemplifies a control tool offering goods and services in case of Design Pattern II. The company does not just monitor household energy consumption, but also serves as a hub to manage energy, as it is described on Smappee's homepage: *"[...] you can use Smappee as a smart energy traffic controller, which decides when to use which energy source and to which appliance it should be assigned. As a result, you keep an overview of all the energy flows in the house, so you can save money more easily [...]."* So, like Avi-on in Design Pattern I, Smappee provides a tool to influence the actions of objects.

However, not only type and form of solutions in Design Pattern I, Design Pattern II and Design Pattern III am identical, all three design patterns utilize identical competitive strategies. Focus on superior performance and turnkey characteristics are the dominant strategies for the three configurations.

Design Pattern II follows Design Pattern I's configuration regarding the core functions its solutions incorporate and the operational focus it sets, as both design patterns concentrate on R&D and Services. However, the key resources that are utilized to carry out core functions differ. Design Pattern II's solutions build on human and organizational resources instead of physical ones. Human resources are of major value for companies like DSP Concepts, a company producing high-tech audio solutions for a variety of applications. For this reason, DSP Concepts' *"[...] engineering team includes some of the world's top talent in numerous audio specialties, including microphone and speaker processing, automotive sound, and telematics systems, IoT applications, wireless technologies and more [...]"* (DSP Concepts,

Services). Organizational resources, which comprise elements such as IT-infrastructure, organization structure and intellectual properties, are essential for companies such as Ushr. The Detroit-based company develops mapping systems to facilitate accurate driving of autonomous vehicles. Ushr's main assets are the software and technologies that are utilized during the mapping process.

Clients of Design Pattern II companies are businesses from diversified settings. The environment Design Pattern II's solutions are applied in however, differ from Design Pattern I's application environment. In Design Pattern II, Smart Environment applications are now offered, such as the solution of Smappee, which ensures that *"[…] you always have a clear overview of the energy flows in the house or company premises, wherever you are […]"* (Smappee, My Technology). Another example is Connected Signals, a company which presents its solution as follows *"[…] I eliminate the complexities of securely gathering real-time signal data and making it readily available in a standard format. I combine this data with map, GPS, and speed limit information, and then apply proprietary analytics and algorithms to predict upcoming traffic light behavior. That information is then delivered to vehicles via cellular networks […]."* The traffic light predictions of Connected Signals are aimed at automobile manufacturers, municipalities, and navigation companies.

Design Pattern II closely interacts with customers. Complex, customized products like DSP Concept's audio solutions, require close collaborations between the company providing the solution and its clients. DSP Concepts consults clients throughout planning and implementation phases of new audio technology. They promote their consulting and training services as follows: "I can explore your goals and constraints to help you decide the number of microphones, microphone topology, size and placement of speakers, and which processor to use. I will even work with you to help develop a proof of concept.

In terms of partners, Design Pattern II exclusively collaborates with Component partners and turns away from Design Pattern I's multi-partner approach. Corporate Venture Capital (CVC) entities are the main financial investors in Design Pattern II companies. DSP Concepts attracted investments from BMW's CVC wing, who are interested in DSP Concepts' audio solutions for automotive

For Design Pattern II companies, one-time sales are the preferred revenue model, instead of Design Pattern I's subscription focus. The design pattern pursues a feature dependent pricing strategy. Smappee's energy monitoring and controlling devices are an example for this configuration. A variety of monitors and smart plugs can be bought. Different monitors are available, depending on whether solar energy is measured additionally or not.

**Design Pattern III**

Design Pattern III's solution and ecosystem characteristics are in my aspects identical to Design Pattern I and Design Pattern II. Besides providing their clients with control tools, Design Pattern III also emphasizes superior performance and turnkey characteristics as basis of its competitive strategies.

While solution properties are very similar to other design patterns, Design Pattern III differs in core functions. These are restricted to monitoring and controlling, the aspect of optimizing, that was present with Design Pattern I and Design Pattern II, is absent. Silvair provides a perfect example for this. The lighting control system provides a platform to set up and control lighting systems and provides the firmware to set up tailored control systems. Organizational resources, such as the Bluetooth mesh-based lighting ecosystem, build the basis for Silvair's offering. Together with human resources, organizational resources build the most important building block for Design Pattern III services.

Along diverging core functions, Design Pattern III also exhibits a novel operational focus. Marketing and Operations are the activities with the highest relevance for Design Pattern III, a shift from R&D and Services that were the key operational focuses so far. The company Meural sets its operational focus on Marketing.  Meural creates a smart frame to seamlessly display art works and photographs. Meural relies on a multitude of social media channels, like Facebook, Instagram, Twitter, and Pinterest to promote its B2C-product. Design Pattern III companies rely on external data instead of data created by the solutions themselves. In case of Meural, external data in the form of pictures can be uploaded to the frame.

Meural remains representative of Design Pattern III companies, also in the field of client interaction intensity. Same as Atmoph, a company selling a smart frame acting as a digital window, Meural does not focus on intense customer interactions. On the contrary, both companies distribute their products through their proprietary online store what results in limited customer interaction. Their focus on private clients however, is an exception for Design Pattern III companies, as representatives of this design pattern primarily target B2B clients.

Component partners remain relevant for Design Pattern III affiliated companies. Meural, Atmoph and the baby monitor Cocoon Cam all produce devices that rely on physical components produced by partners. Besides component partners, network and security partners gain relevance for companies again, like Design Pattern I. CVCs represent the main investors in Design Pattern III companies.

Feature based subscriptions are the revenue model of choice for Design Pattern III companies.

## Design Pattern IV

Design Pattern IV introduces an entirely novel type of solution. Instead of providing Control tools, Design Pattern IV offers Execution and Improvement tools. Execution tools enable the performance of an activity or the creation of a good or service. Seebo for example is an industrial IoT platform that empowers customers to *"[…] create digital prototypes of […] industrial IoT use cases, then bring them to life with code-free tools for data connectivity, predictive analytics, automated root-cause analysis, and digital twin visualization […]"* (Seebo, Platform Overview) Seebo's platform acts as an enabler for users to carry out this multitude of functions. scriptr acts as another example for a platform as an execution tool. Like Seebo, scriptr provides developers the tools to build IoT applications.

Improvement tools on the other hand, increase the value of products and services. Passkit improves the process of building a wide array of mobile applications, by offering an easy to use online platform. They promote their service as follows: *"[…] With expertise in mobile wallet, beacons, chatbots, blockchain, and CRM/POS systems, PassKit ensures that businesses of all sizes are able to access the latest innovations in technology through the PassKit platform, making it even easier for brands to have real engagement with customers that build real loyalty […]"*

Design Pattern IV utilizes a competitive strategy that differs from the characteristics of other design patterns. Design Pattern IV competes based on turnkey and integration characteristics. While It is the purpose of integration characteristics to allow the user to easily integrate other solutions to his Design Pattern IV tool. This seamless incorporation is an essential feature of both Seebo's and scriptr's platforms. Thanks to integration features, Seebo's customers can *"[…]  cut sourcing time in half and reduce risk by working with pre-screened partners, ready and available for collaboration within the Seebo IoT Marketplace […]"* This integration characteristic increases the value of execution tools, as it simplifies the connection to other devices.

Design Pattern IV engages in close interactions with its B2B customers. Solutions like Sensorberg require collaboration between clients and solution provider. Sensorberg equips buildings with connected appliances, which essentially enable the digital controlling of building processes and creates *"smart spaces"*. Close interaction between Sensorberg and real estate owners during the planning phase, results in clients being able to *"[...] ditch the extra keys or access cards and start using your smartphone to interact with digitally responsive environments: open doors, book meeting rooms and organize meetings with your colleagues [...]"* (Sensorberg, smart workspace).

Design Pattern IV concentrates on one type of partnership – component partners – and receives financial support from CVCs. Moreover, Design Pattern IV charges subscription fees for its solutions. Differences in fees are based on features of offered products.

### 6.3.7. Design Choice Relevance for Success

Finally, for analysing the large number of cases and identify pattern within them I used another machine learning technique. Therefore, I use a classification tree. Tree-based machine learning approaches fit a relatively simple model on partitions of the feature space divided by a set of rectangles. Although, this approach is conceptually quite simple, it is very powerful in terms of both performance and interpretability of the model and its results (James et al. 2013).

The aim of this approach is to use the cluster membership for each firm along the sub-dimensions of my business model taxonomy (i.e. solution, ecosystem, market, customer relation, activities, resources, partners, revenue, and cost) to define a model that discriminates between successful and non-successful business models. In other words, this means predicting if a business model is successful or not based on its configuration of dimensions and characteristics. For this purpose, I used Random Forests with 1000 estimators (Breiman 2001).

To draw conclusions from this approach, I used the measure of feature importance. The higher, the more important is the feature. The importance of a feature, also known as Gini importance, is computed as the (normalized) total reduction of the feature space caused by that feature (Breiman 2001). For my context, this measure indicates how important a configuration is to predict if a business model is successful.

Configurations within the whole data set of business models and their relationship to determine firm performance. For this purpose, I use binary dummy variable where 0 indicates that the start-up did not receive Series A funding, while 1 indicates the presence of Series A funding (i.e. success). Using Series A funding is a common success proxy for start-up firms and frequently applied in management research (e.g. Baum and Siverman 2004). As I focused on firms that are not older than four years and controlled for rivalry explanations for achieving Series A funding, I assume that the business model configuration can be used as explanatory factor for firm performance. The successful and non-successful business models did not indicate significant differences in firm age, technology trend, and location.

I used the configuration of business model design choices of each start-up along its components as input features for the decision trees. This means that each business model is characterized by its membership along the nine dimensions. Thus, each business model is a configuration of nine clusters (i.e. configuration of single design choices). For instance, the business model of the venture Accerion (https://accerion.tech) is characterized through having the solution configuration 3, ecosystem configuration 1, market configuration 5, customer relation configuration 4, resource configuration 2, partner configuration 2, activities configuration 2, revenues configuration 2, and cost configuration 4. The random forest consists of an ensemble of decision trees each of which is an inductive discriminative model that divides the feature space between successful and non-successful business models.

| Component Configuration | Feature Importance |
|---|---|
| Customer Relation | **0.20** |
| Revenue | **0.20** |
| Market | **0.18** |
| Solution | **0.15** |
| Cost | 0.09 |
| Resources | 0.04 |
| Partners | 0.05 |
| Activities | 0.04 |
| Ecosystem | 0.04 |

*Relevance of Design Choices*

These results can be interpreted as the most important component configuration of design choices to separate successful from non-successful business models. The result show that the component configuration of customer relation design choices and the revenue model are the most important features to define successful business models. Those features are followed by the market addressed by the business model and the solution provided to the customer. Consequently, the four customer centric component configurations of design choices are most relevant in separating successful from non-successful business models in the IoT industry.

## 6.3.8. Discussion and Conclusion

Within this paper, I inductively explored what configuration of components constitute successful archetypes of business models, and what component configurations are most important in defining business model success. To the best of my knowledge this study is the first research that aims at examining a theoretical rational for designing business models on a large scale. Thereby, my study contributes to different streams of business model research in strategic management (Massa et al. 2017).

For this thesis, this examination provides several contributions. First, it develops a context specific exploration of design choices (i.e. a taxonomy) that allows the application in real-world decisional guidance for entrepreneurial decision making. Therefore, it constitutes a cognitive schema that I use in Section 6.5 to create a shared understanding between the entrepreneur and the crowd as well as formal conceptual representation that allows to translate this cognitive schema in a data model to also create a common understanding between humans and machines for the HI-DSS.

Second, the identified design patterns allow me to examine decision model design choices as attributes of real firms, thus, providing ML supported analytical guidance for entrepreneurial decision making.

Finally, the investigated feature relevance of design choices in predicting entrepreneurial success highlight the relevance of stakeholder interaction such as the market and customer in business model design, as those are the most important factors in discriminating between successful and non-successful entrepreneurial ventures. Therefore, the conceptual arguments for including not only supply-side knowledge of experts but also demand-side knowledge of users in providing decisional guidance for entrepreneurs (see Chapter III) is empirically validated.

## 6.4. Designing Hybrid Intelligence Guidance for Entrepreneurship

The findings of this chapter have been previously published as (Dellermann et al. 2017d). This part of the dissertation focuses on developing a novel method to combine human and machine intelligence for making predictions under uncertainty. By conducting a DSR project, I propose a novel method that can be applied for creating guidance for entrepreneurial decision-making. The findings of this study provide the technically core for the design of the HI-DSS in Section 6.5.

### 6.4.1. Introduction

AI is an emerging topic and will soon be able to perform administrative decisions faster, better, and at a lower cost than humans. Machines can consistently process large amount of unstructured data to identify pattern and make predictions on future events (e.g. Agrawal and Dhar 2014; Baesens et al. 2016). In more complex and creative contexts such as innovation and entrepreneurship, however, the question remains whether machines are superior to humans. Machines fail in two kinds of situations: processing and interpreting *"soft"* types of information (information that cannot be quantified) (Petersen 2004) and making predictions in *"unknowable risk"* situations of extreme uncertainty that require intuitive decision-making. In such situations, the machine does not have representative information for a certain outcome and overfits on training data at cost of the live performance of a learner (Attenberg et al. 2015).

One example where both *"soft"* information signals as well as *"unknowable risk"* are crucial, is predicting the success of early stage start-ups. In this case, entrepreneurs face the challenge to decide if the start-up at hand is worth working on or not. Entrepreneurial decision makers often make decisions before neither the feasibility of a new product nor the existence of a market is proven (Maxwell et al. 2011). In such contexts, entrepreneurs do simply not have enough information to assess the quality of a start-up and thus predict the probability of its future success (Dutta and Folta 2016). Moreover, such information might simply not exist under conditions of extreme uncertainty and make the outcome thus unknowable. Consequently, predictions are made for ideas that serve markets, which do not yet exist or novel technologies, where feasibility is still unknown but may provide great returns of investment (Alvarez and Barney 2007). Nevertheless, identifying such unicorns, start-ups that are highly innovative, disrupt traditional industries, and offer tremendous return is highly relevant.

In these situations, humans are still the *"gold standard"* for processing *"soft"* signals that cannot easily be quantified into models such as creativity, innovativeness etc. (Baer and McKool 2014) and make use of an affective judgment tool to recognize pattern in previous decisions: intuition (Huang and Pearce 2015). Using oneś gut feeling proved to be a valuable strategy to deal with extreme uncertainty. However, individual human judges are tainted by bounded rationality in making predictions, which emphasizes that instead of optimizing every decision, humans tend to rely on heuristics (i.e. mental shortcuts) and thus rather focus on highly accessible information (Simon 1955; Kahneman 2011). This often leads to biased interpretation (cognitive processes that involve erroneous assumptions) and may finally result in disastrous predictions (Busenitz and Barney 1997). To solve this problem, research in the field of human computation provides a valuable solution: utilizing the *"wisdom of crowds"* through collective intelligence (e.g. Brynjolfsson et al. 2016; Larrick et al. 2011; van Bruggen et al. 2010). This is a suitable approach to leverage the benefits of humans in prediction tasks, such as providing subjective evaluation of variables that are difficult to measure objectively through machines (e.g. innovativeness) (Colton and Wiggins 2012) or using their prior domain-specific knowledge to make intuitive decision (Blattberg and Hoch 1990). The aggregation of knowledge and resulting predictions than eliminates the statistical errors of individual human decision makers (Larrick et al. 2011). While each of the methods might work well in separation, we argue that combining the complementary capabilities of humans and machines in a Hybrid Intelligence approach allows to make predictions in contexts of extreme uncertainty such as the case of early start-up success through applying formal analysis of *"hard"* information as well as intuitive decision-making processing also *"soft"* information

The aim of this research is to develop a method to predict the probability of success of early stage start-ups. Therefore, I follow a DSR approach (Hevner 2007; Gregor and Hevner 2013) to develop a Hybrid Intelligence method that combines the strength of both machine intelligence such as ML techniques to access, process, and structure large amount of information as well as collective intelligence, which uses the intuition and creative potential of individuals while reducing systematic errors through statistical averaging in an ensemble approach (Shmueli and Koppius 2011). I, thus, intend to show that a hybrid approach improves predictions for the success of start-ups under extreme uncertainty compared to machine or human only methods.

Within the scope of this paper, I first developed a taxonomy of signals that are potential predictors for the success of early stage start-ups based on previous work and domain knowledge (Shmueli and Koppius 2011).

I then designed a method that uses these predictors as input for both ML algorithms as well as collective intelligence to individually assess the probability of success and then weights and aggregates the results to a combined prediction outcome. Moreover, I provide an outlook on the next steps of my research project.

This work thus contributes to several important streams of IS and management research. First, I provide a taxonomy of potential predictors that can be generalized for modelling start-up success predictions (e.g. Böhm et al. 2017). Second, this research adds to literature on predictive research in IS and data analytics (e.g. Chen et al. 2012) by introducing a new method for predicting uncertain outcomes under limited information and unknowable risk by combining collective and machine intelligence in a Hybrid Intelligence Method. This approach allows to complement formal analysis of *"hard"* information and intuitive predictions based on *"soft"* information. Consequently, my research offers prescriptive knowledge in this vein (Gregor and Jones 2007). Third, I contribute to previous work on collective intelligence (e.g. Malone et al. 2009; Wooley et al. 2010) by proposing novel applications of the crowd. Finally, I provide a practical solution that offers entrepreneurial decision makers a useful way to support their innovation decisions.

## 6.4.2. Predicting Start-up Success under Extreme Uncertainty

One way towards understanding predictions in uncertain situations is to examine the mental processes that underlies the cognitive decision-making process. A theory that is particularly helpful in this context is the dual process theory of decision-making. The underlying assumption of this theory is that people make use of two cognitive modes, one is characterized by intuition (system 1) and one by deliberate analytical predictions (system 2) (Tversky and Kahneman 1983; Kahneman 2011).

Predicting the success of early stage ventures is extremely complex and uncertain because frequently just vague ideas are prevalent, prototypes do not yet exist and thus the proof of concept is still pending. Moreover, such ideas might even not have a market yet, but offer great potential of growth in the future (Alvarez and Barney 2007). Consequently, the decision-making context is highly uncertain as neither possible outcomes nor the probability of such are known. This fact can be explained through two concepts: information asymmetry and unknowable risk (Alvarez and Barney 2007; Huang and Pearce 2015).

Information asymmetry describes situation, in which forecasters have incomplete information to decide. When perfect information is absent, decision makers tend to search for various indicators that signal the likeliness of future outcomes (Morris 1987). In my context, such signals include both *"hard"* signals that can be easily quantified and categorized (e.g. industry, technology, team size) as well as *"soft"* signals (e.g. innovativeness, personality of entrepreneur). Humans then try to apply formal analysis to gather signals that support them in making deliberate, rule-based system 2 decisions (Kahneman 2011). On the other hand, unknowable risk defines situations in which a decision maker cannot gather information that signal a potential outcome or make decisions based on formal analysis because the simply not exist. This may be best compared to the error term of a statistical Bayesian model. Unknowable risk covers unexpected events that describe a deviation from status quo (Diebold et al. 2010); (Kaplan and Garrick 1981). In my context, this means for instance identifying a unicorn start-up that gains enormous return that only few would have expected. Formal analytics are not working in these contexts, as representative cases might be missing in previous experience. In such situations, where humans *"don't know what they don't know"*, decision-making is mainly based on intuition (system 1) rather than formal analysis (Tversky and Kahneman 1983; Huang and Pearce 2015). Thus, predicting the success of early stage start-ups is a challenging task and the costs of misclassification are high as they might lead to disastrous decisions or missing valuable chances for return (Attenberg et al. 2015). Previous research in the context of early stage ventures provides strong evidence the best performance in terms of accuracy are provided by combining both types of predictions: analytical (system 2) and intuitive (system 1) (Huang and Pearce 2015).

### 6.4.3. Machine Intelligence for Predicting Start-up Success

For analytically predicting future events, computational statistical methods become particularly valuable due to progresses in ML and AI. They can identify, extract and process various forms of data from different sources (Böhm et al. 2017; Carneiro et al. 2017). Statistical modelling allows to make highly accurate and consistent predictions in the context of financing decisions (Yuan et al. 2016b), financial return ,or bankruptcy of firms (Olson et al. 2012) by identifying patterns in the prior distribution of data and thus predict future events. Machine intelligence is, thus, particularly valuable as biases or limited capacity of human decision makers does not taint it. Statistical models are unbiased, free of social or affective contingence, consistently integrate empirical evidence and weigh them optimally and they are not constraint by cognitive resource limitations (Blattberg and Hoch 1990). Consequently, machine intelligence is a suitable approach for making statistical inference based on prior data and they can learn as the data input grows (Jordan and Mitchell 2015). While such machine intelligence approaches are superior in analytically predicting uncertain outcomes by minimizing the problems of information asymmetries and bounded rationality based on prior distributions of *"hard"* objective variables (e.g. firm age, team size), they are neither able to explain the remaining random error term of such distributions, which I conceptualized through unknowable risk (Aldrich 1999) nor the *"soft"* and subjective signals of new ventures such as innovativeness, the vision or the fit of the team, or the overall consistency of a new venture (Petersen 2004). Both limitations of machines might lead to costly misclassifications (Attenberg et al. 2015). While progresses in the field of AI provide evidence for the applicability of machines in making subjectivity decisions intuitive predictions remain the advantage of humans and require the completion of machine capabilities.

## 6.4.4. Complementary Capabilities with Hybrid Intelligence

In this vein, the benefits of human decision makers come into play. Humans are still the *"gold standard"* for assessing *"soft"* signals that cannot easily be quantified into models such as creativity and innovativeness (Baer and McKool 2014). Humans are talented at making intuitive predictions by providing subjective judgement of information that is difficult to measure objectively through statistical models (Einhorn 1974). Moreover, human decision makers can have highly organized domain knowledge that enables them to recognize and interpret very rare information. Such information might lead to outcomes that are difficult to predict and would rather represent outliers in a statistical model (Blattberg and Hoch 1990). Consequently, using human intuition proved to be a valuable strategy for anticipating start-up success at an early stage (Huang and Pearce 2015).

However, individual decision makers make errors due to their bounded rationality (Simon 1997; (Kahneman 2003). This assumption considers the capacity of the human mind for solving complex problems as rather constraint. Instead of optimizing every decision, individuals tend to engage in limited information accessing to reduce cognitive effort (Hoenig and Henkel 2015). Consequently, they use cognitive heuristics (i.e. mental shortcuts) and simplifying knowledge structures for reducing information-processing demands. One is for example drawing conclusions from a small amount of information or using easily accessible signals (Tversky and Kahneman 1974). Moreover, humans have several biases (cognitive processes that involve erroneous assumptions) that guide the interpretation of information to make predictions (Busenitz and Barney 1997).

Research on human computation provides a solution for these problems (Quinn and Bederson 2011). Collective intelligence leverages the "*wisdom of crowds*" to aggregate the evaluations of a large group of humans, thereby, reducing the noise and biases of individual predictions (Atanasov et al. 2016; Cowgill and Zitzewitz 2015). The value of crowds compared to individuals underlies two basic principles: error reduction and knowledge aggregation (Larrick and Feiler 2015; Mellers et al. 2015). Error reduction is because although individual decision makers might be prone to biases and errors, the principle of statistical aggregation minimizes such errors by combining multiple perspectives (Armstrong 2001). Second, knowledge aggregation describes the diversity of knowledge that can be aggregated by combining the experience of multiple decision makers. Such knowledge aggregation enables to capture a fuller understanding of a certain context (Keuschnigg and Ganser 2016). Thus, collective intelligence can assess the probability of uncertain outcomes by accessing more diverse signals and reduce the threat of biased interpretation.

Using collective intelligence enables to complement a machine model by assessing unknowable risk, which cannot be explained through prior distribution but rather from the combined intuition of humans. ML provides advantages in making analytical predictions based on *"hard"* information while collective intelligence offers benefits in making intuitive predictions taking also *"soft"* information into account. Previous work emphasizes the complementary nature of humans and statistical models in making predictions about future events in various settings such as sports, politics, economy, or medicine (Nagar and Malone 2011; Meehl 1954; Dawes et al. 1989; Einhorn 1972; Ægisdóttir et al. 2006). I thus argue that a Hybrid Intelligence Method that combines the complementary capabilities of analytical and intuitive predictions is most accurate for predicting the success of early stage start-ups.

| | Access of Signals | Process of Signals | Interpretation of Signals |
|---|---|---|---|

| Limitations of Individual Decision Makers | *Bounded Rationality* | | |
|---|---|---|---|
| | Resource Constraints | Use of Heuristics | Biases |
| **Benefits of Machine Intelligence** | Automated Access to High Volume of Hard Information | Consistent Processing of Hard Information | Identifying Pattern in Prior Distribution of Hard Information |
| **Benefits of Collective Intelligence** | Aggregation of Diverse Hard and Soft Information | Aggregated Processing of Hard and Soft Information | Error Reduction of Individual Bias in Interpreting Hard and Soft Information |

*Rational for Complementary Capabilities*

## 6.4.5. Methodology

To develop a method that capitalizes the benefits of both machine and collective intelligence, I followed a DSR approach (Hevner et al. 2004; Gregor and Hevner 2013) guided by schematic steps of (Shmueli and Koppius 2011). The result of this design science project constitutes a hybrid intelligence method as a new and innovative artefact that helps to solve a real-world problem. Following Hevner's (2007) three cycle view of relevance and rigor I combine inputs from the practical problem domain (relevance) with the existing body of knowledge (rigor). Abstract theoretical knowledge thus has a dual role. First, it addresses the suggestions for a potential solution. Second, the abstract learning from my design serves as blueprint to provide prescriptive knowledge for solving similar problems in the future (Gregor and Jones 2007).

## 6.4.6. Development of a Solution

As outlined above, both collective and machine intelligence approaches have specific benefits to predict uncertainty future events. To develop a novel method to predict the future success of early stage start-ups, I decided to combine an ensemble of machine and collective intelligence due to several reasons. First, combining multi methods has a long tradition in research on forecasting as combined methods and sources should be at least similar or better than the best individual prediction method (Armstrong 2001). Second, previous research considered the strengths of human decision makers and statistical models and anticipated their complementary capabilities in making predictions (Blattberg and Hoch 1990; Nagar and Malone 2011). Third, I argue that for contexts of extreme uncertainty due to both information asymmetry and unknowable risk, a hybrid intelligence method consisting of collective and machine intelligence, is superior to one or the other approach alone. A hybrid intelligence method for predicting the success of new ventures enables to statistically model a prediction by consistently accessing and processing *"hard"* signals based on Bayesian inference, while collective intelligence allows to access more diverse *"soft"* signals through aggregation and reducing the systematic errors of individual humans, which also helps to reduce the random error term of predictions that reflects unknowable risk through the benefits of human intuition.

A fully automated system with a large flow of information requires data mining trough web crawling approaches and pre-processing the raw data. Within the scope of this paper I focus on the main part of the hybrid intelligence method by explaining input metrics, the automation process and expert weighting to predict the success of early stage start-ups.

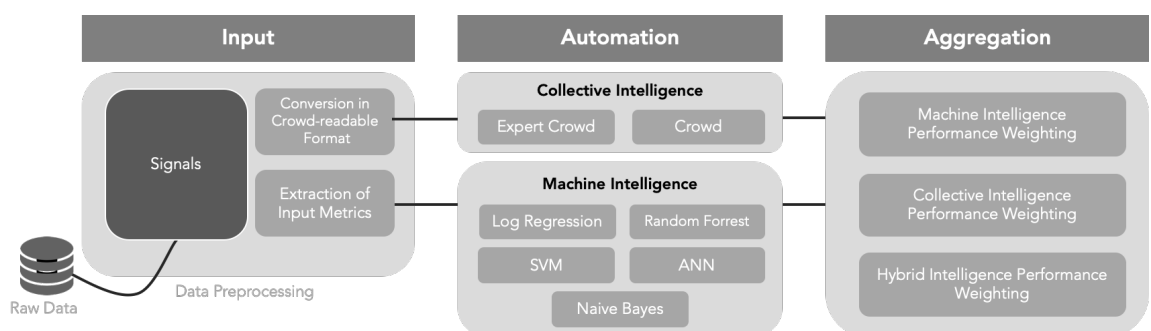

## 6.4.7. Predicting Success of Early Stage Start-ups

The objective of my solution is to accurately predict the success of early stage start-ups and thus make accurate predictions under extreme uncertainty. To define the success of early stage start-ups, I assessed whether they received Series A funding and used this funding as a proxy. This is the commonly accepted indicator for success in this context (Baum and Silverman 2004; Spiegel et al. 2016). Series A funding defines the first venture capital backed funding that allows new ventures to grow. The objective is thus to provide a probabilistic classification of the binary categorical variable Series A funding.

## 6.4.8. Taxonomy of Signals

To identify relevant metrics for the input of the hybrid intelligence automation I first conducted a literature analysis (Webster and Watson 2002) and then conducted interviews (n=15; average duration 60 min) and focus group workshops (n=6; 3-7 participants; average duration 90 min) with experienced angel investors to iteratively combine the findings from the literature review with practical relevant factors following the procedure of Nickerson et a. (2013). Based on these findings, I developed a taxonomy of relevant input metrics for a hybrid intelligence method in the context of predicting the success of early stage ventures (Song et al. 2008) (see Table 15). As I focus on early stage start-ups, I do not include financial metrics, which are barely relevant in this stage (e.g. Maxwell et al. 2011).

## 6.4.9. Machine Input Metrics

From this taxonomy, I choose the *"hard"* signals that can be quantified and translated into machine readable format. I, therefore, used the metrics as displayed in Table 15.

# Conversion in Crowd-readable Format

To automate the prediction through collective intelligence, the signals need to be converted into crowd-readable format. Therefore, I designed a GUI (graphical user interface) that consist of all *"hard"* and *"soft"* signals of my taxonomy (Table 14). This ensures that each human can make a prediction based on a comprehensive representation of the start-up. A graphical approach was chosen to improve the information extraction capabilities that are necessary for making these predictions (e.g. Todd and Benbasat 1999). The GUI is based on a standardized ontology that was developed from my taxonomy of signals (Burton-Jones and Weber 2014).

## 6.4.10. Automation

### Machine Intelligence

To apply machine intelligence, I use the above-mentioned metrics as input for probabilistic classification of Series A funding. I therefore choose the following learning algorithms due to their different methodological benefits:

**1. Logistic Regression:** a very well-known linear regression algorithm used as the baseline algorithm. Frequently applied for binary choice models (Magder and Hughes 1997).

**2. Naive Bayes:** Bayesian parameter estimation problem based on some known prior distribution (Valle et al. 2012).

**3. Support Vector Machine (SVM):** classification algorithm based on a linear discriminant function, which uses kernels to find a hyperplane that separates the data into different classes (Vapnik 1998).

**4. Artificial Neural Network (ANN):** a biologically inspired, non-parametric learning algorithm that can model extremely complex non-linear function (Haykin 2009).

**5. Random Forests:** a popular ensemble method that minimizes variance without increasing bias by bagging and randomizing input variables (Breiman 2001).

**Collective Intelligence**

The collective intelligence automation is designed as judgment task (Riedl et al. 2013). This asks each human participant of the crowd to rate the start-up on a multi-dimensional 7-point Likert scale, which provides the most accurate results in collective intelligence prediction (Blohm et al. 2016). Each start-up is then assessed along the most relevant dimensions to indicate success: feasibility, scalability, and desirability. To execute the actual rating process, I use two instances of collective intelligence: a crowd of non-experts and an expert crowd. This is consistent with previous research that indicates that a larger group of non-experts as well as a smaller group of domain experts are both capable to predict uncertain outcomes (Klein and Garcia 2015; Mannes et al. 2014).

**1. Crowd:** open call to a crowd of non-experts. 16-20 humans judge each start-up to predict its success and leverage the benefits of collective intelligence in non-expert samples (Keuschnigg and Ganser 2017). The non-expert crowd is requested through a crowd working platform (Amazon Mechanical Turk). Language proficiency is ensured through choosing ME residents as my start-up descriptions were generated in English and quality selection is conducted through a minimum HIT approval rate of 95% as selection criteria. Each participant is financially compensated.

**2. Expert crowd:** restricted open call to humans that have high expertise and domain knowledge (i.e. business angels, start-up mentors, consultants) 5-7 experts judge each start-up to predict its success and leverage the benefits of collective intelligence in expert samples (Mannes et al. 2014). The expert crowd is requested through a start-up mentor network and is not be not financially compensated.

The collective intelligence automation is then calculated as unweighted average that combines each crowd memberś ordinal evaluations into a single score indicating the probabilistic classification of Series A funding success, which provides the most accurate combined predictions (Keuschnigg and Ganser 2016).

## 6.4.11.Aggregation

To benefit from the hybrid intelligence method, I finally need to aggregate the predictions of each individual human and machine intelligence approach to generate the output of probability of Series A funding. I thus apply two types of aggregation. First, I use simple unweighted averaging as baseline approach which proved to be accurate for many types of predictions (Armstrong 2001; Keuschnigg and Ganser 2016). Second, I follow the idea of combining different individual predictions of the crowd and ML algorithms according to their performance (Archak et al. 2011). For this purpose, I propose three of this weighting algorithms to identify the most accurate one. Machine intelligence performance weighting that is based on the predictions of only ML algorithms. Collective intelligence performance weighting that is based on the prediction of the crowd. I argue that in contexts involving consumer product start-ups, non-expert crowds might be superior as they represent the voice of potential customers and providing demand side knowledge whereas expert crowds are more accurate in B2B settings (e.g. Magnusson et al. 2016). Finally, hybrid intelligence performance weighting combines the predictions of machines and humans to predict the success outcome.

## 6.4.12. Implementation and Evaluation

To evaluate my method, I implement the input, automation, and aggregation procedures and evaluate their performance. I started by creating a data set of start-ups by combining information on the signals in my taxonomy from several databases (CrunchBase, Mattermark, and Dealroom). Thereby, I focused on tech start-ups as the high-tech context is particularly tainted with extreme uncertainty. The data sample consists of 1500 start-ups from different industries and all extracted signals from the taxonomy. Part of the sample received Series A funding while the other part did not. All start-ups in my dataset are labelled accordingly. I pre-processed the *"hard"* signals for the machine algorithms and converted all *"hard"* and *"soft"* signals into crowd-readable format (an ontology-based GUI). In the next steps, I standardized the input metrics and test to avoid overfitting (Carneiro et al 2017). The automation process of the *"hard"* and *"soft"* signals then be conducted as described in the previous section.

The dataset is then randomly split into two sub-sets, one for training and one for testing. The training data is used to create the prediction models. For my purpose, I choose a 10-fold cross validation approach to split my data set into ten mutually exclusive sub-sets of approximately equal size. The idea behind 10-fold cross validation is to minimize the bias associated with the random sampling of the training and holdout data samples. Each of my proposed prediction approaches (i.e. machine algorithms and the crowd) is then trained and tested ten times with the same ten folds, which means the algorithm is trained on nine folds and tested on the remaining single fold. Cross validation then estimates the overall accuracy of an algorithm by calculating the mean accuracy (e.g. Olson et al. 2012).

The evaluate the performance of my method I use the Matthews correlation coefficient (MCC) that is a well-known balanced performance measure for binary classification, when the classes within the data are of very different sizes (Matthews 1975). I choose this measure as my data set is biased towards successful start-ups although I also identified a large amount of failed ventures, which is a common limitation of such databases (e.g. Böhm et al. 2017). I then use logarithmic regression as baseline algorithm and compare the performance of each individual machine intelligence algorithm, crowd prediction, and weighting algorithm through a two-way analysis of variance (ANOVA) (Bradley 1997). I therefore intend to identify the best combined approach and aim at showing that Hybrid Intelligence approach provides superior results than a machine or human only prediction.

## 6.4.13. Conclusion

Predicting the success of early stage start-ups is a challenging task and the costs of misclassification is high as it might lead to disastrous funding decisions are missing valuable chances for return. To make predictions in such contexts, I propose to combine the complementary capabilities of machine and human intelligence. While machines are particularly beneficial in consistently processing large number of *"hard"* signals that indicate the success of a new venture humans are superior in interpreting *"soft"* signals such as the personality of an entrepreneur or the innovativeness of a new product. Moreover, humans can leverage their intuition to identify valuable start-ups that cannot be find by relying on previous data. To overcome the constraints of bounded rationality of individuals, I thus suggest leveraging collective intelligence. To reach my aim, I developed a preliminary Hybrid Intelligence method that I initially evaluate as I proceed my research. In the next steps, I will then also test its applicability for other outcome variables in the context of start-ups (e.g. growth, survival rate etc.) and other contexts of extreme uncertainty (e.g. innovation in general).

Moreover, I intend to assess the relevance of accuracy and transparency with potential users of this method and if they are more willing to take advice when human sources are included (e.g. Önkal et al. 2009). I expect my research to make several contributions to both academia and practice. First, I provide a taxonomy of potential predictors that can be generalized for modelling start-up success predictions (e.g. Böhm et al. 2017). Second, this research adds to literature on predictive research in IS and data analytics (e.g. Chen et al. 2012) by introducing a new method for predicting uncertain outcomes under limited information and unknowable risk by combining collective and machine intelligence in a Hybrid Intelligence method. This approach allows to complement formal analysis of *"hard"* information and intuitive predictions based on *"soft"* information. Such hybrid method might be valuable for other settings of extreme uncertainty as well. Consequently, my research offers prescriptive knowledge in this vein that might be generalizable for data science methods in general (Gregor and Jones 2007). Third, I contribute to previous work on collective intelligence (e.g. Malone et al. 2009; Wooley et al. 2010) by proposing novel applications of machines and crowd. I argue that my proposed approach can augment the capabilities of collective intelligence in general. While I use a parallel approach in this paper, further research might explore how machine intelligence might be leveraged as feedback for the crowd and thus point towards more collaborative interactive approaches (Dellermann et al. 2019). Finally, I provide a useful solution for a practical prediction problem that may support entrepreneurs in making decisions and potentially reduce the frequency of failure.

## 6.5. Designing a Hybrid Intelligence System for Guiding Entrepreneurial Decisions

The findings of this chapter have been previously published as (Dellermann et al. 2018a).This part of the dissertation finally combines the findings from the previous sections to design and evaluate a DSS that provides guidance for entrepreneurial decision-making. By conducting a DSR project, I merge the findings of using crowdsourcing as a mechanism to integrate collective ecosystem intelligence in entrepreneurial decision-making as well as the idea of hybrid intelligence that can improve such guidance in highly uncertain contexts that require the complementary capabilities of human and machines. The results of this Section are one of the core contributions of my thesis and the web application can be accessed at www.ai.vencortex.com.

### 6.5.1. Introduction

The rapid digital transformation of businesses and society generates great possibilities for developing novel business models that are highly successful in creating and capturing value. Many Internet start-ups such as Hybris, Snapchat, and Facebook are achieving major successes and quickly disrupting whole industries. Yet, most early-stage ventures fail. Nearly 90% of technology start-ups do not survive the first five years. One reason for this is that entrepreneurs face tremendous uncertainties when creating their business models. Consequently, entrepreneurs must constantly re-evaluate and continuously adapt their business models to succeed (Ojala 2016). This task is characterized by high levels of uncertainty concerning market and technological developments. In addition, entrepreneurs cannot be sure whether their competencies and internal resources are suitable to successfully run the new venture (Andries and Debackere 2007). Therefore, entrepreneurs try to collect information that might support them in their decision-making. Such information includes the following:

analytical data such as market or financial data; feedback from customers and other stakeholders; and guidance from associate mentors, business angels, and incubators. This information is used to assess the validity of their assumptions and make decisions that are necessary to succeed (e.g. Shepherd 2015; Ojala 2016).

However, ways to get decision support in the process of business model validation are limited (Dellermann et al. 2017c). One is the use IT-supported tools to provide guidance for incumbent firms, as shown by some existing research (Gordijn et al. 2000; Gordijn et al. 2001; Gordijn and Akkermans 2003; Gordijn and Akkermans 2007; Haaker et al. 2017; Daas et al. 2013; Euchner and Ganguly 2014). These tools frequently rely on formal analysis of financial data and forecasts that might work well for established companies. However, the applicability of these approaches in the start-up context remains challenging because business ideas are vague, prototypes do not yet exist, and thus the proof of concept is still pending. Moreover, early-stage start-ups might not have a market yet but offer great potential of growth in the future (Alvarez and Barney 2007). Another way to deal with uncertainty during the validation of a business model is the validation of the entrepreneur's assumptions by testing them in the market or with other stakeholders (Blank 2013). Such validation allows the entrepreneur to gather feedback, test the viability of the current perception of a business model, and adapt it as necessary. This approach includes both formal analysis and the intuition of human experts, which has proven to be a valuable combination in such uncertain settings (Huang and Pearce 2015). For this purpose, traditional offline mentoring is the state of the art in both theory and practice. However, such offline mentoring provides only limited possibilities for scalable and iterative decision support during the design of a business model (e.g. Hochberg 2016).

Therefore, the purpose of this study is to develop a DSS (DSS) that allows the iterative validation of a business model through combining both social interaction with relevant stakeholders (e.g. partners, investors, mentors, and customers) and formal analysis for the extremely uncertain context of business model development in early-stage start-ups. Such a combination proved to be most valuable for decisions in this setting (e.g. Huang and Pearce 2015). I propose a HI-DSS (HI-DSS) that combines the strength of both machine intelligence to handle large amount of information as well as collective intelligence which uses the intuition and creative potential of individuals while reducing systematic errors through statistical averaging. I follow a design science approach (Hevner 2007; Peffers et al. 2007), making use of both knowledge from previous research that proved to be valuable in various contexts of uncertain decision-making, as well as practical insights, to develop principles for an IT artefact.

My contribution is threefold. First, my research provides prescriptive knowledge that may serve as a blueprint to develop similar DSSs for business model validation in the context of early-stage start-ups (Gregor and Jones 2007). In fact, my research provides prescriptive knowledge regarding DPs (i.e. form and function) as well as implementation principles (i.e. my proposed implementation). Second, I contribute to research on decision support for business model validation by augmenting formal analysis of data through iterative social interaction with stakeholders. Third, I propose a novel approach to support human decision-making by combining machine and collective intelligence and thus contribute to recent research on ensemble methods (McAfee and Brynjolfsson 2017; Brynjolfsson et al. 2015) .

## 6.5.2. Business Models and Business Model Design Guidance

To formulate the problem for my design research approach, I reviewed current literature on business model validation. The concept of the business model has gathered substantial attention from both academics and practitioners in recent years (Veit et al. 2014; Al-Debei and Avison 2010). In general, it describes the logic of a firm to create and capture value (Zott et al. 2011). The business model concept provides a comprehensive approach toward describing how value is created for all engaged stakeholders and the allocation of activities among them (Bharadwaj et al. 2013). In the context of early-stage start-ups, business models become particularly relevant as entrepreneurs define their ideas more precisely in terms of how market needs might be served. A business model reflects the assumptions of an entrepreneur and can therefore be considered as a set of *"[…] hypotheses about what customers want, and how an enterprise can best meet those needs and get paid for doing so […]"* (Teece 2010: 191). Thereby, entrepreneurs make several decisions regarding the design of a business model such as how a revenue model, value proposition, and customer channels should be constructed. Thus, a business model can be used as a framework for constructing start-ups and to conduct predictive what-if scenario analysis to determine the feasibility of an entrepreneur's current pathway (Morris et al. 2005).

However, such scenario analysis concerning an entrepreneur's assumptions about what might be viable and feasible are mostly myopic in terms of the outcome because entrepreneurs are acting under high levels of uncertainty (Alvarez and Barney 2007). Consequently, entrepreneurs must start a sensemaking process to gather information for validating and refining their initial beliefs and guiding future decision-making. During this process, the entrepreneurs refine the business model through iterative experimentations and learning from both successful and failed actions. These design decisions determine how a business model is configured along several dimensions (Alvarez et al. 2013; Blank 2013). When the entrepreneurs' assumptions contradict with the reaction of the market, this might lead to a rejection of erroneous hypotheses and require a reassessment of the business model to test the market perceptions again. Thus, the business model evolves toward the needs of the market and changes the assumptions of entrepreneurs (Ojala 2016). The success of start-ups, thus, depends heavily on the entrepreneurs' ability to develop and continuously adapt their business models to the reactions of the environment by making adequate decisions (Spiegel et al. 2016).

## 6.5.3. Decision Support for Business Model Design

Decision support can assist in making business model design decisions (i.e. how a business model should be constructed) in several ways. One, previous research on decision support and validation in the context of business model analysis mainly focuses on analytical methods such as modelling and simulation (e.g. Gordijn et al. 2001; Haaker et al. 2017; Daas et al. 2013; Euchner and Ganguly 2014). Business model simulations provide a time-efficient and cost-efficient way to help decision makers understand the consequences of business model adaptions without requiring extensive organizational changes (Osterwalder et al. 2005).

In this vein, previous research applies quantitative scenarios analysis to predict the viability of design decisions in the context of business model design for platforms (Zoric 2011) and mobile TV (Pagani 2009), as well as scenario-planning methods for IP-enabled TV business models (Bouwman et al. 2008). Another way of providing guidance in the design of a business model focuses on stochastic analysis of financial models to identify the most important drivers of financial performance in incumbent firms, such as Goodyear (Euchner and Ganguly 2014). A third popular approach evaluates business model design choices against a potential scenario of changes in stress-testing cases (Haaker et al. 2017).

Although most of this research considers the importance of the consistency of causal business model structures and the complex interrelations of components, existing methods do not consider how the effects of changes in a business model unfold dynamically over time and the iterative process of developing business models especially for new ventures (Cavalcante et al. 2011; Demil and Lecocq 2010). Most of these approaches are rather static and thus only few can capture the dynamics that underlie the complex interactions of business model design decisions in practice (Möllers et al. 2017). Such analytical methods to support decisions in the context of business model validation lack the capability to identify complex pattern of components that lead to success. While these methods are valuable for incumbent firms, they are not very applicable for early-stage start-ups. Predicting the success of early-stage ventures'business models is extremely complex and uncertain. This is since neither possible outcomes nor the probability of achieving these outcomes are known, i.e. situations of unknowable risk (Alvarez and Barney 2007). Little data is available and quantifying the probability of certain events remains impossible. In such contexts, formal analysis is a necessary but not sufficient condition to assess if a certain business model design might be viable in the future (Huang and Pearce 2015).

In such situations, entrepreneurs pursue two strategies. First, they seek and gather available information that they can process to guide analytical decision-making (e.g. Shepherd et al. 2015). Second, entrepreneurs rely on their experience and gut feeling to make intuition-based decisions. Such intuition has proven to be a valuable strategy for decision-making under uncertainty (Huang and Pearce 2015). While relying on gut feeling and intuitive decision-making is the purview of successful entrepreneurs during the validation of their business models, I argue that the assessment, processing, and interpretation of additional information that reduces uncertainty and guides decision-making needs support due to the limitations of bounded rationality (March 1978; Simon 1955). This is because entrepreneurs are limited in their capability to access and process information extensively, and therefore not able to optimize their decisions. Moreover, the interpretation of accessed and processed information is constrained by biases and heuristics, frequently leading to bad decisions (Bazerman and Moore 2008; Bazerman and Chugh 2006; Kahneman and Frederick 2002; Kahneman and Egan 2011; Thaler and Sunstein 2008).

Because entrepreneurial decision makers are constrained by bound rationality, start-up mentoring has emerged as strategy to support entrepreneurs in making the required decisions. Mentors (i.e. experienced consultants, experts, or successful entrepreneurs) attempt to help the early-stage start-up team to gain problem-solution fit by conducting one-to-one support initiatives (such as workshops) and offer entrepreneurs methods to develop their idea into a novel venture (Cohen and Hochberg 2014). Such social interaction with relevant stakeholders is frequently offered by service providers such as incubators and more recently accelerators (Pauwels et al. 2016).

However, neither academia nor managerial practice are offering IT-based solutions to iteratively provide such guidance. This is unfortunate, since IT-based solutions have the potential to provide scalable and cost-efficient solutions by leveraging the wisdom of multiple and diverse mentors, iterate the validation and adaption process, and allow the transference of many entrepreneurs' experiences to a single entrepreneur, thereby increasing the learning rate of the individual entrepreneur.

## 6.5.4. DSR Project Methodology

Novel solutions are needed to address the limitations of individual decision makers resulting from their bounded rationality and the lack of scalable solutions for providing guidance in business incubators and accelerators. To provide IT-supported forms of guidance to entrepreneurs, I conduct a DSR project (Peffers et al. 2007; Gregor and Hevner 2013) to design a new and innovative artefact that helps to solve a real-world problem by providing a high-quality and scalable tool for decision guidance. To conduct my research, I followed the iterative DSR methodology process of Peffers et al. (2007) consisting of six phases:

*(1) problem identification and motivation*

*(2) objectives of a solution*

*(3) design and development*

*(4) demonstration*

*(5) evaluation*

*(6) communication*

I used a multi-step formative ex-ante and summative ex-post evaluation in a naturalistic setting with domain experts and potential users to ensure the validity of my results (Sonnenberg and Vom Brocke 2012; Venable et al. 2016). My research starts with phase 1: i.e. the formulation of the problem that is perceived in practice. To ensure both relevance and rigor, I use inputs from the practical problem domain (relevance) and the existing body of knowledge (rigor) for my research project (Hevner 2007).

Abstract theoretical knowledge has a dual role. First, it addresses the suggestions for a potential solution. Second, the abstract learning from my design serves as prescriptive knowledge to develop a similar artefact in the future (Gregor and Jones 2007).

Therefore, I conducted a literature review on decision support in the context of business model validation. To refine and validate the relevance of this problem, I conducted an explorative study within the problem domain using qualitative interviews (e.g. Dul and Hak 2007; Yin 2017). I collected data concerning the business model validation process within business accelerators and incubators. I conducted a series of expert interviews with executives at business incubators and accelerators (n = 27), entrepreneurs (n = 32), and mentors (n = 16). I gained access to interviewees in the context of a project funded by the German Ministry of Research and Education.

My project partners then helped me with a snow sampling approach to gain access to additional participants in their network. The statements of the interviewees were coded and analysed by two of the researchers to identify common themes. My coding procedure was structured and guided by the limitations derived from literature (Strauss and Corbin 1990). This approach allowed me to justify the research gap in practical relevance before designing an artefact (Sonnenberg and Vom Brocke 2012).

In a second step, I analysed previous research on decision support to identify a body of knowledge that provides suggestions for a potential solution resulting in a scientifically grounded version of DPs. To evaluate my design, I used a combination of exploratory and confirmatory focus groups (Hevner and Chatterjee 2010). Originated in the field of psychology, the focus group gained increasing popularity as a knowledge elicitation technique in the field of software engineering (Massey and Wallace 1991; Nielsen 1997). I used exploratory focus groups to gather feedback for design changes and refinement of the artefact. This was used as formative evaluation procedure to iteratively improve the design. Moreover, a confirmatory focus group was applied to demonstrate the utility of my artefact design in the application domain (Tremblay et al. 2010).

The initial version of the tentative DPs was demonstrated, validated, and refined using eight focus-group workshops (6–8 participants; average duration 60 min) with mentors, executives, and software developer. The DPs were visualized, explained, and discussed to formatively evaluate the completeness, internal consistency, and applicability of my ex-ante design (Sonnenberg and Vom Brocke 2012; Venable et al. 2016).

In the next steps, I instantiated the tentative DPs of form and function into an IT artefact. I then applied a summative ex-post evaluation of the design through a qualitative evaluation in a naturalistic setting with potential users. Therefore, I conducted eight focus group workshops with mentors, executives at incubators and corporate accelerators, and entrepreneurs (2–4 participants; average duration: 60 min). The instantiated artefact was explained to the participants and demonstrated by a click-through. The participants then had the possibility to use the artefact and were then asked to assess its effectiveness, efficiency, and fidelity with the real-world phenomenon, which leads to the final version of principles of form and function (Venable et al. 2016).

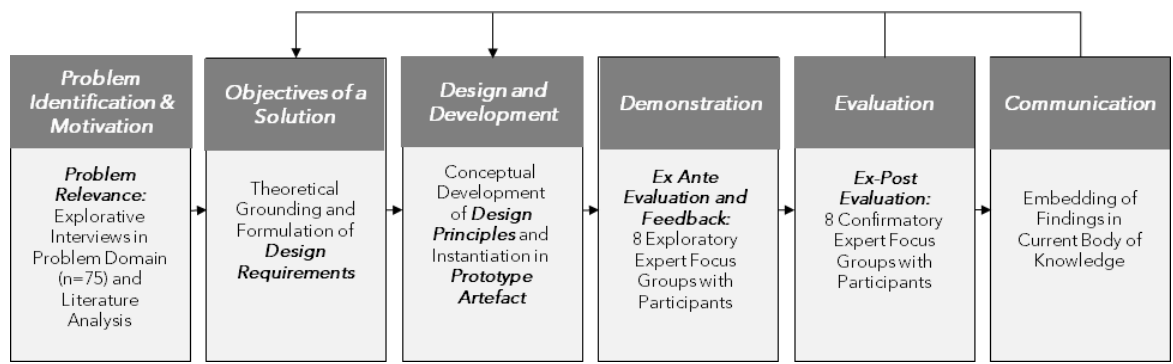

**DSR Project Process**

## 6.5.5. Problem Verification (Phase 1)

To enable a two-sided perspective on the problem and to ensure the practical relevance of the identified gap (Sonnenberg and Vom Brocke 2012), I conducted a total of 75 exploratory interviews with entrepreneurs and members of incubators. The interviews were guided by the central question of how incubators as service providers typically support the design decisions of entrepreneurs' business models and the perceived limitations of these approaches. Therefore, the interviews were used for understanding the problem domain and gathering deep insights in the real-world phenomenon. By analysing these interviews, I gained a deeper understanding of the process of business model validation for start-ups. In sum, it turned out that many entrepreneurs face three types of problems when trying to access the quality of their business models. One type involves the bounded rationality of humans that prevents them from extensively searching for the required information and leads to biased decisions. A second type concerns the limitations of current forms of decisional guidance provided to entrepreneurs (e.g. mentoring in business incubators) that prevent the mentors from providing optimal. These limitations include a limited domain of knowledge especially concerning novel technologies, a lack of cognitive flexibility, and the subjectivity of the evaluations.

Moreover, resource constraints of institutional mentoring organizations in general make it nearly impossible to find the perfect decisional guidance for each business model case and constrain iterative development. The third type of problems deals with the limitations of existing IT-based tools as discussed in the previous section. Table 16 summarizes the aggregated finding of my literature analysis as discussed in the Related Work section, combined with the findings of the interviews. Based on this, I draw conclusions for consequential objectives. In sum, the interviews delivered a detailed overview of the problems that entrepreneurs face when trying to assess the quality of their business model. However, I was not able to identify knowledge on how to support this decision process with the help of an IT tool.

## 6.5.6. Objectives of a Solution (Phase 2)

I therefore investigated existing design theories (i.e. kernel theories) that have been used to solve similar problems (Gregor and Hevner 2013).

As outlined, when making decisions regarding their business model, entrepreneurs must improve decision quality to make successful business model design decisions (Sosna et al. 2010). This is obvious as making the most appropriate decisions at a certain point of time is maybe the most challenging task for entrepreneurs (Alvarez et al. 2013). Such design decisions regarding certain business model components can support or prevent the entrepreneur from achieving milestones such as gaining a viable market position or receiving funding (Morris et al. 2005). In the context of business incubators and other support activities for entrepreneurs, providing guidance through mentoring proved to be the most suitable approach for helping entrepreneurs to design business models.

One other domain where decisional guidance has proved to be a suitable approach is research on DSSs in various contexts of IS research (Silver 1991; Morana et al. 2017; Parikh et al. 2001; Limayem and DeSanctis 2000). Such guidance – which supports and offers advice to a person regarding what to do – was examined especially in the context of DSSs (Silver 1991). In this vein, decisional guidance describes design features that enhance the decision-making capabilities of a user (Morana et al. 2017). The adaption of this finding to the context of this research project (i.e. entrepreneurial decision-making) is promising.

To support decision makers in executing their tasks, they must be provided with design features of decisional guidance differentiated along ten dimensions (Morana et al. 2017; Silver 1991). First, the target of guidance supports the user in choosing which activity to perform or what choices to make. Second, the invocation style defines how the decisional guidance is accessed by the user, such as automatically, user invoked, or intelligently (Silver 1991). Third, guidance can be provided at different timings such as during, before, or after the actual activity (Morana et al. 2017). The timing for the context of business model design choices should not be time-specific but rather should be accessible during various time points during the process (Silver 1991).

Fourth, decisional guidance can be provided for novices and experts (Gregor and Benbasat 1999). This dimension of design features is also not relevant for my context as entrepreneurs vary in their expertise from novices to experts that are serial entrepreneurs. Fifth, the trust building of decisional guidance is not explicitly covered for business model design decisions (Silver 1991). As this is not an issue in traditional offline mentoring either, I propose that trust issues – though highly relevant for future research – be outside the scope of this paper. Finally, the content dimension of decisional guidance is defined as the purpose of the guidance provision (Morana et al. 2017). As this is highly interrelated with the intention of decisional guidance, I did not explicitly highlight this design feature as a requirement for my solution.

## Deriving Objectives from Literature on Decisional Guidance

For the scope of this research, I focused on four of these guidance dimensions that are particularly relevant for business model decision support: the form (or directivity), the mode, the intention, and the format of decisional guidance. Decisional guidance can be divided into two forms of guidance: informative guidance, which provides the user with additional information; and suggestive guidance which offers guidance for a suitable course of action.

Informative guidance is used as expert advice to provide declarative or definitional knowledge, thereby helping the user to increase the understanding of a decision model (Limayem and DeSanctis 2000). Suggestive guidance makes specific recommendations on how the user should act (Arnold et al. 2004). This form of guidance can improve decision quality and reduce resource requirement for making decisions (Montazemi et al. 1996).

The mode of guidance describes how the guidance works. This covers the design feature of how the guidance is generated for the user. It can be predefined by the designer and thus be statically implemented into a system a priori, dynamically learned from the user and generate decisional guidance on demand, and participative depending on the user's input (Silver 1991). Dynamic guidance is particularly useful to improve decision quality, user learning, and decision performance, while participative guidance also increases the performance of users in solving complex tasks (Parikh et al. 2001; Morana et al. 2017).

The intention of guidance describes why guidance is provided to users. This might for instance include clarification, knowledge, learning, or recommendation (Arnold et al. 2004; Gönül et al. 2006). Decisional guidance can be provided with the intention to provide specific recommendations on how to act or expert advice to help users solve problems and make decisions.

Finally, the format of guidance pertains to the manner of communicating the guidance to the user. Decisional guidance can be formatted for instance as text or multimedia (images, animation, audio) to make it more accessible for the user. The format of guidance be should selected depending of the characteristics of the underlying task (Gregor and Benbasat 1999; Morana et al. 2017).

## Formulation of DRs

In general, a system that supports the entrepreneur in making design decisions during business model validation needs to support the entrepreneur in executing her task. This requires a certain combination of guidance that is specific for the context (Silver 1991). For my research, I structured my DRs along the four dimensions that were identified as suitable for the class of decision problem.

First, the form of guidance provided needs to include information about the probability of success of the current version of the business model. This means providing a forecast on the probability of having success in the future such as receiving funding, survival, growth etc.: i.e. informative guidance (Morana et al. 2017; Silver 1991). Therefore, I formulate the following DR:

**DR1:** *Business model validation should be supported by a DSS that provides informative guidance to signal the value of the business model.*

Second, the HI-DSS should guide the entrepreneur's adaption of the business model by providing concrete advice on how the elements of the business model should be adapted: i.e. suggestive guidance (Montazemi et al. 1996). Therefore, I formulate the following DR:

**DR2:** *Business model validation should be supported by a DSS that provides suggestive guidance to advice on concrete future actions.*

Third, the DSS should learn from the user and generate guidance on user demand as the task of business model validation is highly uncertain and dynamic and does not allow the offering of predefined guidance (dynamic guidance). Therefore, I formulate the following DR:

**DR3:** *Business model validation should be supported by a DSS that allows dynamic changes and learns from users' input.*

Fourth, users should be able to determine the guidance provided. In the context of business model validation, this means providing direct guidance through mentors: i.e. participative guidance. Both modes of decisional guidance (participative and dynamic) are particularly effective in improving decision quality, user learning, and decision performance in highly complex tasks such as business model validation (e.g. Parikh et al. 2001). Therefore, I formulate the following DR:

**DR4:** *Business model validation should be supported by a DSS that allows participation of users (i.e. mentors) in providing the guidance.*

Fifth, the DSS should provide additional knowledge to the entrepreneur to give her guidance on how to improve the business model: i.e. knowledge (e.g. Schneckenberg et al. 2017). Therefore, I formulate the following DR:

**DR5:** *Business model validation should be supported by a DSS that provides the user with predictive and prescriptive knowledge on the business model.*

Sixth, it is central that the user learns from the actions in the long term: i.e. learning (e.g. Alvarez and Barney 2007). Therefore, I formulate the following DR:

**DR6:** *Business model validation should be supported by a DSS that allows the user to learn from iterations.*

In previous studies, combining the above six dimensions of decisional guidance proved to be most suitable when trying to overcome limitations in individual decision-making (Mahoney et al. 2003; Montazemi et al. 1996).

Finally, the user needs to properly visualize the decisional guidance to be able to draw inferences from it; i.e. visualization. Therefore, combining different formats of presenting the results are needed, such as text-based and multimedia (Gregor and Benbasat 1999). The formats should match the characteristics of the task (i.e. business model design decisions) (Vessey 1991). Therefore, I formulate the following DR:

**DR7:** *Business model validation should be supported by a DSS that provides the user with visualization of the guidance.*

Implementing decisional guidance in highly uncertain contexts. To support decision-making, it is essential to provide high-quality guidance to the user (Gregor and Benbasat 1999; Silver 1991; Montazemi et al. 1996). Two recently popular approaches for providing high-quality guidance in decision support in uncertain settings are statistical methods and collective intelligence (e.g. Surowiecki 2005; Malone et al. 2009).

Computational methods have become particularly valuable due to progress in ML and machine intelligence to identify, extract, and process various forms of data from different sources to make predictions in the context of financing decisions (Yuan et al. 2016), financial return, and bankruptcy of firms (Olson et al. 2012). Statistical models are unbiased, free of social or affective contingence, able to consistently integrate empirical evidence and weigh them optimally, and not constrained by cognitive resource limitations (Blattberg and Hoch 1990). ML is a paradigm that enables a computer program (i.e. an algorithm) to learn from experience (i.e. data) and thus improves the program's performance (e.g. accuracy) in conducting a certain class of tasks (e.g. classification or regression). Consequently, machine intelligence can make statistical inferences based on patterns identified in previously seen cases and learning as the data input grows (Jordan and Mitchell 2015). In addition, such procedures allow the identification of complex patterns in business model configurations and the interrelation between single components and thus extend methods such as business model simulations and financial scenarios.

Although ML techniques might be generally suitable for predicting the consequences of certain business model design choices based on prior data distributions of easily quantifiable features (e.g. firm age, team size), they are often biased and fail in highly uncertain settings, when for instance data shifts over time and the data that was previously used to train the model is no longer representative or patterns emerge that were never seen before by the algorithm (Attenberg et al. 2015;

Dellermann et al. 2017b). Furthermore, they are not able to predict the soft and subjective evaluations of new ventures such as innovativeness of a value proposition, the vision or the fit of the team, or the overall consistency of a business model, which makes it impossible for machines to annotate such types of data (Cheng and Bernstein 2015; Petersen 2004).

Due to these limitations, machine intelligence systems require the augmentation of human intuition to successfully guide the design of business models (Attenberg et al. 2015; Kamar 2016). Human decision makers bring several benefits. Humans are still the gold standard for assessing data that cannot easily be annotated and trained for ML models such as creativity and innovativeness (Baer and McKool 2014; Cheng and Bernstein 2015). Humans are particularly good at providing subjective judgements of data that is difficult to measure objectively through statistical techniques (e.g. Einhorn 1974; Cheng and Bernstein 2015).

Additionally, human experts have highly organized domain knowledge that enables them to recognize and interpret very rare information. Such data might lead to specific outcomes that are difficult to predict a priori and would rather represent outliers in a statistical model (Blattberg and Hoch 1990). Using humans for augmenting machine intelligence proved to be valuable in many other settings (Cheng and Bernstein 2015; Kamar 2016). Pertaining to the context of this research, using human intuition proved to be a valuable strategy for anticipating start-up business model success at the early-stage (Huang and Pearce 2015).

While individual humans still have the cognitive limitations discussed in previous chapters, these can be minimized through the mechanism of collective intelligence (Mannes et al. 2012). This approach aggregates the judgments of a larger group of humans to reduce the noise and bias of individual evaluations (Klein and Garcia 2015; Blohm et al. 2016;

Leimeister et al. 2009). For making judgments in uncertain settings, the value of crowds over individual experts can be explained by two basic principles: error reduction and knowledge aggregation (Mannes et al. 2012). While an individual decision maker might be prone to biases and errors (such as individual entrepreneurs or mentors in my context), the principle of statistical aggregation minimizes such errors by combining several judgements (Armstrong 2001). Furthermore, aggregating the judgement of several individuals is informative as it aggregates heterogenous knowledge about a certain problem and allows the capture of a fuller understanding of a decision-making problem (Ebel et al. 2016a; Ebel et al. 2016b; Soukhoroukova et al. 2012; Keuschnigg and Ganser 2016).

Consequently, I argue that collective intelligence represents a proper way to augment ML systems by accessing more diverse domain knowledge, integrating it into an algorithm, and reducing the threat of biased interpretation. Due to these complementary capabilities, I decided to combine machine and collective intelligence for providing decisional guidance to entrepreneurs. I call such combined systems Hybrid Intelligence Decision Support Systems (HIDSS).

For this research project, I define HIDSS as a computerized decisional guidance to enhance the outcomes of an individual's decision-making activities by combining the complementary capabilities of human and machines to collectively achieve superior results and continuously improve by learning from each other. HI-DSS might be especially suitable to solve my research problem due to three reasons.

First, machines are better at processing analytical information and providing consistent results, especially when data is dispersed and unstructured (Einhorn 1972). In the context of business model validation, this becomes evident through the unstructured data regarding market demands, technological developments, etc.

Second, human decision-makers are particularly useful in interpreting and evaluating soft information as they are superior in making judgments about factors like creativity or imagining the future (which are required for start-up business models) or providing comprehensive guidance on which action to take (Colton and Wiggins 2012; McCormack and d'Inverno 2014).

Third, in highly uncertain and complex situations such as setting up a business model for a new venture, humans can use their intuition and gut feeling which augments statistical methods (Huang and Pearce 2015; Dellermann et al. 2018a). In this regard, collective intelligence is applied to capitalize on the benefits of humans while simultaneously minimize the drawbacks of individual decisionmakers such as bias or random errors (Mannes et al. 2015).

### 6.5.7. Design and Development (Phase 3)

**Principles of Form and Function**

To develop the HI-DSS, I translated the required types of decisional guidance (organized along the conceptually identified dimensions of decisional guidance; Morana et al. 2017) into DPs (principles of form and function) thereby combining mechanisms of ML and collective intelligence.

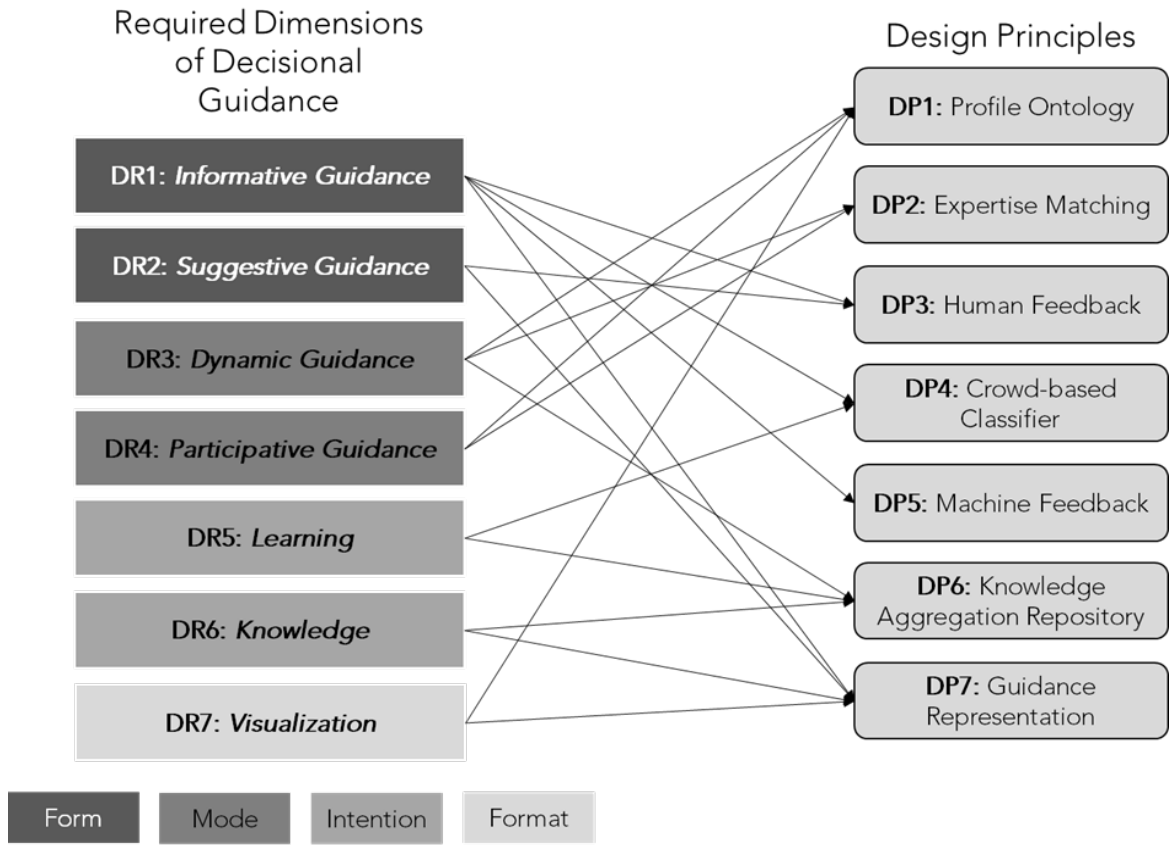

**Translation Process of DRs in DPs**

To apply HI-DSS, entrepreneurs must transfer their implicit assumptions to both human mentors as well as the ML algorithm to create a shared understanding. Business models are mental representations of an entrepreneur's individual beliefs that should be made explicit by transferring them into a digital object (Bailey et al. 2012; Carlile 2002). Approaches to transfer such knowledge into a common syntax are required to make the knowledge readable for both humans and algorithms (Nonaka and Krogh). Therefore, ontologies can be used to leverage knowledge sharing through a system of vocabularies. Such ontologies represent popular solutions in the context of business models and include descriptions of a business model's central dimensions, such as value proposition, value creation, and value capture mechanisms (Osterwalder et al. 2005). This allows the user to provide input in the DSS dynamically and participially (**DR3** and **DR4**). Previous work on human cognition has shown that the representation of knowledge in such an object (i.e. digital representation of the business model) should fit the corresponding task (i.e. judging the business model) to enhance the quality of the human guidance (John and Kundisch 2015). Since judging a business model is a complex task, a visual representation is most suitable as it facilitates cognitive procedures to maximize the decision quality (Speier and Morris 2003). This allows human judges to visualize the business model (**DR7**). To make the business model readable for the ML algorithm, the ontology requires also a machine-readable format that can be achieved through standardizing the representation of design choices (e.g. pattern) or natural language processing (John 2016). Therefore, I propose the following **DP**.

**DP1:** *Provide the HI-DSS with an ontology-based representation to transfer an entrepreneur's assumptions and create a shared understanding among the mentors, the machine, and the entrepreneur.*

Past literature shows that a judge who is qualified for providing decisional guidance on business model validation is also an expert in the respective context (Ozer 2009). Such appropriateness results in a higher ability to provide valuable feedback and enables the prediction of the potential future success of a business model even in highly dynamic contexts. Therefore, to be suitable as a judge and provide more accurate guidance, an individual human mentor should have two types of expertise: demand- and supply-side knowledge (Magnusson et al. 2016; Ebel et al. 2016). Demand-type knowledge is necessary to understand the market side of a business model (e.g. Customer, competitors, sales channels, value proposition), which indicates the desirability of a business model. Supple-type knowledge consists of knowledge on feasibility (e.g. resources, activities) and profitability (e.g. cost structure, revenue model) of a business model configuration (Osterwalder et al. 2005; Magnusson et al. 2016). Consequently, it is central for a HI-DSS to match certain business models with specific domain experts to ensure high human guidance quality (**DR3** and **DR4**). Therefore, I propose the following DP:

**DP2:** *Provide the HI-DSS with expertise matching through a recommender system in order that the entrepreneur obtains access to expertise.*

To provide guidance for entrepreneurs, humans need adequate feedback mechanisms to evaluate the developed assumptions (Blohm et al. 2016). From the perspective of behavioral decision-making, this feedback can be categorized as a judgment task in which a finite set of alternatives (i.e. business model dimensions) is evaluated by applying a defined set of criteria by which each alternative is individually assessed by using rating scales. Using multi-attribute rating scales for judging and thus providing informational guidance are most appropriate in this context (Riedl et al. 2013). The multi-criteria rating scales should cover the desirability, feasibility, and profitability of a business model by assessing dimensions which are strong predictors for the future success, such as the market, the business opportunity, the entrepreneurial team, and the resources (Song et al. 2008) (**DR1**). Additionally, the human mentor should be able to provide qualitative feedback to guide the entrepreneur's future action and point towards possible directions. This allows the human to provide additional suggestive guidance (Silver 1991) (**DR2**). Therefore, I propose the following DP:

**DP3:** *Provide the HI-DSS with qualitative and quantitative feedback mechanisms to enable the humans to provide adequate feedback.*

In addition to that, the input of the human can also be used to train a ML classifier to assess the probability of achieving milestones within the version of the presented business model. This procedure allows consistent processing and weighting of the collective human judgement, which is required to achieve high quality evaluation through collective intelligence (Keuschnigg and Ganser 2016) (**DR1** and **DR4**). Therefore, I propose the following DP:

**DP4:** *Provide the HI-DSS with a crowd-based classifier to predict the outcomes of business model design choices based on human assessment.*

Every created business model consists of different design choices (e.g. types of value proposition, revenue models, etc.). This allows supervised ML approaches to provide machine feedback concerning the probability of success regarding a certain business model element (e.g. Jordan and Mitchell 2015; John 2016). Supervised algorithms allow a machine to learn from training data (i.e. the user's input) to predict which configuration leads to a favourable outcome (i.e. achieving a milestone) (**DR1**). For this purpose, the user must provide information, so-called labelling, when a business model achieves a milestone (e.g. funding). This procedure allows training of the machine's ability to evaluate new business model configurations to predict the probability of success of a certain business model version and thus validate (or reject) an entrepreneur's assumptions. Therefore, I propose the following DP:

**DP5:** *Provide the HI-DSS with machine feedback capability to predict the outcomes of business model design choices based on statistical assessment.*

Business model validation is an iterative process consisting of validating the existing model, adapting it, and then revalidating it. Therefore, HI-DSS should aggregate the results of each validation round to transient domain knowledge to show how the human and the machine feedback changes an entrepreneur's assumptions and how such changes lead to a certain outcome (e.g. John 2016). The accumulation of such knowledge can trigger cognitive processes that restructure the entrepreneur's understanding of the domain (Sengupta and Abdel-Hamid 1993). My proposed HI-DSS needs to accumulate in a repository the knowledge created during use, to continuously improve the guidance quality through ML, and to learn not only from the iterations of the individual validation process but also from other users of the system (e.g. other entrepreneurs) (Jordan and Mitchell 2015). The knowledge repository can then store general patterns of how changes in a business model are evaluated by humans or the ML algorithm, and how they lead to achieving certain milestones (e.g. receiving funding) (**DR3, DR5** and **DR6**). Therefore, I propose the following DP:

**DP6:** *Provide the HI-DSS with a knowledge aggregation repository to allow it to learn from the process.*

Finally, knowledge in the form of additional information and the learning of the entrepreneur must be achieved through a representation of the guidance in a dashboard (Benbasat et al. 1986). Following previous work on business intelligence and decision support visualizations, I argue that the best quality of guidance is achieved when the representation fits the task (Vessey and Galletta 1991; Vessey 1991). To achieve high decision quality for the user, a cognitive link between the highly complex task (i.e. making business model design decisions) and the guidance should be made by providing visual guidance representation (**DR1, DR2, DR6,** and **DR7**). Moreover, reducing the user's effort for understanding and retrieving the guidance should be achieved by structuring the guidance along the business model dimensions (Baker et al. 2009). Therefore, I propose the following DP:

**DP7:** *Provide the HI-DSS with a visual guidance representation in order that the entrepreneur obtains access to informative and suggestive guidance.*

## 6.5.8. Implementation of the HI-DSS

When implementing my HI-DSS, I mapped the identified DPs to concrete design features that represent specific artefact capabilities to address each of the DPs. To implement my DPs into a prototype version of the HI-DSS artefact, I created a web-application.

The prototype of the artefact consists of a graphic user interface (GUI) that allows the input and visualization of the entrepreneur's business model. For this purpose, I developed a web application in Angular (https://angular.io/). A business model was represented in a standardized and dynamically adaptable format, allowing the entrepreneur to make categorical choices for each element along the value proposition, value delivery, value creation, and value capture dimensions of the business model (Osterwalder and Pigneur 2013) (**DP1**).

The expertise matching is achieved through a simple tagging of expertise (i.e. market, technology, or finance). These tags are then matched with a SQL table that consists of a list of categorized mentors (**DP2**).

The feedback mechanisms that allow human judgement are implemented by using the same tool. I implemented Likert rating scales (1 to 10) covering the desirability, implementability, scalability, and profitability of a business model that are commonly applied in practice. Moreover, I provided a textbox for providing concrete qualitative guidance on how to improve the business model. This guidance was structured in terms of the value proposition, value delivery, value creation, and value capture mechanisms of the business model (**DP3**).

To gather initial data, I collected publicly available information on start-up business models and their respective success to train the ML algorithm. The ML part of the prototype was developed based on the open source ML framework TensorFlow (www.tensorflow.org) in the programming language Python (www.python.org). For the crowd-based classifier I utilized a Classification and Regression Tree (CART) as it provides both good performance and interpretability of results through replication of human decision-making styles (Liaw and Wiener 2002) (**DP4**).

I applied the same learning algorithm for analysing the complex interactions between business model components. Therefore, the success probability of a certain business model is calculated (**DP5**).

All the results (i.e. business model components, profile data of mentors, and human judgement) are stored in a relational PostgreSQL (www.postgresql.org) database on an Ubuntu SSD server (**DP6**).

The final visualization of results (informative and suggestive guidance) is provided through the dashboard implemented in Angular (https://angular.io). This represents aggregated results along the dimensions of desirability, feasibility, and profitability as well as the predicted probability of success along the outcome dimensions of survival and Series A funding, which represent commonly accepted proxies for successfully early-stage business models (**DP7**).

.

## 6.5.9. Demonstration of the Artefact (Phase 4)

The first evaluation of my HI-DSS serves as lightweight and formative ex-ante intervention to ensure that the IT artefact is designed as an effective instrument for solving the underlying research problem (Venable et al. 2016).

For this purpose, I decided to make use of exploratory focus groups to refine the artefact design based on feedback from participants. When conducting the focus groups, I followed the process proposed by Tremblay et al. (2010). Within a total of eight focus groups, my DPs were demonstrated, validated, and refined by entrepreneurs and mentors, as well as by developers to validate the technical feasibility of the DPs in a naturalistic setting.

During this ex-ante evaluation, I focused on the clarity, completeness, internal consistency, and applicability to solve the practical problem (Sonnenberg and Vom Brocke 2012). The tentative version of the DPs was then adapted before being instantiated into my prototype artefact. The required changes were especially related to the expertise matching and the business model ontology.

The participants suggested a switch from well-known business model visualizations (i.e. the business model canvas) to a novel form of start-up profiles because mentors require more in-depth information on a certain start-up. Especially for an IT based and time-location asynchronous solution, such an approach is mandatory to ensure high-quality guidance.

Moreover, the evaluation revealed the need for expert matching on a fine granular level. Apart from matching industry experts with a start-up in a certain domain (e.g. a FinTech start-up with a banking industry expert), it is crucial to get feedback from a certain type of expert on each dimension of the business model (e.g. a finance expert for evaluating the value-capture mechanisms).

Finally, the participants in the workshop requested the possibility to provide in-depth qualitative feedback to not only point towards suggestions of improvement such as changing the proposed revenue model, but on how to proceed and achieve this goal.

## 6.5.10. Evaluation (Phase 5)

For the ex-post evaluation of my instantiated DPs into a concrete IT artefact, I applied a qualitative evaluation method to test proof of applicability in the real-world context and to assess the feasibility, effectiveness, efficiency, and reliability against the real-world phenomenon of supporting business model design decisions (Sonnenberg and Vom Brocke 2012; Venable et al. 2016).

I conducted confirmatory focus-group workshops with decision makers and potential users of my HI-DSS in practice. I chose this evaluation approach as a confirmatory method for several reasons. First, the flexibility of the method enabled me to adapt the procedure if necessary. Second, this approach allowed me to directly interact with the potential users of the system, which ensured that the artefact was understood unambiguously. Finally, the focus-group method provided huge amounts of rich data, providing a deeper understanding of the effectiveness and efficiency of the artefact to solve a real-world problem in an actual business environment (Hevner and Chatterjee 2010; Tremblay et al. 2010).

For conducting a total of eight focus-group workshops, I recruited 24 participants from business incubators and accelerators as well as independent start-up mentors. Four of the focus groups consisted of participants from business incubators, two from accelerators and three with independent mentors. I presented the HI-DSS via a click-through approach and explained each of the DPs in detail. Then the workshop was guided by the effectiveness of solving the real-world problems and the identified research gap. As the results of my evaluation show, the HI-DSS overcomes the limitations of previous solutions by combining the analytical processing of interaction between complex business model patterns and the input provided by human intuition.

The focus groups reveal that decisional guidance in general helps entrepreneurs to deal with the highly complex and uncertain task of making business model design decisions and overcoming their individual limitations. Moreover, an IT-based solution that aggregates the collective judgement of individual mentors allows for reducing subjectivity while aggregating knowledge that can be stored through ML.

Finally, the hybrid nature of my proposed design allows it to deal with soft factors and extreme uncertainty by having human intuition in the loop. In addition, using ML to identify the complex interaction between different business model elements allows it to deal with the complexity of start-up business models. I then continued the summative ex-post evaluation by assessing each DP in detail.

All the proposed DPs and their instantiations were perceived as useful and effective in solving the problems that are faced during the task execution of decision support for business model validation. The participants argued that the HI-DSS is particularly suitable to improving decision quality and efficiency of the entrepreneurs (**DR6: knowledge**) and helping them to learn (**DR5: learning**).

The digital nature of the tool was perceived as saving time and resources and allowed mentors to provide guidance independent of time and location, thus providing high-quality guidance (**DR1: informative** and **DR2: suggestive**).

The executives of business incubators and accelerators praised the possibility of accumulating knowledge on the business model design of start-ups and the implicit sharing of such knowledge through the ML approach (**DR3: dynamic**). The experts agree that this might increase the survival rate of new ventures at an early stage. The participants liked the possibility to use external mentors from different industries.

While the public-funded incubators evaluated the applicability in this context as very high, profit-oriented accelerator mentioned that compensation methods for external mentors beyond intrinsic rewards should be defined. Although altruistic mentoring works well in practice (e.g. business plan competitions and feedback), reward mechanism should be considered to apply my artefact in practice.

Furthermore, the experts see great potential in improving decisional guidance through ML. They indicate that due to the human component of the HIDSS, acceptance of the guidance might be higher among entrepreneurs than with only statistical modelling and simulations (**DR4: participative guidance**).

Finally, the dashboard for visualizing the decisional guidance through graphs and feedback text was perceived as favourable to make the decisional guidance easily accessible for entrepreneurs (**DR7: visualization**).

However, the results of the evaluation also reveal two criticisms that should be resolved before use in a real-life setting. First, the participants highlighted the need for creating trust in AI-based DSSs. While providing highly accurate decisional guidance is crucial, there is a trade-off between accuracy and transparency, which was highlighted by most of the participants. Future research could examine this issue when applying HI-DSS in business contexts.

Second, the participants indicated that such IT-based guidance might be perceived as missing the in-depth support of personal mentors. Although the value of the HI-DSS was obvious for all participants, they argued that for communicating with the users of such systems (i.e. entrepreneurs and mentors), the human should still be the focus, while augmented by machine intelligence.

## 6.5.11. Conclusion

Determining business models for start-ups is a highly challenging and uncertain task for entrepreneurs and requires various decisions regarding the design of the business model. Due to limitations of individual human decision-makers, this process is frequently tainted by poor decision-making, leading to substantive consequences and sometimes even failure of the new venture. As most DSS for business model validation rely on simulations or modelling rather than human intuition, there is an obvious gap in literature on such systems.

Using DSR project methodology, I analysed problems in making decisions about business model design in uncertain environments. I then developed and refined DPs for a HI-DSS that combines the specific benefits of machine and collective human intelligence to steer entrepreneurial decision-making by providing decisional guidance. I then instantiated my DPs into a prototype artefact and evaluated them using several focus-group workshops with domain experts.

### Contributions

My study makes several contributions, both theoretical and practical. First, my research provides prescriptive knowledge that may serve as a blueprint to develop similar DSSs for business model validation (Gregor and Jones 2007). The findings of this paper reveal prescriptive knowledge about form and function (i.e. DPs) as well as principles of implementation (i.e. my proposed instantiation).

Due to utilizing justificatory knowledge from the body of knowledge on decisional guidance and the justification of the research gap in both theory and practice, I provide meaningful interventions in the form of DPs to solve a real-world problem and contribute to the discussion of decision support in business model validation.

Second, my results indicate a possible application of collective intelligence in more complex and knowledge-intensive tasks. While previous work (e.g. Blohm et al. 2016; Klein and Garcia 2015) utilized the wisdom of the crowd in rather basic decision support settings such as filtering novel product ideas without considering explicit expertise requirements, my findings indicate the potential of applying collective intelligence in uncertain decision tasks. Addressing the concrete expertise requirements of humans, decisional guidance is based on the heterogenous domain knowledge of experts and reduces misleading biases and heuristics.

Third, I propose a novel approach to support human decision-making by combining machine and collective intelligence into a hybrid intelligence system. My results show that this form of decisional guidance is particularly relevant in situations of extreme uncertainty where a combination of formal analysis through ML techniques and human intuition through collective intelligence is most valuable. Thus, my research contributes to recent work on combined applications in different domains (Brynjolfsson et al. 2016; Nagar and Malone 2011; Nagar et al. 2016).

Fourth, I contribute to research on decision support for business model validation by augmenting formal analysis of data to iterative social interaction with stakeholders (e.g. Gordijn et al. 2001; Haaker et al. 2017; Daas et al. 2013; Euchner and Ganguly 2014). This research takes human guidance and judgement into account to help decision makers to design business models. Moreover, the findings start a novel discussion in the field of research on DSS: how can such systems be designed for situations of extreme uncertainty where no objective truth exists.

Finally, my proposed prototype artefact offers an actual solution for helping service providers such as business incubators and accelerators to extend their service offering beyond solely offline mentoring to a digital solution and thus provides a first step towards a practical solution in this context. Based on the results of this paper, further research is focusing on the provision of Hybrid Intelligence services in real-world applications.

## Limitations and Future Research

Despite its various contributions to theory and practice, my work is not without limitations. First, I focused my research on the context of business incubator and accelerators to provide a DSS that helps them to provide decisional guidance to entrepreneurs. This setting implies that access to a network of mentors is already available and that advise is mainly offered with altruistic motives. Such DSS might require adaption for attracting experts to participate and provide advise via the system. Therefore, further research might explore the motives of such mentors and how DSS might be extended through activation supporting components (e.g. Leimeister et al. 2009).

Second, I chose a qualitative evaluation procedure to assess the applicability and effectiveness of a HI-DSS in providing decision support for the business model design process. Although I intended to evaluate in a naturalistic setting with potential users and domain experts, my evaluation procedures were not capable of testing the actual quality of guidance provided by the HI-DSS or its value during long-term use. Further research might therefore develop hybrid prediction algorithms to evaluate the performance (e.g. accuracy) of HI-DSS, particularly compared to other methods. Moreover, a longitudinal study of the use of a HI-DSS in a real-world context might be useful for determining the value of such a system.

Finally, my study is limited to the field of business model validation for start-ups. However, it starts a discussion on a valuable novel form of DSS that combines humans and machines, and as such, encourages exploration of HI-DSS applicability in other settings of uncertainty such as medicine, job applications, and innovation contexts.

# Chapter V

## Contributions and Further Research

# 7. Contributions and Further Research

## 7.1. Summary of Findings

Within this thesis, I examined the entrepreneurial decision-making context (Chapter II) and then developed the design paradigms of crowd-based decisional guidance (Chapter III) as well as hybrid intelligence (Chapter IV) and the relevant design principles for building DSS that are based on those paradigms.

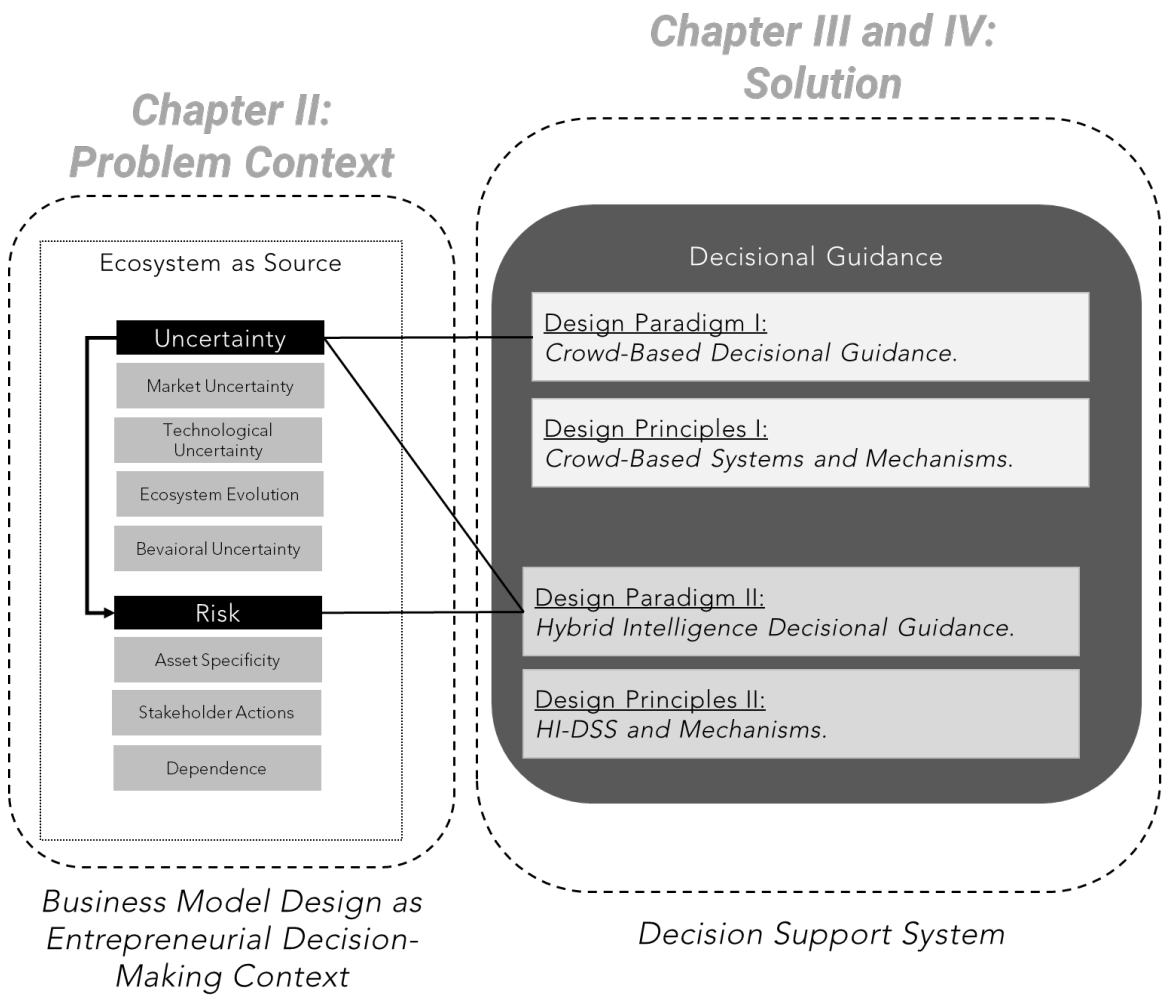

**Summary of Key Findings**

First, **Chapter II** focuses on exploring the entrepreneurial decision-making context and identifying the ecosystem as source of both risk and uncertainty in designing new business models. I identified the generativity of ecosystem evolution (Section 4.1), market uncertainty, technological uncertainty, and the (potentially opportunistic) behaviour of other stakeholders (Section 4.2) as the most important dimensions of uncertainty and explore how such are created through the ecosystem. I then identified the specificity of assists, the dependence on actors of the ecosystem as well as the actions of such stakeholder as sources of risk (Section 4.3). This combination of uncertainty and risk, thus, define the decision-making context and influence the success probability of an entrepreneurial venture and can be best managed through an integration of stakeholders combined with analytical risk assessment methods (Section 4.3).

Second, **Chapter III** explores a novel design paradigm for decisional guidance in entrepreneurship: collective intelligence and the mechanism of crowdsourcing that is particularly valuable for dealing with uncertainty. By highlighting limitations of previous approaches of decisional guidance such as feedback from mentors or peers, I subsequently develop crowd-based decisional guidance as novel design paradigm (Section 6.1) and provide requirements for adapting existing crowdsourcing endeavours (Section 6.2). Based on these findings, I propose design principles for CBMV systems (Section 6.3) that aim at guiding entrepreneurial decision-making and provide deep insights into the design of related mechanisms (Section 6.4). The findings of this Chapter provide a way for integration human intuition and social interaction in scalable IT tools that can guide entrepreneurial decision-making. Those kinds of guidance are most suitable for dealing with uncertainty.

Third, in **Chapter IV** my dissertation aims at further developing decisional guidance for entrepreneurial decision-making and provides a novel and innovative form for creating such guidance: hybrid intelligence (Section 6.1). This design paradigm for decisional guidance proposes the combination of the complementary strengths of human (i.e. intuitive thinking) and artificial intelligence (i.e. analytical pattern identification) as valuable approach for dealing with both uncertainty and risk. Based on this rational of hybrid intelligence as new design paradigm, I derive design knowledge for developing such systems (Section 6.2) and propose novel methods and mechanisms to create guidance for complex decision-making problems under uncertainty and risk (Section 6.3 and 6.4). Finally, this thesis concludes with developing design principles for a HI-DSS as an innovative artefact for a new class of complex decision problems under risk and uncertainty.

## 7.2. Theoretical Contributions

### 7.2.1. The Mechanisms of the Ecosystem in Creating Risk and Uncertainty in Entrepreneurial Decisions

The first theoretical contribution of my thesis is related to the mechanisms of the ecosystem in creating uncertainty and risk and thus explaining both ecosystem dynamics (Um et al. 2013); Ravasz and Barabási 2003) and innovation evolution (Audretsch 1995; Nelson 2009) as well as the myopic outcomes of entrepreneurial efforts (Alvarez and Barney 2007; Alvarez et al. 2013; Alvarez et al. 2014; Wood and McKinley 2010) in Chapter 5.

By examining this role of the ecosystem in this vein, I contribute to previous work by providing a deeper understanding of the mechanisms that create myopic innovation outcomes and limit the predictability of entrepreneurial success. I further contribute to the body of knowledge by investigating how risk and uncertainty are related to innovation success. Consequently, those findings provide a better understanding of the context of entrepreneurial decision-making, the reason many ventures fail, and an explanation of why predicting the success of entrepreneurial efforts is so difficult.

The following table provides an overview of the risks and uncertainties that are created by the ecosystem of an entrepreneurial opportunities and that were identified in during the studies in this thesis. Furthermore, it provides a definition for each risk/uncertainty and proposes related literature.

| Type | | Definition |
|---|---|---|
| **Uncer-tainty** | **Market Uncertainty** | Uncertainty regarding market demand, customer dynamics, and product-market fit (e.g. the existence of a market for a certain product). |
| | **Technological Uncertainty** | Uncertainty regarding technological process and volatility (e.g. the progress of technological capabilities or the emergence of new standards). |
| | **Ecosystem Evolution** | Uncertainty regarding ecosystem dynamics (e.g. the co-evolution of technical and social actors based on external influences). |
| | **Behavioral Uncertainty** | Uncertainty regarding the behavior of partners, investors, or other stakeholders (e.g. unpredictable opportunistic behavior). |
| **Risk** | **Asset Specificity** | Risk regarding the amount of investments to be made for a specific ecosystem membership (e.g. technological platform standards). |
| | **Stakeholder Actions** | Risk regarding actions of stakeholders such as competitors (e.g. predictable changes of pricining behavior of competitors). |
| | **Dependence** | Risk regarding the dependence on a third-party (e.g. dependence on investment due to capital intensive R&D). |

*Summary of Ecosystem Uncertainty and Risk for Entrepreneurs*

## 7.2.2. The Cognitive Rational of Collective Intelligence for Guidance in Entrepreneurial Decisions

The second theoretical contribution is related to examining the cognitive rational for applying collective intelligence for guiding entrepreneurial decisions in Chapter 5.

My contribution is noteworthy for several reasons. First, I contribute to the discourse in entrepreneurship how opportunities emerge from the interactions between entrepreneurs and ecosystem (e.g. Alvarez and Barney, 2007; Alvarez et al., 2013). I also contribute to the cognitive perspective of opportunity creation and enactment (e.g. Gregoire et al., 2011) by highlighting the role of leveraging external heterogeneous social resources in objectifying and enacting an opportunity. Building on previous work on the role of social resources in this process (e.g. Tocher et al. 2015), I argue that crowdsourcing facilitates opportunity objectification by providing entrepreneurs with social resources to engage in a sense-making process. I show that such heterogeneous feedback provides several benefits compared to the knowledge of peers and facilitates the iterative development of an opportunity. After the opportunity is objectified, I argue that crowdsourcing supports the opportunity enactment by signalling the market viability of an opportunity, therefore reducing stakeholders' opportunity-related uncertainty, which arises in the consensus-building stage

| Entrepreneurial Actions | Rational of Collective Intelligence | Overcoming Limitations of Previous Approaches |
|---|---|---|
| **Social Interaction** | ▪ Access to anonymous and heterogeneous social resources<br>▪ Access to heterogenous knowledge and error reduction | ▪ Limited social resources<br>▪ Homogeneity of social resources<br>▪ Social influence<br>▪ Limited expertise/market knowledge |
| **Uncertainty** | ▪ Evaluating the opportunity idea<br>▪ Reducing uncertainty about the value of the idea by signalling reaction of the market | ▪ Limited social resources<br>▪ Homogeneity of social resources |
| **Iterative Development** | ▪ Providing feedback<br>▪ Co-creation of opportunity<br>▪ Enabling iterative development | ▪ Limited social resources<br>▪ Homogeneity of social resources<br>▪ Limited knowledge |
| **Learning** | ▪ Integration of feedback on value of the idea<br>▪ Learning about stakeholders' perception Integration of novel market knowledge | ▪ Limited social resources<br>▪ Limited market knowledge<br>▪ Limited expertise |

*Rational for Collective Intelligence in Guiding Entrepreneurial Decisions*

Finally, I posit that crowdsourcing facilitates extended access to resources such as human capital or funding to fully enact an opportunity. I, therefore, provide a theoretical rational for the value of collective intelligence in the cognitive processes of entrepreneurial agents. For this purpose, I show how crowdsourcing may overcome the cognitive constrains and bounds of previous approaches, such as interacting with peers to open the boundaries of entrepreneurs' existing social networks or integrating demand-side knowledge (e.g. Nambisan and Zahra, 2016) into the creation of entrepreneurial opportunities, and provide applications during different stages of the creation process.

Second, I introduce the topic of crowdsourcing for opportunity creation as a promising field for further research in the field of digital entrepreneurship (e.g. Nambisan 2016) and propose a research agenda that may guide future efforts. I particularly argue for interdisciplinary research that might include the fields of strategy, information system, as well as cognitive entrepreneurship scholars and suggest design-oriented research (e.g. Hevner et al, 2004) on crowdsourcing for opportunity creation. Such design-oriented research might especially interesting to ensure the practical relevance of the discourse.

### 7.2.3. The Requirements of Crowdsourcing as Guidance Design Paradigm for Entrepreneurial Decisions

My thesis contributes to research on crowdsourcing for iteratively developing an idea over time in general and entrepreneurial decisions in specific. Therefore, my thesis offers several contributions to the body of knowledge on crowdsourcing in Chapter 5.

First, as in previous studies on crowdsourcing, the crowd represents a source of creative ideas for problem solving that can be objectively discovered through distant search (e.g. Leimeister et al. 2009). Thus, linear, and one-directional social interactions with the crowd constitute an accelerator for recognizing ideas than collaboratively co-creating innovative value propositions. Consequently, the crowd is incentivized for posting new ideas rather than refining an existing one (Majchrzak and Malhotra 2013).

Second, I show that crowdsourcing in the context of entrepreneurial opportunity creation requires multi-directional interactions with the crowd. From a constructivist's perspective, this is crucial to foster feedback-based idea evolution (Alvarez et al. 2013). Apart from an open call to the crowd, it requires further and intensive exchange between the initiator and the crowd. The crowd is therefore not the source of an initial idea but provides feedback on the correctness of an entrepreneur's assumptions and refines an idea. The initiation of innovation in this process, however, is to the entrepreneur, who starts the interaction with the crowd by showing her beliefs and ideas about an opportunity (Alvarez and Barney 2007). This is a central limitation of previous IS research on crowdsourcing architectures that led to lots of failures in creating solutions that could be implemented by sponsoring firms (Majchrzak and Malhotra 2013).

Third, I highlight the difference between traditional crowdsourcing efforts to foster innovation and the context of entrepreneurial opportunity creation is the level of task complexity. Contrary to previous research that focuses on using the crowd on the fuzzy front end of innovation (e.g. Poetz and Schreier 2012), the support of the opportunity creation represents a more complex task. The development of an opportunity goes far beyond the creation of early-stage ideas or product innovation as it includes the complete process including an initial idea of the entrepreneur, prototypes, and finally the development of a business model and an entire start-up (Ojala 2016). This contrasts with previous IS research that has focused on participation architectures and platforms for modular and closed problems solving tasks and leveraging the crowd for suggesting ideas while leaving the subsequent steps in the innovation process inside the boundaries of the sponsoring firm (Leimeister et al. 2009).

Fourth, I show that identifying a suitable crowd that represents an entrepreneur's potential stakeholders (e.g. investors, customers) is different from crowdsourcing in existing innovation communities that foster the discovery of novel ideas among existing users (e.g. Poetz and Schreier 2012). In this context, the selection of crowd members should balance heterogeneity and expertise in the entrepreneur's technological and industrial domain. However, required application contexts and markets are frequently not known a priori but emerging (Alvarez and Barney 2007). Therefore, the requirements for crowd members' supply- and demand-side knowledge might also change over time and recruiting the crowd from a firm-specific community might be misleading. This contrasts with the widespread principles of an open call.

Finally, I propose that opportunity creation is an evolutionary and iterative process to develop an initial idea into a new venture. In the context of crowdsourcing for innovation, however, participation architectures and platforms focus on the contribution of creative ideas while they provide only limited support for the evolution of an idea or the generative co-creation to further develop such ideas into novel value propositions and business models (Majchrzak and Malhotra 2013). In general, there is frequently minimal collaboration among the innovating firm and the crowd. Therefore, the current architectures of crowdsourcing platforms for innovation emphasize the generation of novel ideas over the evolution of opportunities that are suggested by either one member of the community or the innovating firm (Madsen et al. 2012). Such participation architectures, however, are required to integrate the crowd to provide feedback and support the opportunity creation of an entrepreneur and point towards directions for developing IS research on crowdsourcing and online communities.

### 7.2.4. Hybrid Intelligence as Guidance Design Paradigm for Entrepreneurial Decisions

The fourth contribution of my research is related to conceptualizing hybrid intelligence as a novel paradigm for making predictions under uncertainty and risk and deal with complex and dynamic decision-making problems in Chapter 6.

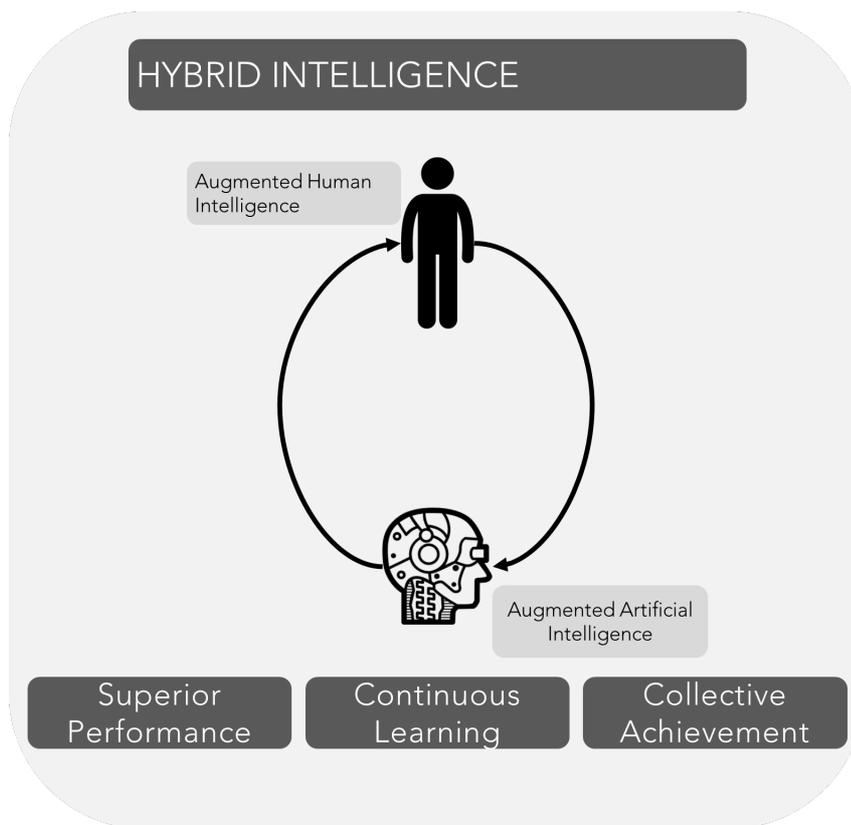

***Conceptualization of Hybrid Intelligence***

In my thesis, I conceptualize hybrid intelligence as *[...] the ability to accomplish complex goals by combining human and artificial intelligence to collectively achieve superior results than each of them could have done in separation and continuously improve by learning from each other [...]"* (Dellermann et al. 2019:3). Moreover, I highlight the three constituting dimensions of this concept: the collective achievement of a goal with human and artificial agents, the superior performance on a socio-technical system level that supersedes the outcome of each single entity, and the continuous learning of the machine and the human component of the system.

Moreover, I conceptually integrate this new design paradigm into existing related work on intelligence in general, human intelligence (Sternberg and Sternberg 1985), collective intelligence (Woolley et al. 2010), and AI (Goertzel and Pennachin 2007) by showing how the complementary benefits of different forms of intelligence can be combined to reduce the limitations of each.

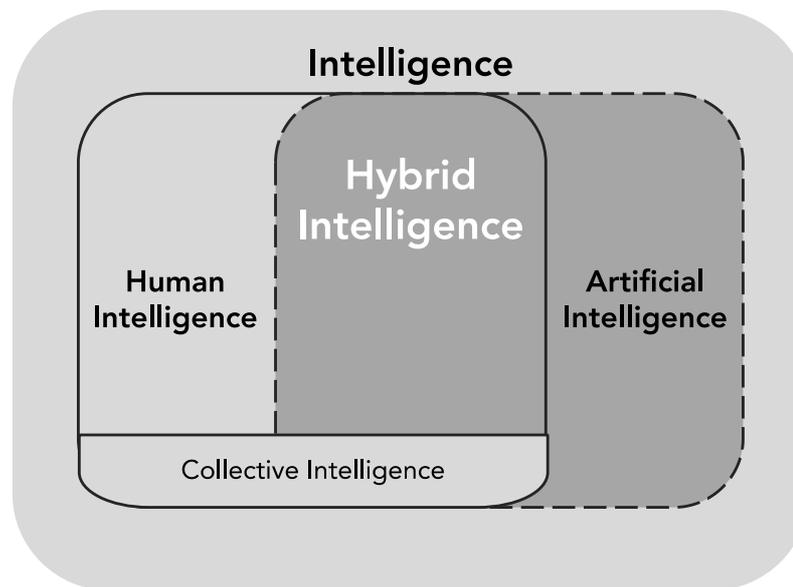

***Conceptual Integration of Hybrid Intelligence***

## 7.2.5. Design Principles for HI-DSSs

Finally, my research contributes to the current body of knowledge on DSS and decisional guidance (Arnolda et al. 2004; Meyer et al. 2014; Remus and Kottemann 1986). My findings imply that the complementary capabilities of formal analysis and pattern recognition provided through ML combined with human intuition provided through collective intelligence is a valuable solution to the extremely uncertain context of iterative business model validation in early-stage start-ups. HI-DSS enables mentors to provide the required decisional guidance to support entrepreneurs in making their decisions. For this purpose, I provide several contributions. Figures 57 and 58 provide an overview of the proposed DPs as knowledge contribution to existing research.

| Design Principles | |
|---|---|
| **DP1** | *Provide the HI-DSS with an ontology-based representation to transfer an entrepreneur's assumptions and create a shared understanding among the mentors, the machine, and the entrepreneur.* |
| **DP2** | *Provide the HI-DSS with expertise matching through a recommender system in order that the entrepreneur obtains access to expertise.* |
| **DP3** | *Provide the HI-DSS with qualitative and quantitative feedback mechanisms to enable the humans to provide adequate feedback.* |
| **DP4** | *Provide the HI-DSS with a crowd-based classifier to predict the outcomes of business model design choices based on human assessment.* |
| **DP5** | *Provide the HI-DSS with machine feedback capability to predict the outcomes of business model design choices based on statistical assessment.* |
| **DP6** | *Provide the HI-DSS with a knowledge aggregation repository to allow it to learn from the process.* |
| **DP7** | *Provide the HI-DSS with a visual guidance representation in order that the entrepreneur obtains access to informative and suggestive guidance.* |

*Design Principles for a HI-DSS*

First, my proposed DPs capture complex interaction between business model design decisions and the dynamic nature of such choices, and thus overcome limitations of traditional analytical methods such as modelling and simulation (e.g. Haaker et al. 2017; Euchner and Ganguly 2014). My findings, thereby, also provide a novel and innovative approach in line with previous research on dynamics and complex interactions (e.g. Moellers et al. 2017).

Second, the HI-DSS augments traditional analytical methods to human intuition. By leveraging collective intelligence rather than individual decision makers, my approach prevents the limitations of individual mentors. Consequently, the HI-DSS benefits from the heterogenous knowledge of several experts and aggregates the evaluations of a larger group for reducing the noise and bias of individual judgements. This procedure is particularly valuable in the uncertain and complex context of supporting business model design decisions, providing not only informative guidance in form of business model evaluations but also suggestive guidance that points entrepreneurs towards direct interventions to improve the business model.

Third, the HI-DSS stores the created knowledge in a knowledge repository. In the long run, this allows both entrepreneurs and mentors to learn from the experience of others. During its use, the system learns what business model design decisions are evaluated positively and negatively by humans and which decisions lead to a specific outcome in a certain context. This may allow the full automation of such decisional guidance in the future.

Fourth, the digital nature of the HI-DSS provides a way to digitize human mentoring. Such IT systems can iteratively validate a business model as well as provide asynchrony and location-independent feedback for resource-efficient mentoring.

Finally, my results point towards a new class of DSS that might be particularly valuable in highly uncertain contexts. With increasing uncertainty, the relative advantages of statistical methods in providing decisional guidance decrease and the value of human intuition increases. As the combinatory nature of formal analysis and intuition during predictions in extremely uncertain contexts is commonly accepted (e.g. Huang and Pearce 2015), such HI-DSS can provide high-quality guidance that might also work in different settings such as innovation or medicine. My proposed DPs provide a first step in this direction.

## 7.3. Practical Contributions

In addition to my contributions to the current body of knowledge, my thesis offers several contributions to managerial and entrepreneurial practice.

### 7.3.1. Design Knowledge on Digital Business Models

The first contribution to practice is design knowledge on digital business models. So far little is known about a managerial rational on domain specific design knowledge on digital business models and how to make certain decisions on those. My findings reveal the most important configurations of design choices and how they should be made to succeed.

Moreover, I propose a standardized conceptual representation of business models that can be used to develop data-models for further business model tooling and offers the potential for standardization of entrepreneurial start-ups. This representation can be for instance used as template for business plans or profiles in web applications that aim at communicating the entrepreneurial venture to stakeholders (e.g. www.crowdserv.eu).

Finally, my taxonomy of business model design choices can be used for mentoring, consultancy, or guiding entrepreneurial business model design. Therefore, I provide a canvas model with specific questions that entrepreneurs need to answer and related design choices that they can make. Moreover, I put this into a processual model that guides entrepreneurs in designing digital business models and validating their design choices along the way. I call this approach **d³**, which stands for *"digital disruption design"* and developed a web tool box that can be used for both workshops and IT-supported business model design.

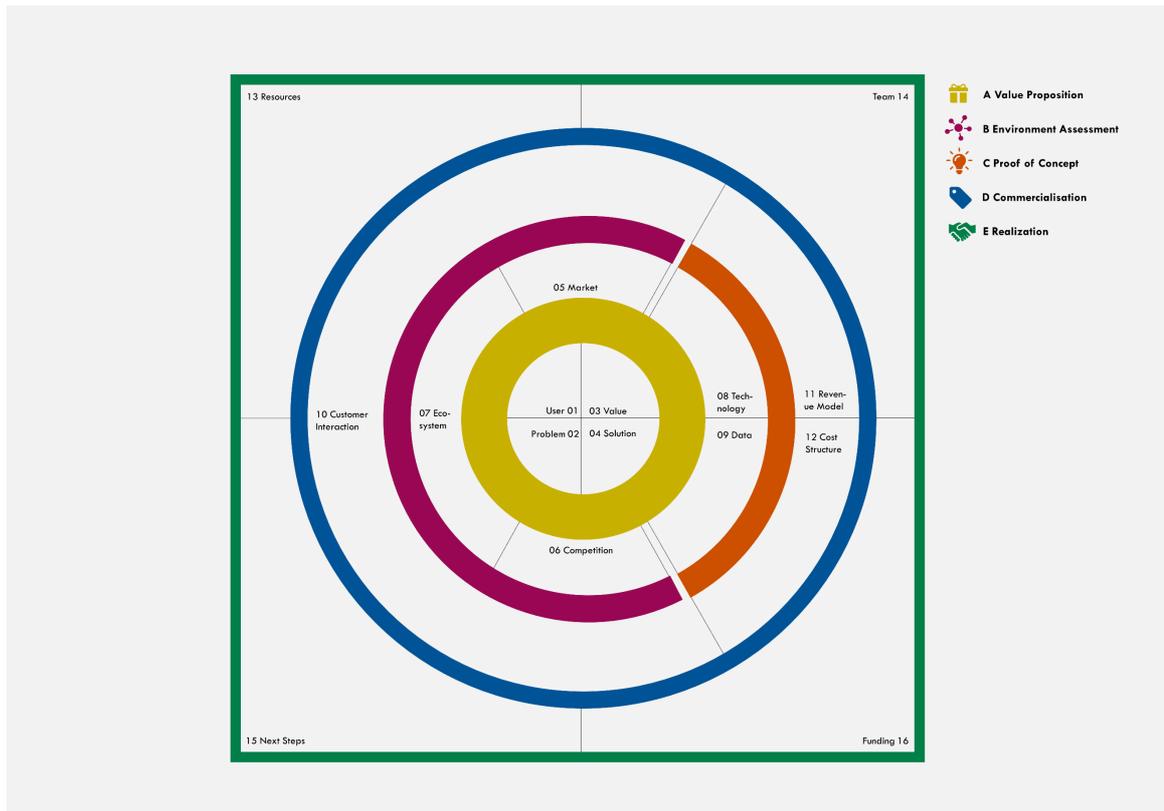

*d3 (Digital Disruption Design) Canvas Model*

### 7.3.2. Design Knowledge on Crowd-based Services

The second practical contribution of my dissertation is related to offering crowd-based services for guiding entrepreneurs. For this context, I focused mostly on how crowd-based services can be used to digitize and improve the current service portfolio of business incubators and accelerators. Within the project CrowdServ (www.crowdserv.de), me and my colleagues developed several concepts for new crowd-based services such as business model validation, user feedback integration, joint development, and crowdfunding. Moreover, our team was able to implement such services in a web platform (www.crowdserv.eu) and test those in the context of business incubators.

| Crowd-based Incubation Service | Traditional Equivalent | Advantages of Crowd |
|---|---|---|
| **Decision support** | Coaching/Mentoring | - Collective intelligence in validating entrepreneurial assumptions<br>- Heterogenous feedback and advice for developing the idea |
| **Co-creation** | Coaching/Mentoring, Access to Talent Network | - Heterogenous knowledge<br>- Distribution of workload<br>- Creative potential of diverse actors<br>- Active co-creation rather than mentoring |
| **Signalling** | Access to Analyst Networks | - Higher credibility of crowd signals to potential investors<br>- Higher credibility of crowd signals to potential customers<br>- Higher credibility of crowd signals to potential partners |
| **Financing** | Access to Funding Networks | - Higher number of investors<br>- Network effect among investors<br>- Dedicated investors<br>- Lower transaction cost |

*Examples of Crowd-based Services and Advantages*

For those services, my investigation of the limitations of previous services, and the cognitive rational for integrating collective intelligence in entrepreneurial decisions provide the general foundation. My findings, thus, provide several directions for managers in business incubators to digitize and enlarge their service portfolio. This will then allow to offer scalable services with higher quality to guide entrepreneurial efforts.

### 7.3.3. Business Model Tools for Entrepreneurial Decisions

Finally, the most valuable practical contribution of this dissertation is the provision of design knowledge for business model DSS as well as the actual implementation of an IT artefact.

First, my findings reveal instructions and concrete DPs that allow to build mechanisms and tools for integrating the collective intelligence of an entrepreneurś ecosystem in her decision-making. Furthermore, I offer DPs on how to design HI-DSS for complex decision-making problems. Therefore, my findings offer concrete construction and design instructions that may guide the development of such systems and mechanisms in the future and, thus, allow service providers such as business incubators offer solution for guiding entrepreneurs in the future.

Second, the implementation of my findings in Section 6.5 and the instantiated DPs are available as web application that can be used by entrepreneurs to validate their business model assumptions and receive decisional guidance based on the design paradigm of hybrid intelligence. This web application can be accessed via www.ai.vencortex.com.

## 7.4. Further Research

My thesis also indicates several directions for further research. While some of these areas for further research arise from limitations of this study, other directions arise due the novelty of this field of research. Especially for the field of hybrid intelligence for making predictions under uncertainty and risk and to solve complex dynamic decision problems, my thesis provides a first step towards several fields of future research opportunities. Those future research directions might guide researchers in conducting studies and shed light on issues that could not be answered by this study so far.

### 7.4.1. Direction 1: IT-based Business Model Tools

The first direction for future research is related to the design of IT-based business model tools. Although such tools were identified as valuable research field for IS (Veit et al. 2014), so far little is known on the design of such systems in general. My thesis focuses on the design of DSS that helps entrepreneurs in designing novel business models. However, various further tools might be relevant for the field as well.

One concrete directions for further research might be related to tools for business model mining (e.g. (Augenstein and Fleig 2017; Augenstein et al. 2018). Such tools can be used for instance to standardize start-up data for VCs, screen new ventures for long-listing in due diligence processes, or to assess the similarity among business models to guide further decision-making.

Such systems can capture structured data about the business model and should be able to (semi-) automatically extract also *"soft"* and latent information such as a companyś value proposition. Therefore, simple information retrieval of a webpage can lead to wrong, too less/much information or false assumptions. Data mining algorithms can provide an objective degree of abstraction of a business model.

Future research in this field might particularly focus on how publicly available data from websites, whitepapers, and business plans can be used to mine rather latent information on the logic of how a firm creates and captures value. Therefore, data mining techniques and hierarchical ontologies are required to learn latent classes from unstructured data and continuously adapt to dynamics. This direction is a valuable path for further work on IT-based business model tools.

### 7.4.2. Direction 2: Interaction Design in Hybrid Intelligence Systems

The second direction for further research is related to designing the interaction between human and computational agents in hybrid intelligence systems. To leverage the complementary benefits of humans and AI such interaction design is a crucial factor in determining the performance of hybrid systems. I suggest two directions for this purpose.

First, future work might focus on the integration of such interaction design and related mechanisms in existing business processes or system landscape. To make use of human input in hybrid intelligence systems the design of interaction that is convenient and user friendly for the human expert is one of the central issues. Thus, researchers might investigate how to integrate the interaction in the real-world business context.

Second, one central point for examining the interaction design between humans and artificial agents is the interdisciplinary field of interpretability. Such interpretability is defined as the degree to which a human user is able to understand the prediction that is provided by a model, thus, being understandable and predictable for the user (Lipton 2018). This allows then to ensure several traits of a ML model such as fairness, privacy, reliability and robustness, causality, and trust (Doshi-Velez and Kim 2017). While research on this topic is recently emerging and mostly covered from an algorithmic perspective of computer science, IS research might focus on the user centric development and requirements of interpretability in human and AI interaction design for hybrid intelligence systems.

### 7.4.3. Direction 3: Incentivation Mechanisms for Human Input

The third direction for future research is related to incentivation mechanisms for receiving human input. This is especially relevant when humans need high requirements on domain expertise, which is the case for various critical domains such as financial services, health care, strategic decisions, etc. For this thesis, I assumed that the existence of altruistic human mentors makes human input available for HI-DSS in entrepreneurial decision-making. However, the results of my evaluation in Section 6.5 reveals that further research is required to design suitable incentivation mechanisms for such contexts in which human input is extremely costly compared to traditional crowd work and well-studied incentivation mechanisms might not work properly (e.g. Leimeister et al. 2009).

One possible solution that I already started to explore and that seems to be valuable for such contexts is the application of blockchain-based tokens (e.g. Lipusch et al. 2019; Lipusch et al under review). Such tokens are characterized through three main characteristics: They reinforce/ incentivize a certain type of behaviour (they are conditional upon certain actions), they work within a clearly defined context and they are bound to access clearly defined privileges. For the context of human input of domain experts in hybrid intelligence systems, this approach provides two advantages. First, it incentivizes human experts for providing input. Second, it allows to bring *"skin in the game".* This means that experts can be committed to the systems output through having stakes (i.e. token) in the hybrid intelligence system, thus, ensuring an up-side and down-side potential of appropriate and inappropriate behaviour. Consequently, I suggest two specific directions for future research.

First, further work on the incentivation of domain experts in providing input to hybrid intelligence systems might investigate the application of curation markets for curating predictive information in complex and dynamic decision-making contexts. Such curation markets *"[…] are a model that allows groups to more effectively coordinate and earn from value they co-create around shared goals […]"* (La Rouviere 2017).

Second, future research should focus on the concrete, domain-specific engineering of tokens for incentivizing experts in hybrid intelligence systems. As blockchain-based technology and smart contracts ensure the automatically conducted, codified incentivation structures, the design of tokens is a central aspect for further investigation. Token engineering can be defined as *"[…] the theory, practice and tools to analyse, design, and verify tokenized ecosystems […]"* (Mc Conaghy 2018).

### 7.4.4. Direction 4: Design of DSS for Complex Problems

Finally, the fourth direction for future research is related to the design of DSS for complex and dynamic decision-making problems. This is probably one of the most interesting challenges for research at the intersection of interdisciplinary fields such as IS, cognitive science, and computer science. I suggest two distinctive directions in this vein.

First, future research might explore alternative applications of ML approaches in DSS that can deal with non-stationary, dynamic problems, self-adapt, and thus solve AI-complete tasks. While research in computer science currently works on the algorithmic side of such systems (Müller-Schloer and Tomforde 2017), future research in the field of IS might focus on the application of such approaches in the context of real-world business problems and concrete use in DSSs.

Second, research should build on the findings of this dissertation to improve and expand the design of HI-DSS. For instance, one direction might be making the human input more resource efficient through techniques such as active learning. On the other hand, research should apply the identified design paradigm of hybrid intelligence for other contexts that are part of the class of complex dynamic problems such as healthcare (e.g. www.viscortex.io). A further investigation of the design paradigm and concrete design principles provides the opportunity to develop generalizable design theories of HI-DSS as a novel class of artefacts in IS research.

# Author Bio

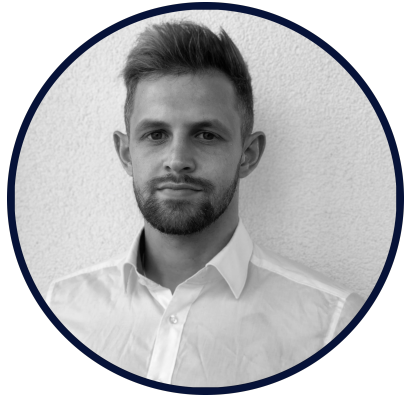

## Dominik Dellermann

Dominik Dellermann is co-founder of Vencortex, an AI-based decision augmentation platform for the Intelligent Enterprise empowering managers to validate business hypotheses in the uncertain, complex, and fast economy. Dominik holds a Ph.D. in Information Systems on inventing hybrid intelligence for augmenting business decisions and has over 5 years of working experience in the field of innovation and over 4 years of experience with R&D of AI technologies. The former professional athlete is a thought leader on combining human intuition and AI for the intelligent enterprise and speaker on topics such as AI, corporate innovation and M&A. Dominik published over 60 articles on those topics in world-leading outlets.